\documentclass[sigconf, nonacm]{acmart}
\pdfoutput=1

\newcommand{\tabincell}[2]{\begin{tabular}{@{}#1@{}}#2\end{tabular}}
\usepackage{color}
\usepackage{times}
\usepackage{bm}
\usepackage{subfigure}
\usepackage{makecell,rotating,multirow,diagbox}
\usepackage{multirow}
\usepackage{slashbox}
\usepackage[section]{placeins}
\usepackage{ragged2e}
\usepackage{float}
\usepackage{tablefootnote}

\renewcommand{\raggedright}{\leftskip=0pt \rightskip=0pt plus 0cm}

\begin{document}
\title{A Comparative Study on Unsupervised Anomaly Detection for Time Series: Experiments and Analysis}

\author{Yan Zhao$^{1}$, Liwei Deng$^{2}$, Xuanhao Chen$^{2}$, Chenjuan Guo$^{1}$, Bin Yang$^{1}$, Tung Kieu$^{1}$, Feiteng Huang$^{3}$, Torben Bach Pedersen$^{1}$, Kai Zheng$^{2}$, Christian S. Jensen$^{1}$}
\affiliation{%
  \institution{$^{1}$Aalborg University \\ $^{2}$University of Electronic Science and Technology of China \\ $^{3}$Huawei Cloud Database Innovation Lab, China}
  \institution{$^{1}$\{yanz, cguo, byang, tungkvt, tbp, csj\}@cs.aau.dk \hspace{5pt} $^{2}$\{deng\_liwei, xhc\}@std.uestc.edu.cn, zhengkai@uestc.edu.cn \hspace{5pt}
  $^{3}$huangfeiteng@huawei.com}
}

\justifying
\begin{abstract}
The continued digitization of societal processes translates into a proliferation of time series data that cover applications such as fraud detection, intrusion detection, and energy management, where anomaly detection is often essential to enable reliability and safety.
Many recent studies target anomaly detection for time series data.
Indeed, area of time series anomaly detection is characterized by diverse data, methods, and evaluation strategies, and comparisons in existing studies consider only part of this diversity, which makes it difficult to select the best method for a particular problem setting.
To address this shortcoming,
we introduce taxonomies for data, methods, and evaluation strategies, provide a comprehensive overview of unsupervised time series anomaly detection using the taxonomies, and systematically evaluate and compare state-of-the-art traditional as well as deep learning techniques.
In the empirical study using nine publicly available datasets, we apply the most commonly-used performance evaluation metrics to typical methods under a fair implementation standard.  
Based on the structuring offered by the taxonomies, we report on empirical studies and provide guidelines, in the form of comparative tables, for choosing the methods most suitable for particular application settings.
Finally, we propose research directions for this dynamic field.

\end{abstract}

\maketitle
\section{Introduction}
The continued, society-wide digitization and the accompanying deployment of sensing technologies generate increasingly massive amounts of time series data.
A time series is a sequence of observations recorded in chronological order.
In time series, a small number of observations may deviate significantly from most observations because they are generated by different processes. Depending on the setting and application domain, such observations are called anomalies, abnormalities, deviants, intrusions, outliers, failures, discordant observations, exceptions, aberrations, 
peculiarities, or contaminants~\cite{kwon2019survey,chandola2009anomaly}.
For simplicity, we use the term ``anomaly'' throughout the paper.

Anomaly detection for time series has been studied in diverse settings, such as credit card fraud detection, health care insurance, intrusion detection in cyber security, and fault detection in safety critical systems~\cite{srivastava2008credit,ganji2012credit,nune2015novel,7420931,chandola2009anomaly,joudaki2015using,govindarajan2014outlier,hong2014integrated,jyothsna2011review,buczak2015survey}.
Anomaly detection methods aim to identify observations that differ significantly from the bulk of observations.
Anomaly detection methods may be supervised or unsupervised.
The former require the availability of labels indicating ground-truth anomaly observations,
which is not the case in many real-world application settings. In contrast, unsupervised methods, the focus on this paper, do not require the availability of ground truth anomaly labels for their functioning and are more generally applicable. In particular, time-consuming and labor-intensive human labeling is not needed, and it becomes possible to identify unanticipated anomalies that may have gone unnoticed during manual labeling.

The wide range of existing unsupervised methods for time series anomaly detection can be classified as traditional methods~\cite{breunig2000lof,kriegel2009loop,tang2001robust,yeh2016matrix,liu2008isolation,tan2011fast,liu2010detecting,manevitz2001one} or as deep learning methods~\cite{Kieu0J18,krizhevsky2012imagenet,MalhotraRAVAS16,hundman2018detecting,ZhouLHCY19,deng2021hifi,chen2017outlier,KieuYGJ19}.
However, a key challenge when attempting to leverage this body of proposals for performing anomaly detection in real-world application settings is the lack of guidance as to which methods are appropriate for use in different settings.
In the case of supervised methods, the labels used offer such guidance, but a comprehensive mapping of unsupervised methods to application settings is needed.
To achieve this, we face two main challenges.

\begin{table*}
\scriptsize
\centering
\caption{Survey Comparisons and Contributions}
\label{tab:survey}
\renewcommand{\arraystretch}{1}
\begin{tabular}{|l|l|l|c|c|c|c|c|c|c|}
\hline
\textbf{Survey and benchmark } & \textbf{Application domain} &
\tabincell{c}{\textbf{Supervised/}\\\textbf{Unsupervised}}&\tabincell{c}{\textbf{Taxonomy}\\\textbf{of data}}& \tabincell{c}{\textbf{Taxonomy}\\\textbf{of methods}}& \tabincell{c}{\textbf{Taxonomy of}\\\textbf{evaluation}}& \tabincell{c}{\textbf{Traditional}\\\textbf{methods}}& \tabincell{c}{\textbf{Deep learning}\\\textbf{methods}}&\tabincell{c}{\textbf{Empirical}\\\textbf{study}}&\tabincell{c}{\textbf{Method}\\\textbf{recommendation}}
\\
\hline
2003~\cite{lazarevic2003comparative}	&	Network intrusion	&Unsupervised&	$\times$	&	$\surd$	&	$\times$	&	$\surd$	&	$\times$	&	$\surd$		&	$\times$	\\ \hline
2004~\cite{hodge2004survey}	&	General anomaly	&Both&	$\times$	&	$\surd$	&	$\times$	&	$\surd$	&	$\surd$	&	$\times$		&	$\times$	\\ \hline
2009~\cite{chandola2009anomaly}	&	General anomaly	&Both&	$\times$	&	$\surd$	&	$\times$	&	$\surd$	&	$\surd$	&	$\times$		&	$\times$	\\ \hline
2013~\cite{gupta2013outlier} &	Temporal data anomaly	&Both&	$\times$	&	$\surd$	&	$\times$	&	$\surd$	&	$\surd$	&	$\times$		&	$\times$	\\ \hline
2017~\cite{2017On}	&	Streaming data anomaly	&Unsupervised &	$\times$	&	$\surd$	&	$\times$	&	$\surd$	&	$\times$	&	$\surd$ &	$\surd$	\\ \hline
2018~\cite{kiran2018overview} &	Video anomaly	&Unsupervised&	$\times$	&	$\surd$	&	$\times$	&	$\times$	&	$\surd$	&	$\surd$		&	$\times$	\\ \hline
2019~\cite{chalapathy2019deep}	&	General anomaly	&Both&	$\surd$	&	$\surd$	&	$\times$	&	$\times$	&	$\surd$	&	$\times$		&	$\times$	\\ \hline
2019~\cite{kwon2019survey}	&	Network anomaly	&Both&	$\times$	&	$\surd$	&	$\times$	&	$\times$	&	$\surd$		&	$\times$	&	$\times$	\\ \hline
2020~\cite{jacob2020anomalybench}		&	Time series anomaly	&Supervised&	$\times$	&	$\times$	&	$\times$	&	$\times$	&	$\surd$	&	$\surd$		&	$\times$	\\ \hline
2020~\cite{pang2020deep}&	General anomaly	&Supervised&	$\surd$	&	$\surd$	&	$\times$	&	$\times$	&	$\surd$	&	$\times$		&	$\times$	\\ \hline
This work	&	Time series anomaly	&Unsupervised&	$\surd$	&	$\surd$	&	$\surd$	&	$\surd$	&	$\surd$	&	$\surd$	&	$\surd$	\\ \hline
\end{tabular}
\end{table*}

\emph{Challenge I: diversity of data, methods, and evaluation strategies.}
It is difficult to select a suitable unsupervised method among the many alternatives for a particular application because this amounts to a multi-criteria decision-making problem. First, it is necessary to take into account the complex and diverse nature of the underlying time series data, such as the dimensionality, the stationarity,
and the temporal correlations among observations.
Second, many unsupervised methods are available, and they use different ways of detecting anomalies, including density-based clustering~\cite{breunig2000lof,kriegel2009loop,tang2001robust}, similarity search~\cite{yeh2016matrix}, tree-based partitioning~\cite{liu2008isolation,tan2011fast,liu2010detecting}, one-class classification~\cite{manevitz2001one}, reconstruction~\cite{Kieu0J18,krizhevsky2012imagenet,MalhotraRAVAS16}, and prediction~\cite{hundman2018detecting,malhotra2015long}.
It is non-trivial to select a suitable method simply based on the description of the method. Third, unlike in time series prediction and classification, where the output is a value that straightforwardly indicates a future value or a class label, the outputs of time series anomaly detection methods are diverse and include density values, distance values, reconstruction errors, etc.
Different ways of deriving anomalies from such values exist, which also yields different strategies for evaluating anomalies. No single evaluation strategy exists that indicates the suitability of a method for an application. Rather, it occurs commonly that a method achieves good results according to one evaluation strategy but performs poorly according to another.
For example, precision and recall, which are used commonly for assessing performance, are often conflicting and may be traded for one another according to application requirements. To illustrate, AIOps (artificial intelligence for IT operations) applications require high precision, as operators do not want to be disturbed by frequent false alarms, while applications in intensive care units that monitor signs of life (e.g., respiratory rate and blood pressure) prioritize recall at the cost of precision~\cite{liu2015opprentice}.

\emph{Challenge II: fair comparison.}
The effectiveness of unsupervised methods depends highly on the settings of hyperparameters, which increases the difficulty of selecting a suitable method. As unsupervised methods train models without labels, no fair or golden standard exists for choosing suitable hyperparameter settings that maximize performance. In other words, it is challenging to select hyperparameters fairly and effectively in an unsupervised manner. This increases the need for being able to compare unsupervised methods fairly in different settings to understand thoroughly the advantages and disadvantages of these methods, such that their suitability for a specific application could be well demonstrated.

Therefore, as more and more methods become available, there is an increasing need for a comprehensive framework and study of the suitability of unsupervised time series anomaly detection methods for different application settings. Table 1 offers an overview of prior surveys and benchmarks, showing that these focus mostly on general anomaly detection, e.g., detecting novelties in images and identifying novel molecular structures in pharmaceutical research~\cite{hodge2004survey,chandola2009anomaly,chalapathy2019deep,pang2020deep}
or a particular application domain (e.g.,
video anomaly)~\cite{lazarevic2003comparative,gupta2013outlier,2017On,kiran2018overview,kwon2019survey}, instead of time series anomaly detection.
In addition, most of these existing studies do not provide comparative taxonomies for data, methods, and evaluation strategies, or they target only specific types of traditional or deep learning methods. Furthermore, there is an unmet need for a comparative empirical study of the fit of methods for given data and application settings. The only exception is the study of Choudhary et al.~\cite{2017On} that targets streaming data anomaly detection rather than time series anomaly detection.
In addition, it only selects a suitable method based on attributes exhibited by datasets from different application domains and latency requirements. It disregards general data attributes, e.g., dimensionality and stationarity, and evaluation strategies.

This work aims to address the above challenges by providing a comprehensive reference for researchers and practitioners seek to utilize unsupervised time series anomaly detection techniques in specific applications.
In particular, \emph{this study focuses on providing guidance on which unsupervised time series anomaly detection methods in different settings, and attempts to explore how to employ the methods on different data}.
Specifically, the paper makes three key contributions.

1) \textbf{Qualitative Analysis.} This paper offers a comprehensive overview of unsupervised time series anomaly detection by proposing taxonomies
for data, methods, and evaluation strategies.

2) \textbf{Quantitative Analysis.} The paper considers
$14$ carefully selected unsupervised anomaly detection methods, including representative traditional and deep learning methods, and evaluates them using nine publicly available datasets
in terms of accuracy, robustness, and running time efficiency while ensuring consistent main hyperparameter settings to ensure fairness.
We provide a comprehensive overview of the results. 
The related source code can be accessed at Github\footnote{https://github.com/yan20191113/ADTS}.

3) \textbf{Guidelines of Method Recommendation.} The experimental results and  accompanying analyses serve as guidelines to suggest an appropriate method for a particular dataset. Specifically, we offer several comparative tables that provide recommendations of techniques suitable for the applications at hand based on the proposed taxonomies and experimental evaluations. This paper can be used as a hands-on guide for understanding, using, or developing unsupervised anomaly detection methods for different real-world applications.

The remainder of the paper is structured as follows. We define taxonomies for time series anomaly detection data, methods, and evaluation strategies in Sec.~\ref{sec:taxonomy}.
Sec.~\ref{sec:expsetup} details the experimental setup,
followed by Sec.~\ref{sec:exp} that gives the experimental results and offers method recommendations.
We finally conclude the paper and provide research directions in Sec.~\ref{sec:conclu}.

\section{Taxonomies}\label{sec:taxonomy}
It is a multi-criteria decision-making problem to choose the best unsupervised time series anomaly detection method for a particular application, where the key criteria include the anomaly detection data, methods, and evaluation strategies. Therefore, we provide three taxonomies for the data, methods, and evaluation strategies, thus achieving a structured basis for method recommendation.

\subsection{Taxonomy of Data}\label{sec:adts}
Challenges in time series anomaly detection method recommendation in terms of data include diverse dimensionality and stationarity, based on which we propose a data taxonomy.
\subsubsection{Univariate vs. Multivariate}
The dimensionality is the number of features captured in each observation~\cite{chandola2009anomaly}.
A time series $T$ is a sequence of $k$-dimensional vectors $T =  \langle S_1,S_2,$ $...,S_n \rangle$, where 
$S_i = (s^1_i, s^2_i,..., s^k_i) \in \mathbb{R}^k$ denotes a $k$-dimensional vector ($k\ge 1$) that describes $k$ features of an observation (also referred to as an instance, record, or data point)
at time $t_i$ ($1\le i\le n$), and $s^j_i$ ($1\le j\le k$) corresponds to
the $j$th feature of $S_i$.

Time series data is classified into univariate (one-dimensional) and multivariate (multi-dimensional) time series.
Formally, $T$ is univariate when $k=1$ and multivariate when $k>1$.
Classical anomaly detection is concerned primarily with point anomalies, anomalies that occur at a single observation, which are also our focus since point anomaly detection is the foundation of other anomaly (like sequence) detection. The point anomaly detection problem is defined as follows.

Given a $k$-dimensional time series $T =  \langle S_1,S_2,...,S_n \rangle$, we aim at assigning each vector $S_i$ an anomaly score, denoted as $\mathit{score(S_i)}$. The higher the anomaly score $S_i$, the more likely it is that $S_i$ is an anomaly.
Based on the anomaly scores, we can rank the observations in a time series. 
We can consider the top $\alpha\%$ (e.g., $5\%$) of the observations, according to their anomaly scores, as anomalies.
Alternatively, given a fixed threshold, we regard observation $S_i$ as anomalous if its anomaly score exceeds the threshold.

For example, given a driver's time series, the anomaly observations may represent the occasions where the driver behaves differently from other occasions, e.g., when the driver encounters a dangerous situation.
In the rest of the paper, we use \emph{anomaly detection} and \emph{time series anomaly detection} interchangeably when the context is clear.

\subsubsection{Stationary vs. Non-stationary}
Time series can also be classified as stationary or non-stationary.
Since strictly stationary time series occur infrequently in practice, weak stationarity conditions are introduced applied~\cite{nason2006stationary,lee2017anomaly}, where the mean of any observation in a time series $T =  \langle S_1,S_2,...,S_n \rangle$ is constant and the variance is finite for all observations.
Also, the covariance between any two observations $S_i$ and $S_j$, $\mathit{cov}(S_i,S_j)$, depends only on their distance $|j-i|$, i.e., $\forall i+r\le n, j+r\le n$ $(\mathit{cov}(S_i,S_j) = \mathit{cov}(S_{i+r},S_{j+r}))$.

In contrast, non-stationary time series have statistical distribution properties that vary across time. 
Such time series have time-varying means and variances, so a time series that exhibits a trend (e.g., increasing or decreasing) across time is non-stationary. 

\subsection{Taxonomy of Methods}\label{sec:method}

\begin{table}
\scriptsize
\centering
\caption{A Taxonomy of Anomaly Detection Methods}
\label{tab:algoTax}
\renewcommand{\arraystretch}{1}
\begin{tabular}{|p{0.88cm}|p{1.15cm}|p{0.35cm}|p{0.94cm}|p{0.85cm}|p{0.7cm}|p{0.3cm}|p{0.3cm}|}\hline
\multirow{2}*{\tabincell{c}{\textbf{Learning}\\\textbf{capabilities}}}& \multirow{2}*{\tabincell{c}{\textbf{Method type}}} &
\multicolumn{2}{c|}{\multirow{2}*{\textbf{Method name}}}&
\multirow{2}*{\tabincell{c}{\textbf{Method}\\\textbf{integration}}} &
\multirow{2}*{\tabincell{c}{\textbf{Anomaly}\\\textbf{score}}}&
\multicolumn{2}{c|}{\tabincell{c}{\textbf{Temporal}\\\textbf{dependency}}} \\
\cline{7-8}
\multirow{2}*{}&\multirow{2}*{}&\multicolumn{2}{c|}{\multirow{2}*{}}&\multirow{2}*{}&\multirow{2}*{}&	\textbf{S2S}	& \textbf{SW} \\
\cline{1-8}
\multirow{2}*{NL}&	Density	&	\multicolumn{2}{l|}{LOF}		&	Single	&LOF & $\times$& $\times$ \\
\cline{2-8}
&	Similarity	&	\multicolumn{2}{l|}{MP}		&	Single	&\tabincell{l}{MP\\distance} & $\times$& $\surd$\\
\cline{1-8}
\multirow{2}*{CL}&	Partition	&	\multicolumn{2}{l|}{ISF}		&	Ensemble		&Anomaly score & $\times$& $\times$ \\
\cline{2-8}
&	Classification	&	\multicolumn{2}{l|}{OC-SVM}		&	Single		&\tabincell{l}{Signed\\distance} & $\times$& $\times$\\
\cline{1-8}
\multirow{10}*{DL}&	\multirow{9}*{Reconstruction}	&	\multirow{6}*{AE}		&	CNN-AE	&Single	&Rec. err.& $\surd$& $\surd$   \\
\cline{4-8}
\multirow{10}*{}&	\multirow{9}*{}&\multirow{6}*{}
&2DCNN-AE 	&	Single		&Rec. err.& $\surd$ & $\surd$ \\
\cline{4-8}
\multirow{10}*{}&	\multirow{9}*{}&\multirow{6}*{}
&LSTM-AE	&	Single		&Rec. err.& $\surd$& $\surd$  \\
\cline{4-8}
\multirow{10}*{}&	\multirow{9}*{}&\multirow{6}*{}
&HIFI	&	Single		&Rec. err.&
$\surd$& $\surd$  \\
\cline{4-8}
\multirow{10}*{}&	\multirow{9}*{}&\multirow{6}*{}
&RN	&	Ensemble		&Rec. err.&
$\times$& $\surd$ \\
\cline{4-8}
\multirow{10}*{}&	\multirow{9}*{}&\multirow{6}*{}
&RAE	&	Ensemble	&Rec. err.&$\surd$&$\surd$ \\
\cline{3-8}
\multirow{10}*{}&	\multirow{9}*{}&\multirow{1}*{VAE}		&	Omni	&Single	&Rec. err.& $\surd$&$\surd$ \\
\cline{3-8}
\multirow{10}*{}&	\multirow{9}*{}&
\multirow{1}*{GAN}		&	BeatGAN	&Single	&Rec. err.& $\surd$& $\surd$ \\
\cline{2-8}
\multirow{10}*{}&\multirow{1}*{Prediction}&	\multicolumn{2}{l|}{LSTM-NDT}		&	Single		&Pre. err. &$\surd$& $\surd$ \\
\cline{1-8}
\end{tabular}
\end{table}

We classify methods according to three main categories: \emph{non-learning (NL)}, \emph{classical learning (CL)}, and \emph{deep learning (DL)} methods, based on their learning capabilities.

Within the three main categories, we identify six fine-grained categories according to the core concepts of the methods, as shown in Table~\ref{tab:algoTax}. It also presents the method integration, output, and temporal dependencies that includes two ways of capturing temporal dependencies, i.e., sequence to sequence (S2S) and sliding windows (SW).
Due to space limitation, we only study the most representative methods in each category. 

\subsubsection{Non-learning (NL) Methods}
NL methods include density and similarity based methods.

\noindent\textbf{Density-based Methods.} 
The density-based methods~\cite{breunig2000lof,kriegel2009loop,tang2001robust} first group observations into clusters and then utilize the density of the clusters to discover anomalies.
We illustrate density-based methods using a classical method, Local Outlier Factor (LOF)~\cite{breunig2000lof}.

For each observation $S_i$ in a time series $T$, LOF first computes a $g$-distance neighborhood of
$S_i$, i.e., a cluster, denoted as $N_g(S_i) = \{S'\in T-S_i \mid d(S_i, S')\le d_g(S_i)\}$, where $d(S_i,S')$ is the distance between $S_i$ and $S'$, and $d_g(S_i)$, the $g$-distance of $S_i$, is the distance between $S_i$ and its $g$th nearest neighbor.
The observations in the $g$-distance neighborhood of $S_i$ are called the $g$-nearest neighbors of $S_i$.
Second, for each neighbor $S'\in N_g(S_i)$, LOF computes its reachability distance to $S_i$ that is the maximum of the $g$-distance of $S'$ and the distance between $S'$ and $S_i$, denoted as $\mathit{rd_g(S_i,S')} =$ $\mathit{\max\{d_g(S')},$ $d(S_i,S')\}$.
Third, LOF computes the local reachability density of $S_i$ by Eq.~\ref{equ:lrd}.
\begin{equation}
\label{equ:lrd}
\mathit{lrd}_{g}(S_i)=1 /\left(\frac{\sum_{S' \in N_{g}(S_i)} \mathit{rd}_{g}(S_i,S')}{\left|N_{g}(S_i)\right|}\right)
\end{equation}

Finally, the LOF value (i.e., anomaly score) of $S_i$ is calculated:
\begin{equation}
\label{equ:lof}
\mathit{LOF}_{g}(S_i)=\frac{\sum_{S' \in N_{g}(S_i)} \frac{\mathit{lrd}_{g}(S')}{\mathit{lrd}_{g}(S_i)}}{\left|N_{g}(S_i)\right|}
\end{equation}

A higher $\mathit{LOF}_{g}(S_i)$ indicates lower density of $S_i$, meaning that $S_i$ is more likely to be anomalous, and being larger than $1$ means that $S_i$ comes from  a low-density area.
LOF 
takes $T$ as a set of points, not as a sequence of points. Therefore, LOF fails to capture temporal dependencies in time series.

\noindent\textbf{Similarity-based Methods.}
In similarity-based methods,
time series that are similar to others are regarded as normal while the rest are reported as anomalous.
Matrix Profile (MP)~\cite{yeh2016matrix} is an efficient and representative similarity-based method.

MP first splits $T$ into overlapped subsequences of length $m$. 
Then, MP computes the distance (similarity) between a pair of subsequences, where
$\langle S_{max\{j-\frac{m}{2},1\}},...,S_{j},...,$ $S_{min\{j+m+\frac{m}{2}-1,|T|\}} \rangle$ is excluded since it almost matches itself, i.e., the number of same observations between two subsequences is not less than half of their length ($\frac{m}{2}$), leading to a trivial distance, where $j=\{1,2,...,|T|\}$, and $|T|$ denotes the length of $T$.
The MP distance of a subsequence is the minimum distance between it and any other subsequences that generate non-trivial distances.
If the MP distance (i.e., anomaly score) of a subsequence is relatively high, it is unique in the time series and is more likely to be an anomaly.
MP detects anomalies based on subsequences, meaning that it is able to capture temporal dependencies in time series.

\subsubsection{Classical Learning Methods}
The classical learning methods include partition-based and classification-based methods.

\noindent\textbf{Partition-based Methods.}
The partition-based methods partition observations into different parts.
Isolation Forest (ISF)~\cite{liu2008isolation} is a typical partition-based method.

ISF first constructs $m$ isolation trees, denoted as \emph{iTree}.
To generate each \emph{iTree}, we randomly sample $\psi$ observations from $T$, denoted as $T(\psi) =\{S_1, S_2, ..., S_{\psi}\}$.
Then we recursively divide $T(\psi)$ into sub-trees by randomly selecting a feature $q$ among the $k$ features of an observation and a split value $p$ between the maximum and minimum values of the selected feature $q$, where the observations with values smaller than $p$ are put in the left sub-tree and the others are put in the right sub-tree, until one of the following conditions is satisfied:

1) The \emph{iTree} reaches a height limit, i.e., $l=\mathit{ceiling}(log_2\psi)$, where $l$ is approximately the average tree height~\cite{donald1999art}.
We are only interested in observations with below average path lengths, as those observations are more likely to be anomalies.

2) There is only one observation in the leaf nodes of  \emph{iTree}.

3) All the observations in the leaf nodes have the same value.

Subsequently, ISF calculates the
anomaly score of a given observation $S_i$, $\mathit{score}(S_i,\psi)$, in the following.
\begin{equation}\label{equ:score}
\mathit{score}(S_i,\psi)=2^{-\frac{E(h(S_i))}{c(\psi)}}
\end{equation}
\begin{equation}\label{equ:cn}
c(\psi)=2 H(\psi-1)-(2(\psi-1) / \psi),
\end{equation}
where $h(S_i)$ is the path length of $S_i$, $E(h(S_i))$ is the average of $h(S_i)$ from a collection of \emph{iTree}s, $c(\psi)$ is an estimate of the average path length given $\psi$, and $H(i)$ is a harmonic number that is estimated by $ln(i) + 0.5772156649$ (Euler's constant).
Since ISF consists of multiple \emph{iTree}s (i.e., multiple classifiers)  built iteratively, it is an ensemble method. Further, it regards $T$ as independent points and cannot capture temporal dependencies in time series.

\noindent\textbf{Classification-based Methods.}
The classification-based methods find a small region that contains most normal observations,
and the observations outside of this region are considered as anomalies.
One-Class Support Vector Machines (OC-SVM)~\cite{manevitz2001one} belongs to this category.

OC-SVM uses a kernel map, $\Phi:S\to H$, to transform each $k$-dimensional vector $S_i\in T$ into a high-dimensional feature space $H$.
Next, OC-SVM separates the mapped vectors from the origin by a hyperplane, and it maximizes the distance of this hyperplane to the origin. This yields a small region that contains most observations.
To achieve this, OC-SVM solves a quadratic programming problem to get parameters, $w$ and $\rho$.
Then it uses the decision function, $f(S_i)=\operatorname{sign}((w \cdot \Phi(S_i))-\rho)$,
to detect anomalies.
The output, $f(S_i)$, of OC-SVM is a positive/negative value, and $S_i$ is likely to an anomaly when $f(S_i)$ is negative. Then the signed distance that is the distance between $S_i$ (with negative $f(S_i)$) and the hyperplane is calculated to denote the anomaly score of $S_i$.
OC-SVM is a single method, and it cannot capture temporal dependencies in time series.

\subsubsection{Deep Learning Methods}

DL methods are primarily reconstruction or prediction based.

\noindent\textbf{Reconstruction-based Methods.}
These methods compress original input data into a compact, hidden representation and then reconstruct the input data from the hidden representation.
Since the hidden representation is compact, it is only possible to reconstruct representative features from the input, not the specifics of the input data, including any anomalies.
This way, the reconstruction errors, i.e., the difference between the
original data and the reconstructed data, indicate how likely it
is that observations in the data are anomalies.

\noindent\underline{AE-based methods.}
The autoencoder (AE) architecture is the main architecture for a variety of reconstruction-based methods that output a reconstruction error for each observation $S_i\in T$ that captures its anomaly degree.

Specifically, an autoencoder, consisting of an encoder and a decoder, outputs a so-called reconstructed $k$-dimensional vector $S_i^{'} = (s_i^{' 1},s_i^{'2},...,$ $s_i^{'k})$. 
The encoder and decoder can be defined by $f_e$ and $f_d$:
\begin{equation}\label{equ:autoencoder}
\begin{array}{l}
\text { Encoder: } f_e(S_i) : \mathbb{R}^{k} \rightarrow \mathbb{R}^{m},\\
\text { Decoder: } f_d (X): \mathbb{R}^{m} \rightarrow \mathbb{R}^{k}
\end{array}
\end{equation}

Here, $X\in \mathbb{R}^{m}$ ($m<k$) denotes an intermediate $m$-dimensional vector mapped from the original input vector $S_i\in \mathbb{R}^{k}$ by the encoder. Then the decoder maps $X$ to a reconstructed vector $S_i^{'} \in \mathbb{R}^{k}$.

By design, an autoencoder is unable to reconstruct anomalies well,
which then yields large reconstruction errors for anomalies.
During training, we minimize the reconstruction
error between $S_i$ and $S_i^{'}$ using a distance measure, e.g., $L1$ and $L2$ distance. Existing learning methods, e.g., gradient descent and back propagation, can be used to learn functions $f_e$ and $f_d$.

In terms of base models adopted in the encoder and decoder, AE-based methods include  CNN-based autoencoders (e.g., CNN-AE and 2DCNN-AE)~\cite{Kieu0J18,krizhevsky2012imagenet} and RNN-based autoencoders (e.g., LSTM-AE)~\cite{MalhotraRAVAS16}.
Recently, as an attention-based AE structure, a Transformer (e.g., HIFI~\cite{deng2021hifi}) is proposed to detect anomalies in time series using
stacked self-attention and point-wise, fully connected layers for both the encoder and decoder.

Ensemble models (e.g., RN~\cite{chen2017outlier} and RAE~\cite{KieuYGJ19}) with AE use a number of different AEs to observe each observation, and a voting mechanism, \emph{stacking}, is often employed to combine the outputs from models, where multiple models are trained 
using the entire training data~\cite{deng2014ensemble}.
RN trains a set of independent AEs, each having some randomly selected constant dropout connections, 
and RAE uses sparsely-connected RNN-based AEs for anomaly detection, where multiple autoencoders are trained independently or jointly in a multi-task learning fashion.

\noindent\underline{VAE-based methods.}
A variational autoencoder (VAE) is an AE variant rooted in Bayesian inference~\cite{kingma2013auto}. Unlike AEs, VAEs model the underlying distribution of the original data and generate reconstructed data by introducing a set of latent random variables in the AEs.
VAEs use a KullbackLeibler (KL)-divergence penalty to impose a prior distribution on the latent variables that capture the new pattern of the input time series (that does not appear in the training set) to some extent.
Omni~\cite{su2019robust} is a typical VAE-based model for time series anomaly detection.

\noindent\underline{GAN-based methods.}
Generative Adversarial Networks (GANs) establish min-max adversarial games between a generator ($G$) and a discriminator ($D$), which are implemented as neural networks.

Following the architecture of a regular GAN~\cite{goodfellow2014generative}, $G$ and $D$ are trained with the following two-player min-max game:
\begin{equation}\label{equ:gan}
\min_{G} \max_{D} \mathbb{E}_{S \sim p_{\mathit{data}}(S)}[\log D(S)]+\mathbb{E}_{z \sim p(z)}[\log (1-D(G(z)))],
\end{equation}
where $G(z)$ defines a probability distribution for the samples generated by the generator $G$, $D(S)$ represents the probability that $S$ comes from the real data, $S$ denotes a real sample from $T$, $p_{\mathit{data}}(S)$ denotes the real data distribution, $z$ denotes a noise sample from a random latent space, and $p(z)$ is a prior on the input noise variables. BeatGAN~\cite{ZhouLHCY19} is a time series anomaly detection method using the GAN architecture.

All the above reconstruction-based methods except RN can capture temporal dependencies in time series in two ways: sequence to sequence (S2S) and sliding widows (SW).
RN can only capture temporal dependencies using SW.

\noindent\textbf{Prediction-based Methods.}
The prediction-based methods 
predict current observation using representations of previous observations. The process captures temporal and recurrent dependencies within a given sequence length~\cite{pang2020deep}.

In particular, given a $t$-length time series $T =  \langle S_1,S_2,...,S_t \rangle$, this method predicts a future observation $\hat{S}_{t+1}$ by making it as close to the ground truth $S_{t+1}$ as possible.

The prediction error (Pre. err.) between $\hat{S}_{t+1}$ and $S_{t+1}$ is defined as the anomaly score for $S_{t+1}$, where a higher prediction error means that $S_{t+1}$ is more likely to be an anomaly.

\subsection{Taxonomy of Evaluation Strategies}\label{sec:ade}
As the interest in time series anomaly detection has grown, evaluation strategies for anomaly detection methods have also received increased attention, mirroring the fact that the effectiveness and efficiency of anomaly detection are of high interest to researchers and practitioners.
The taxonomy of evaluation strategies and their metrics is shown in Table~\ref{tab:taxEvaluation}.

\begin{table}[!htbp]
\scriptsize
\centering
\vspace{-0.2cm}
  \caption{Taxonomy of Evaluation Strategies}
  \label{tab:taxEvaluation}
  \vskip -9pt
  \renewcommand{\arraystretch}{1}
  \begin{tabular}{|l|l|l|}\hline
    \multicolumn{2}{|c|}{\textbf{Evaluation strategies}}& \tabincell{c}{\textbf{Metrics}} \\
    \cline{1-3}
    \multirow{2}*{Effectiveness}  &     \multirow{1}*{Accuracy} & \multirow{2}*{\tabincell{l}{No thresholds:  \emph{ROC-AUC}, \emph{PR-AUC}\\  Specified thresholds:   \emph{precision}, \emph{recall}, \emph{$\mathit{F1}$ score}}} \\
    \cline{2-2}
    \multirow{2}*{} &\multirow{1}*{Robustness}& \multirow{2}*{}\\
\cline{1-3}
    \multirow{2}*{Efficiency}  &     \multirow{1}*{Training efficiency}  & \multirow{1}*{\emph{training time}}\\
\cline{2-3}
\multirow{2}*{} &\multirow{1}*{Testing efficiency}  & \multirow{1}*{\emph{testing time}}   \\
\cline{1-3}
\end{tabular}
\vspace{-0.5cm}
\end{table}

\subsubsection{Effectiveness}\label{sec:evalumetrics}
The effectiveness of anomaly detection algorithms encompasses aspects such as accuracy and robustness.
We first introduce the metrics, which are used throughout the
following accuracy and robustness.

\noindent\textbf{Metrics.}
Two situations, i.e., no and specified thresholds, exist when evaluating effectiveness.

\noindent\underline{No Thresholds.}
Anomaly detection methods assign an anomaly score to each observation in the testing data that captures the degree to which that observation is considered as an anomaly.
Thus, the output of such methods is a ranked list of anomalies.
Two metrics that consider all possible thresholds are adopted to evaluate these methods, including Area Under the Curve of Receiver Operating Characteristic (\emph{ROC-AUC}) and Area Under the Curve of Precision-Recall curve (\emph{PR-AUC}), where the \emph{ROC} and \emph{PR} curves are evaluation tools for binary classification that enable the visualization of performance at any threshold. Specifically, in the ROC curve, we take the anomaly score of each observation as a threshold and compute the False Position Rate (FPR) and the True Positive Rate (TPR) for each threshold, where FPR and TPR are X-axis and Y-axis, respectively, as follows.
\begin{equation}
\label{equ:FPR}
\operatorname{\mathit{FPR}}=\frac{\mathit{FP}}{\mathit{FP}+\mathit{TN}},
\operatorname{\mathit{TPR}}=\frac{\mathit{TP}}{\mathit{TP}+\mathit{FN}},
\end{equation}
where $\mathit{FP}$ is the number of false positives, $\mathit{TN}$ is the number of true negatives,
$\mathit{TP}$ is the number of true positives, and $\mathit{FN}$ is the number of false negatives.

In the PR curve, we also take the anomaly score of each observation as a threshold and compute the corresponding \emph{precision} and \emph{recall} ($\mathit{TPR}$) as the X-axis and the Y-axis, respectively.
\begin{equation}
\label{equ:precision}
\operatorname{\mathit{precision}}=\frac{\mathit{TP}}{\mathit{TP}+\mathit{FP}}
\end{equation}
\begin{equation}
\label{equ:recall}
\operatorname{\mathit{recall}}=\mathit{TPR}=\frac{\mathit{TP}}{\mathit{TP}+\mathit{FN}}
\end{equation}

\emph{ROC-AUC} (\emph{PR-AUC}) is a single number that summarizes the information in the \emph{ROC} (\emph{PR}) curve.
The two metrics do not depend on a specific threshold and reflect the full trade-off among true positives, true negatives, false positives, and false negatives.

\noindent\underline{Specified Thresholds.}
When a threshold is specified,
a binary label (normal or anomalous) is assigned to each observation according to whether its anomaly score exceeds the threshold. We adopt \emph{precision}, \emph{recall}, and \emph{$\mathit{F1}$ score} as metrics to measure the method accuracy, where 
\emph{$\mathit{F1}$ score}, $\mathit{F1}$, is calculated in follows.
\begin{equation}
\label{equ:f1}
\mathit{F1}=2 * \frac{\mathit{precision} * \mathit{recall}}{\mathit{precision }+\operatorname{\mathit{recall}}}
\end{equation}
The \emph{$\mathit{F1}$ score} is the harmonic mean of \emph{precision} and \emph{recall}, which represents a balance between the two.

Three commonly-used unsupervised threshold selection methods~\cite{yang2019outlier}, i.e., Standard Deviation (SD)\footnote{The observations in $[\mathit{mean}-\alpha * \mathit{SD}, \emph{mean}+\alpha * \mathit{SD}]$ are regarded as normal, where $\mathit{mean}$ and $\mathit{SD}$ are the mean and standard deviation of the anomaly scores, respectively, and $\alpha$ is often set to $3$.}, Median Absolute Deviation (MAD)\footnote{The observations in $[\mathit{median}(X)-\alpha * \mathit{MAD}, \mathit{median}(X)+\alpha *\mathit{MAD}]$ are regarded as normal, where $\mathit{median}(X)$ and $\mathit{MAD}$ are the median and median absolute deviation of the anomaly scores, respectively, $X$ is the anomaly scores of the observations, $\alpha$ is specified by the user and usually set to $3$,  $\mathit{MAD}=b*\mathit{median}(|X-\mathit{median}(X)|)$, and $b$ is suggested to be $1.4826$~\cite{simmons2011false}.}, and Interquartile Range (IQR)\footnote{The IQR is the difference between the values ranking in $25\%$ and $75\%$ among the whole anomaly scores, where the values are denoted as $Q1$ and $Q3$, respectively. The observations in $[Q1-c*\mathit{IQR}, Q3-c*\mathit{IQR}]$ are regarded as normal, where $c=1.5$, and $\mathit{IQR}= Q3-Q1$.}, are often adopted for threshold selection in time series anomaly detection.

\noindent\textbf{Accuracy.}
The accuracy of anomaly detection methods can simply be measured based on the anomaly scores or labels in testing data, when ground truth anomaly labels of
observations are available. However, due to the unsupervised
setting, we do not employ such labels for training, but use them only for evaluating accuracy.
We use the five metrics, \emph{ROC-AUC}, \emph{PR-AUC}, \emph{precision}, \emph{recall}, and \emph{$\mathit{F1}$ score} to measure the accuracy of anomaly detection methods. Higher values of the five metrics indicate higher accuracy.

\noindent\textbf{Robustness.}
Generally, an anomaly detection method is characterized as being robust if it is able to retain its detection accuracy with noisy
data. Therefore, when it comes to robustness analysis,
the accuracy metrics of methods are generally assessed by applying them to noisy data. Robustness is important because applications typically are faced with noisy data. Noise represents fluctuations in the reported values which may not be significant to the overall structure of the data as a whole and may be caused by minor variations in the sensitivity of sensors, or by unrelated events occurring within the vicinity of a sensor.

\subsubsection{Efficiency}
The efficiency of anomaly detection methods is studied extensively in the literature.
Due to real-time requirements, anomaly detection efficiency consists of training efficiency (i.e., \emph{training time}) and testing efficiency (i.e., \emph{testing time}), which indicate whether the methods can support frequent periodic training and real-time anomaly detection in streaming settings.
The evaluation of efficiency depends on
many different factors, such as the dataset size, the data dimensionality, and the choice of parameters, in addition to implementation aspects~\cite{kriegel2017black}.
In the experiments (cf. Sec.~\ref{sec:expEfficiency}), we evaluate the efficiency of different methods on datasets of varying size and dimensionality and with different parameter settings in a fair implementation environment.

\section{Experimental Setup}\label{sec:expsetup}
\subsection{Datasets}
We conduct experiments on nine publicly available datasets of different dimensions, stationarity, and sizes.

\subsubsection{Univariate Datasets}
\begin{enumerate}
\item \textbf{NAB.} Numenta Anomaly Benchmark (NAB)\footnote{https://github.com/numenta/NAB} is a benchmark for evaluating algorithms of anomaly detection in streaming, and real-time applications. It is comprised of $58$ univariate real and artificial time series, with $1,000$ to $22,000$ vectors per time series. The time series are collected from a wide variety of applications, primarily focusing on real-time anomaly detection for streaming data~\cite{lavin2015evaluating}, which range from network traffic monitoring and CPU utilization in cloud services to industrial machine operation monitoring and social media activities. For each time series, each vector is associated with a manually supplied Boolean outlier label, indicating whether the vector is an outlier.
We remove the ``machine\_temperature\_system\_failure.csv'' subset, which is unreadable, and use the remaining $57$ subsets to conduct the experiments. 

\item \textbf{S5.}
Yahoo's Webscope S5 (S5)\footnote{https://webscope.sandbox.yahoo.com/} consists of real and synthetic time series with tagged anomaly points. The dataset tests the detection accuracy of various anomaly types including outliers and change-points.
S5 contains $371$ time series that are organized into four groups, where two groups (i.e., \emph{A1Benchmark} and \emph{A2Benchmark}) are for outlier detection, and the remaining two groups are for change-point detection. We only use 
time series of the two outlier detection groups.
These time series are univariate with lengths that range from $1,482$ to $2,922$. Similar to NAB, each vector in the time series has a Boolean outlier label.

\item \textbf{KPI.}
The $2018$ AIOps's Key Performance Index (KPI)\footnote{https://github.com/NetManAIOps/KPI-Anomaly-Detection} anomaly detection dataset consists of KPI time series data from many real scenarios of Internet companies with ground truth labels.

\end{enumerate}

\subsubsection{Multivariate Datasets}

\begin{enumerate}
\item \textbf{ECG.}
The Electrocardiograph (ECG) dataset is from MIT-BIH Arrhythmia Database\footnote{https://physionet.org/content/mitdb/1.0.0/}, which contains $48$ half-hour excerpts of two-channel ambulatory ECG recordings.
ECG comprises seven two-dimensional time series from seven patients, where each time series has $3,750$ to $5,400$ observations, and the ground truth labels of anomaly observations are available.

\item \textbf{NYC.}
The New York City (NYC) taxi passenger data stream was provided by the New York City Transportation Authority, which was preprocessed (aggregated at $30$ mininus intervals) by Cui et al.~\cite{cui2016comparative}.

\item \textbf{SMAP.}
The Soil Moisture Active Passive satellite (SMAP) dataset, a public dataset from NASA~\cite{hundman2018detecting}, has a training and a testing dataset, where anomalies in each testing subset have been labeled. It has $55$ subsets and includes $25$-dimensional data, the anomaly ratio of which is $13.13\%$.
The observations in the dataset are equally-spaced $1$ minute apart, and we remove the ``P\-2'' subset since it is labeled twice.

\item \textbf{Credit.}
In the Credit Card Fraud Detection (marked as Credit)\footnote{http://www.ulb.ac.be/di/map/adalpozz/data/creditcard.Rdata} dataset, anonymous credit card transactions are labeled as fraudulent or genuine.
The dataset contains transactions made by credit cards in September $2013$ by European card-holders.
This dataset presents transactions occurred in two days, where we have $492$ frauds out of $284,807$ transactions. The dataset is highly unbalanced, the positive class (frauds) account for $0.172\%$ of all transactions.

\item \textbf{SWaT.}
For Secure Water Treatment (SWaT)\footnote{http://itrust.sutd.edu.sg/research/dataset}, the system takes $5$--$6$ hours to reach stabilization. So we eliminate the first 21,600 instable samples generated in the first $5$--$6$ hours.

\item \textbf{MSL.}
Same with SMAP, the Mars Science Laboratory rover (MSL) dataset is also a public dataset from NASA~\cite{hundman2018detecting}.
It has $27$ subsets and consists of $55$-dimensional data, which has its own training set and testing set. The anomaly ratio is $10.72\%$.
The observations in the dataset are equally-spaced $1$ minute apart, and we remove the ``P\-2'' subset since it is labeled twice.
\end{enumerate}

Dataset statistics are provided in Tables~\ref{tab:data} and~\ref{tab:DimAndSta}. 
As datasets NAB, S5, ECG, and Credit do not specify their training and testing data, we use the first $70\%$ of the data for training, and the rest for testing. 
KPI, NYC, SMAP, SWaT, and MSL have specified training and testing sets.
For all datasets, the last $30\%$ of the training set is chosen as the validation set to train DL methods by generating validation errors, which is illustrated in Sec.~\ref{sec:expDL}.
The stationarity of the subsets in NAB is shown in Table~\ref{tab:DimAndSta}, which are group according to their applications and stationarity.
Since the stationarity of the subsets in NAB varies, we choose four stationary subsets including ``NAB-art-stationary'', ``NAB-cpu-stationary'', ``NAB-ec2-stationary'', and ``NAB-elb-stationary''
and four non-stationary subsets including ``NAB-ambient-nonstationary'', ``NAB-ec2-nonstationary'', ``NAB-exchange-nonstationary'', and ``NAB-grok-nonstationary'' for use in experiments.
The subsets of each dataset represent different applications and are trained separately.

\subsection{Methods Compared}\label{sec:expmethod}
Based on the taxonomy in Table~\ref{tab:algoTax}, we select thirteen state-of-the-art methods for time series anomaly detection, covering both traditional and DL methods.
We have tried to establish fair comparisons among the methods by setting the main hyperparameters using a grid search method (i.e., setting the hyperparameters to several typical values) and
maintaining other hyperparameters at their default settings (as in the original studies) or by keeping the settings as consistent across the methods as possible.

\subsubsection{Traditional Methods}
The traditional methods include non-learning (LOF and MP) and classical learning (e.g., ISF and OC-SVM) methods. 

In LOF~\cite{breunig2000lof}, 
we vary the number of neighbors among $10$, $20$, $30$, $40$, and $50$, and the default value is $50$. The leaf size is set to $50$.

In MP~\cite{yeh2016matrix},
we vary the subsequence length among $16$, $32$, $64$, $128$, and $256$, where the default value is $128$.

In ISF~\cite{liu2008isolation},
we vary the number of base estimators 
among $20$, $40$, $60$, $80$, and $100$, and the default value is $100$.

In OC-SVM~\cite{manevitz2001one},
the upper bound on the fraction of training errors is set to $0.1$, and the tolerance for the stopping criterion is $1e-4$.

\begin{table}
\scriptsize
  \caption{Statistics of Datasets}
  \label{tab:data}
  \vskip -9pt
  \renewcommand{\arraystretch}{1}
  \begin{tabular}{|l|l|l|l|l|l|}\hline
    \tabincell{c}{\textbf{Name}}  & \tabincell{c}{\textbf{Subset}\\\textbf{count}}&\tabincell{c}{\textbf{Training set}\\\textbf{cardinality}} &\tabincell{c}{\textbf{Testing set}\\\textbf{cardinality}}&\tabincell{c}{\textbf{Anomaly ratio,}\\ \textbf{training set ($\%$)}}& \tabincell{c}{\textbf{Anomaly ratio,}\\\textbf{testing set ($\%$)}} \\\hline
    NAB  &57&239,982&102,881&8.5&10.36  \\\hline
	S5& 167&165,763&71,203&0.66&1.47  \\\hline
KPI	&1&2,102,846&	901,220		&3.29&1.14\\\hline
ECG&	7	&21,838	&9,364		&4.51&5.75\\\hline
NYC&	1&	13,104&	4,416&	5.07&	2.24\\\hline
SMAP&	55&	135,183&	427,617&	0&	13.13\\\hline
Credit	&1&	199,364	&85,443		&0.19&0.13\\\hline
SWaT&	1	&475,200&	449,919	&0&	12.14\\\hline
MSL	&27	&58,317&	73,729&	0&	10.72\\\hline
\end{tabular}
\vspace{-0.1cm}
\end{table}

\begin{table}[!htbp]
\scriptsize
\centering
  \caption{Dimensionality and Stationarity of Datasets}
  \label{tab:DimAndSta}
  \vskip -9pt
  \renewcommand{\arraystretch}{1}
  \begin{tabular}{|l|l|l|l|}\hline
    \multicolumn{2}{|c|}{\textbf{Name}}& \tabincell{c}{\textbf{Dimensionality}} & \multicolumn{1}{c|}{\textbf{Stationarity}} \\
    \cline{1-4}
    \multirow{4}*{NAB}

    & NAB-art-stationary &1&Stationary  \\
    \cline{2-4}
    & NAB-cpu-stationary &1&Stationary  \\
    \cline{2-4}
    & NAB-ec2-stationary &1&Stationary  \\
     \cline{2-4}
    & NAB-elb-stationary &1&Stationary  \\
    \cline{2-4}
     & NAB-nyc-stationary &1&Stationary  \\
    \cline{2-4}
    & NAB-occupancy-stationary &1&Stationary  \\
    \cline{2-4}
    & NAB-rds-stationary &1&Stationary  \\
        \cline{2-4}
    & NAB-rogue-stationary &1&Stationary  \\
    \cline{2-4}
    & NAB-speed-stationary &1&Stationary  \\
    \cline{2-4}
    & NAB-TravelTime-stationary &1&Stationary  \\
    \cline{2-4}
    & NAB-Twitter-stationary &1&Stationary  \\
    \cline{2-4}
& NAB-exchange-stationary &1&Stationary  \\
    \cline{2-4}
    & NAB-ambient-nonstationary &1&Non-stationary  \\
\cline{2-4}
    & NAB-ec2-nonstationary &1&Non-stationary  \\
    \cline{2-4}
      & NAB-exchange-nonstationary &1&Non-stationary  \\
    \cline{2-4}
    & NAB-grok-nonstationary &1&Non-stationary  \\
    \cline{2-4}
    & NAB-rds-nonstationary &1&Non-stationary  \\
    \cline{2-4}
    & NAB-iio-nonstationary &1&Non-stationary  \\
    \cline{2-4}
    & NAB-rogue-nonstationary &1&Non-stationary  \\
    \cline{1-4}
	\multicolumn{2}{|c|}{S5}&1&Non-stationary  \\
\cline{1-4}
	\multicolumn{2}{|c|}{KPI}&	1&Non-stationary	\\
\cline{1-4}
	\multicolumn{2}{|c|}{ECG}&	2&Non-stationary\\
\cline{1-4}
	\multicolumn{2}{|c|}{NYC}&	3&Non-stationary\\
\cline{1-4}
	\multicolumn{2}{|c|}{SMAP}&	25&Non-stationary\\
\cline{1-4}
	\multicolumn{2}{|c|}{Credit}	&	29&Non-stationary\\
\cline{1-4}
	\multicolumn{2}{|c|}{SWaT}&	51&Non-stationary\\
\cline{1-4}
	\multicolumn{2}{|c|}{MSL}	&	55&Non-stationary\\
\cline{1-4}
\end{tabular}
\vspace{-0.42cm}
\end{table}

\subsubsection{DL Methods}\label{sec:expDL}
We particularly focus on the Deep Learning (DL) methods that represent the recent state of the art in the following.

\begin{enumerate}
\item \textbf{CNN-AE.}
CNN-AE is a 1D CNN autoencoder.

\item \textbf{2DCNN-AE.}
2DCNN-AE~\cite{krizhevsky2012imagenet} also takes time series as images and uses a 2D CNN autoencoder to reconstruct image.

\item \textbf{LSTM-AE.}
LSTM-AE~\cite{MalhotraRAVAS16} is a combination of a LSTM and an autoencoder.

\item \textbf{HIFI.}
HIFI~\cite{deng2021hifi} is a Transformer-based anomaly detection model for multivariate time series with HIgh-order Feature Interactions
(HIFI).

\item \textbf{RN.}
Rand Net (RN)~\cite{chen2017outlier} adopts feedforward autoencoder ensembles for non-sequential data. 

\item \textbf{RAE.}
RAE~\cite{KieuYGJ19} is an ensemble-based model that relies on the availability of multiple RNN autoencoders with  an independent
framework training multiple autoencoders independently in a multi-task learning fashion.

\item \textbf{Omni.}
Omni~\cite{su2019robust} is a VAE-based model that adopts the stochastic variable connection and planar normalizing flow to reconstruct input data.

\item \textbf{BeatGAN.}
BeatGAN~\cite{ZhouLHCY19} employs the GAN architecture to reconstruct data.

\item \textbf{LSTM-NDT.}
LSTM-NDT~\cite{hundman2018detecting} is a prediction-based model, which uses LSTM to predict data based on the historical observations.
\end{enumerate}

For a fair comparison, the main hyperparameters for the DL methods have the same settings, shown in Table~\ref{tab:hyperparameter1}.

Following grid search, we vary both hidden sizes and sliding window sizes among $16$, $32$, $64$, $128$, and $256$, whose default values are $64$ and $128$, respectively, indicating that there are $5*5$ hyperparameter combinations. 
2DCNN-AE and RN run out of memory in certain combinations, such as when the hidden size and the slide window size are both set to $256$. We therefore ignore their results for these settings.

\noindent \textbf{Training process.} We train each model based on the validation set using early stopping. For each hyperparameter combination in a model, we proceed as follows.

1) train a model on the training set and record the validation errors (i.e., the reconstruction or prediction errors) of the trained model on the validation set in each epoch;

2) when the difference between the current validation error and current minimal validation error is below the validation loss difference (i.e.,  $3e-4$) for successive $10$ times (i.e., $\mathit{Patience} = 10$), or the maximum number of epochs is reached (i.e., $200$), the training stops;

3) we use the parameters with the least validation error as the final parameters.

\begin{table}
\scriptsize
  \caption{Hyperparameter Settings of DL Methods}
  \label{tab:parameter}
  \vskip -9pt
  \renewcommand{\arraystretch}{1}
  \begin{tabular}{|l|l|}\hline
    \tabincell{c}{\textbf{Hyperparameters}} & \tabincell{c}{\textbf{Settings}} \\\hline
	Learning rate& $1e-3$\\\hline
    Maximum epochs & $200$ \\\hline
    Milestone epochs\tablefootnote{Epochs of learning rate decay}& $50$ \\\hline
    Decay rate of learning rate & $0.9$5 \\\hline
    Batch size& $64$ \\\hline
    Weight decay\tablefootnote{Weight decay is to prevent overfitting} & $1e-8$ \\\hline
    Gradient clip\tablefootnote{Gradient clip is to prevent gradient exploding} & $10$ \\\hline
    Validation loss difference  & $3e-4$ \\\hline
    Patience& $10$ \\\hline
\end{tabular}
\vspace{-0.4cm}
\label{tab:hyperparameter1}
\end{table}

\subsection{Metrics}\label{sec:metrics}
\noindent\textbf{\emph{Precision}, \emph{Recall}, \emph{$\mathit{F1}$ score}, \emph{ROC-AUC}, and \emph{PR-AUC}.}
When a threshold is given, we use
\emph{precision}, \emph{recall}, and \emph{$\mathit{F1}$ score} to evaluate the detection accuracy.
As does the AnomalyBech~\cite{jacob2020anomalybench}, we use three unsupervised threshold selection methods, i.e., SD, MAD and IQR, which are covered in Sec.~\ref{sec:evalumetrics}.
We apply these thresholds to the testing set and report the one with the best \emph{$\mathit{F1}$ score}, based on which
\emph{precision} and \emph{recall} are also reported.
When no thresholds are given, \emph{ROC-AUC} and \emph{PR-AUC} are used to evaluate detection accuracy.
The anomaly score of each method is normalized to the unit interval using \emph{min}-\emph{max} scaling.

\begin{figure*}
\centering
\subfigure[S5: \emph{one-dimensional}] {\includegraphics[width=0.21\textwidth]{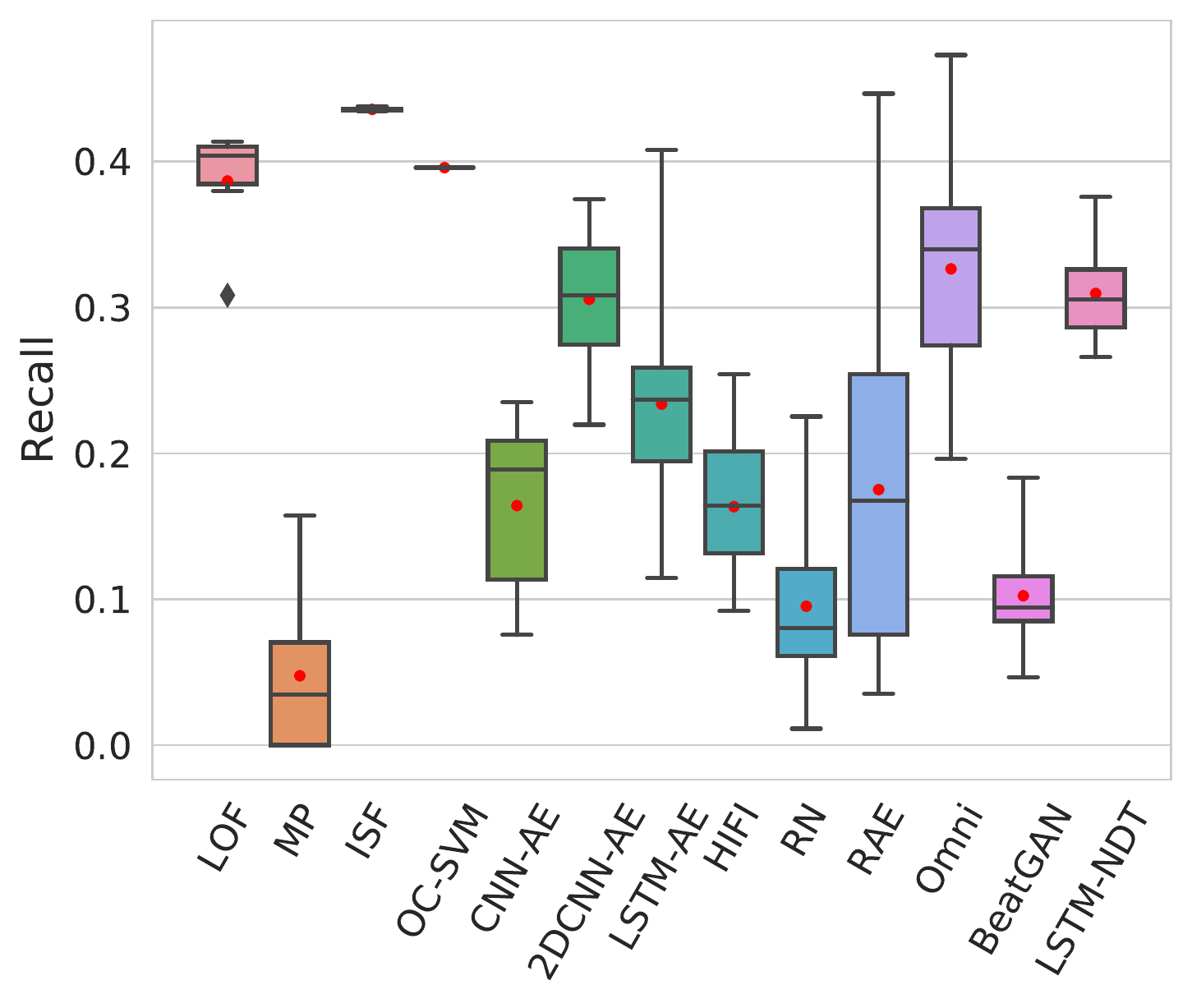}\label{fig:S5-rec}}
\subfigure[KPI: \emph{one-dimensional}] {\includegraphics[width=0.21\textwidth]{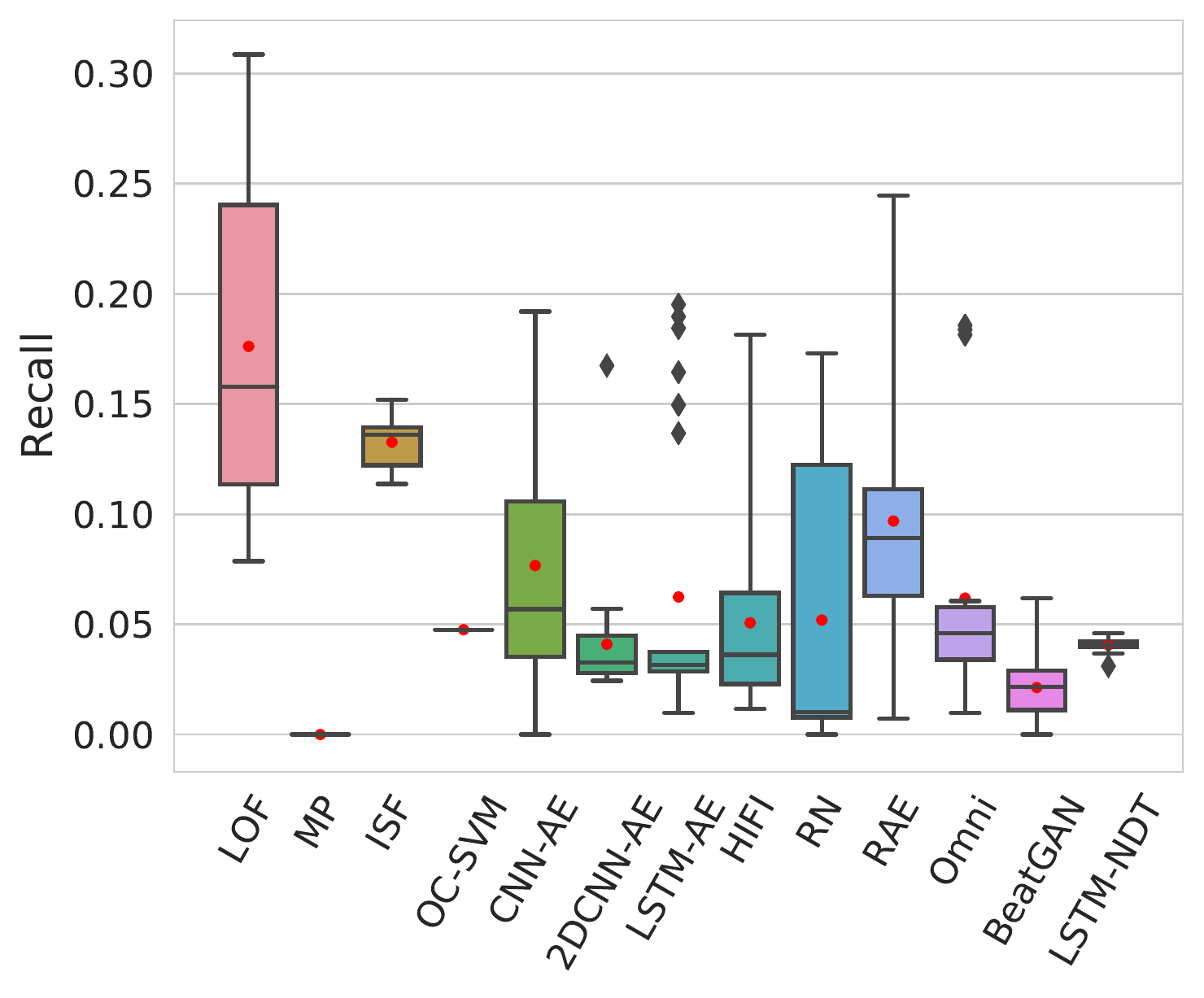}\label{fig:KPI-rec}}
\subfigure[ECG: \emph{2-dimensional}] {\includegraphics[width=0.21\textwidth]{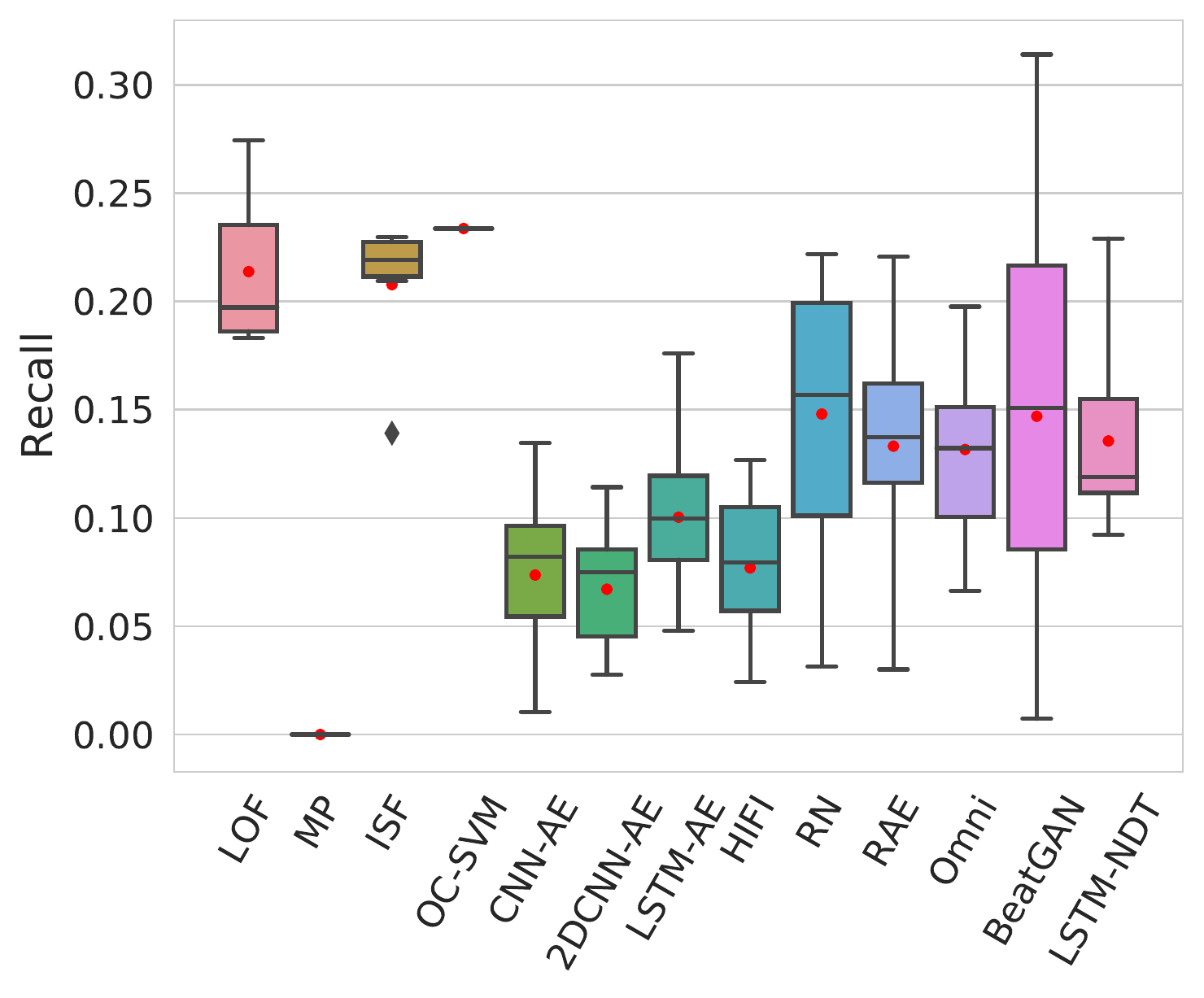}\label{fig:ECG-rec}}
\subfigure[NYC: \emph{3-dimensional}] {\includegraphics[width=0.21\textwidth]{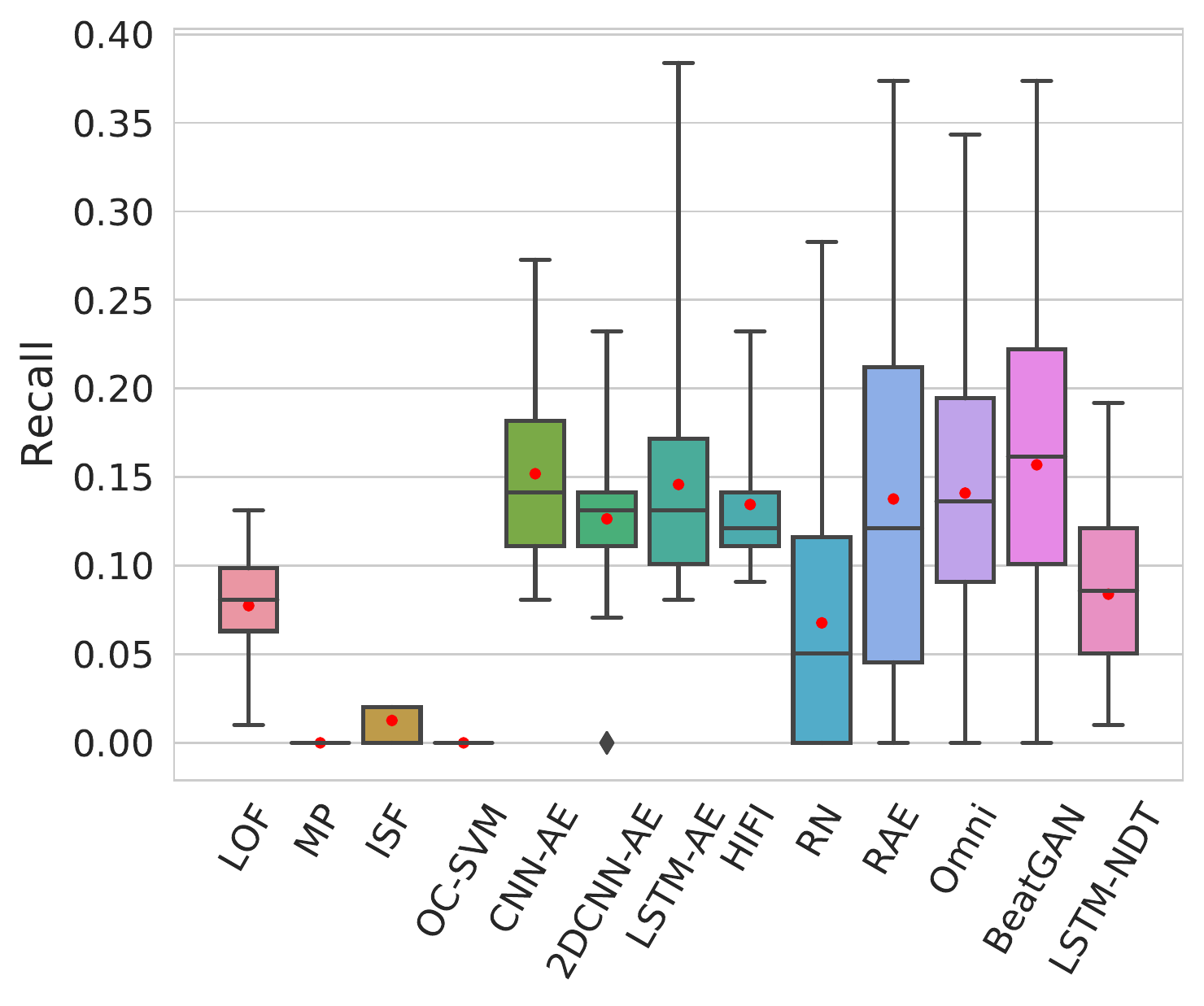}\label{fig:NYC-rec}}
\subfigure[SMAP: \emph{25-dimensional}] {\includegraphics[width=0.21\textwidth]{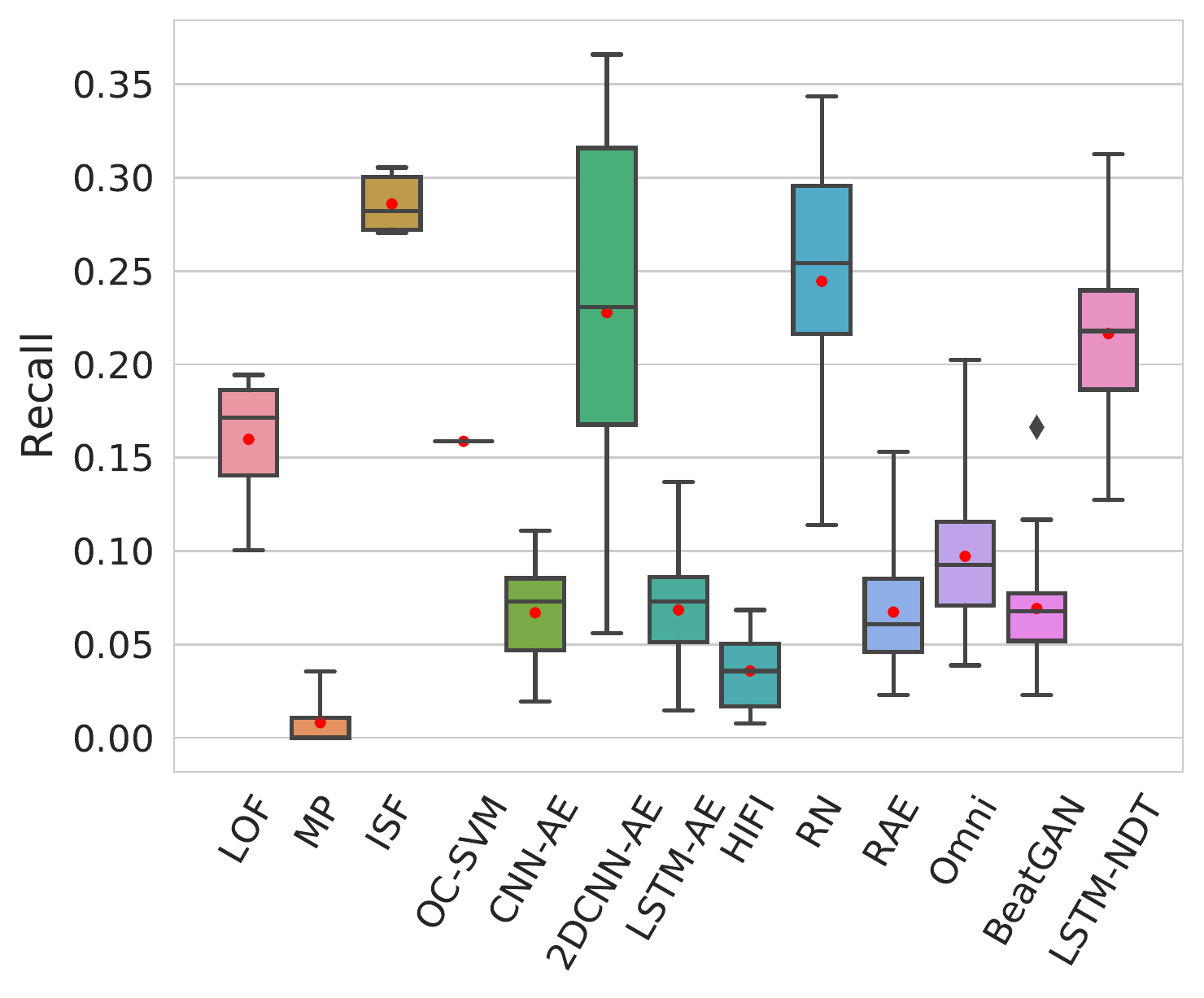}\label{fig:SMAP-rec}}
\subfigure[Credit: \emph{29-dimensional}] {\includegraphics[width=0.21\textwidth]{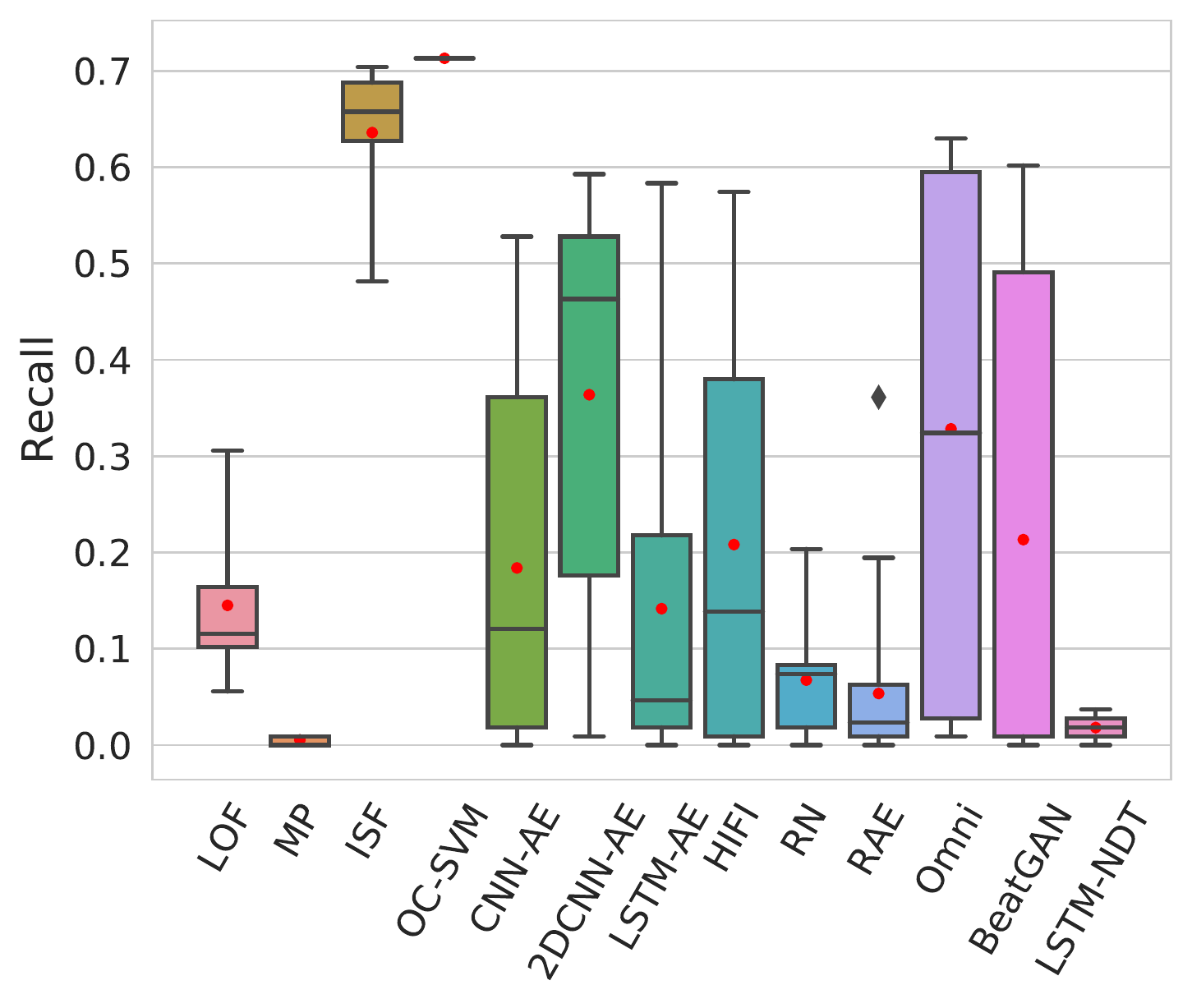}\label{fig:Credit-rec}}
\subfigure[SWaT: \emph{51-dimensional}] {\includegraphics[width=0.21\textwidth]{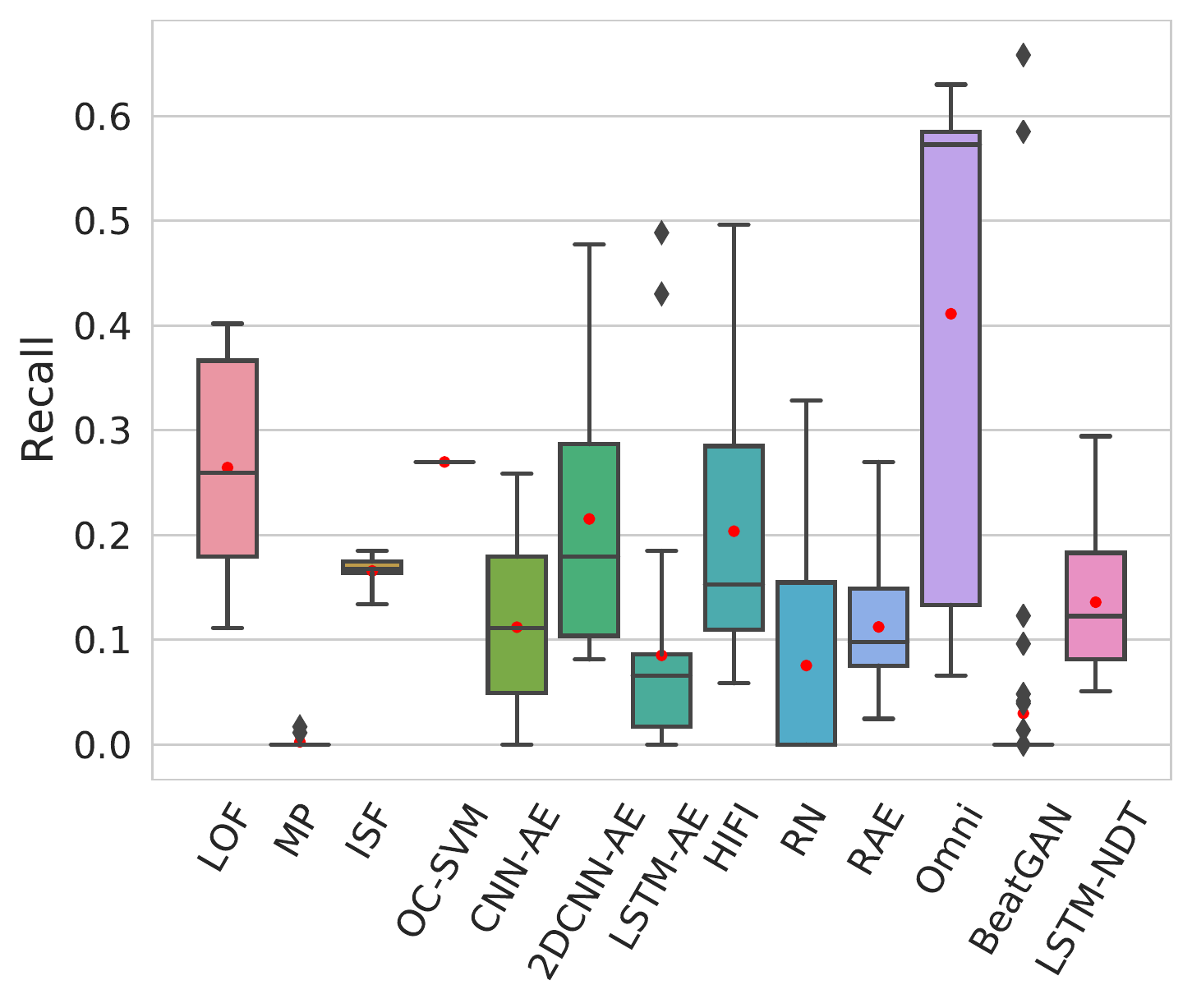}\label{fig:SWaT-rec}}
\subfigure[MSL: \emph{55-dimensional}] {\includegraphics[width=0.21\textwidth]{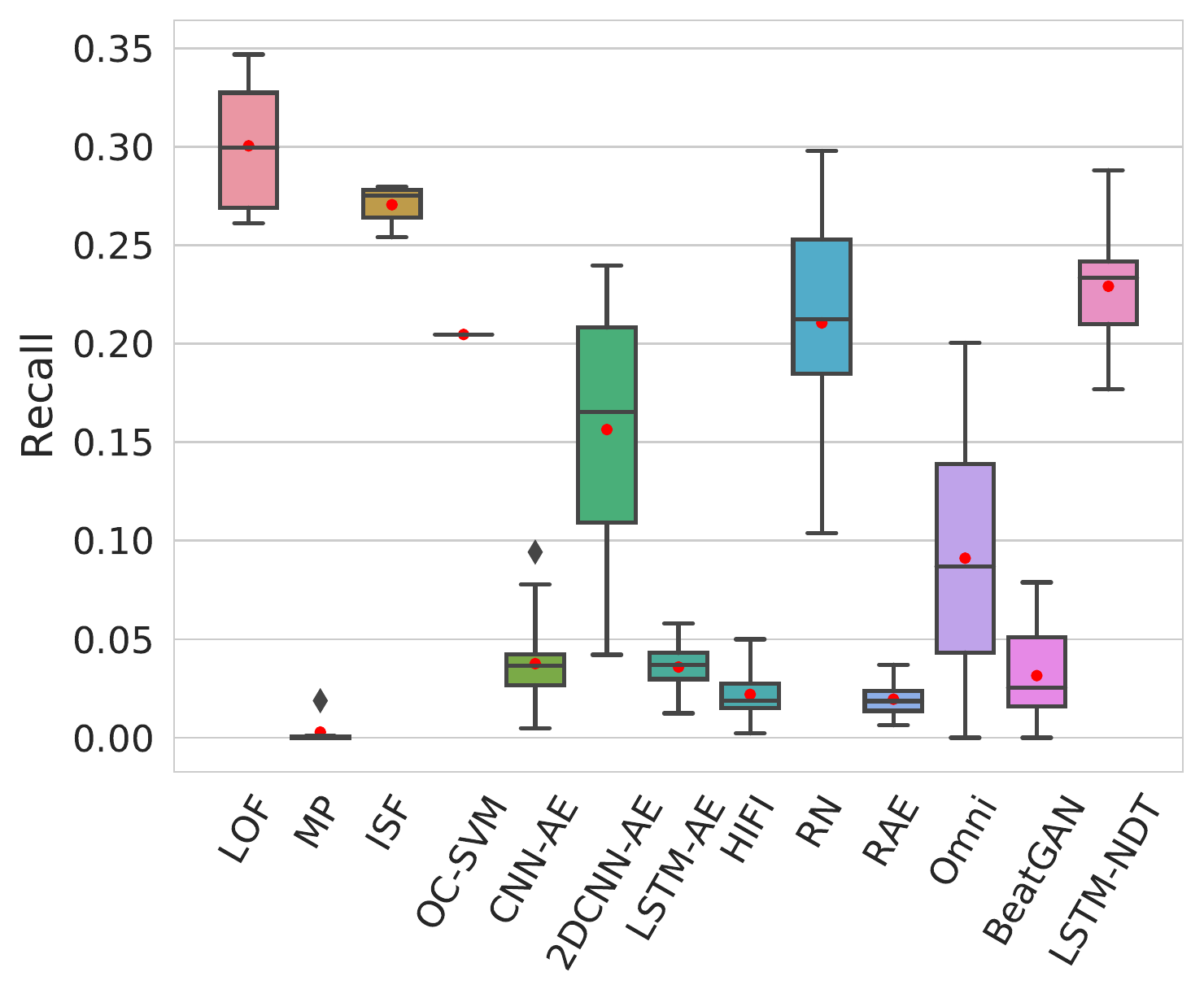}\label{fig:MSL-rec}}
\vskip -9pt
\caption{Recall (Univariate vs. Multivariate)}
\label{fig:rec}
\end{figure*}

\begin{figure*}
\centering
\subfigure[S5: \emph{one-dimensional}] {\includegraphics[width=0.21\textwidth]{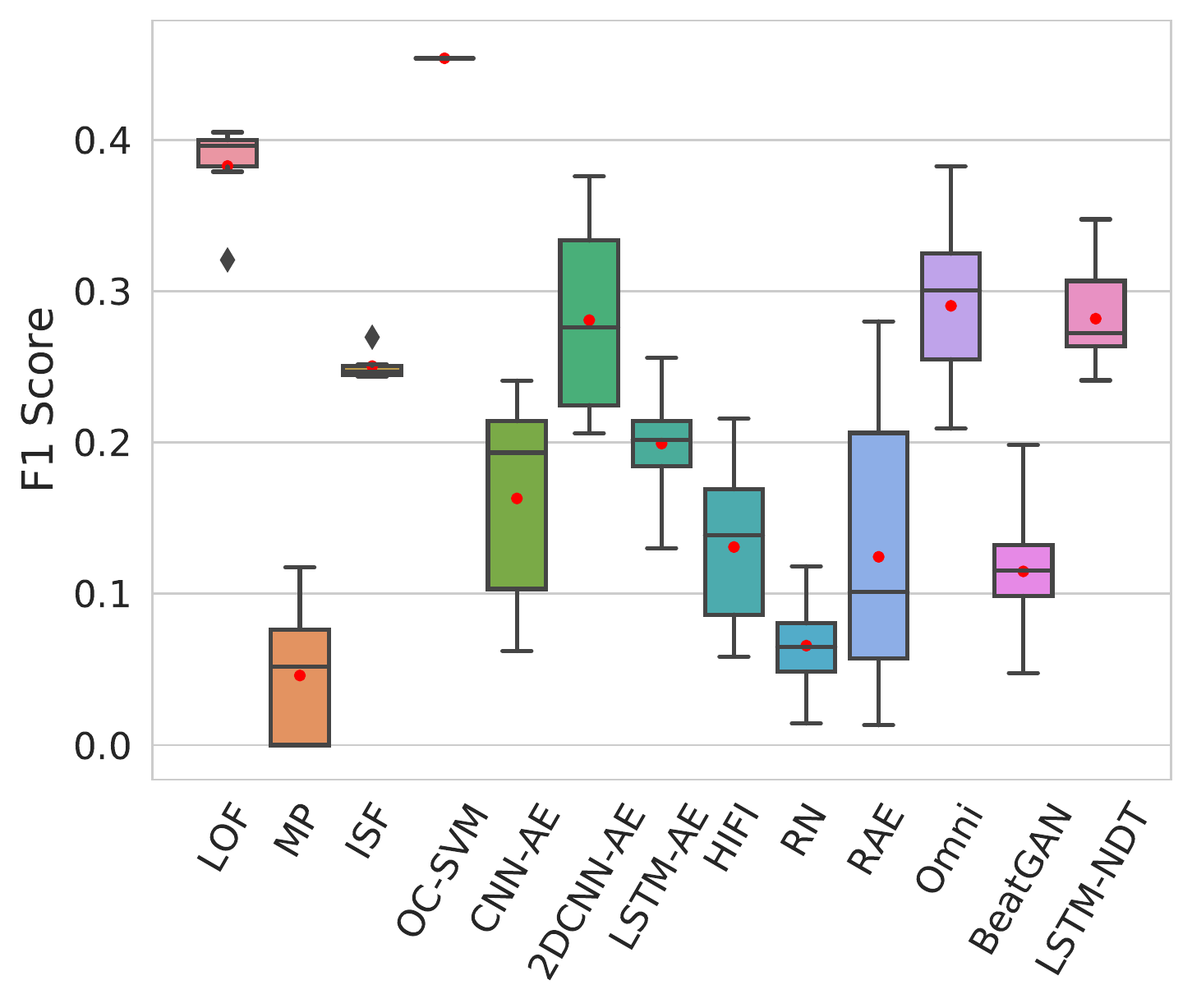}\label{fig:S5-f1}}
\subfigure[KPI: \emph{one-dimensional}] {\includegraphics[width=0.21\textwidth]{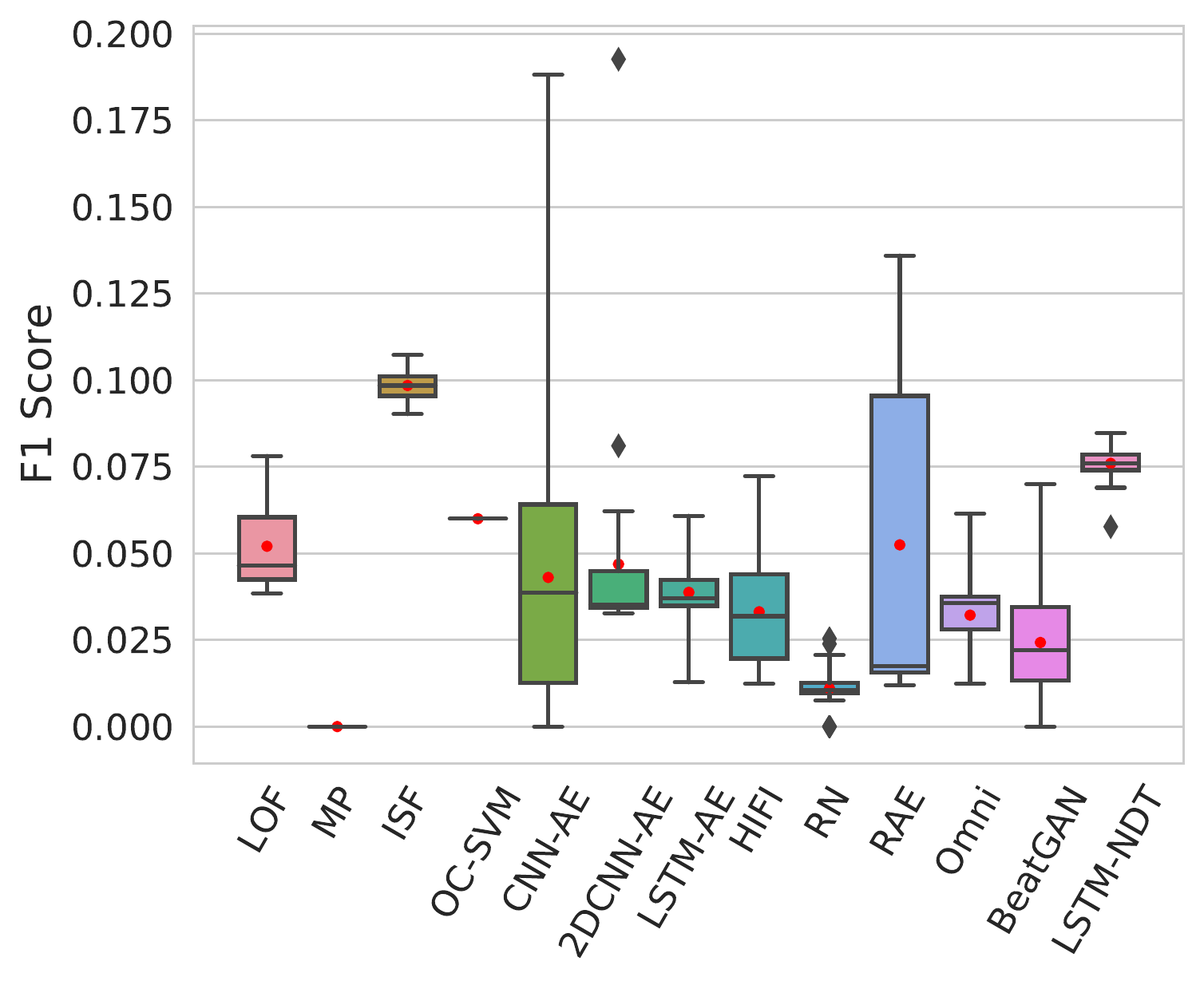}\label{fig:KPI-f1}}
\subfigure[ECG: \emph{2-dimensional}] {\includegraphics[width=0.21\textwidth]{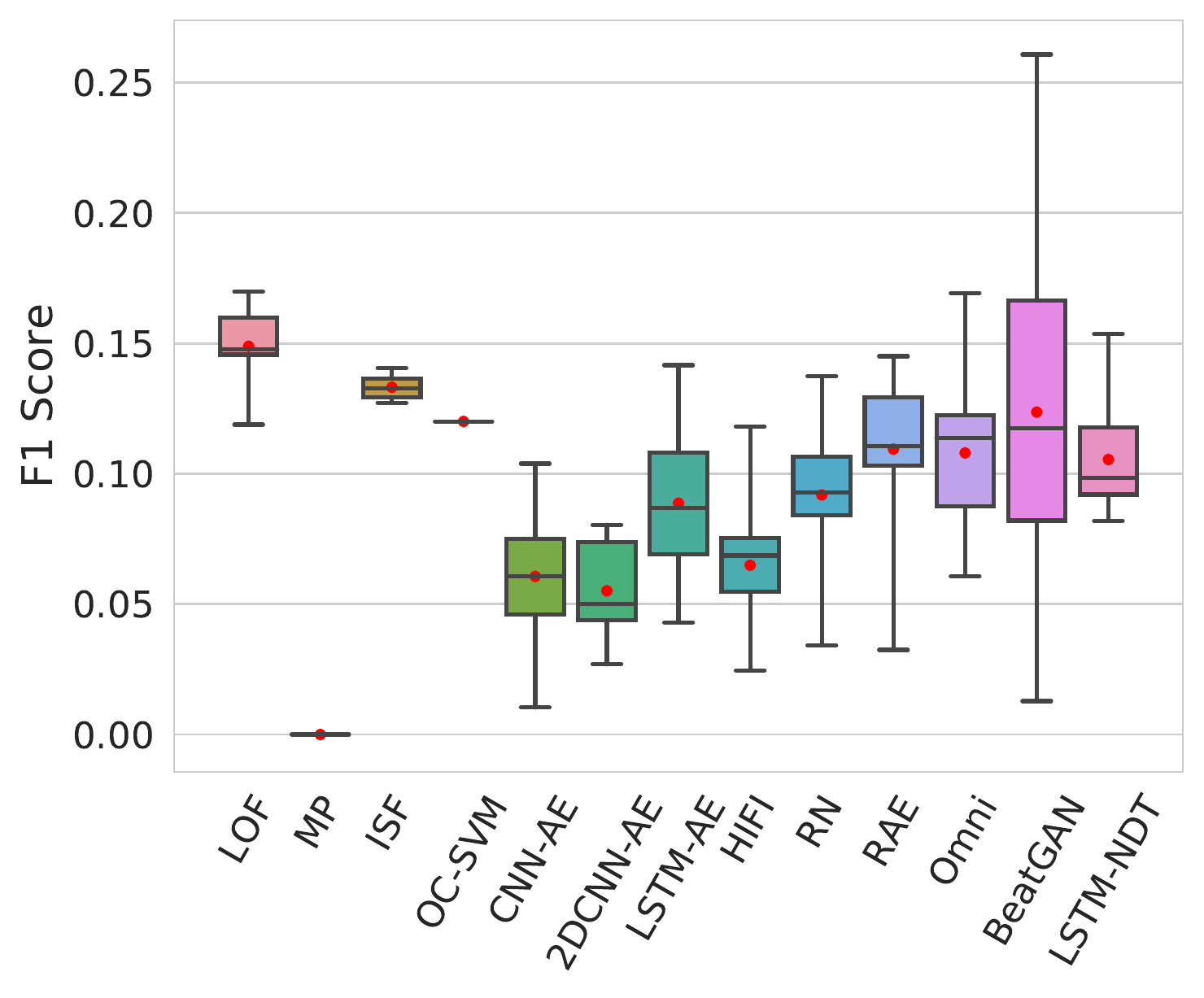}\label{fig:ECG-f1}}
\subfigure[NYC: \emph{2-dimensional}] {\includegraphics[width=0.21\textwidth]{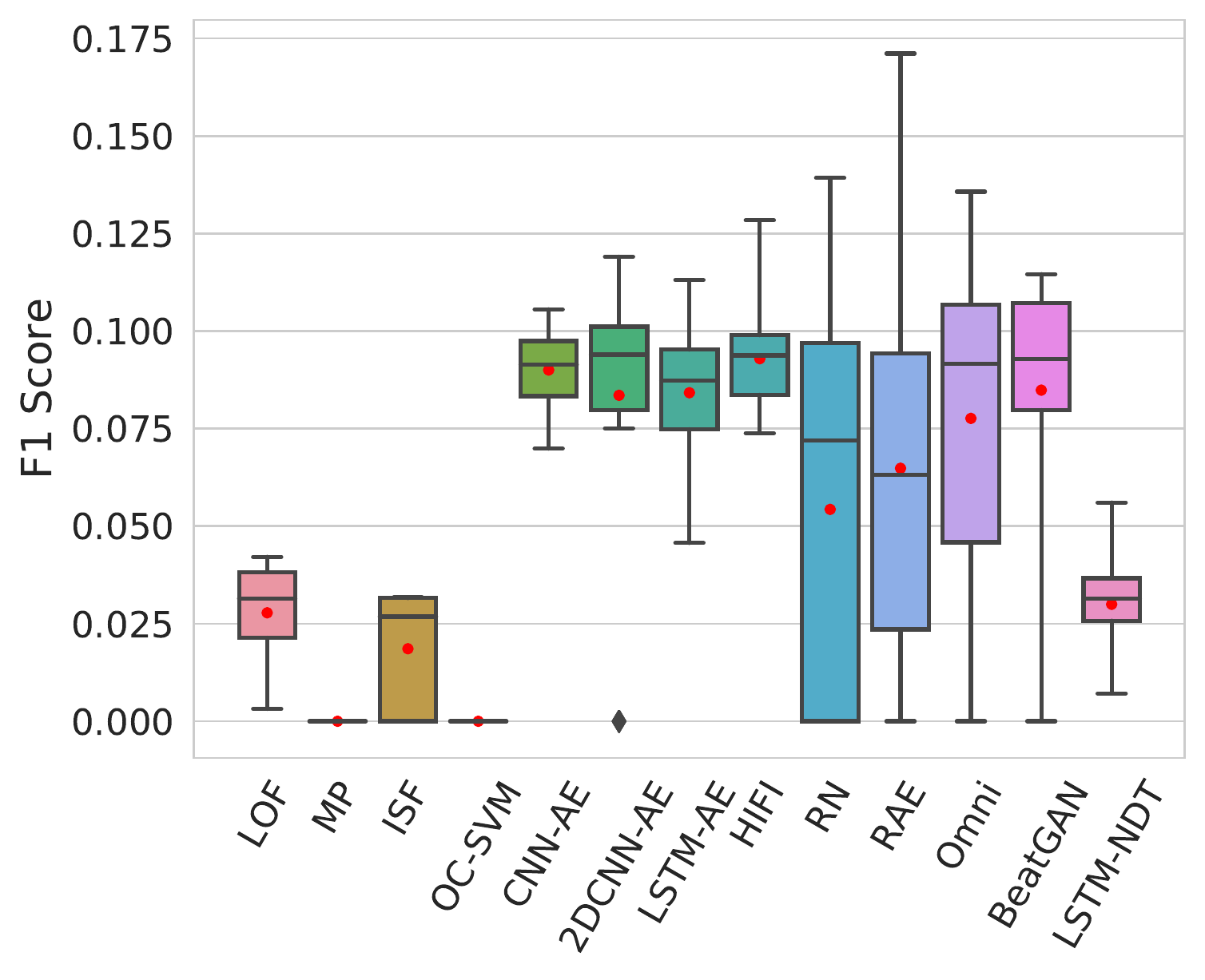}\label{fig:NYC-f1}}
\subfigure[SMAP: \emph{25-dimensional}] {\includegraphics[width=0.21\textwidth]{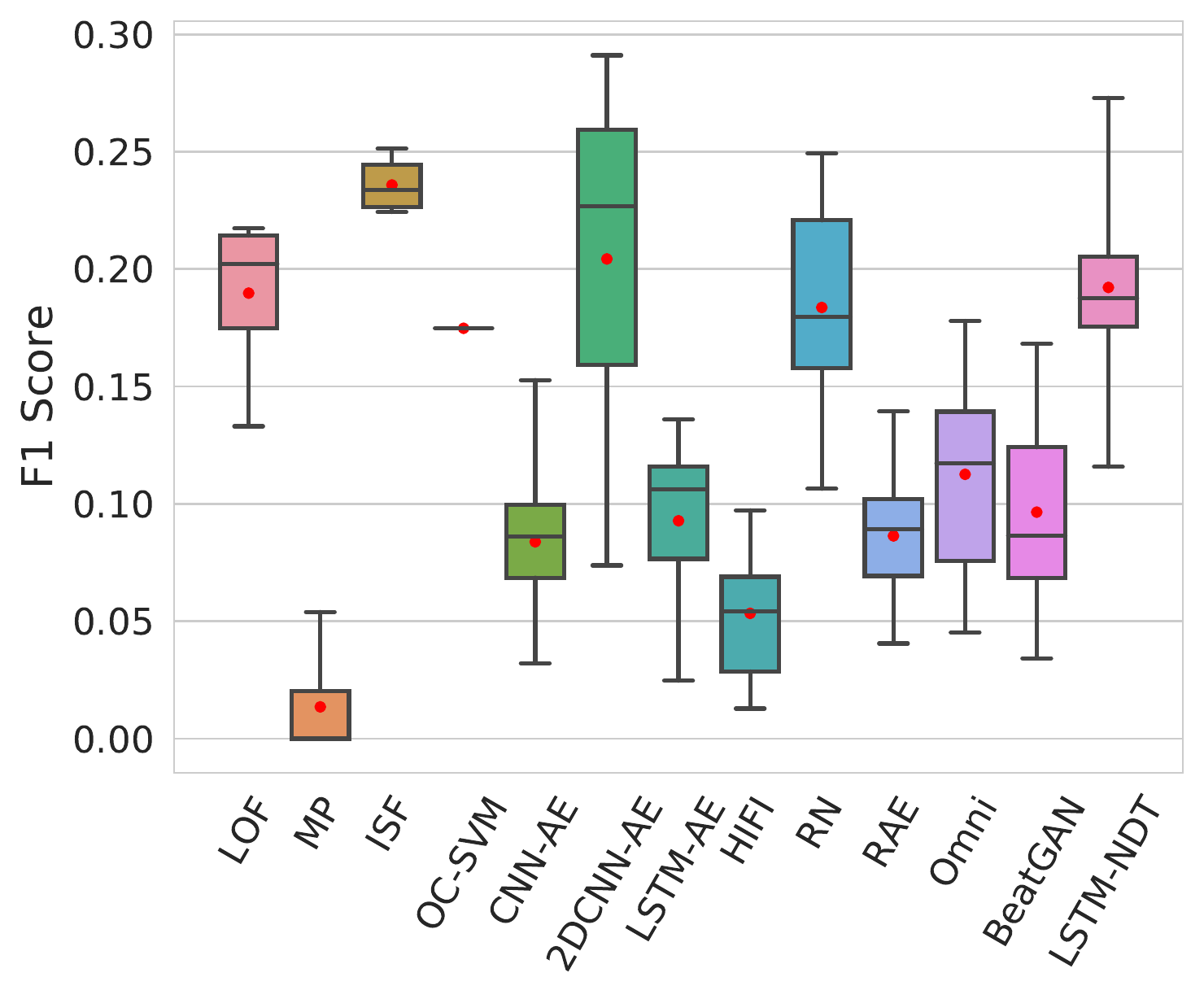}\label{fig:SMAP-f1}}
\subfigure[Credit: \emph{29-dimensional}] {\includegraphics[width=0.21\textwidth]{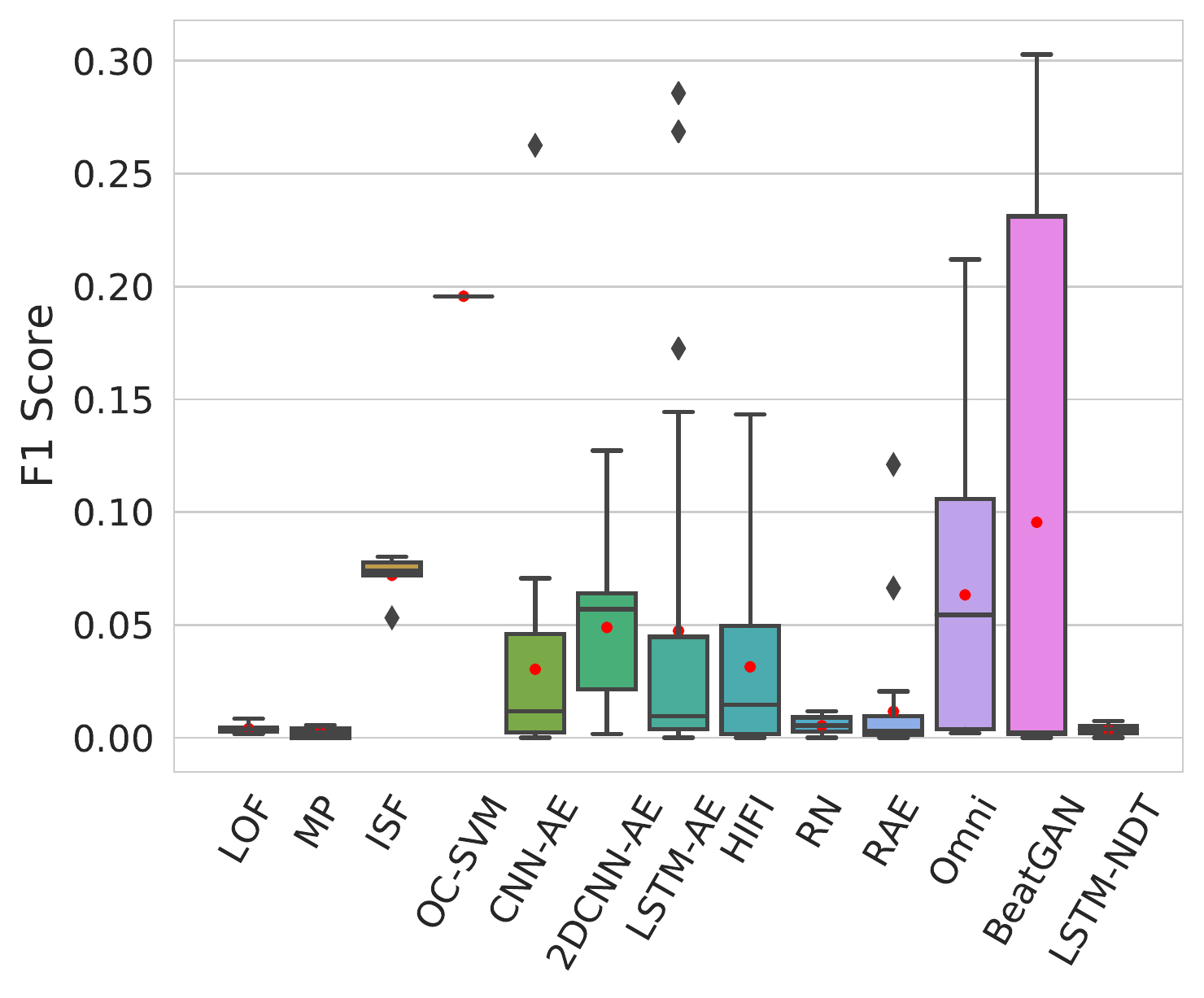}\label{fig:Credit-f1}}
\subfigure[SWaT: \emph{51-dimensional}] {\includegraphics[width=0.21\textwidth]{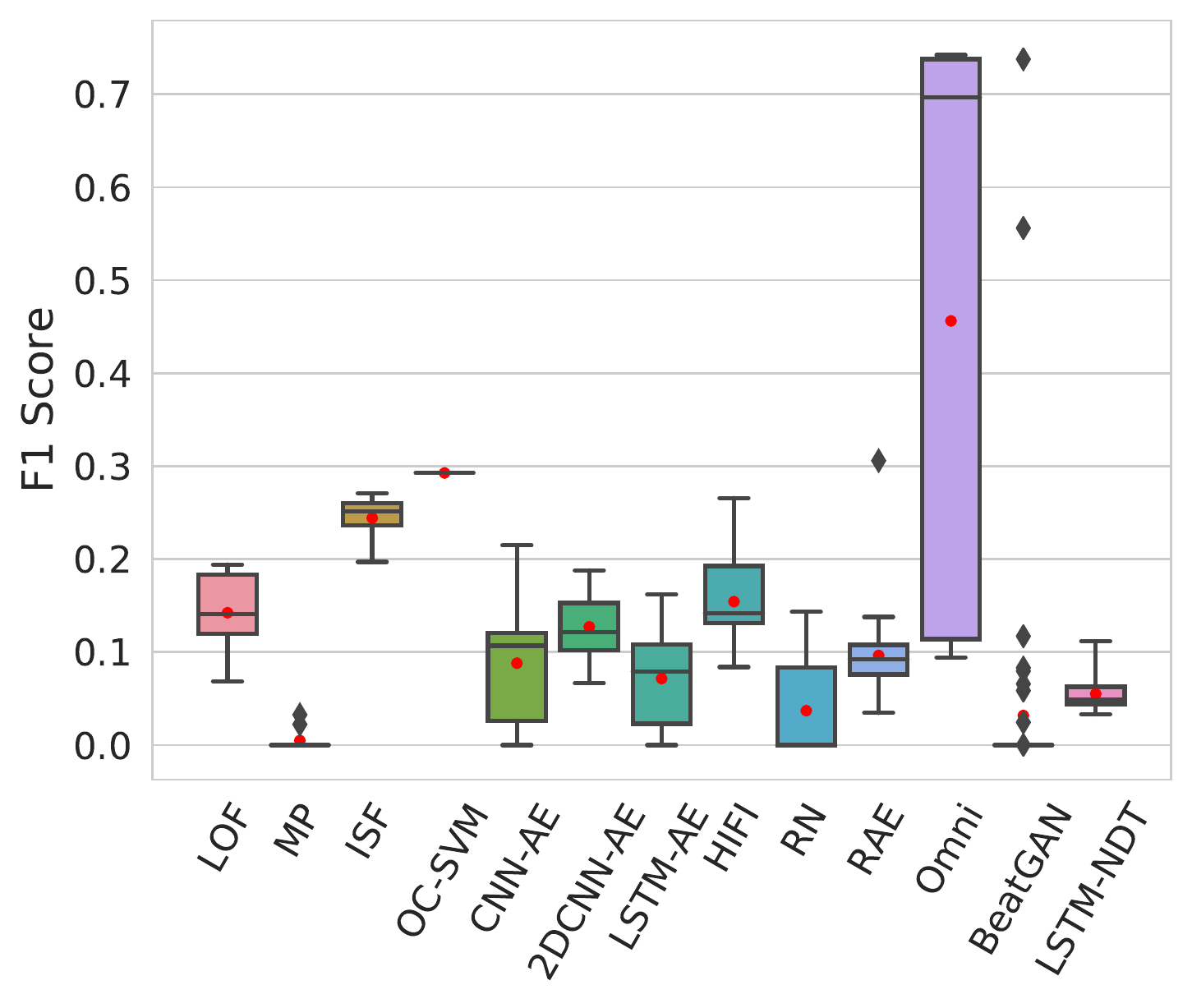}\label{fig:SWaT-f1}}
\subfigure[MSL: \emph{55-dimensional}] {\includegraphics[width=0.21\textwidth]{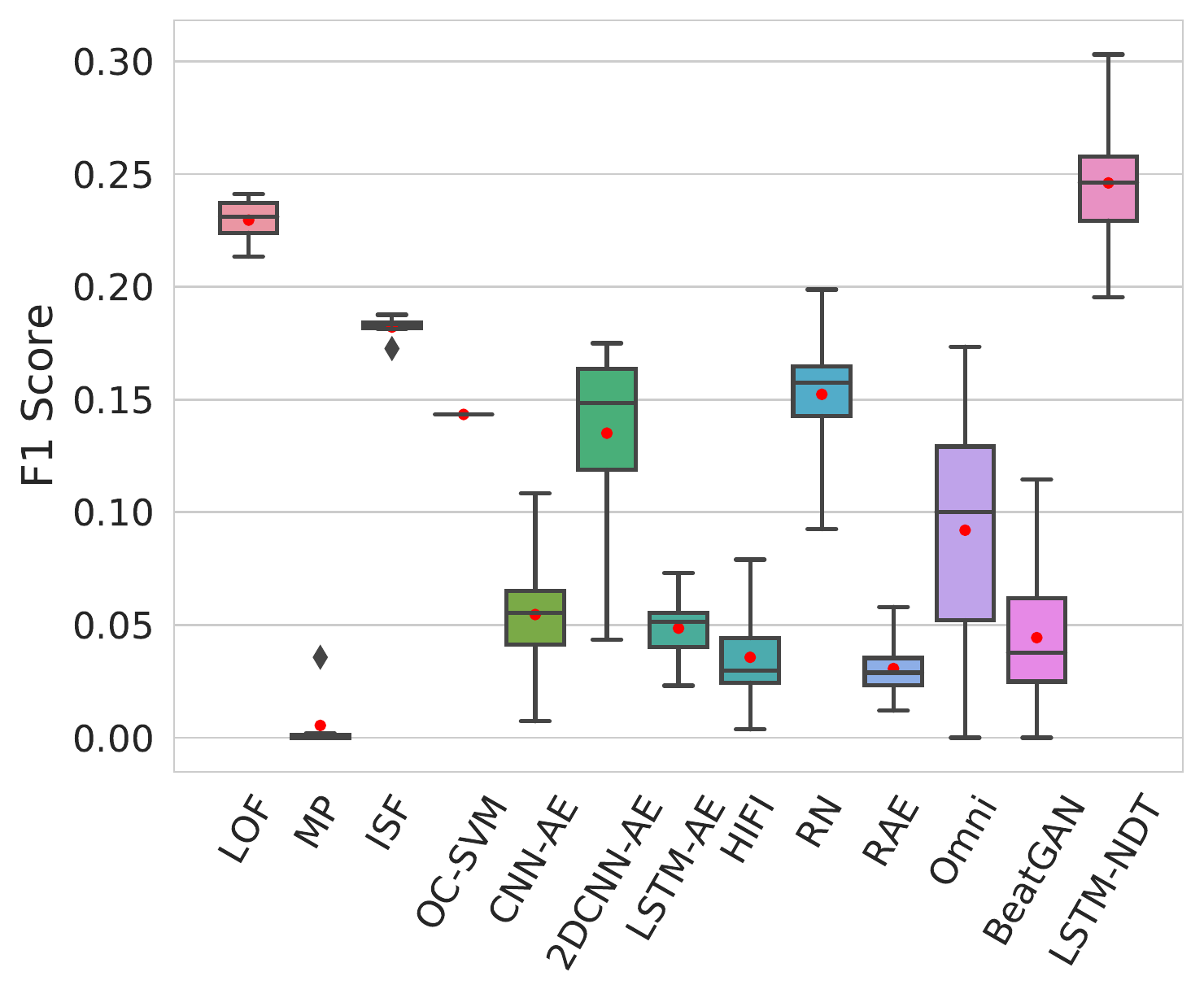}\label{fig:MSL-f1}}
\vskip -9pt
\caption{F1 Score (Univariate vs. Multivariate)}
\label{fig:f1}
\end{figure*}

\begin{figure*}
\centering
\subfigure[S5: \emph{one-dimensional}] {\includegraphics[width=0.21\textwidth]{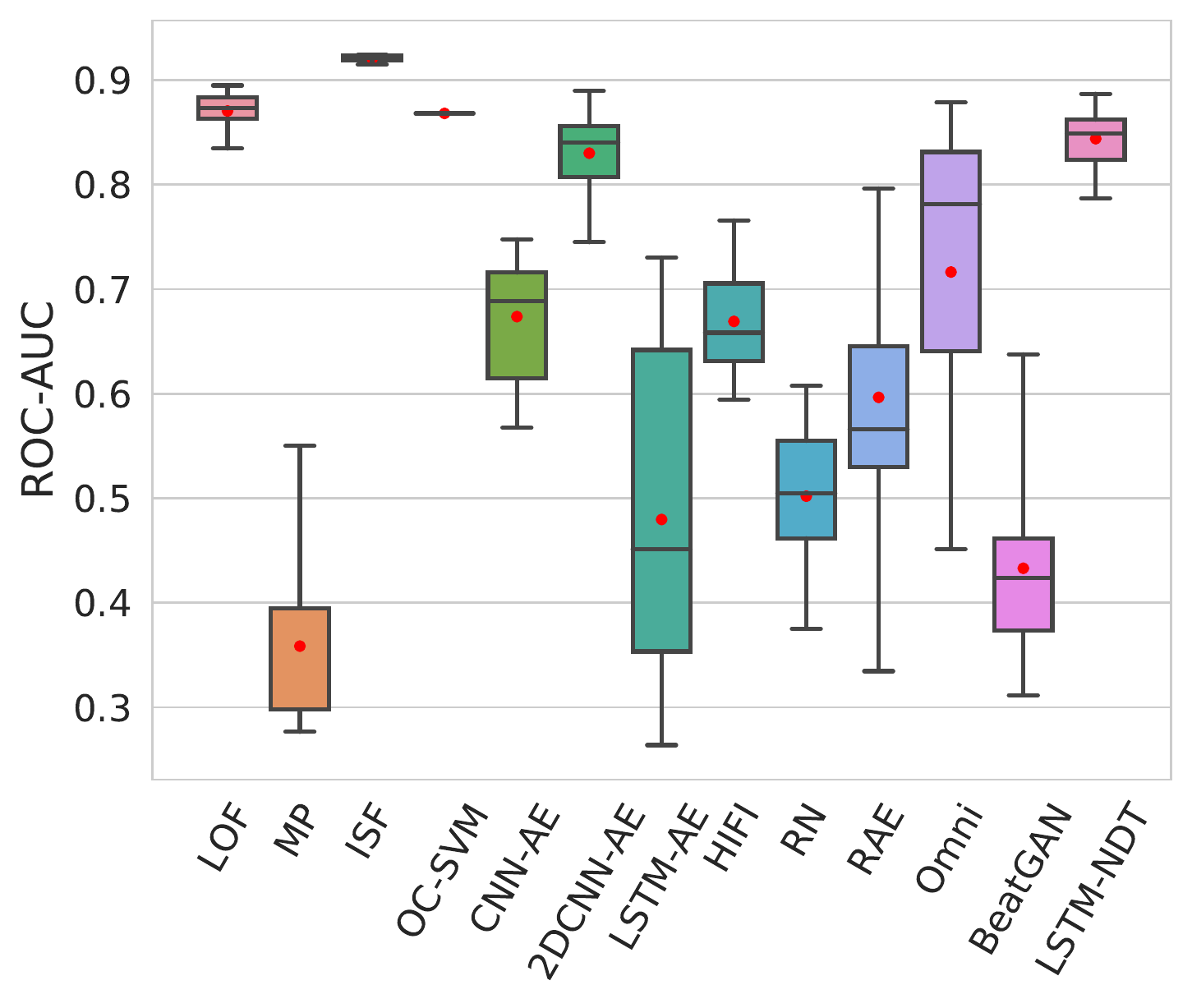}\label{fig:S5-ROC_AUC}}
\subfigure[KPI: \emph{one-dimensional}] {\includegraphics[width=0.21\textwidth]{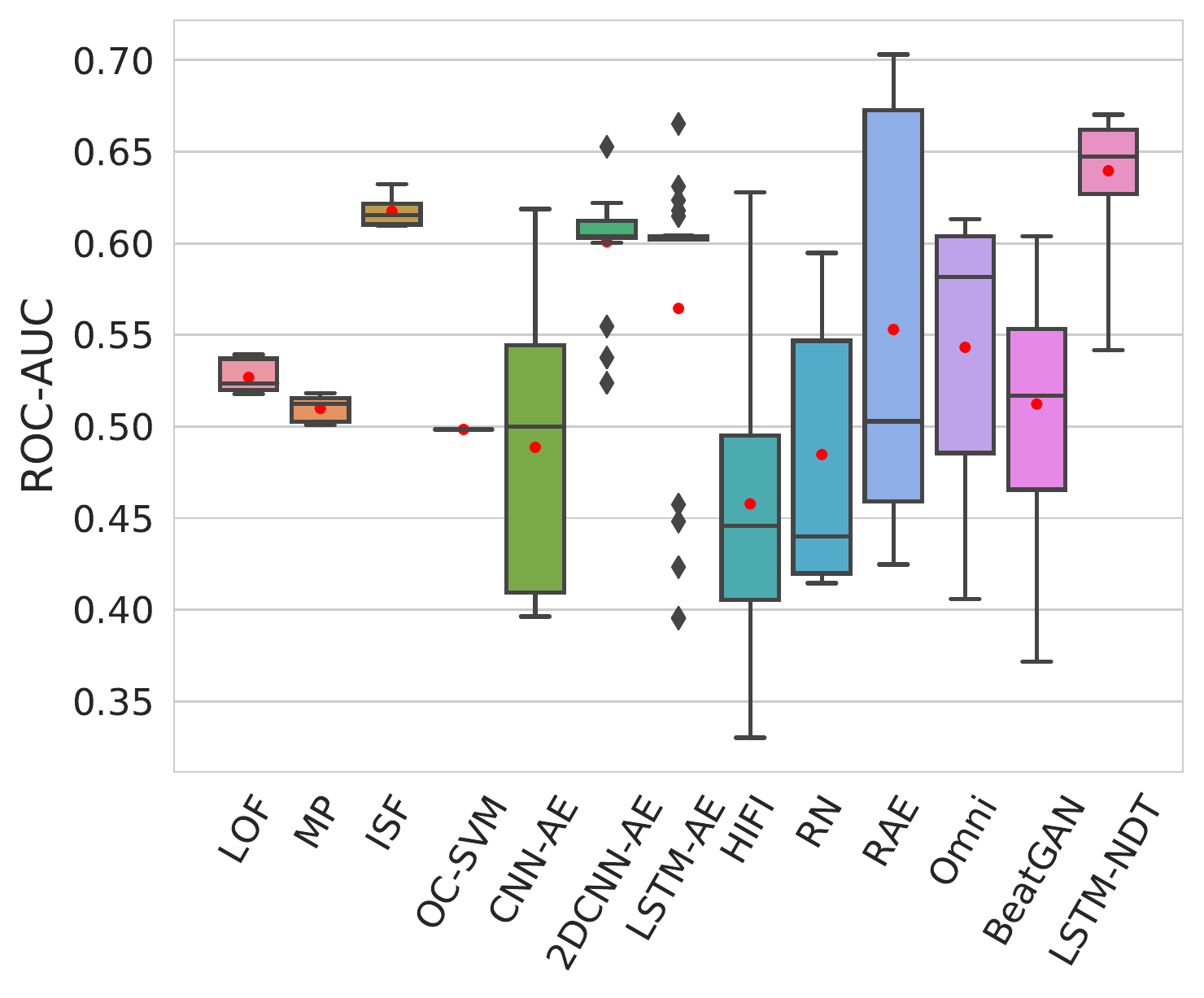}\label{fig:KPI-ROC_AUC}}
\subfigure[ECG: \emph{2-dimensional}] {\includegraphics[width=0.21\textwidth]{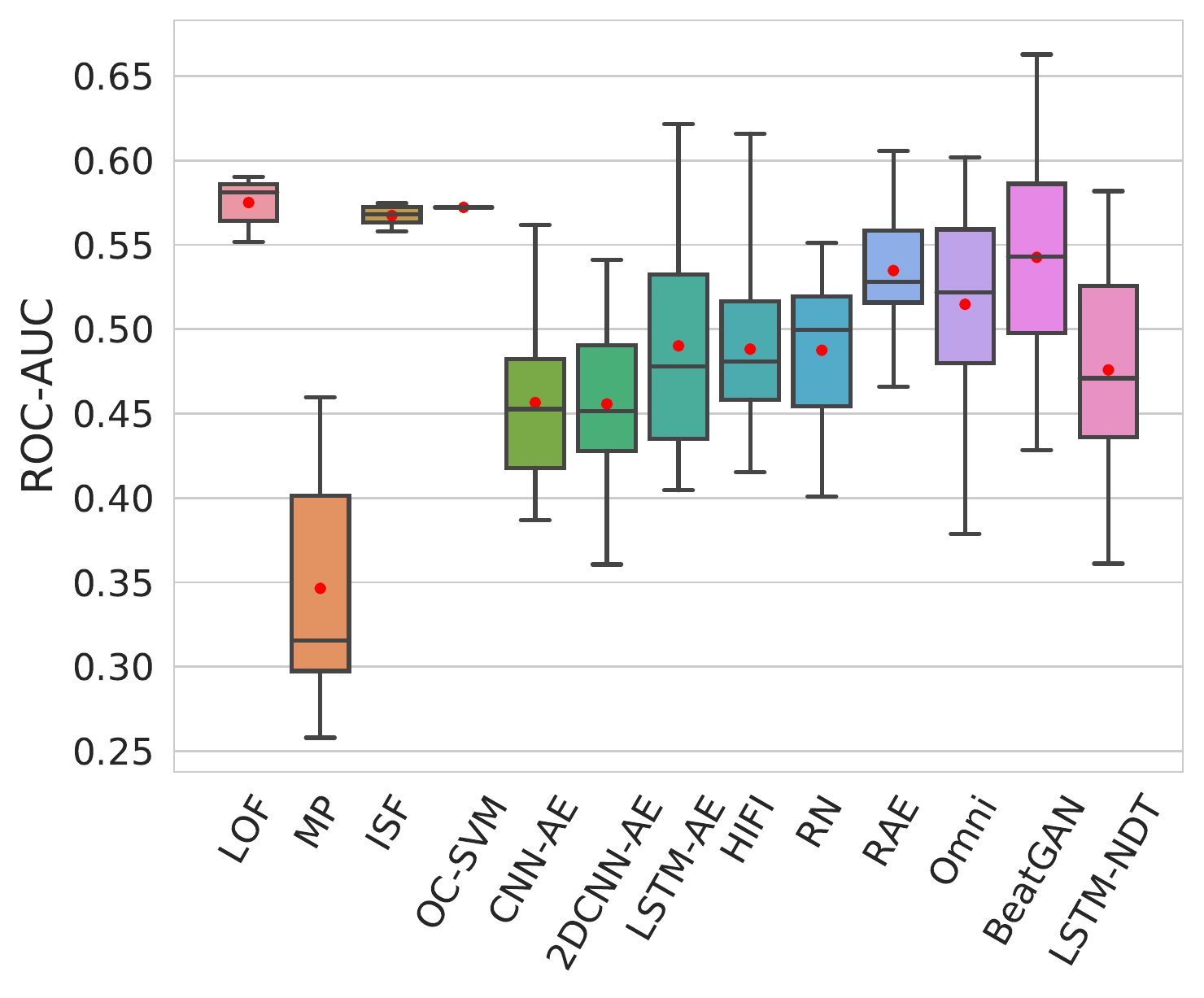}\label{fig:ECG-ROC_AUC}}
\subfigure[NYC: \emph{3-dimensional}] {\includegraphics[width=0.21\textwidth]{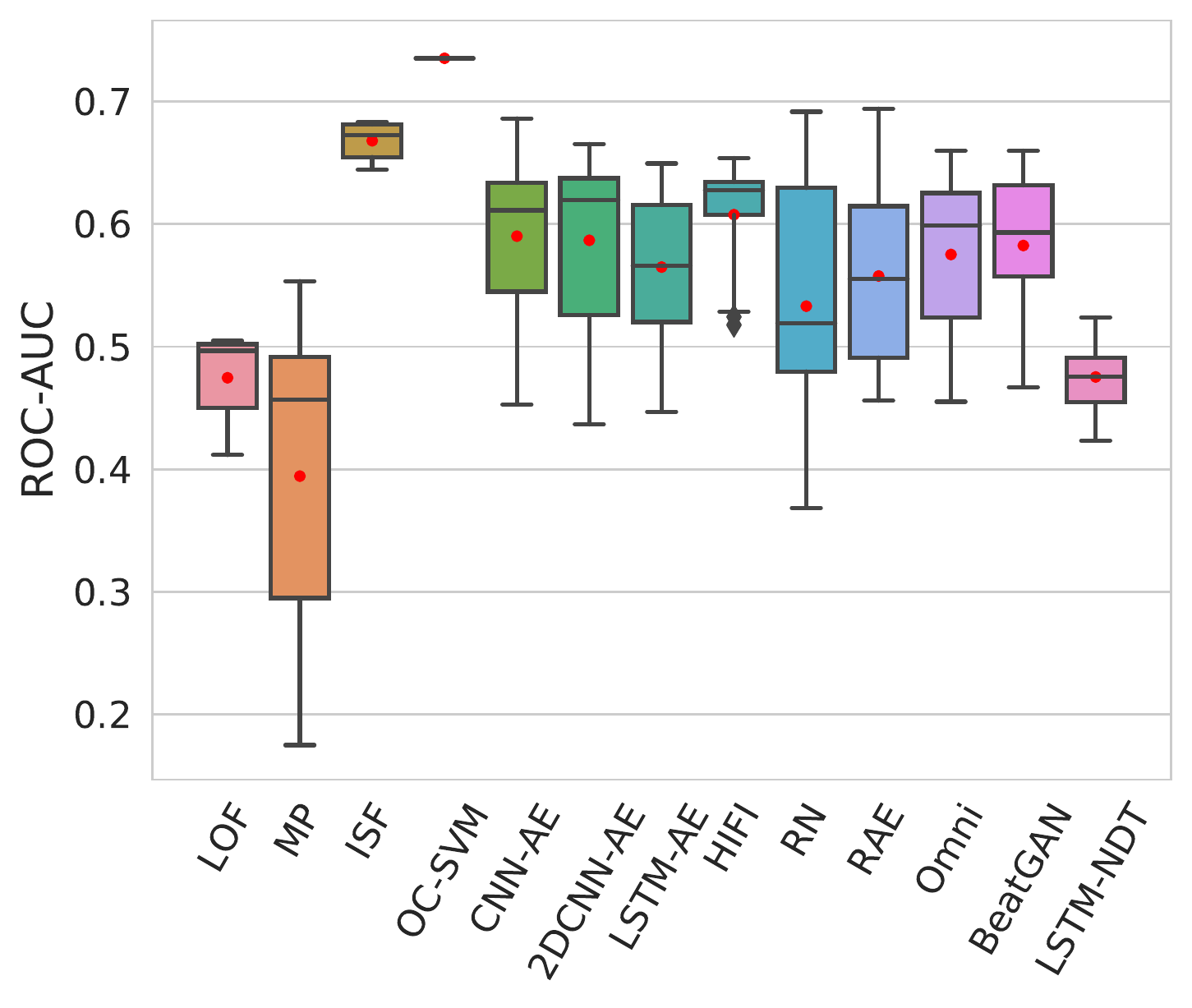}\label{fig:NYC-ROC_AUC}}
\subfigure[SMAP: \emph{25-dimensional}] {\includegraphics[width=0.21\textwidth]{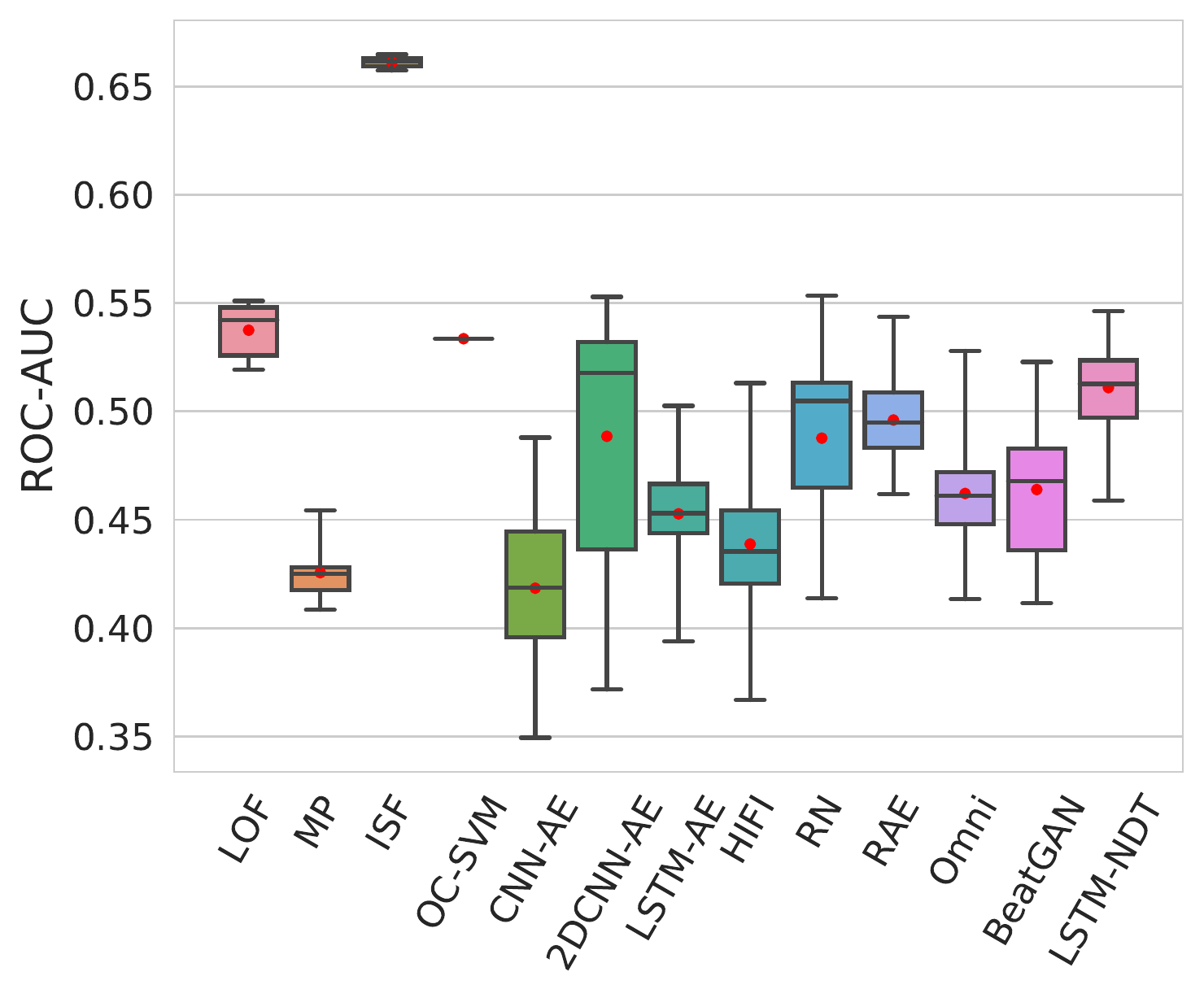}\label{fig:SMAP-ROC_AUC}}
\subfigure[Credit: \emph{29-dimensional}] {\includegraphics[width=0.21\textwidth]{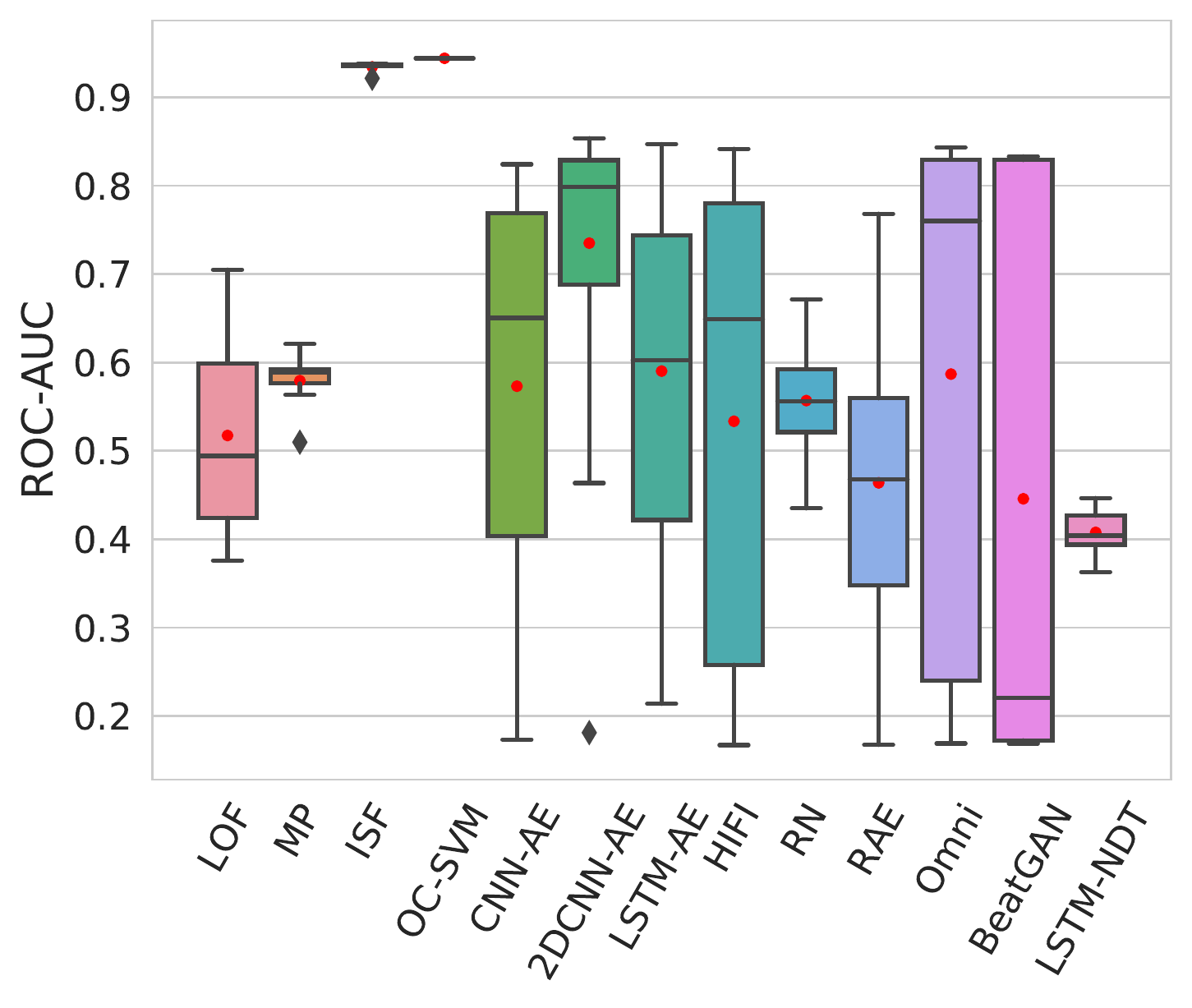}\label{fig:Credit-ROC_AUC}}
\subfigure[SWaT: \emph{51-dimensional}] {\includegraphics[width=0.21\textwidth]{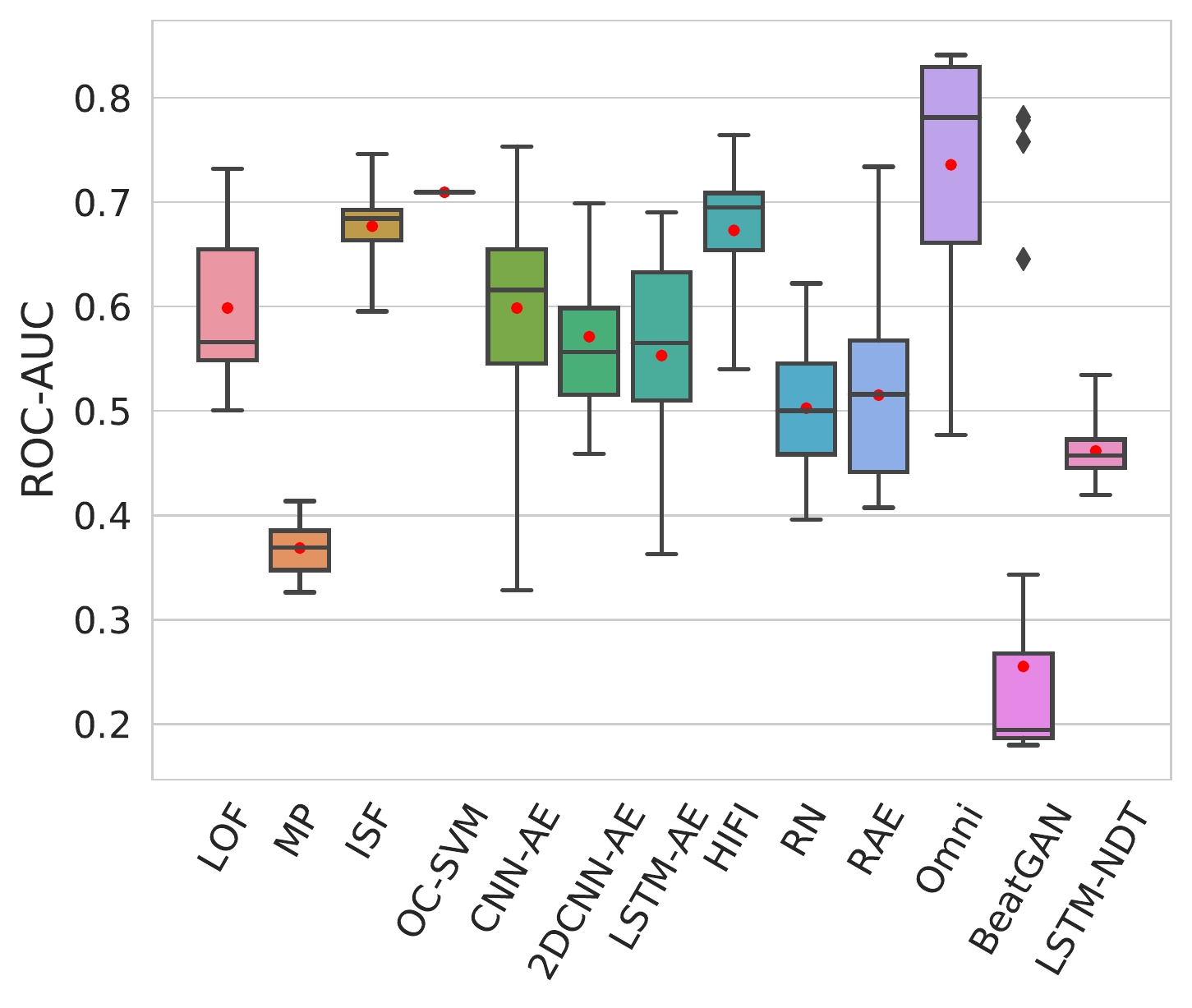}\label{fig:SWaT-ROC_AUC}}
\subfigure[MSL: \emph{55-dimensional}] {\includegraphics[width=0.21\textwidth]{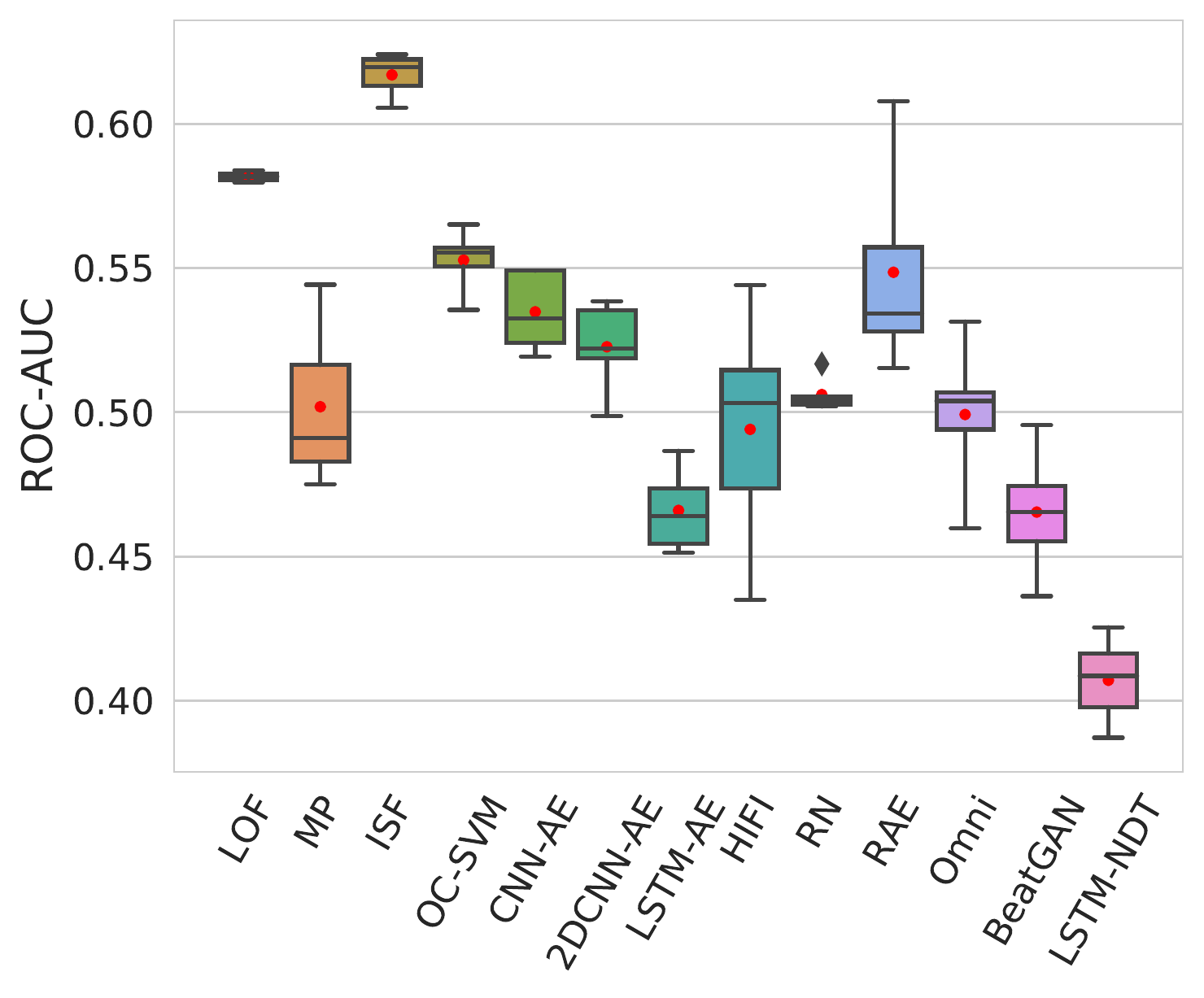}\label{fig:MSL-ROC_AUC}}
\vskip -9pt
\caption{ROC-AUC (Univariate vs. Multivariate)}
\label{fig:roc}
\end{figure*}

\begin{figure*}
\centering
\subfigure[S5: \emph{one-dimensional}] {\includegraphics[width=0.21\textwidth]{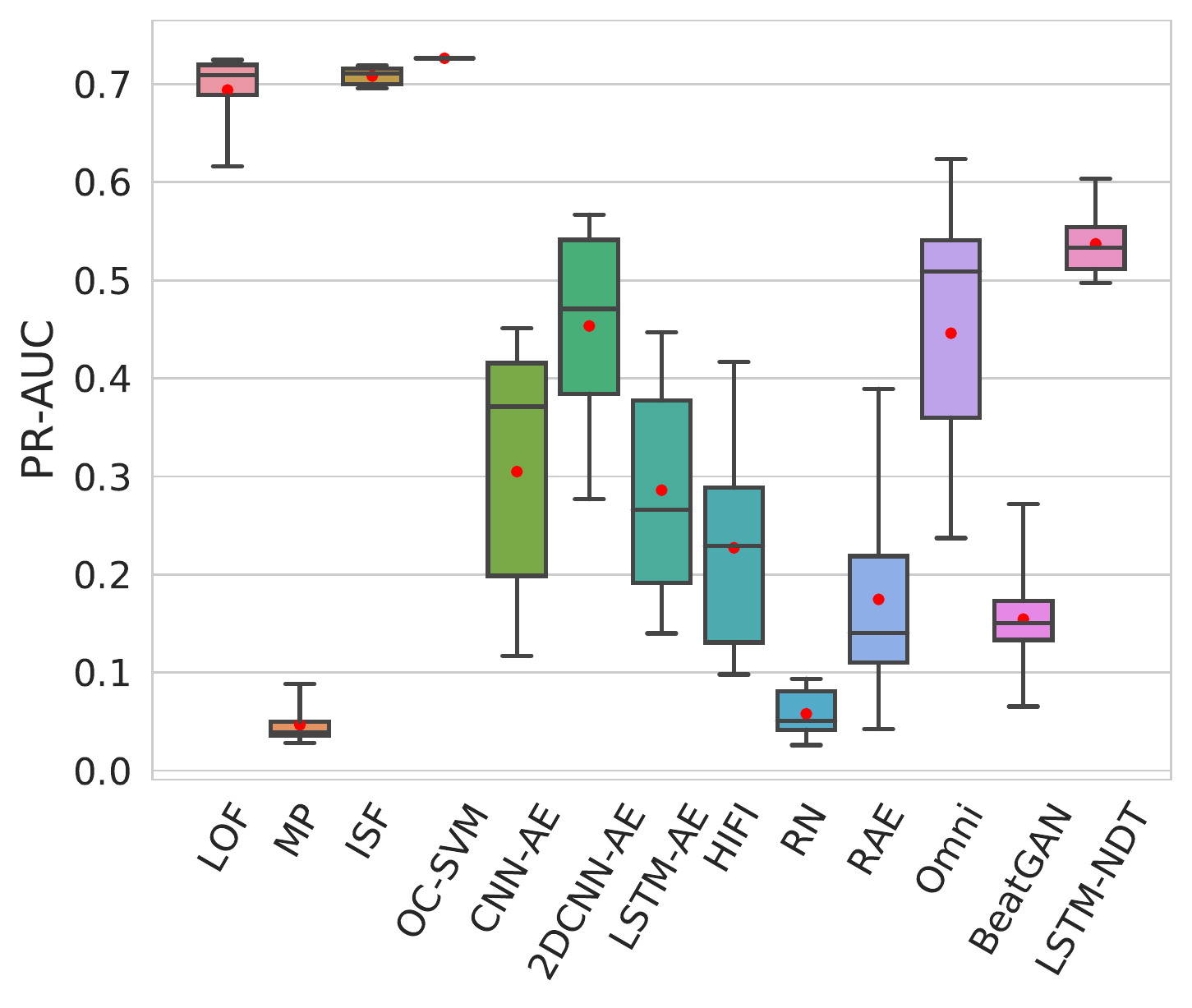}\label{fig:S5-PR_AUC}}
\subfigure[KPI: \emph{one-dimensional}] {\includegraphics[width=0.21\textwidth]{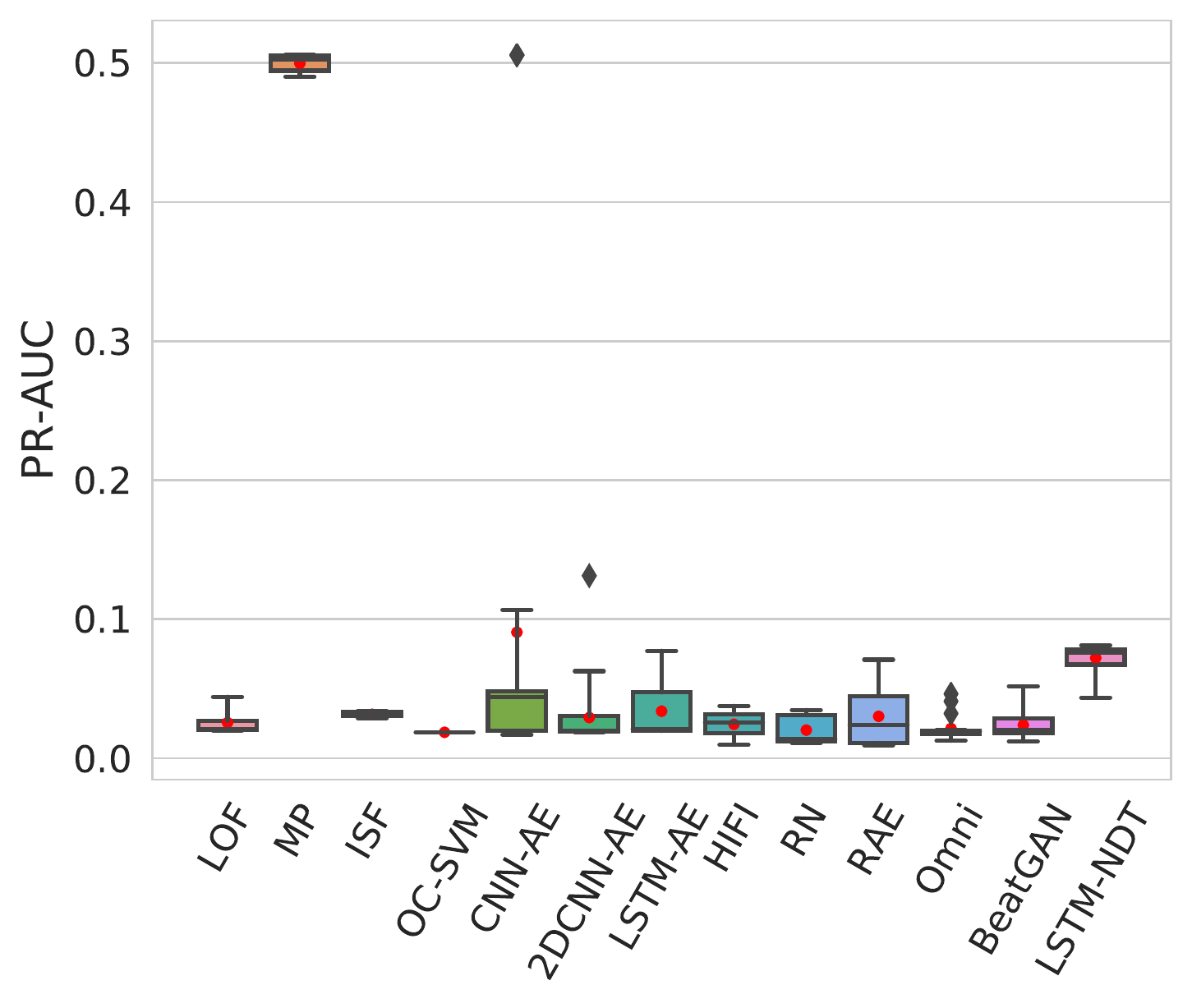}\label{fig:KPI-PR_AUC}}
\subfigure[ECG: \emph{2-dimensional}] {\includegraphics[width=0.21\textwidth]{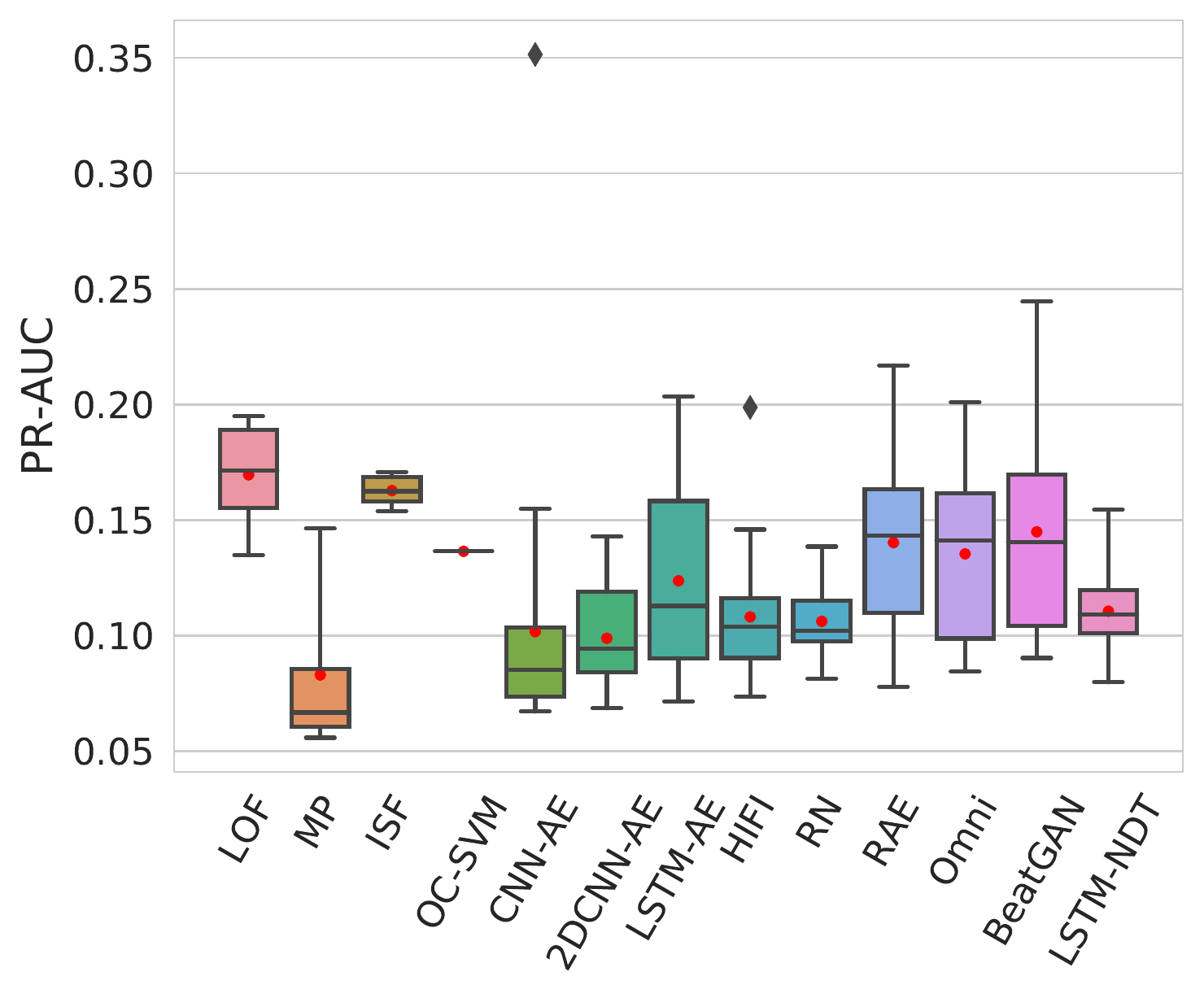}\label{fig:ECG-PR_AUC}}
\subfigure[NYC: \emph{3-dimensional}] {\includegraphics[width=0.21\textwidth]{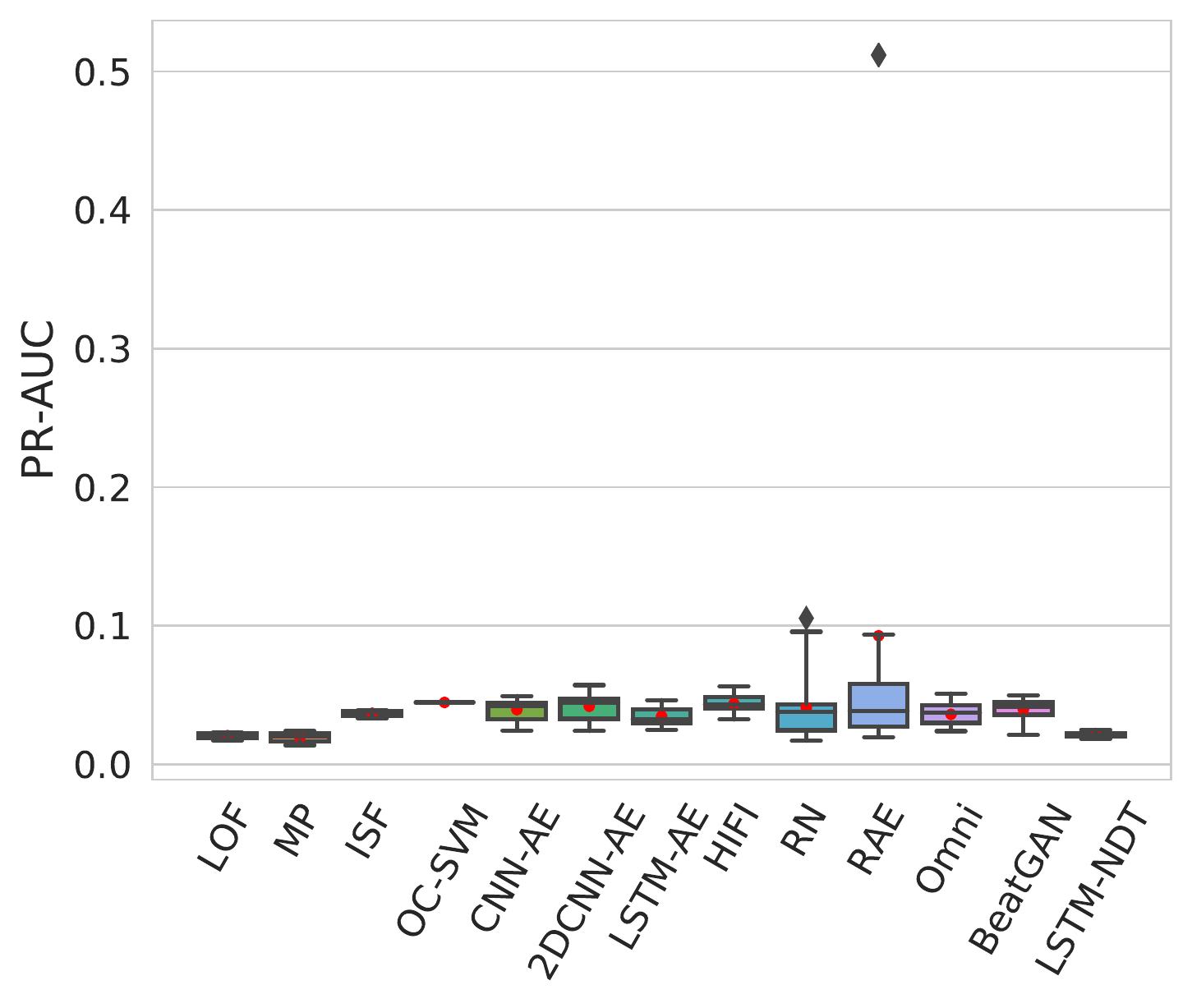}\label{fig:NYC-PR_AUC}}
\subfigure[SMAP: \emph{25-dimensional}] {\includegraphics[width=0.21\textwidth]{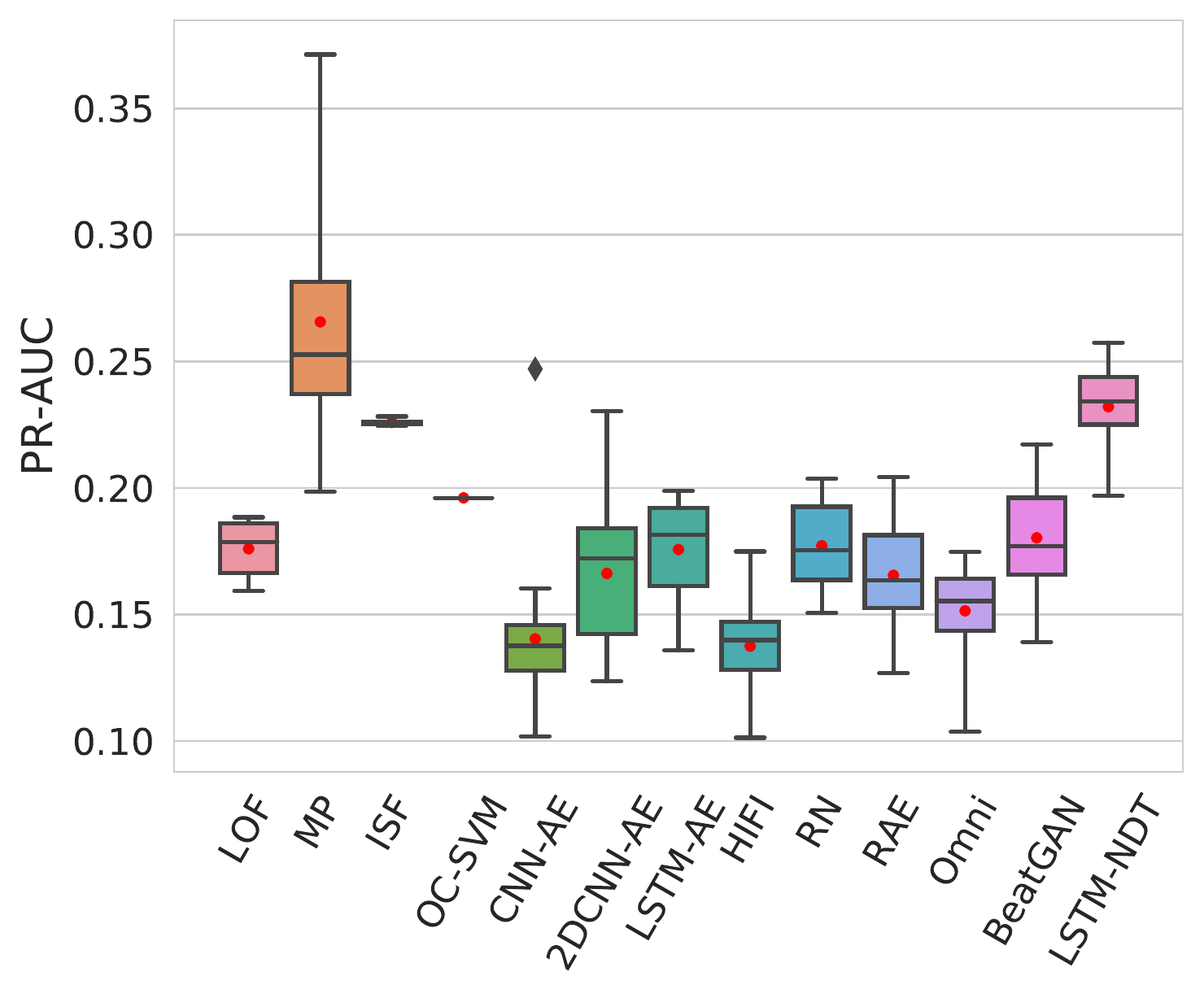}\label{fig:SMAP-PR_AUC}}
\subfigure[Credit: \emph{29-dimensional}] {\includegraphics[width=0.21\textwidth]{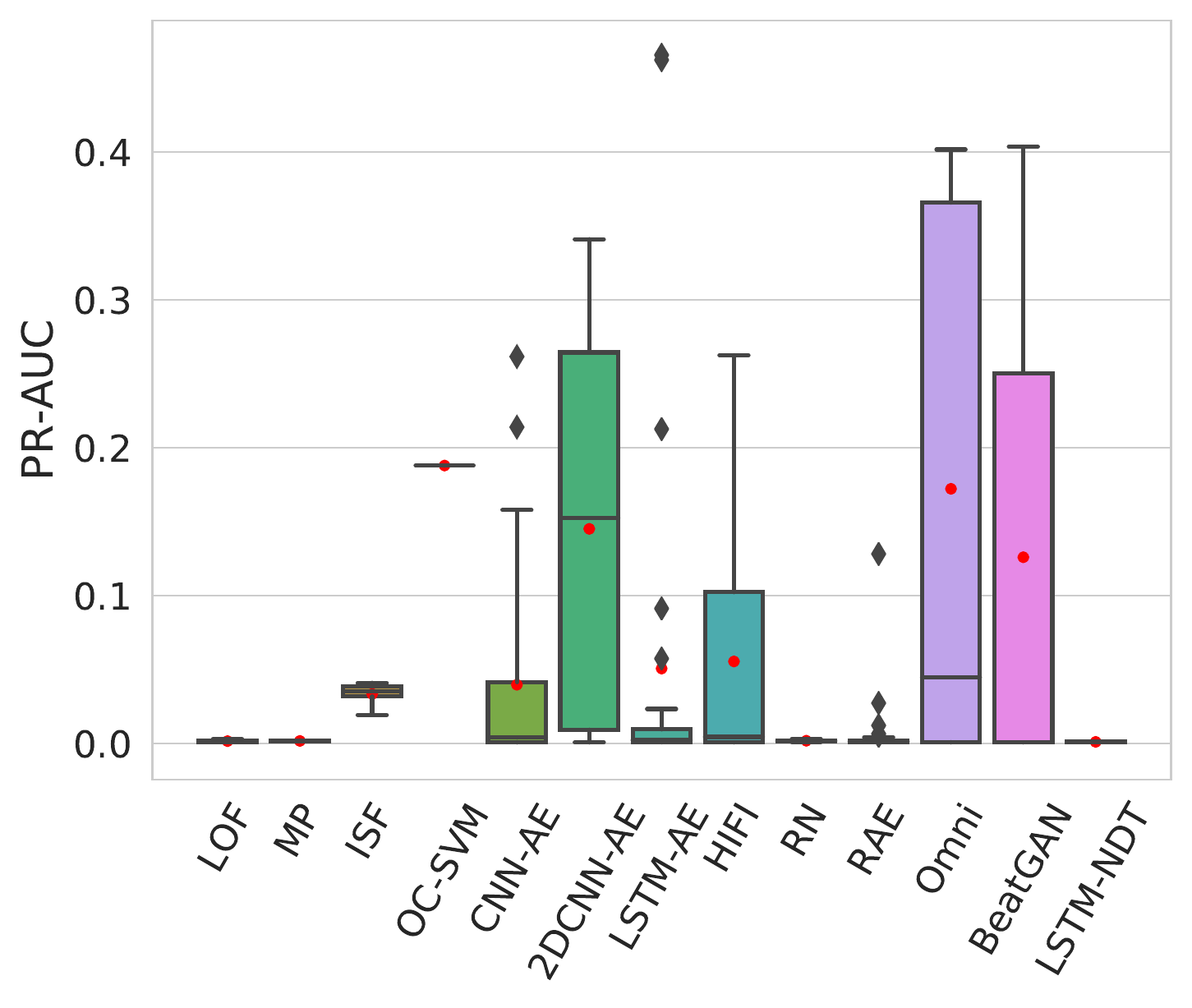}\label{fig:Credit-PR_AUC}}
\subfigure[SWaT: \emph{51-dimensional}] {\includegraphics[width=0.21\textwidth]{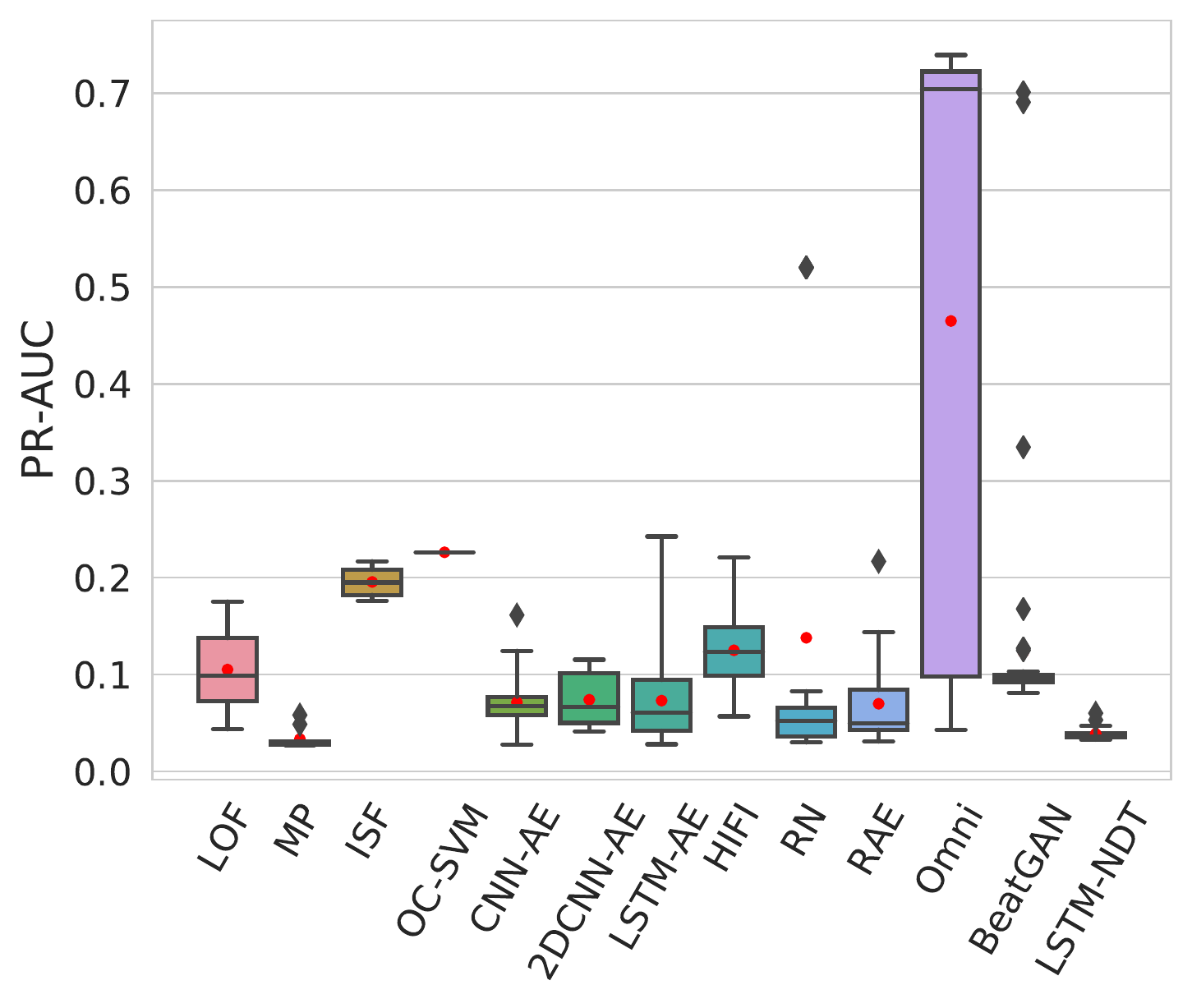}\label{fig:SWaT-PR_AUC}}
\subfigure[MSL: \emph{55-dimensional}] {\includegraphics[width=0.21\textwidth]{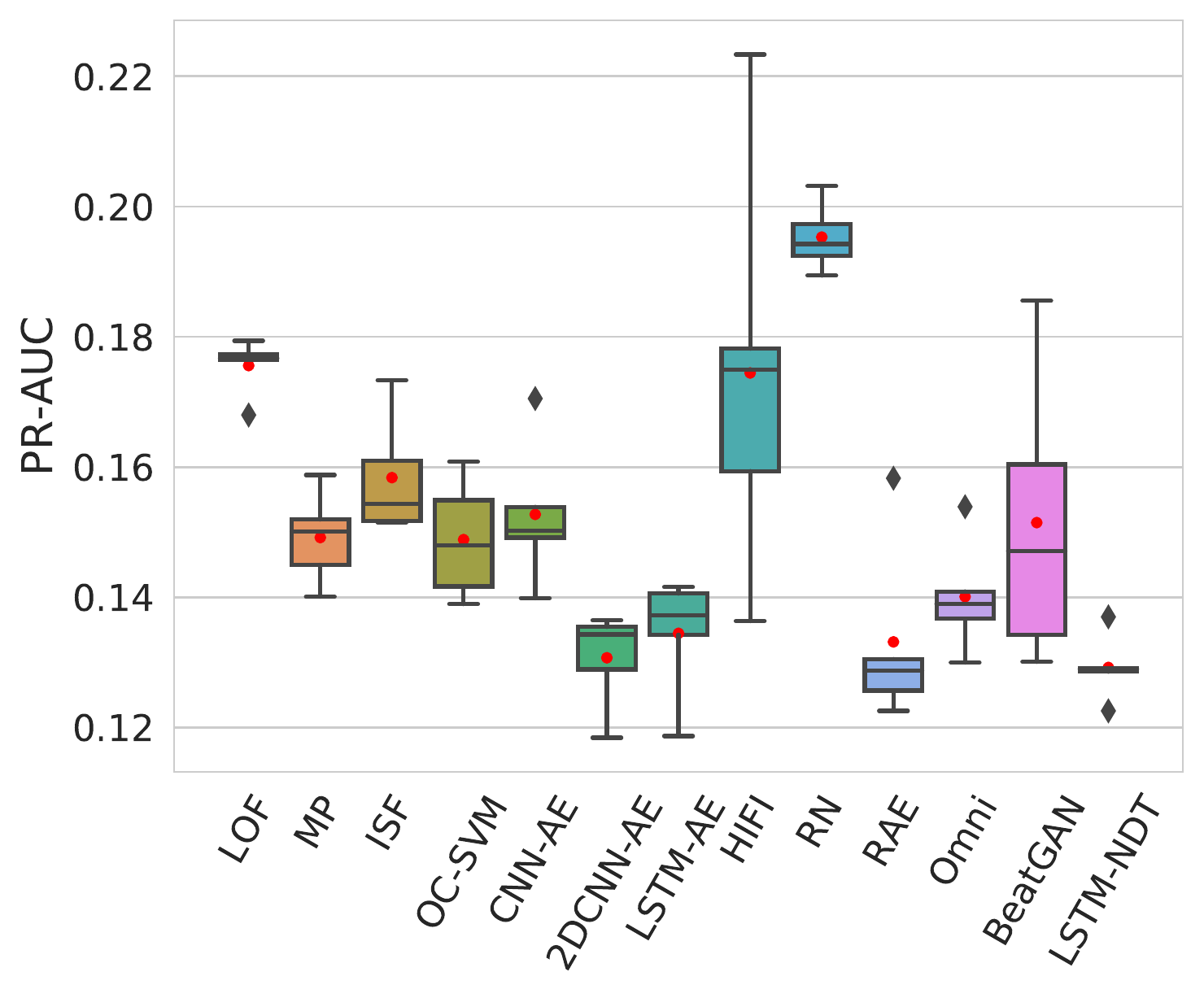}\label{fig:MSL-PR_AUC}}
\vskip -9pt
\caption{PR-AUC (Univariate vs. Multivariate)}
\label{fig:pr}
\end{figure*}

\begin{figure*}
\centering
\subfigure[NAB-art-stationary] {\includegraphics[width=0.21
\textwidth]{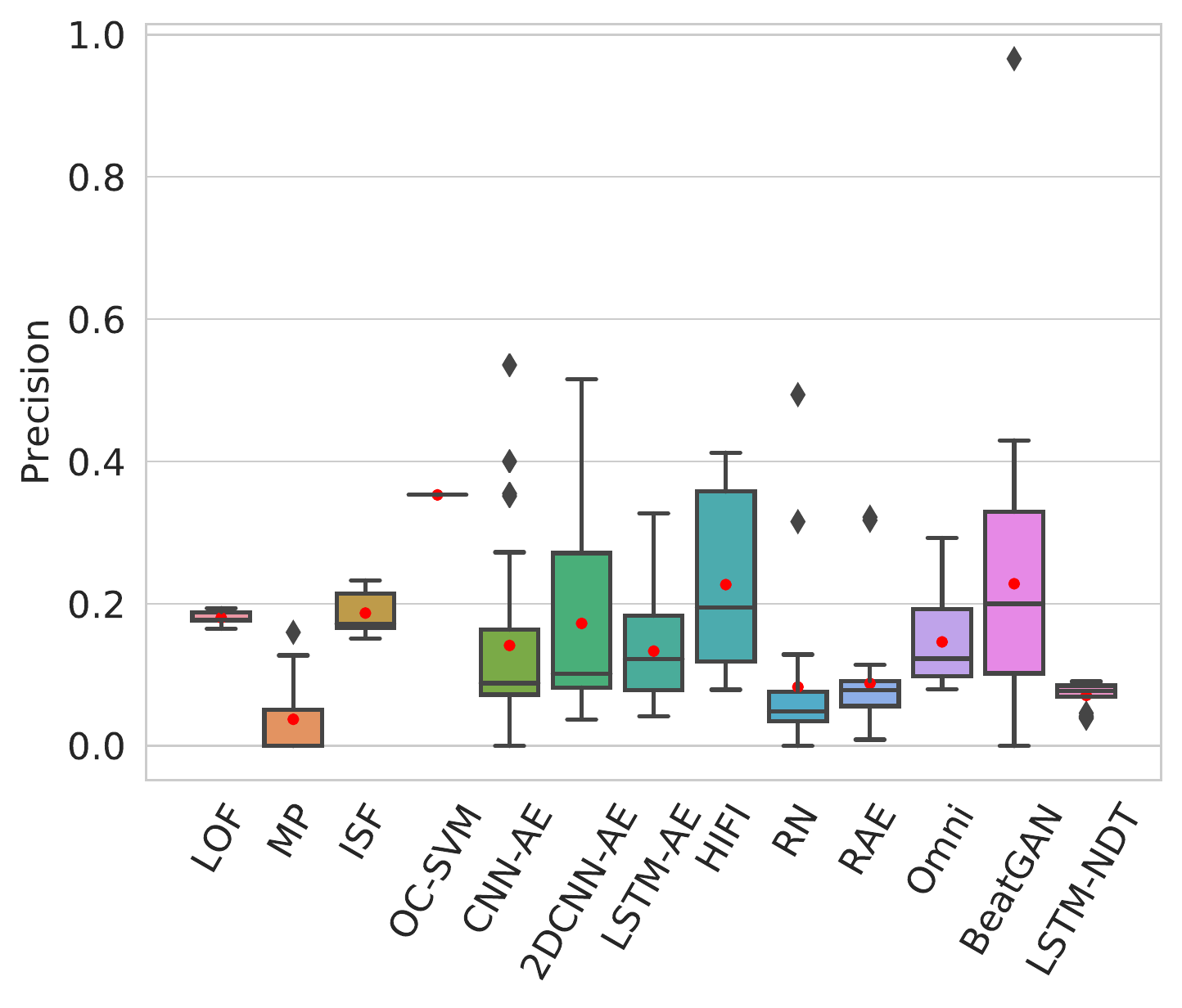}\label{fig:NAB-art-stationary-precision}}
\subfigure[NAB-cpu-stationary] {\includegraphics[width=0.21
\textwidth]{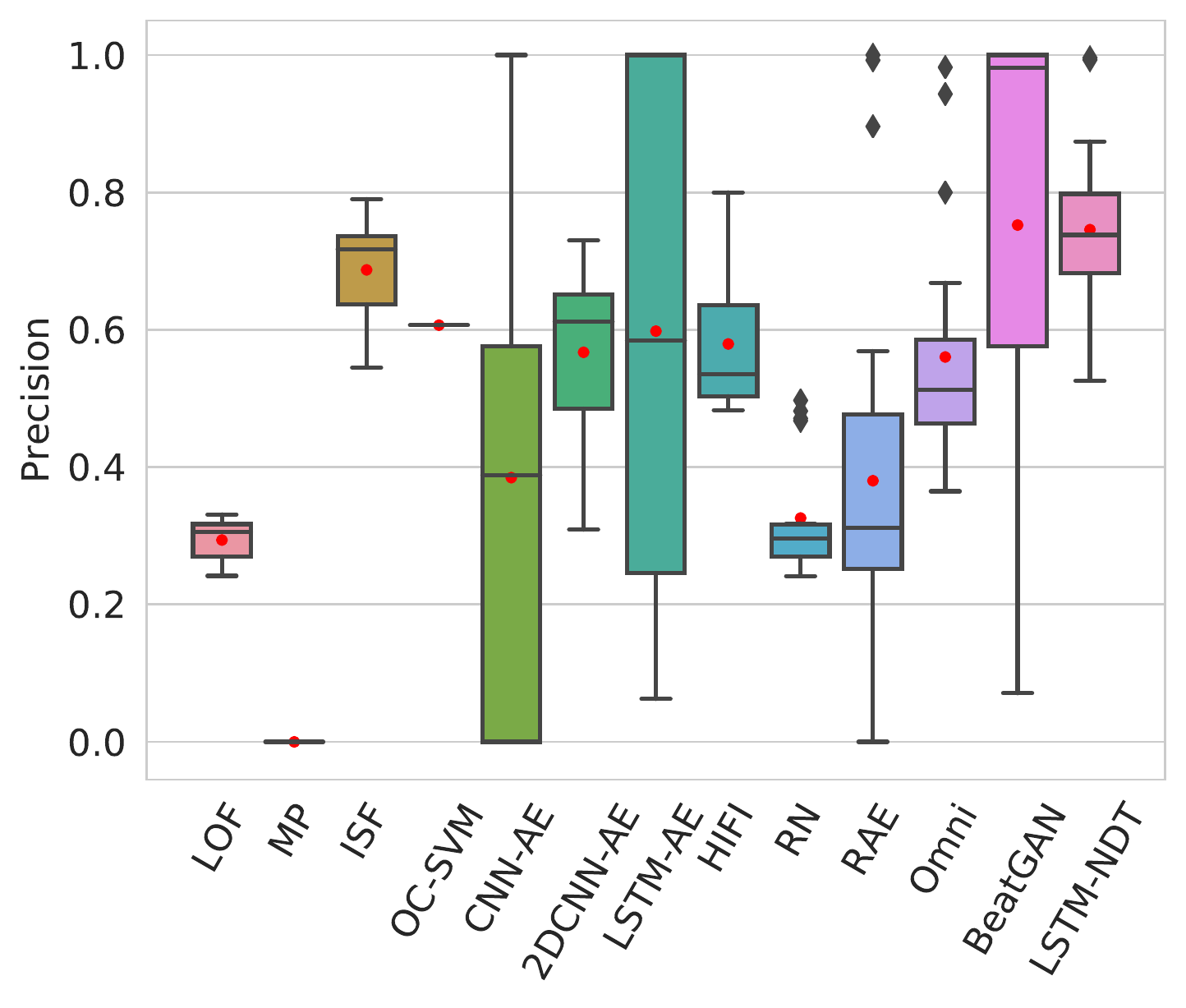}\label{fig:NAB-cpu-stationary-precision}}
\subfigure[NAB-ec2-stationary] {\includegraphics[width=0.21
\textwidth]{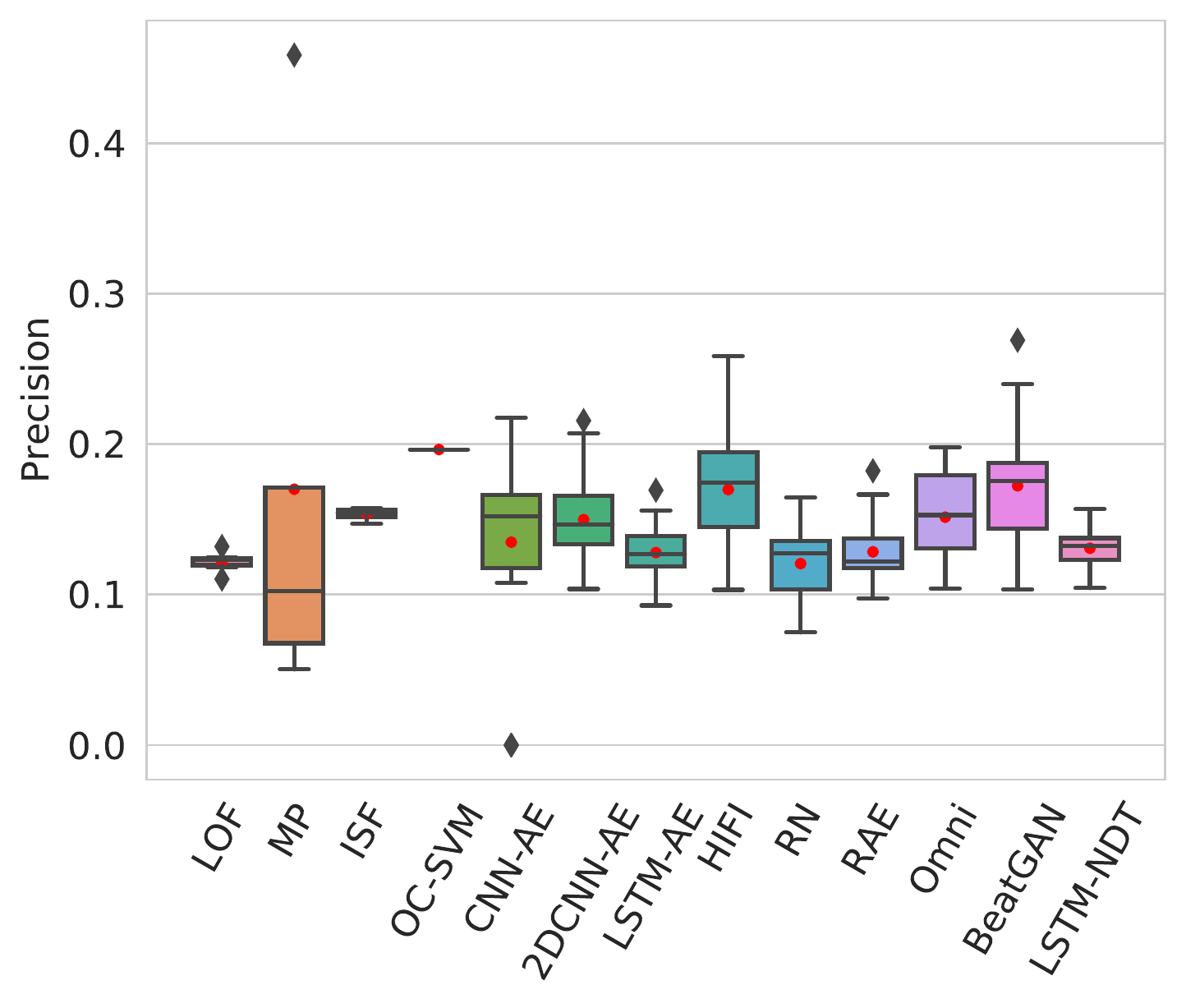}\label{fig:NAB-ec2-stationary-precision}}
\subfigure[NAB-elb-stationary] {\includegraphics[width=0.21
\textwidth]{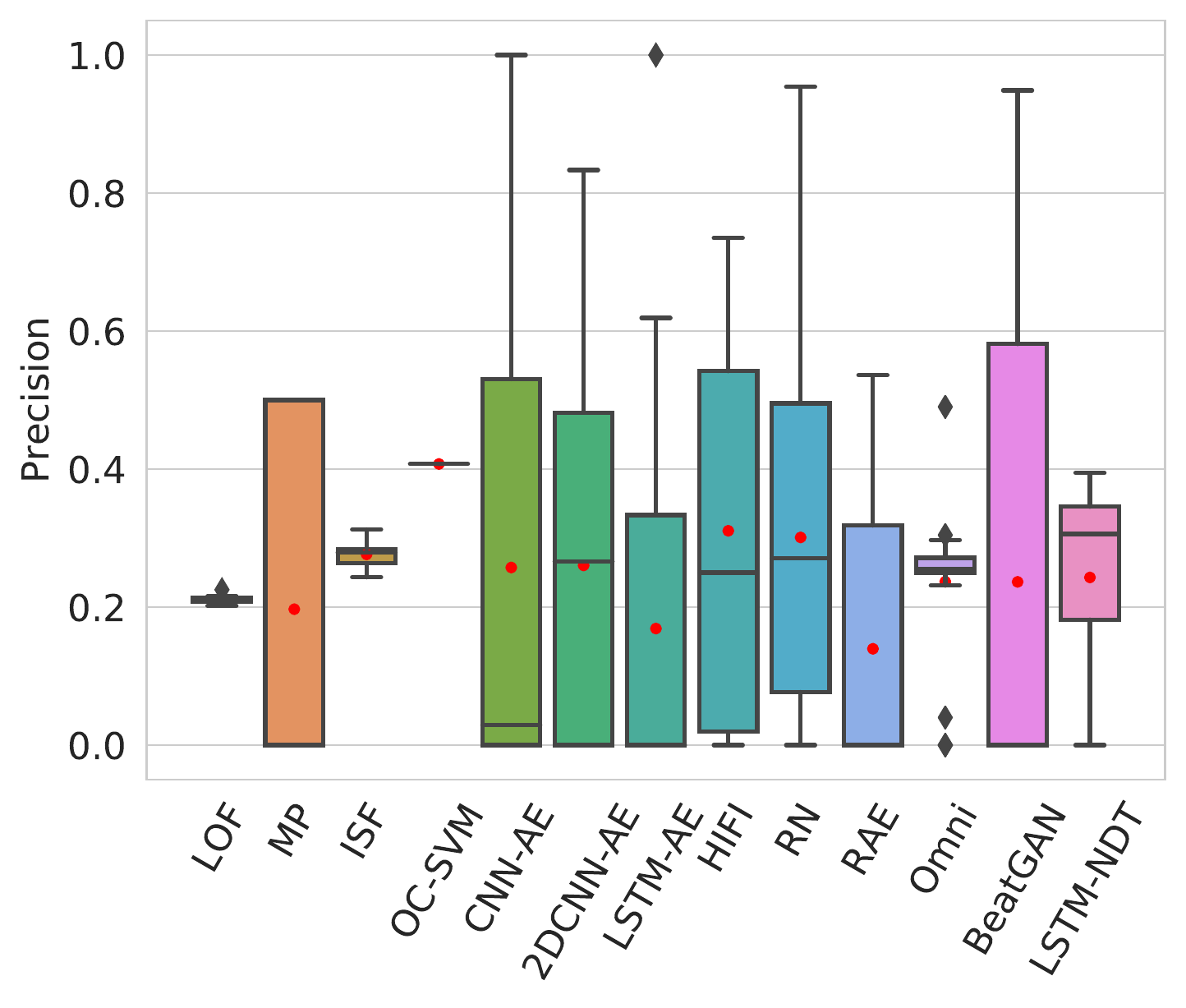}\label{fig:NAB-elb-stationary-precision}}
\subfigure[NAB-ambient-nonstationary] {\includegraphics[width=0.21
\textwidth]{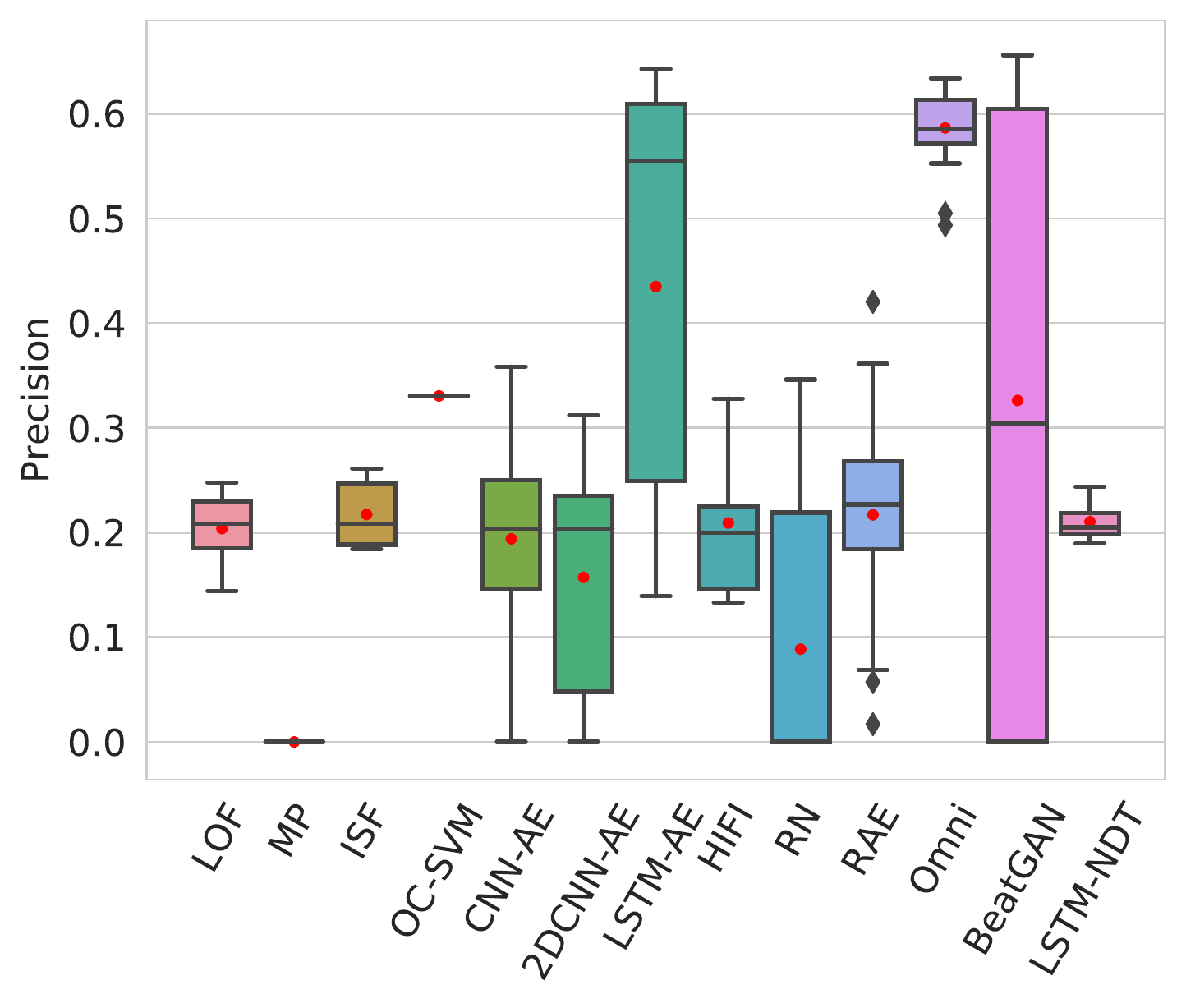}\label{fig:NAB-ambient-Nonstationary-precision}}
\subfigure[NAB-ec2-nonstationary] {\includegraphics[width=0.21
\textwidth]{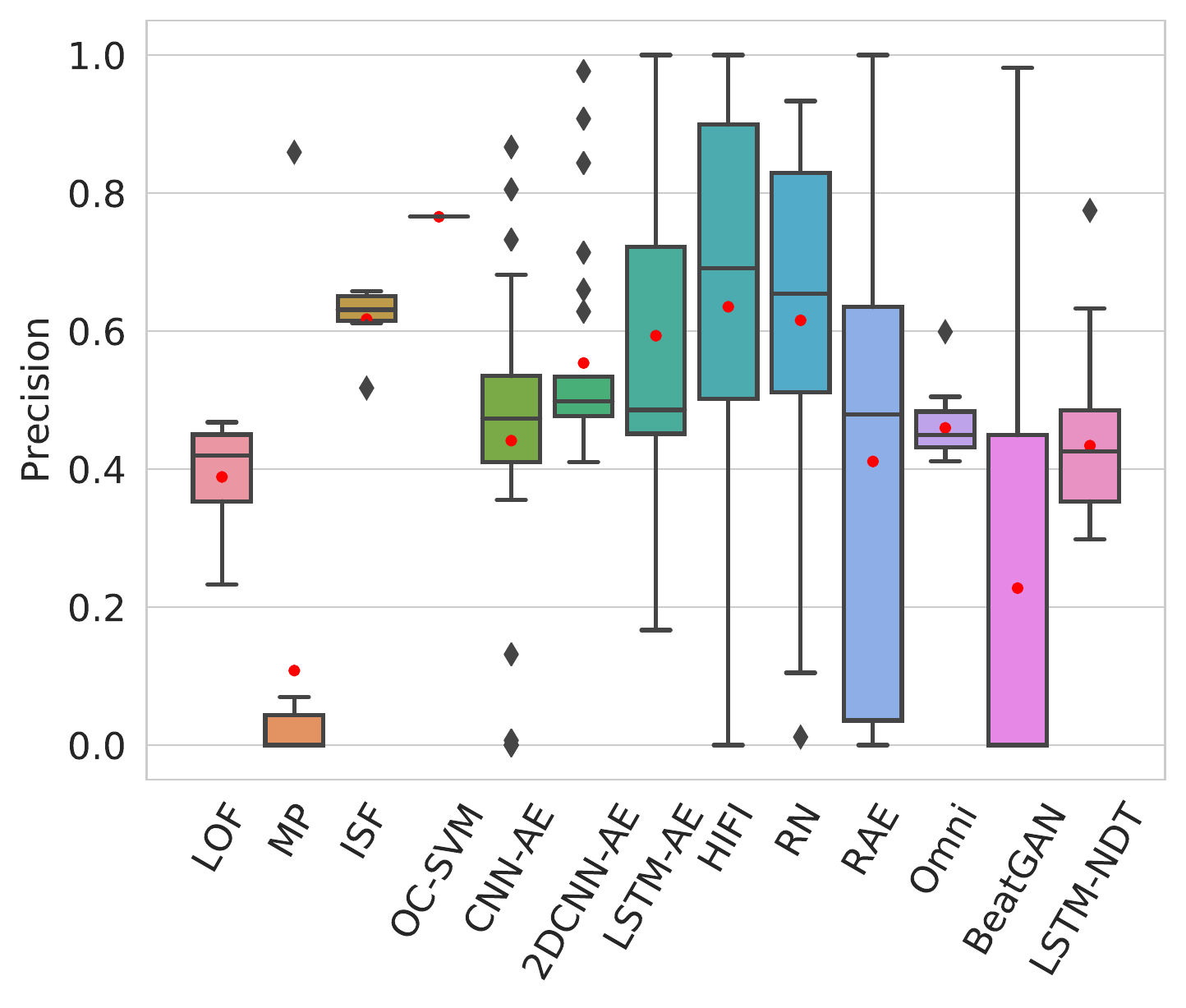}\label{fig:NAB-ec2-Nonstationary-precision}}
\subfigure[NAB-exchange-nonstationary] {\includegraphics[width=0.21
\textwidth]{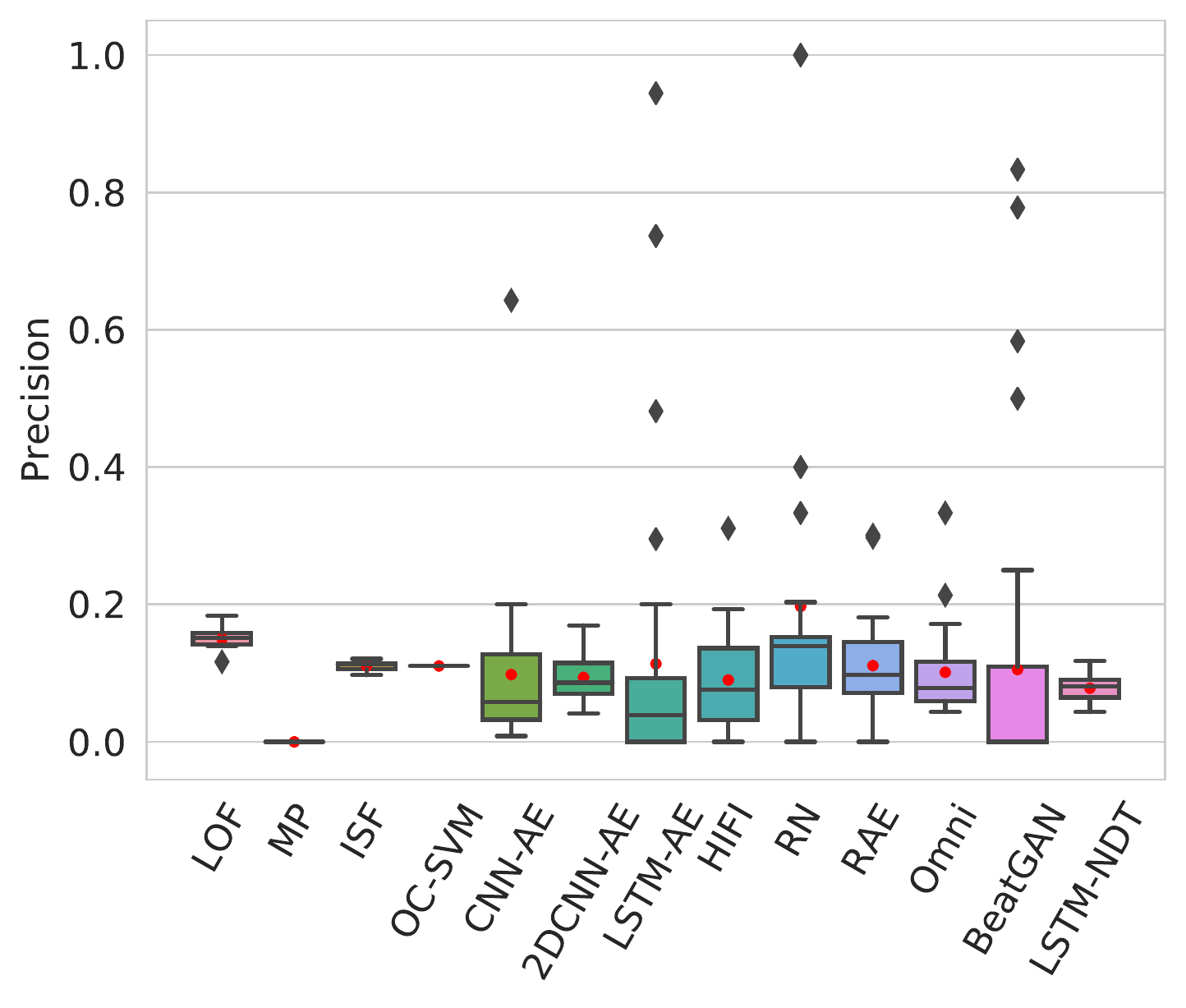}\label{fig:NAB-exchange-Nonstationary-precision}}
\subfigure[NAB-grok-nonstationary] {\includegraphics[width=0.21
\textwidth]{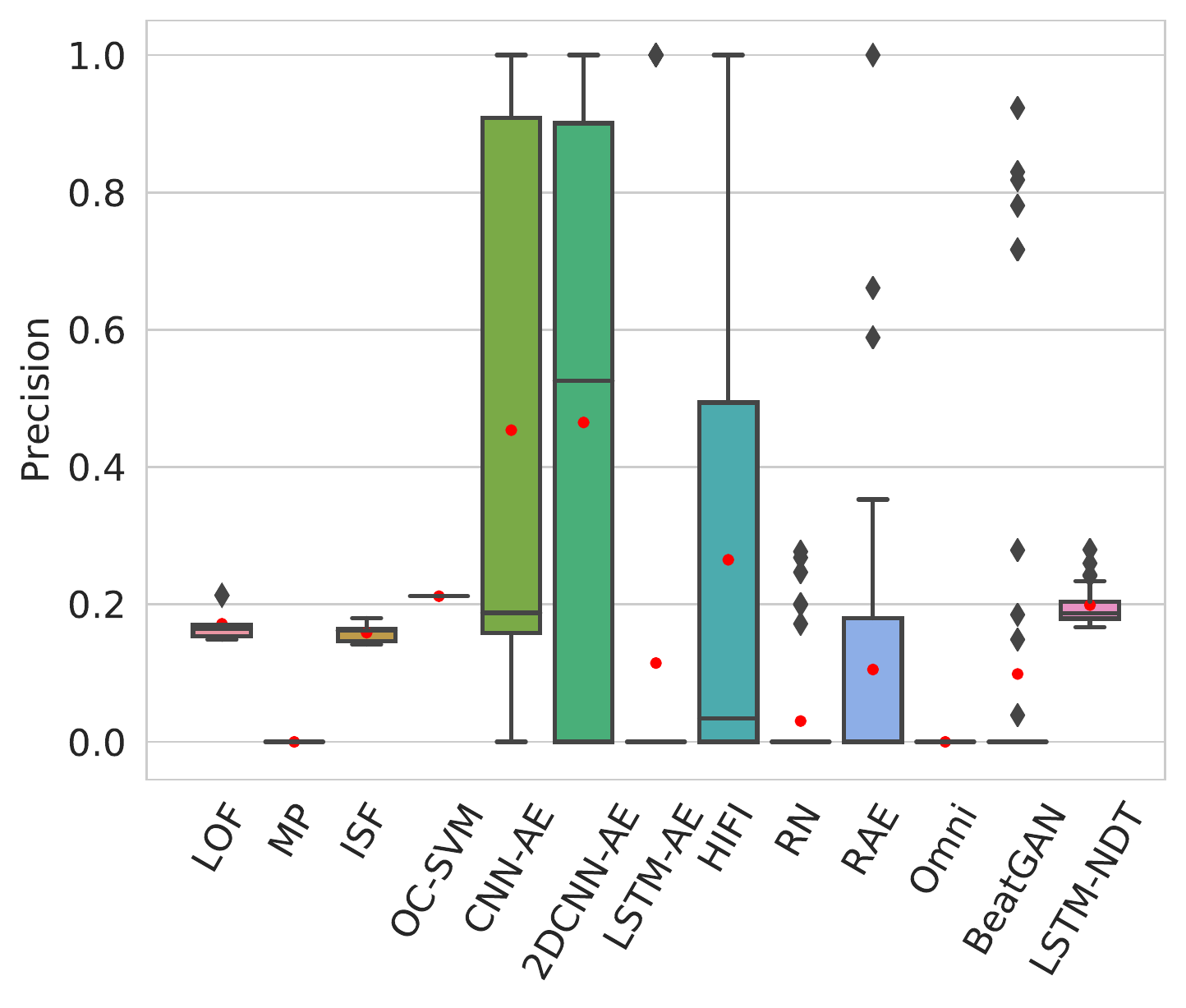}\label{fig:NAB-grok-Nonstationary-precision}}
\vskip -9pt
\caption{Precision (Stationary vs. Non-stationary)}
\vspace{-0.3cm}
\label{fig:NAB_pre}
\end{figure*}

\begin{figure*}
\centering
\subfigure[NAB-art-stationary] {\includegraphics[width=0.21
\textwidth]{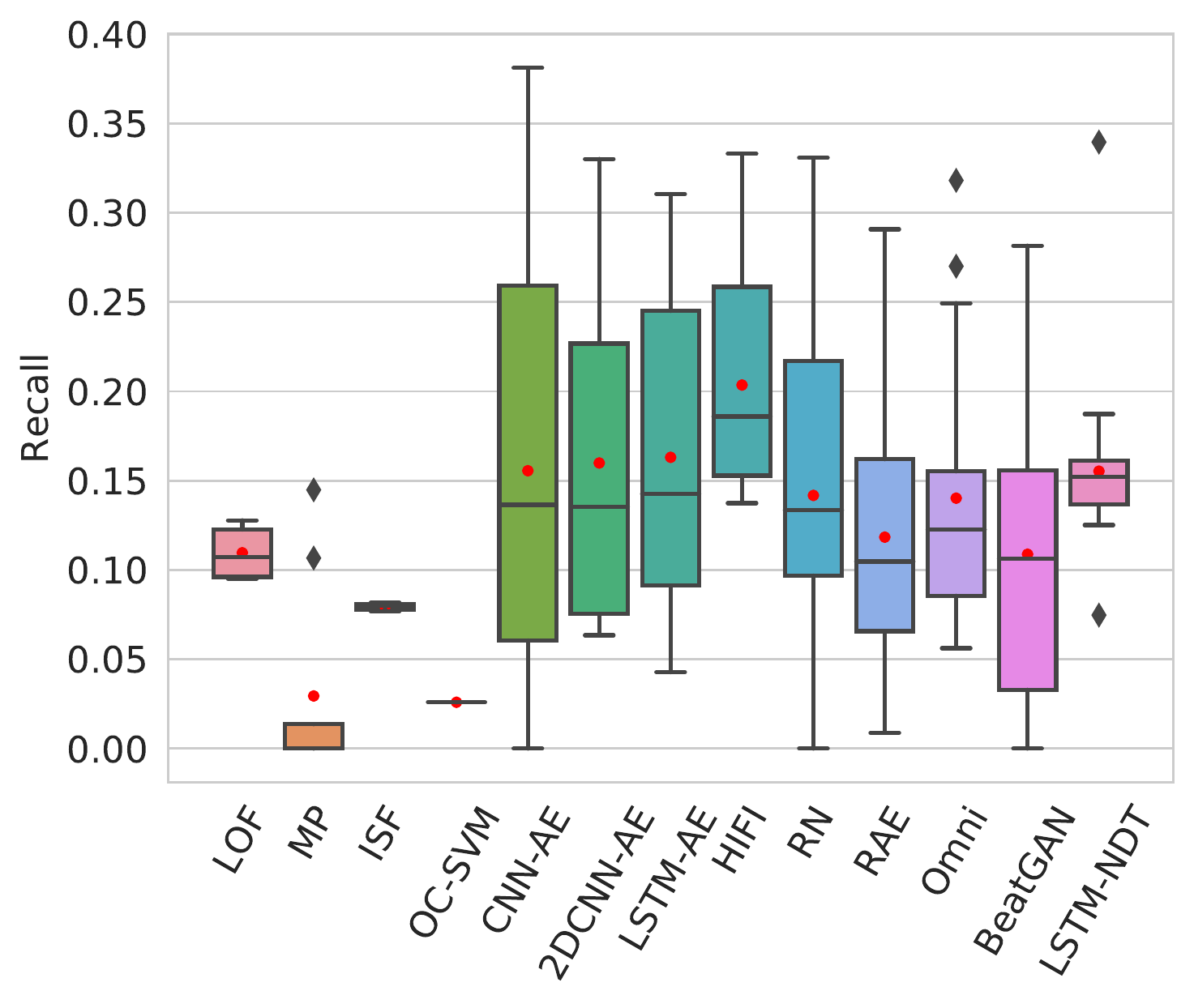}\label{fig:NAB-art-stationary-recall}}
\subfigure[NAB-cpu-stationary] {\includegraphics[width=0.21
\textwidth]{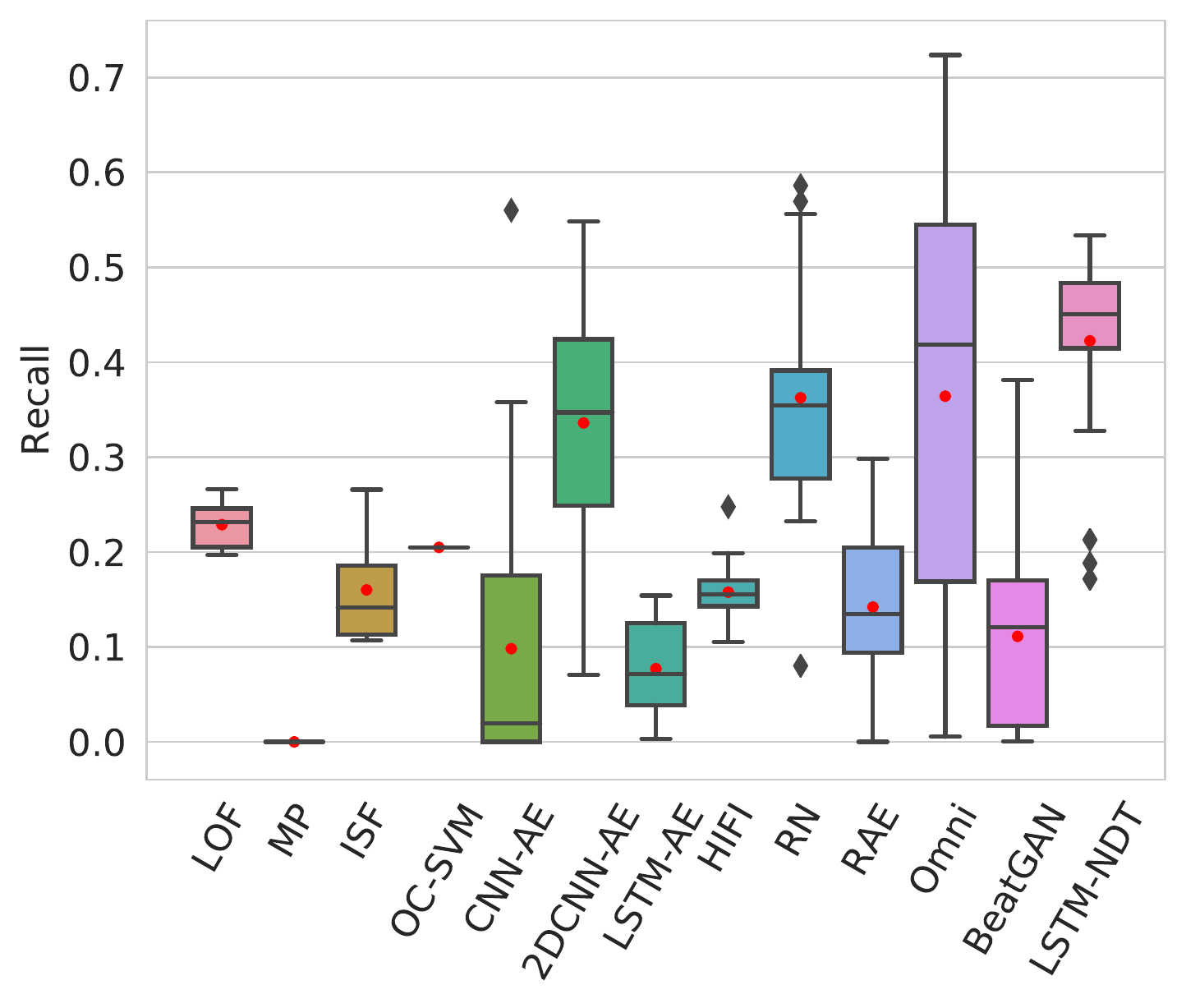}\label{fig:NAB-cpu-stationary-recall}}
\subfigure[NAB-ec2-stationary] {\includegraphics[width=0.21
\textwidth]{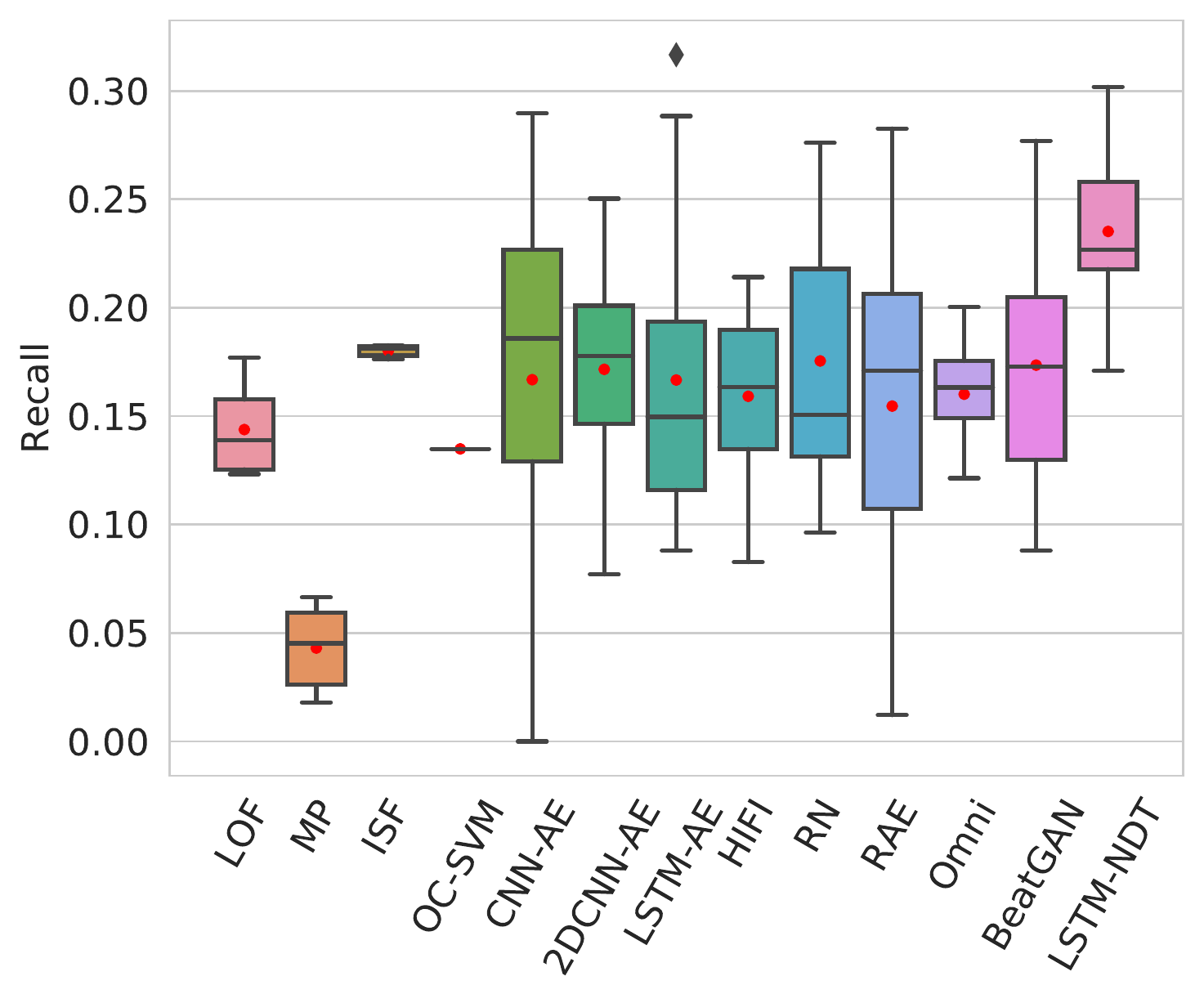}\label{fig:NAB-ec2-stationary-recall}}
\subfigure[NAB-elb-stationary] {\includegraphics[width=0.21
\textwidth]{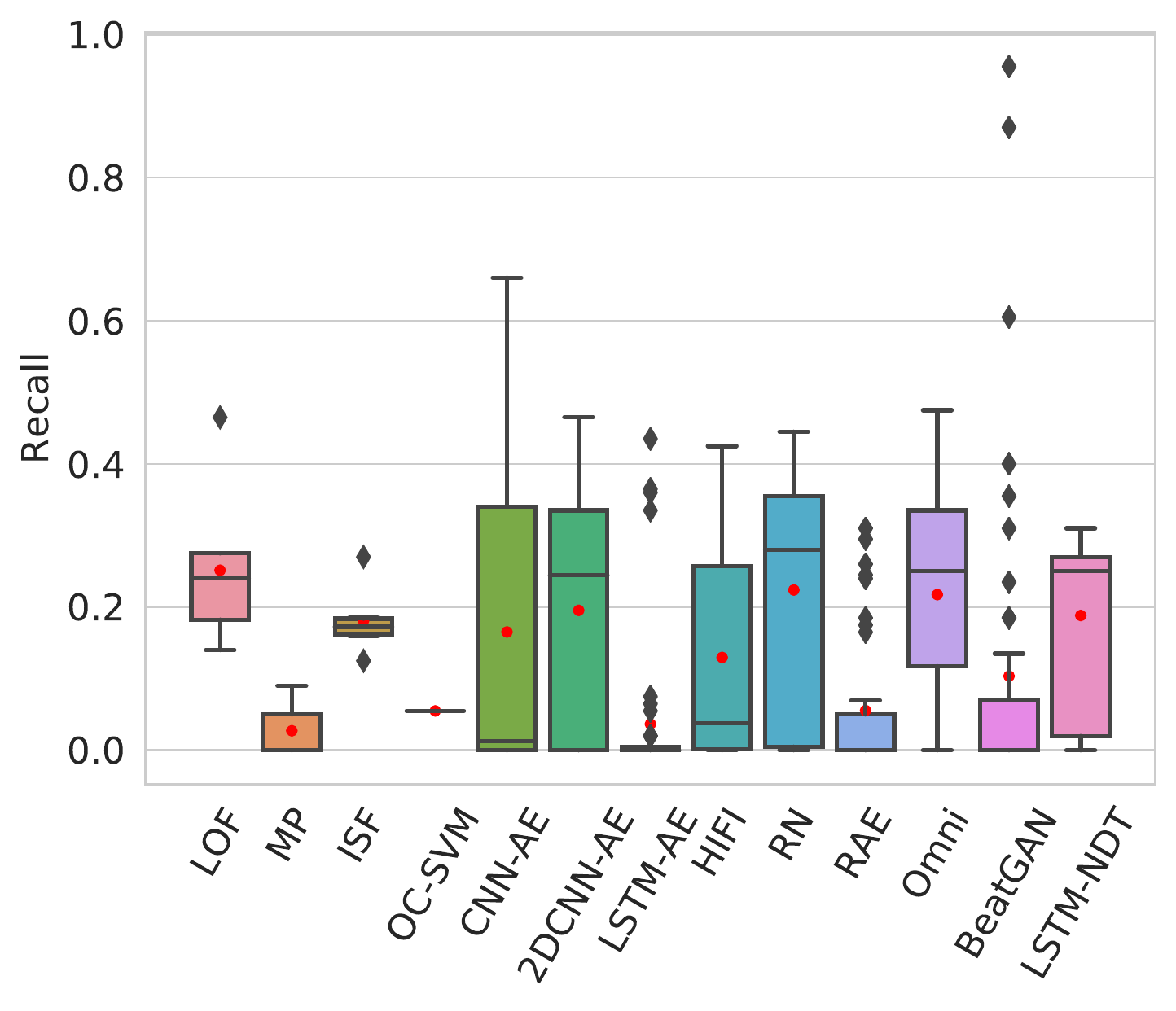}\label{fig:NAB-elb-stationary-precision}}
\subfigure[NAB-ambient-nonstationary] {\includegraphics[width=0.21
\textwidth]{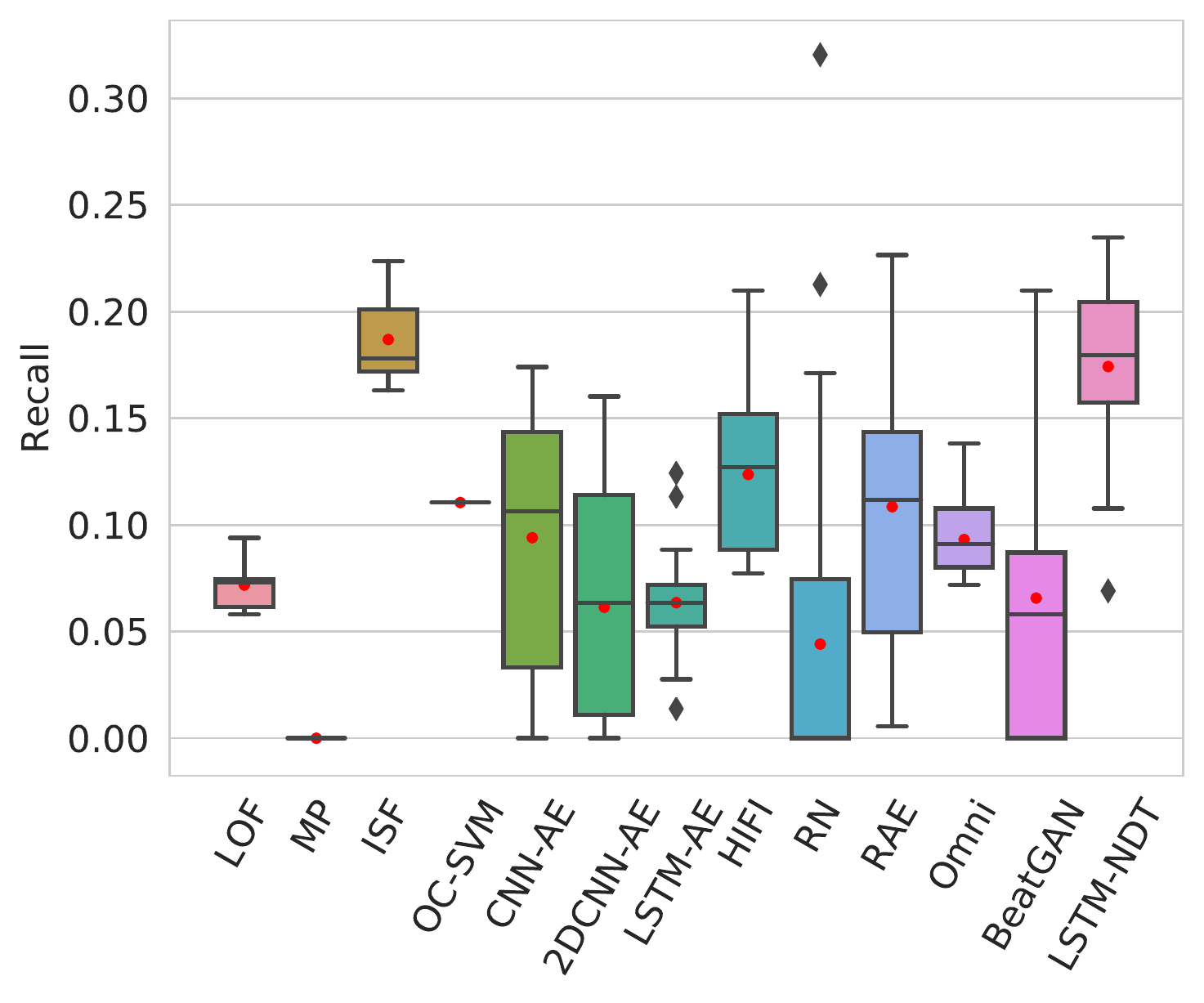}\label{fig:NAB-ambient-Nonstationary-recall}}
\subfigure[NAB-ec2-nonstationary] {\includegraphics[width=0.21
\textwidth]{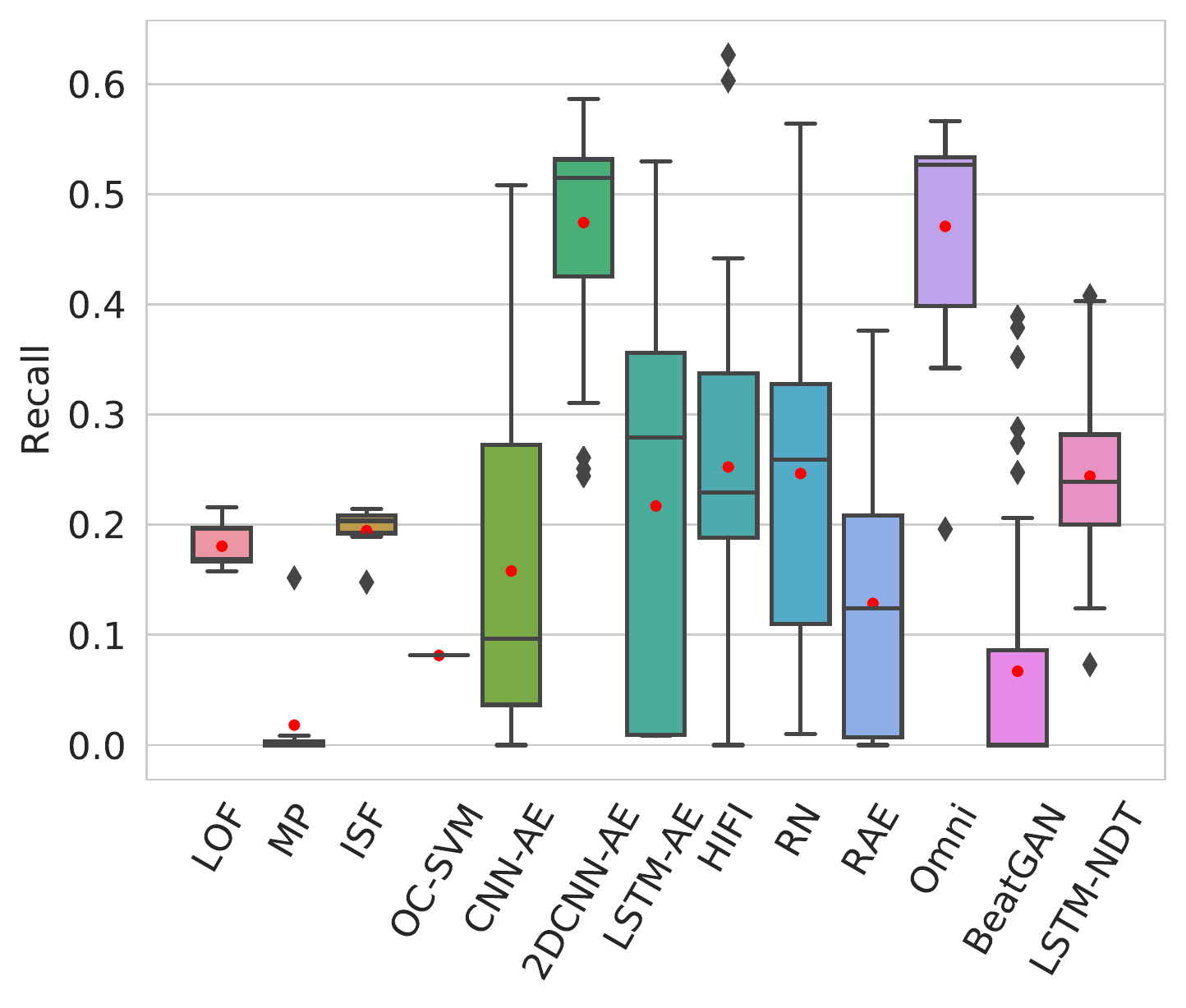}\label{fig:NAB-ec2-Nonstationary-recall}}
\subfigure[NAB-exchange-nonstationary] {\includegraphics[width=0.21
\textwidth]{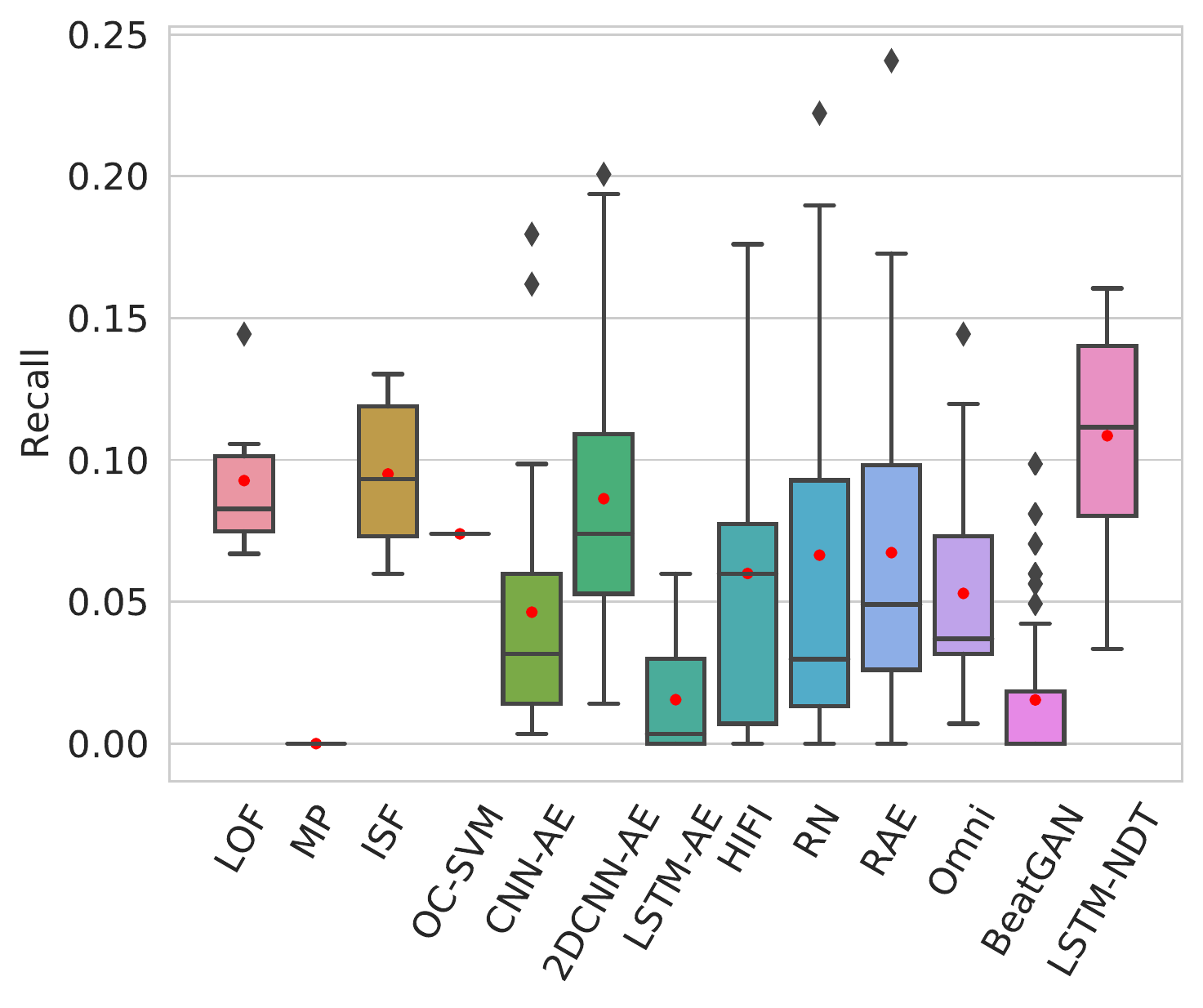}\label{fig:NAB-exchange-Nonstationary-recall}}
\subfigure[NAB-grok-nonstationary] {\includegraphics[width=0.21
\textwidth]{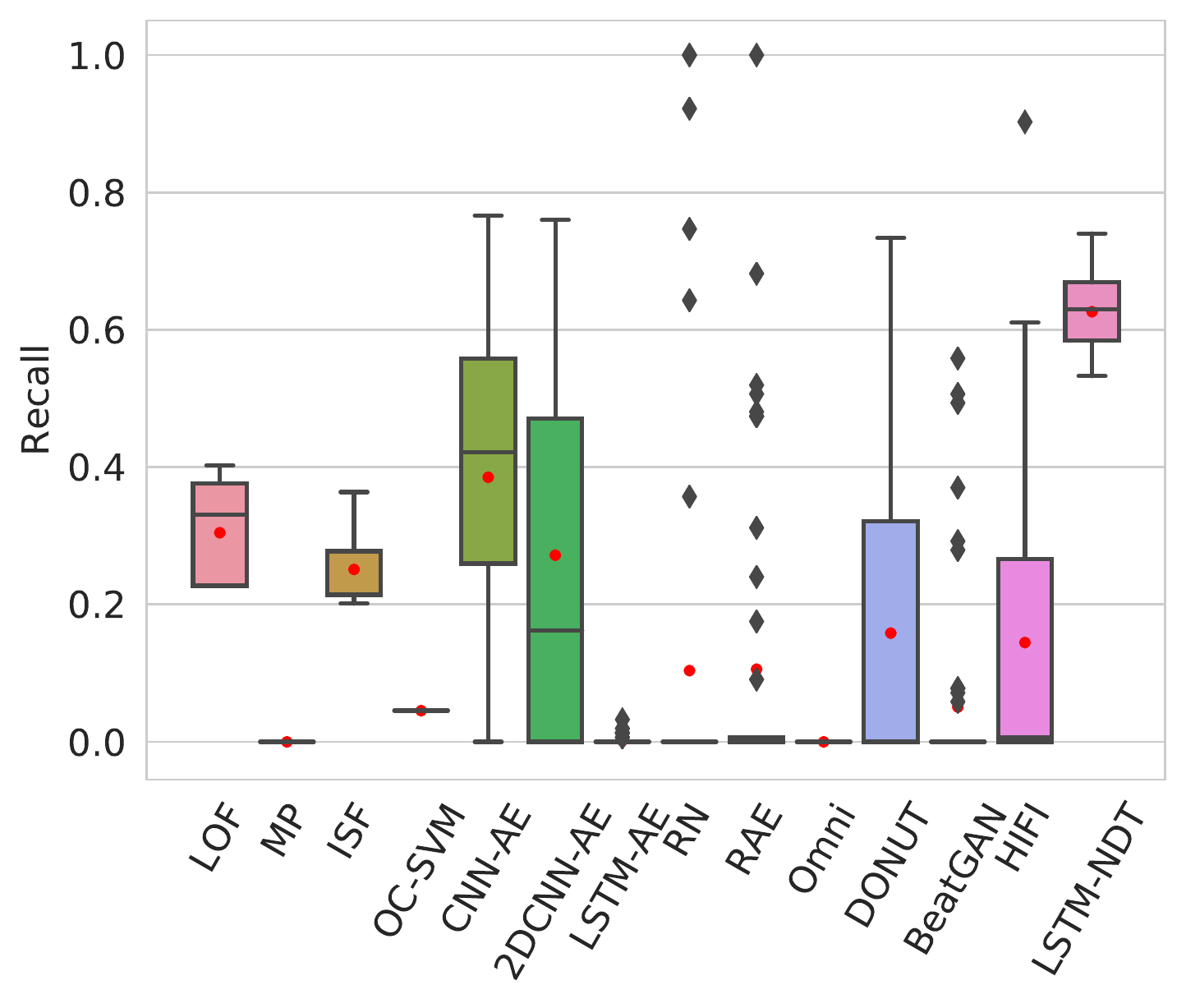}\label{fig:NAB-grok-Nonstationary-recall}}
\vskip -9pt
\caption{Recall (Stationary vs. Non-stationary)}
\vspace{-0.3cm}
\label{fig:NAB_rec}
\end{figure*}

\begin{figure*}
\centering
\subfigure[NAB-art-stationary] {\includegraphics[width=0.21
\textwidth]{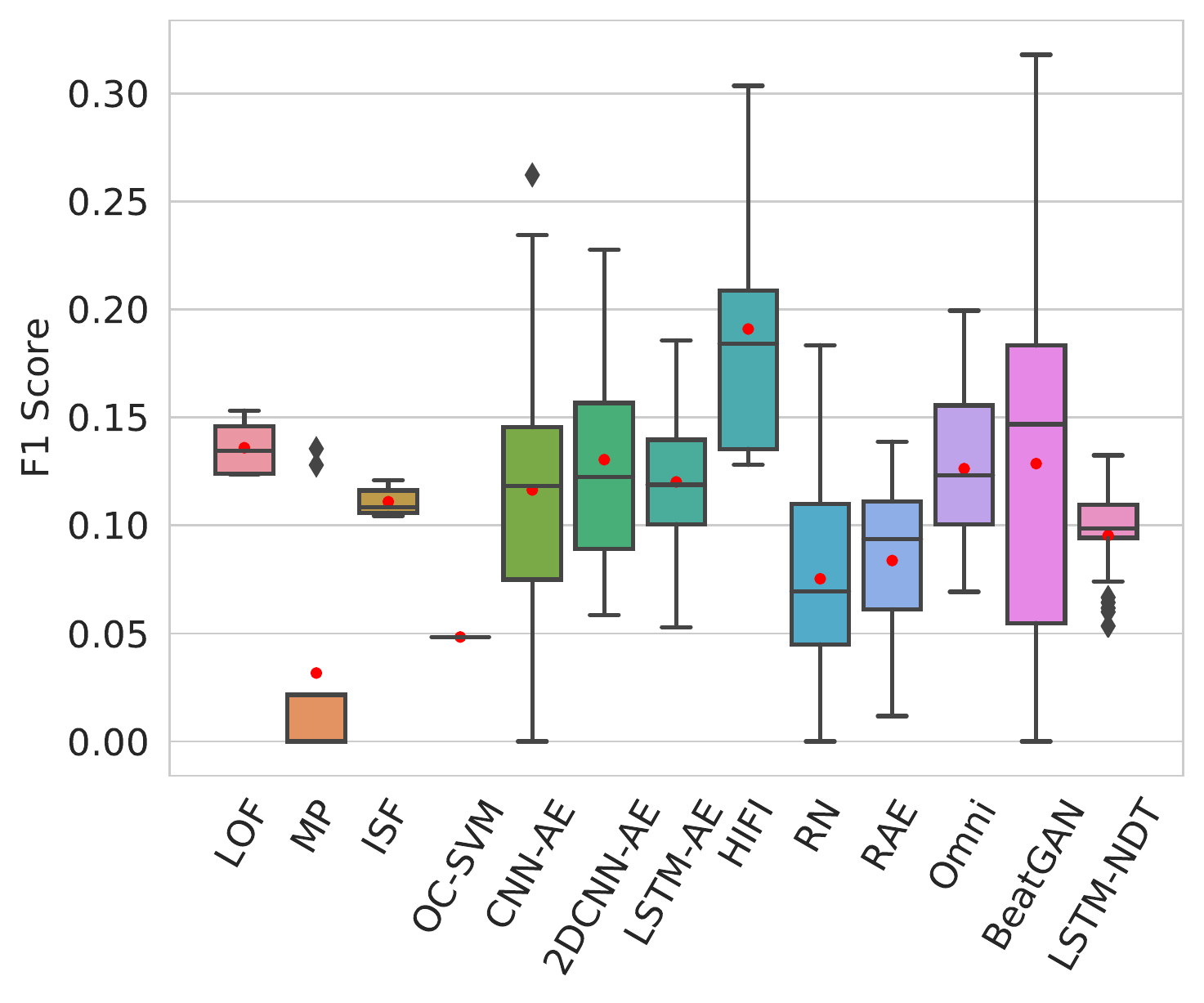}\label{fig:NAB-art-stationary-f1}}
\subfigure[NAB-cpu-stationary] {\includegraphics[width=0.21
\textwidth]{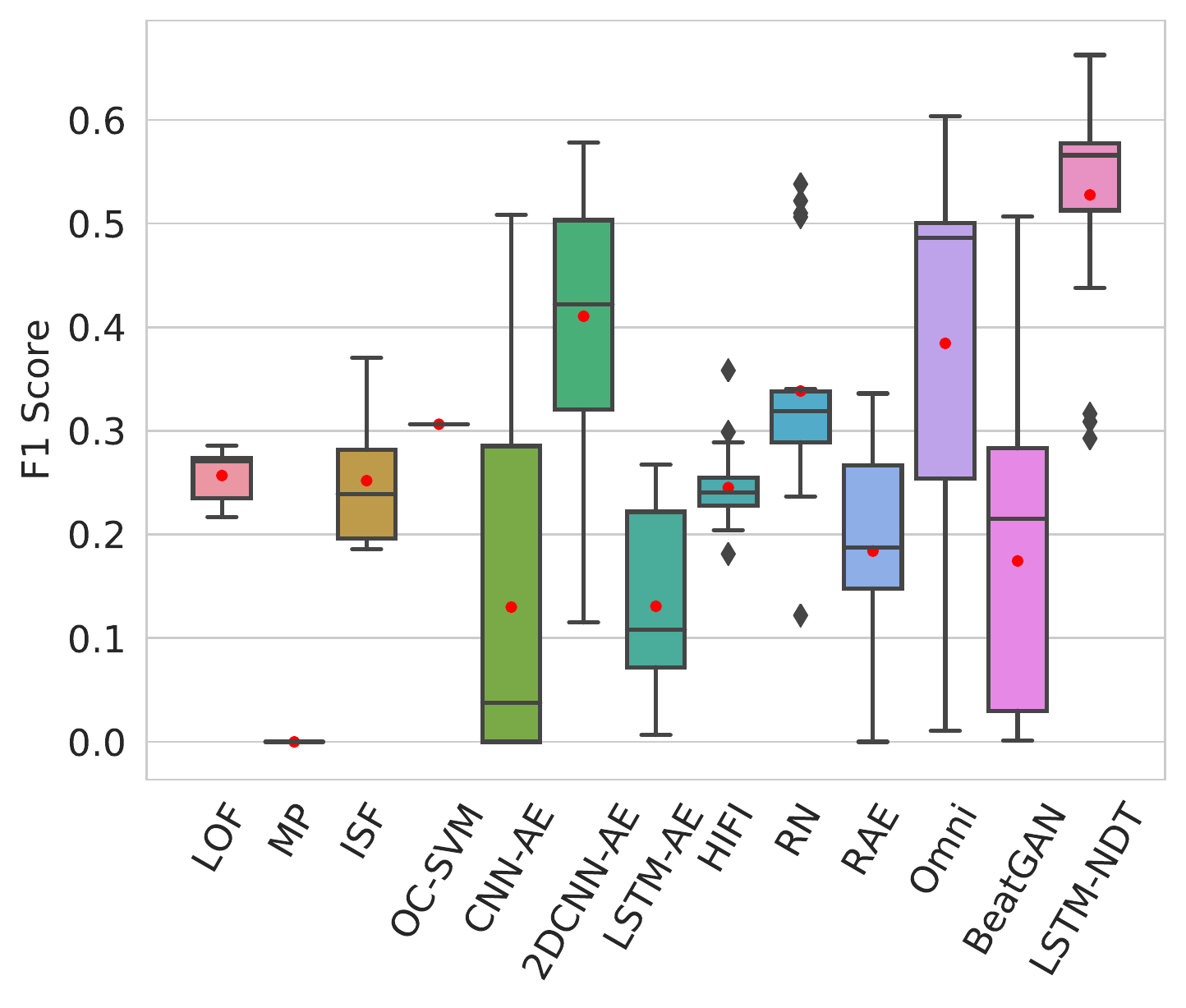}\label{fig:NAB-cpu-stationary-f1}}
\subfigure[NAB-ec2-stationary] {\includegraphics[width=0.21
\textwidth]{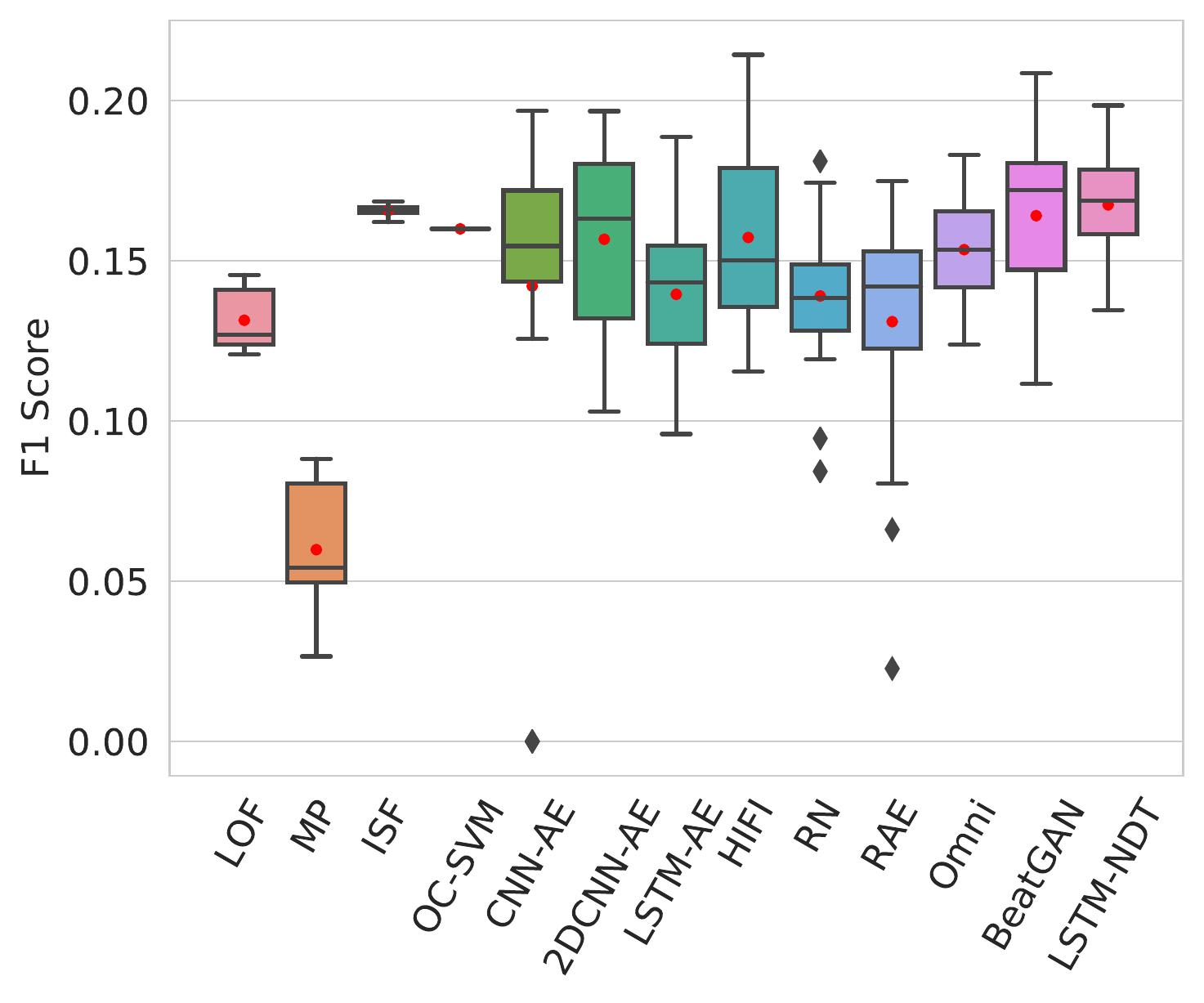}\label{fig:NAB-ec2-stationary-f1}}
\subfigure[NAB-elb-stationary] {\includegraphics[width=0.21
\textwidth]{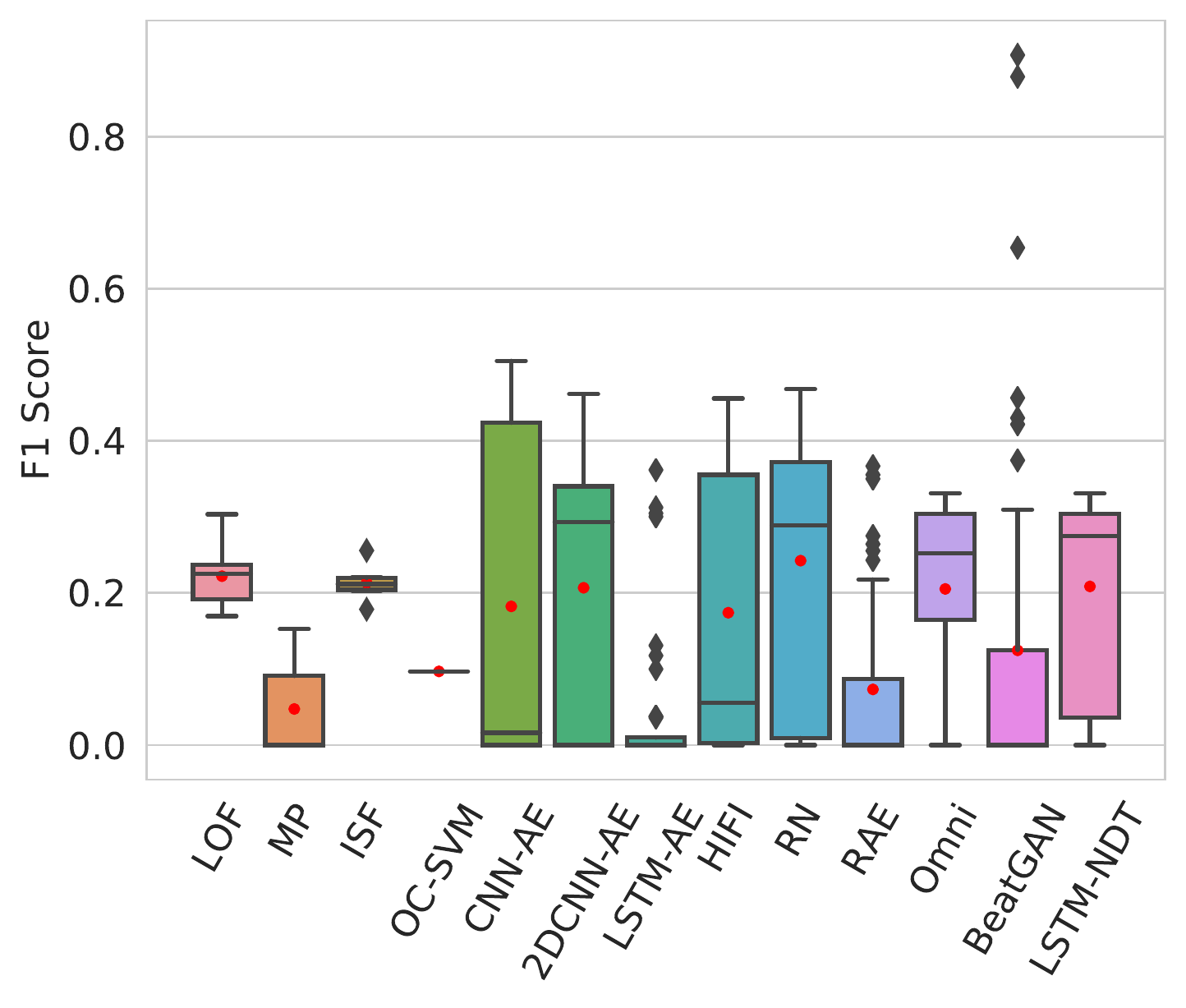}\label{fig:NAB-elb-stationary-f1}}
\subfigure[NAB-ambient-nonstationary] {\includegraphics[width=0.21
\textwidth]{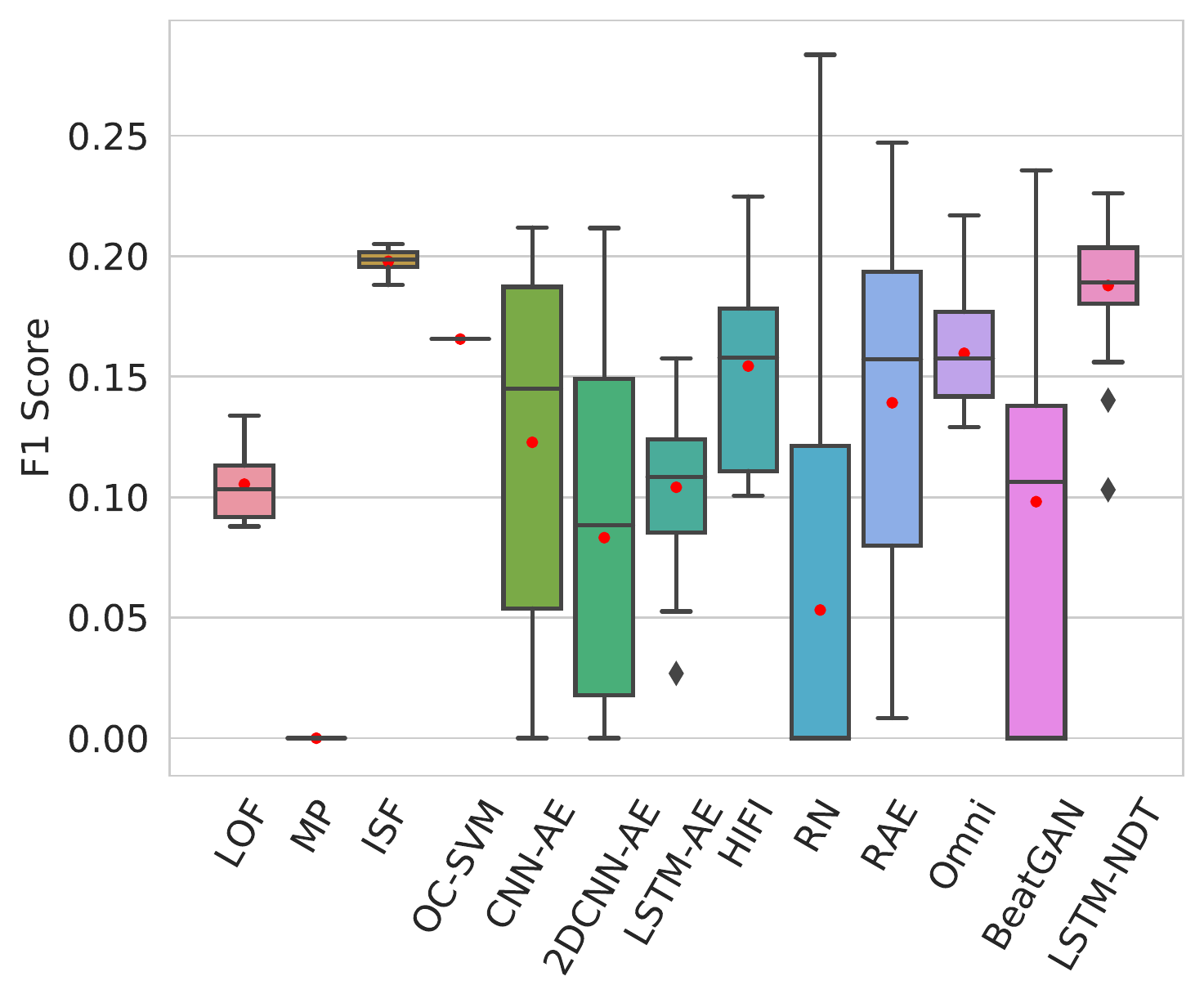}\label{fig:NAB-ambient-Nonstationary-f1}}
\subfigure[NAB-ec2-nonstationary] {\includegraphics[width=0.21
\textwidth]{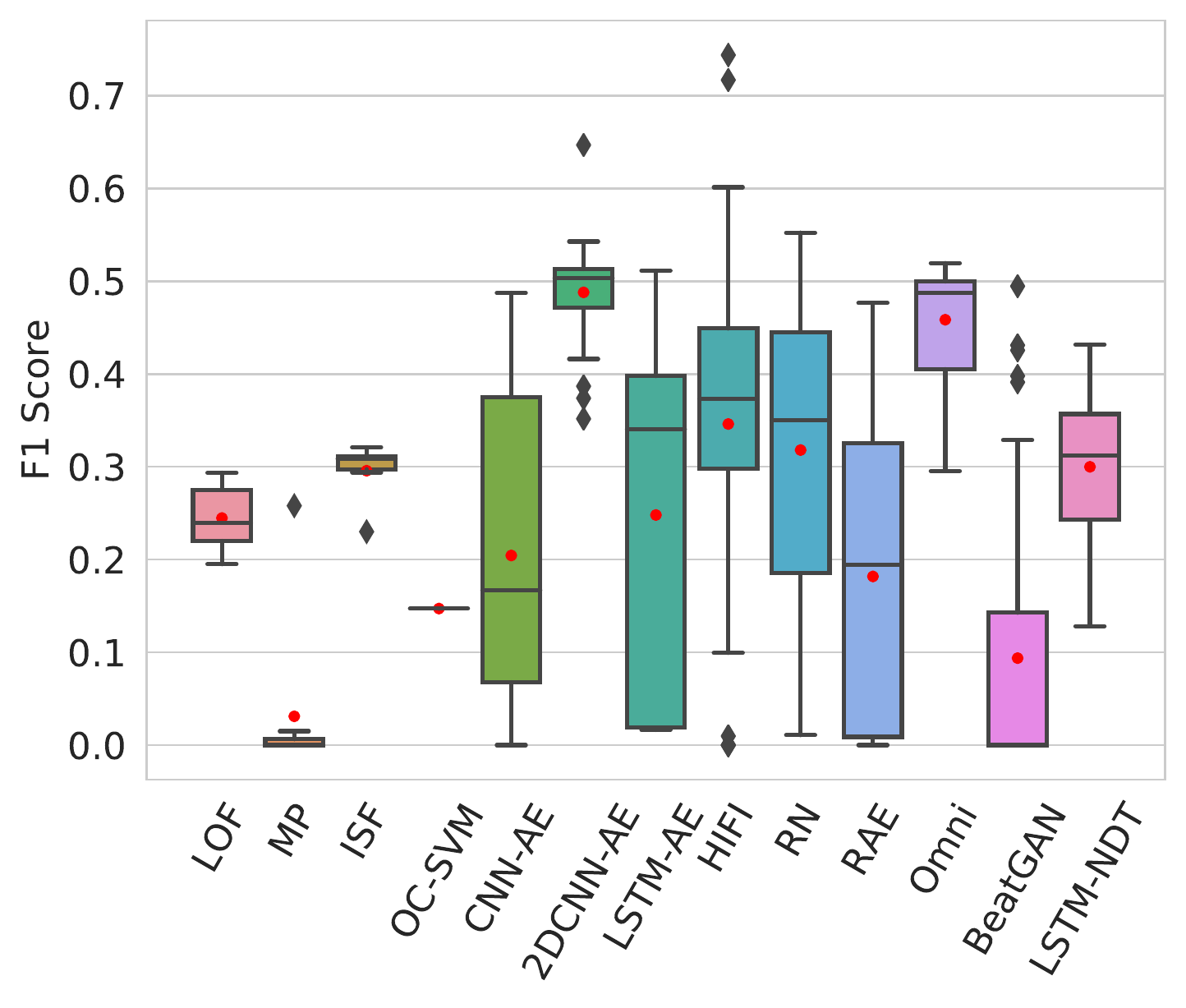}\label{fig:NAB-ec2-Nonstationary-f1}}
\subfigure[NAB-exchange-nonstationary] {\includegraphics[width=0.21
\textwidth]{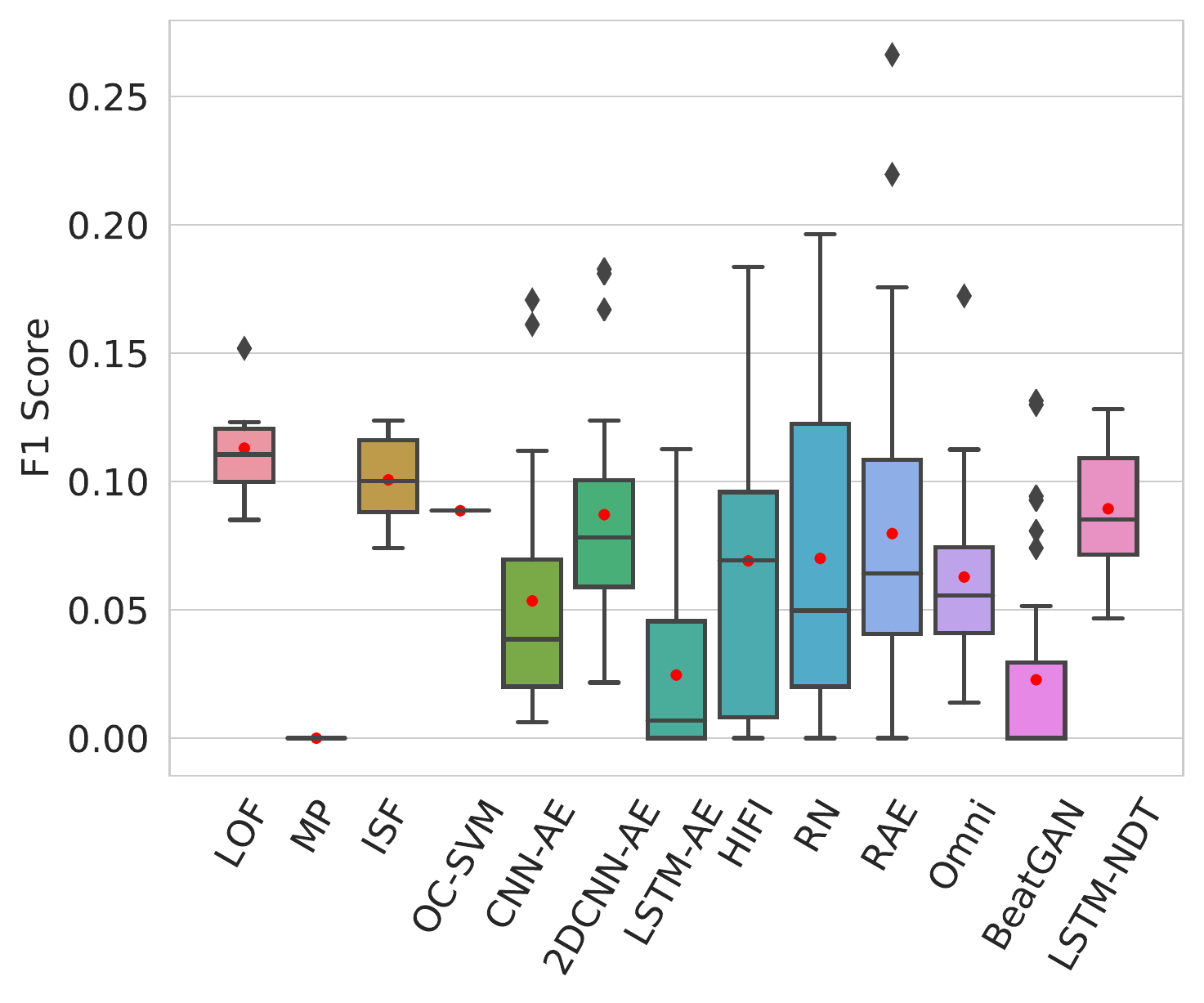}\label{fig:NAB-exchange-Nonstationary-f1}}
\subfigure[NAB-grok-nonstationary] {\includegraphics[width=0.21
\textwidth]{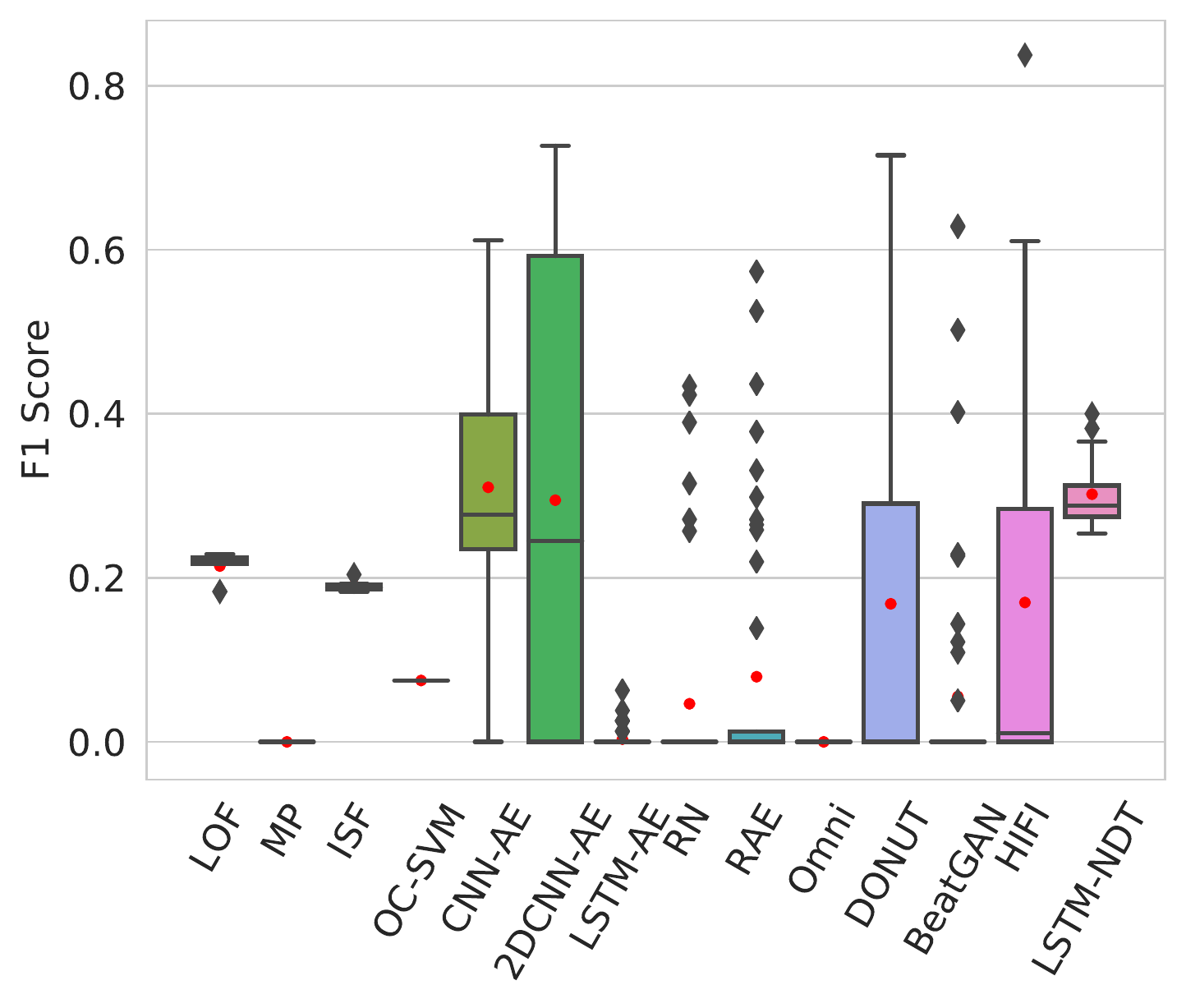}\label{fig:NAB-grok-Nonstationary-f1}}
\vskip -9pt
\caption{F1 Score (Stationary vs. Non-stationary)}
\vspace{-0.3cm}
\label{fig:NAB_f1}
\end{figure*}

\begin{figure*}
\centering
\subfigure[NAB-art-stationary] {\includegraphics[width=0.21
\textwidth]{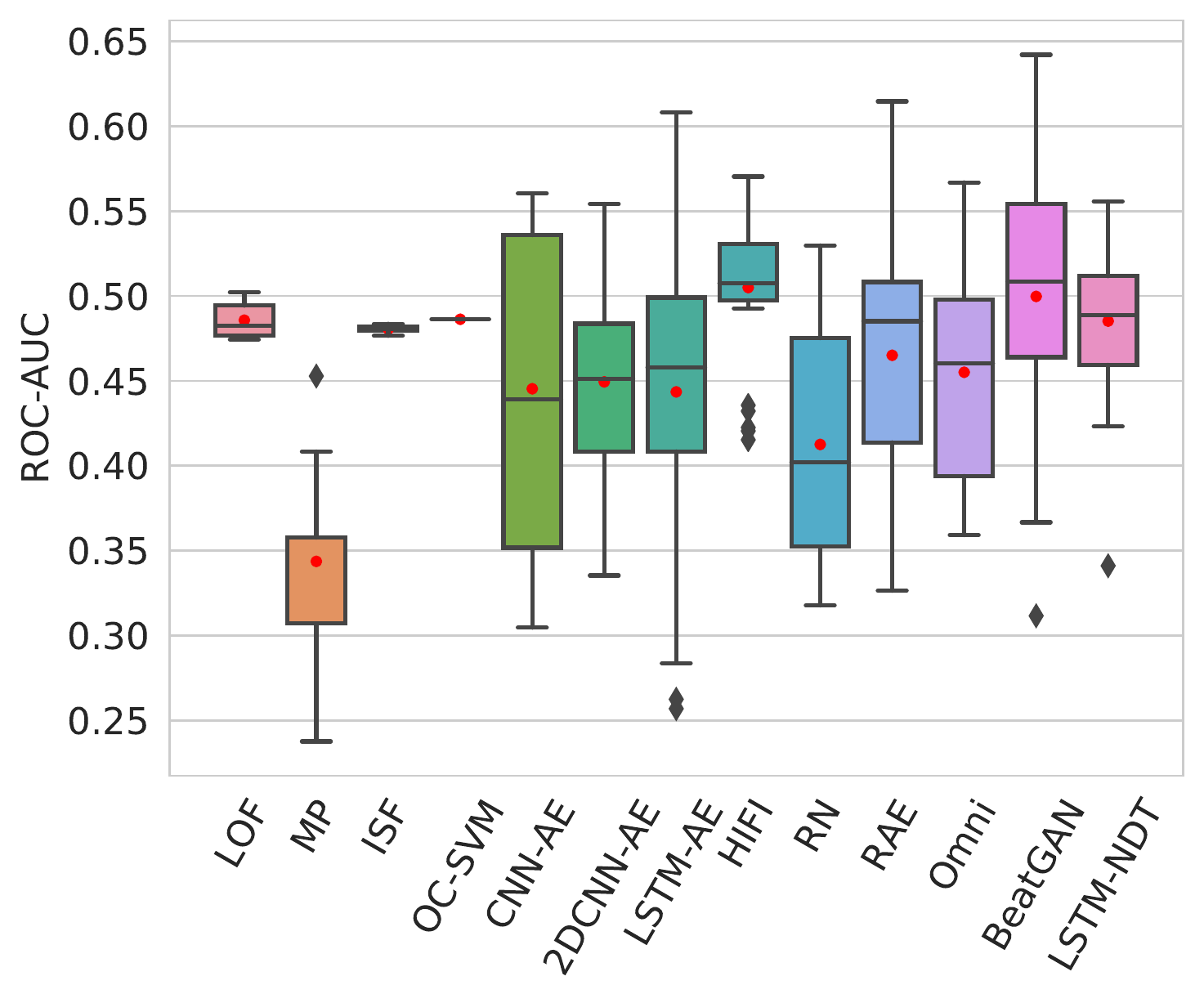}\label{fig:NAB-art-stationary-ROC}}
\subfigure[NAB-cpu-stationary] {\includegraphics[width=0.21
\textwidth]{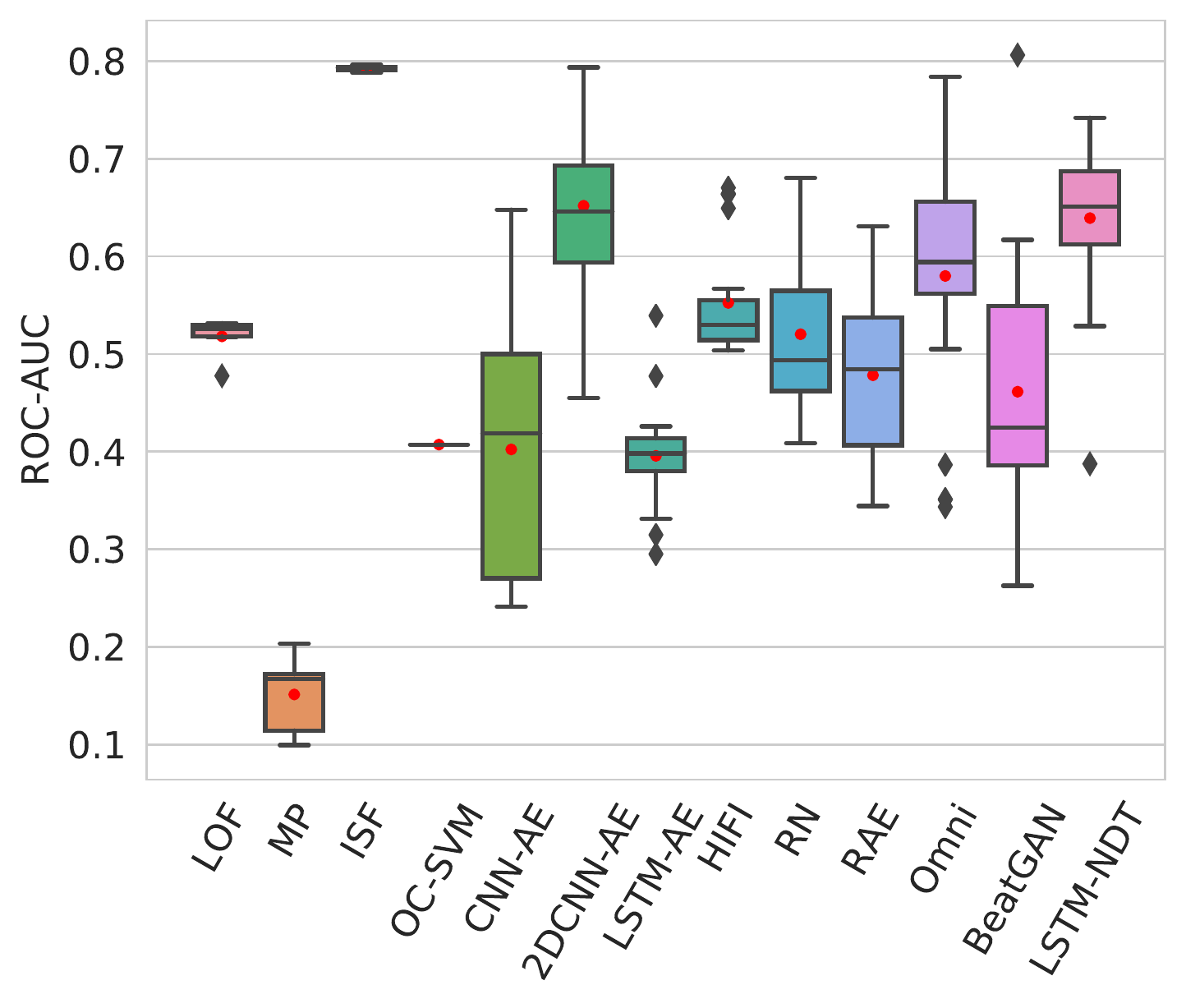}\label{fig:NAB-cpu-stationary-ROC}}
\subfigure[NAB-ec2-stationary] {\includegraphics[width=0.21
\textwidth]{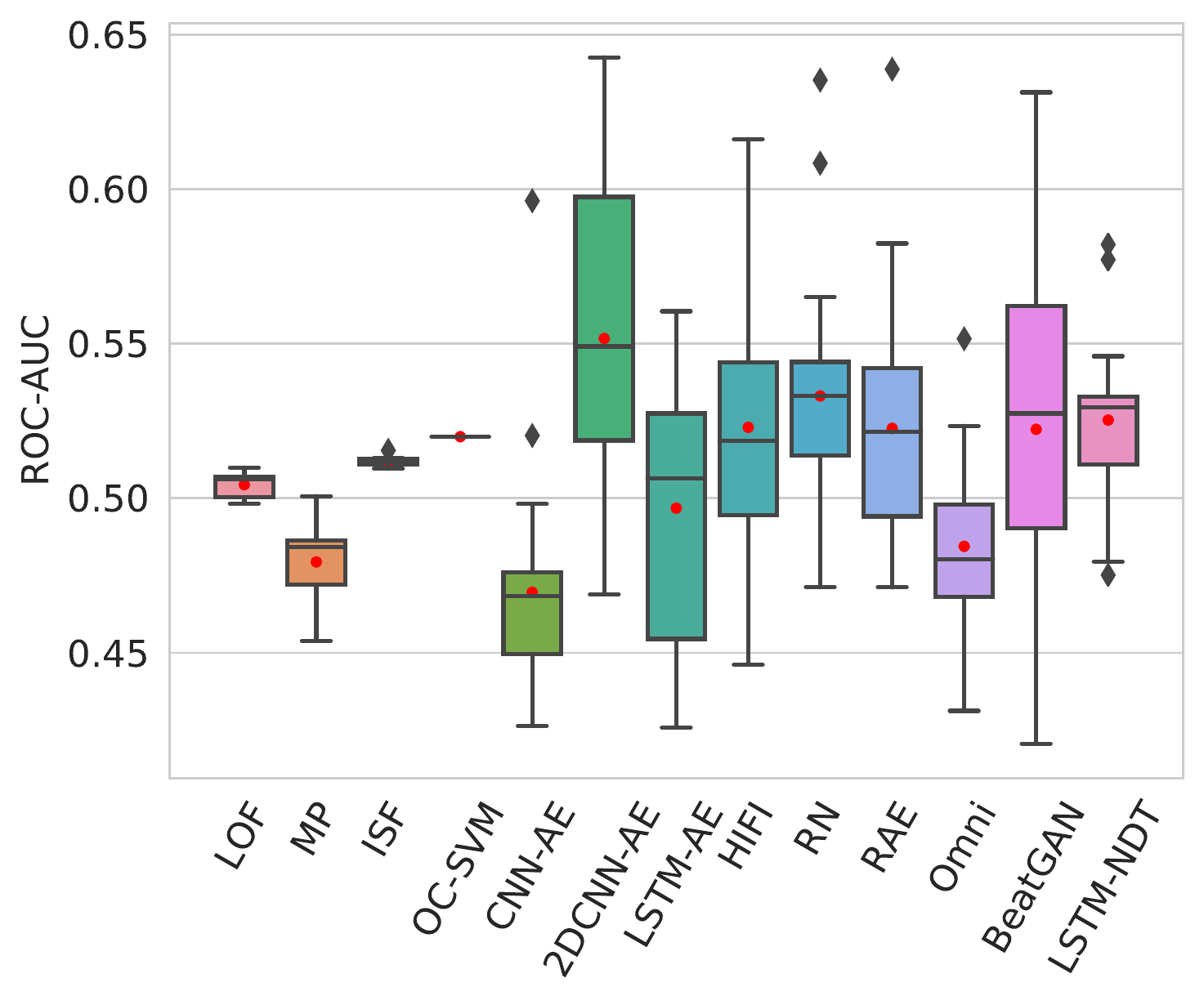}\label{fig:NAB-ec2-stationary-ROC}}
\subfigure[NAB-elb-stationary] {\includegraphics[width=0.21
\textwidth]{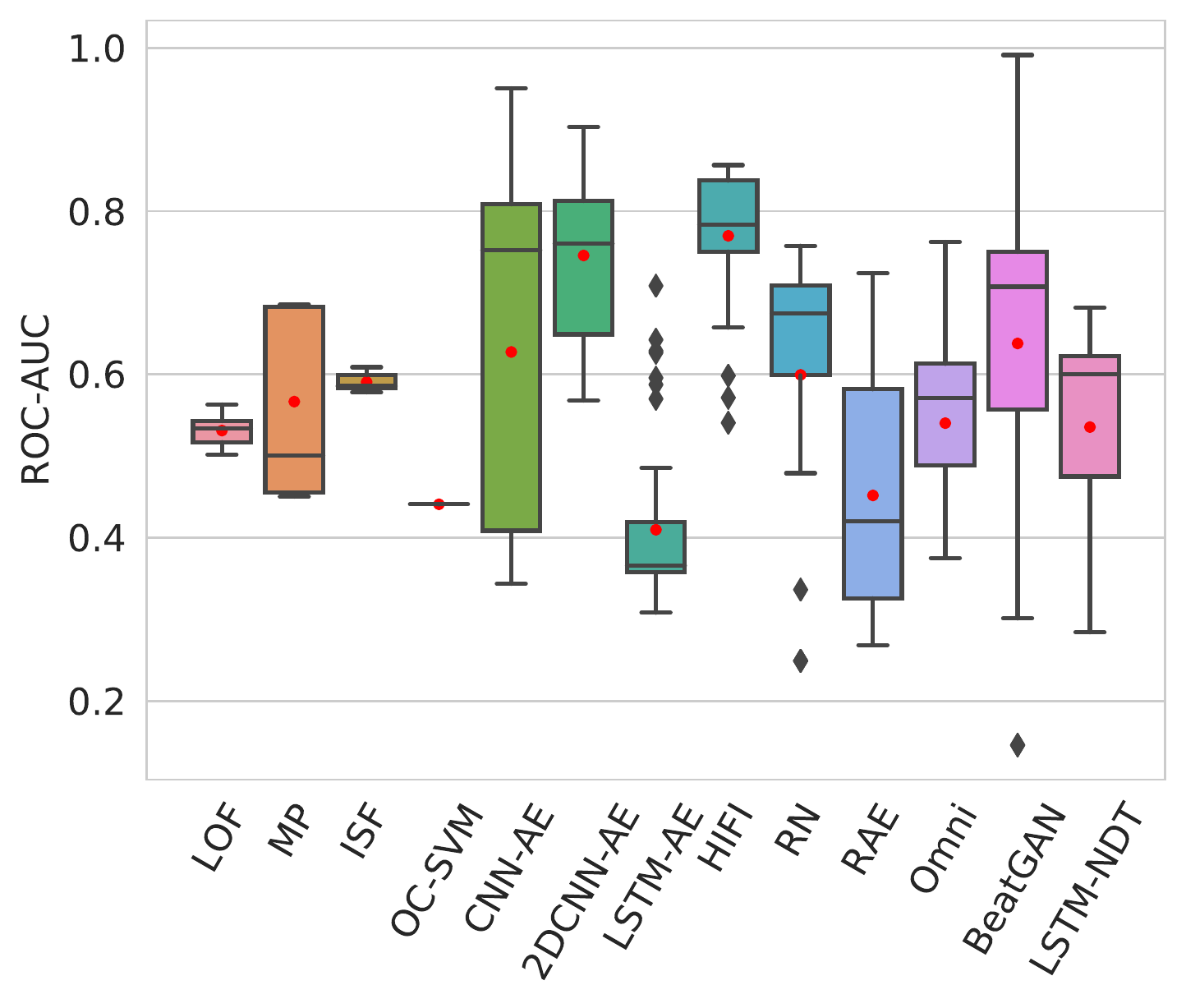}\label{fig:NAB-elb-stationary-ROC}}
\subfigure[NAB-ambient-nonstationary] {\includegraphics[width=0.21
\textwidth]{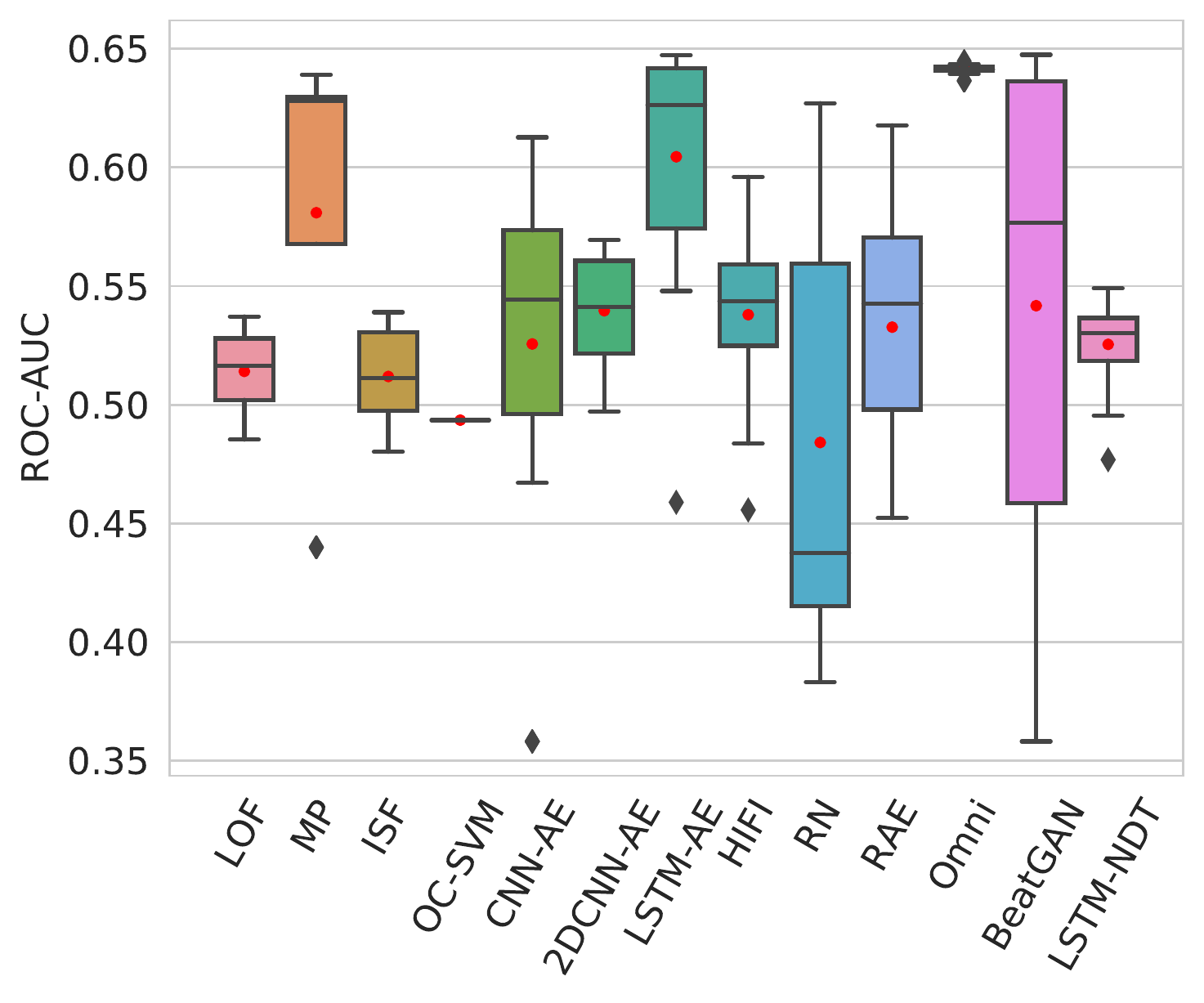}\label{fig:NAB-ambient-Nonstationary-ROC}}
\subfigure[NAB-ec2-nonstationary] {\includegraphics[width=0.21
\textwidth]{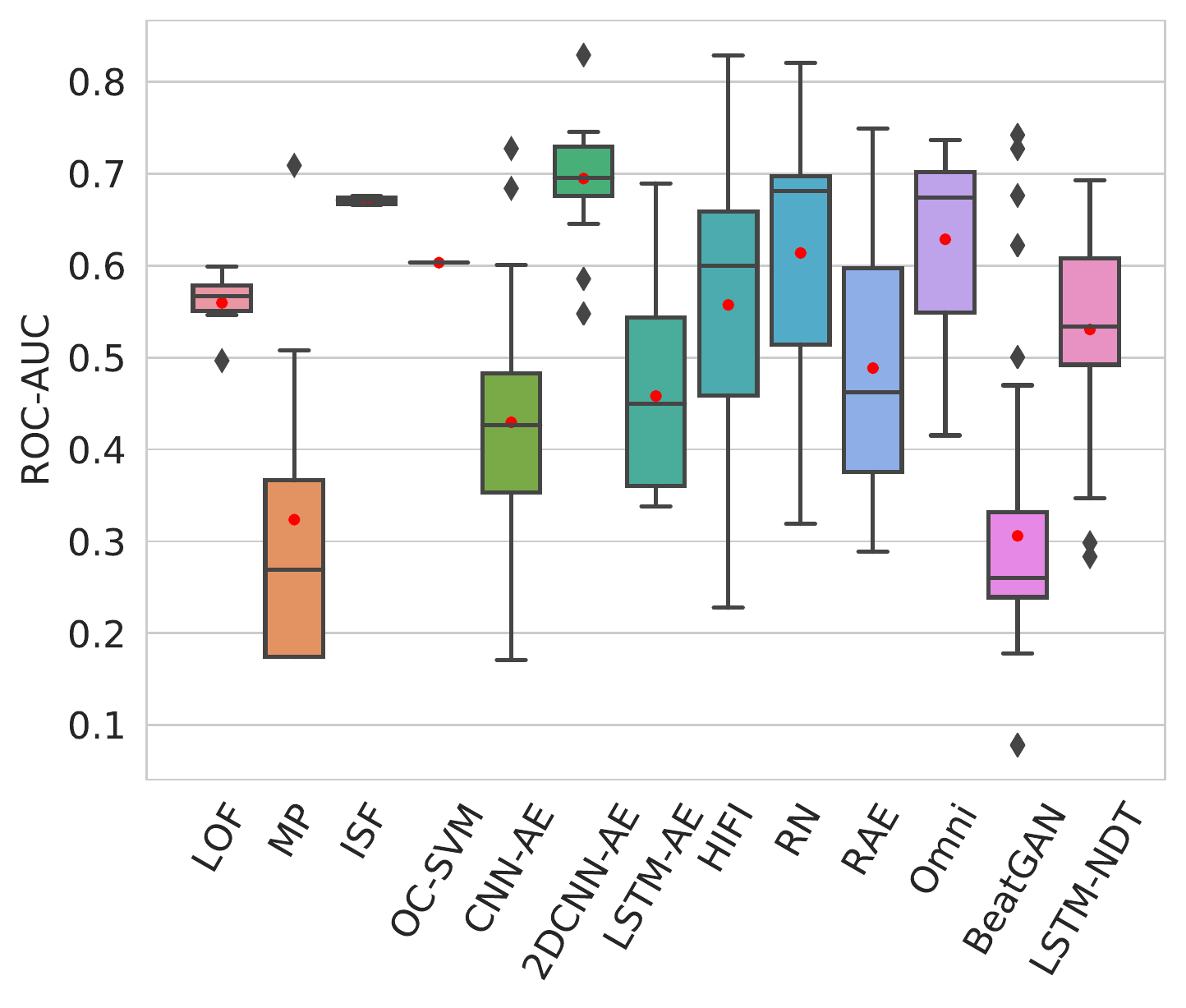}\label{fig:NAB-ec2-Nonstationary-ROC}}
\subfigure[NAB-exchange-nonstationary] {\includegraphics[width=0.21
\textwidth]{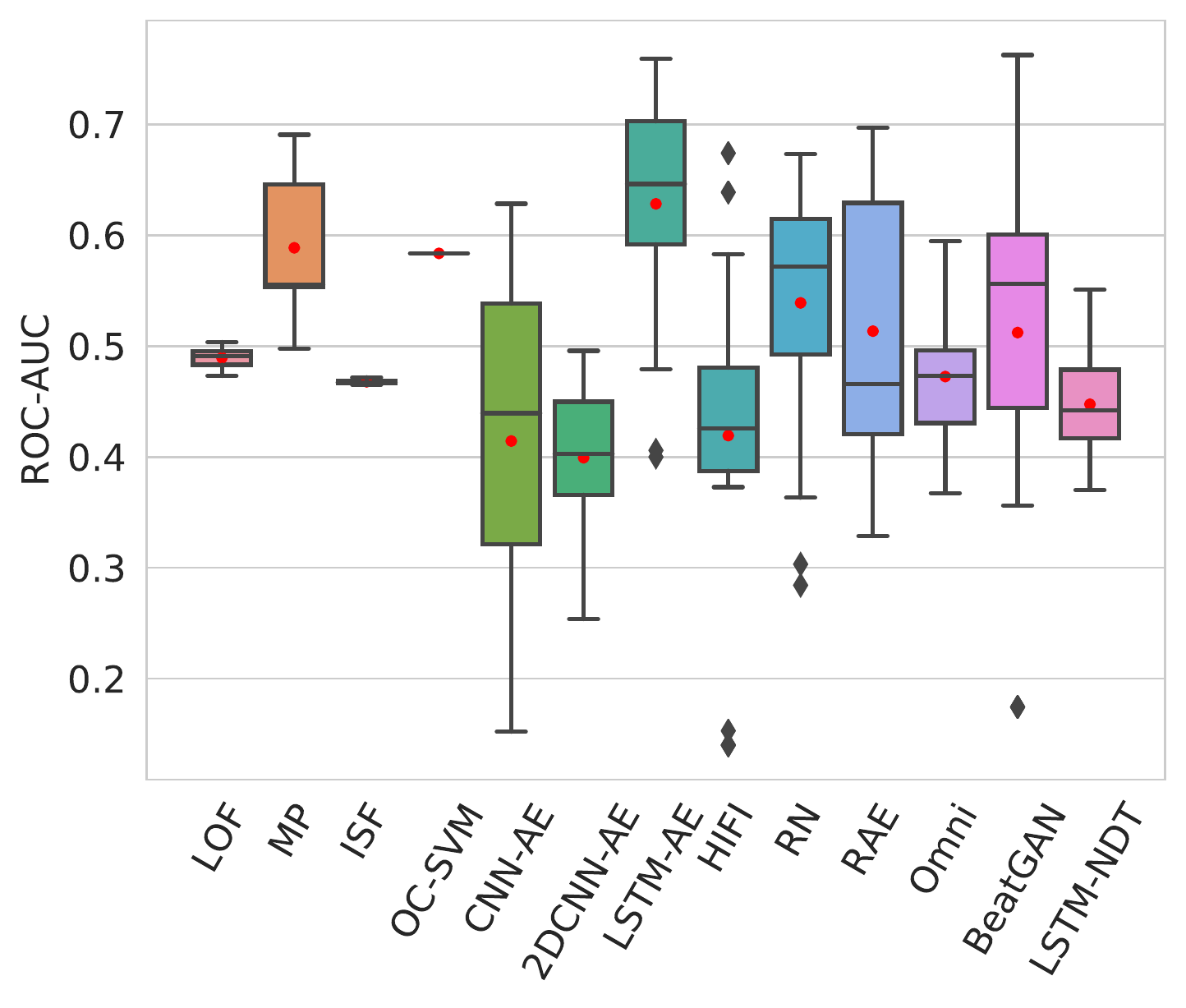}\label{fig:NAB-exchange-Nonstationary-ROC}}
\subfigure[NAB-grok-nonstationary] {\includegraphics[width=0.21
\textwidth]{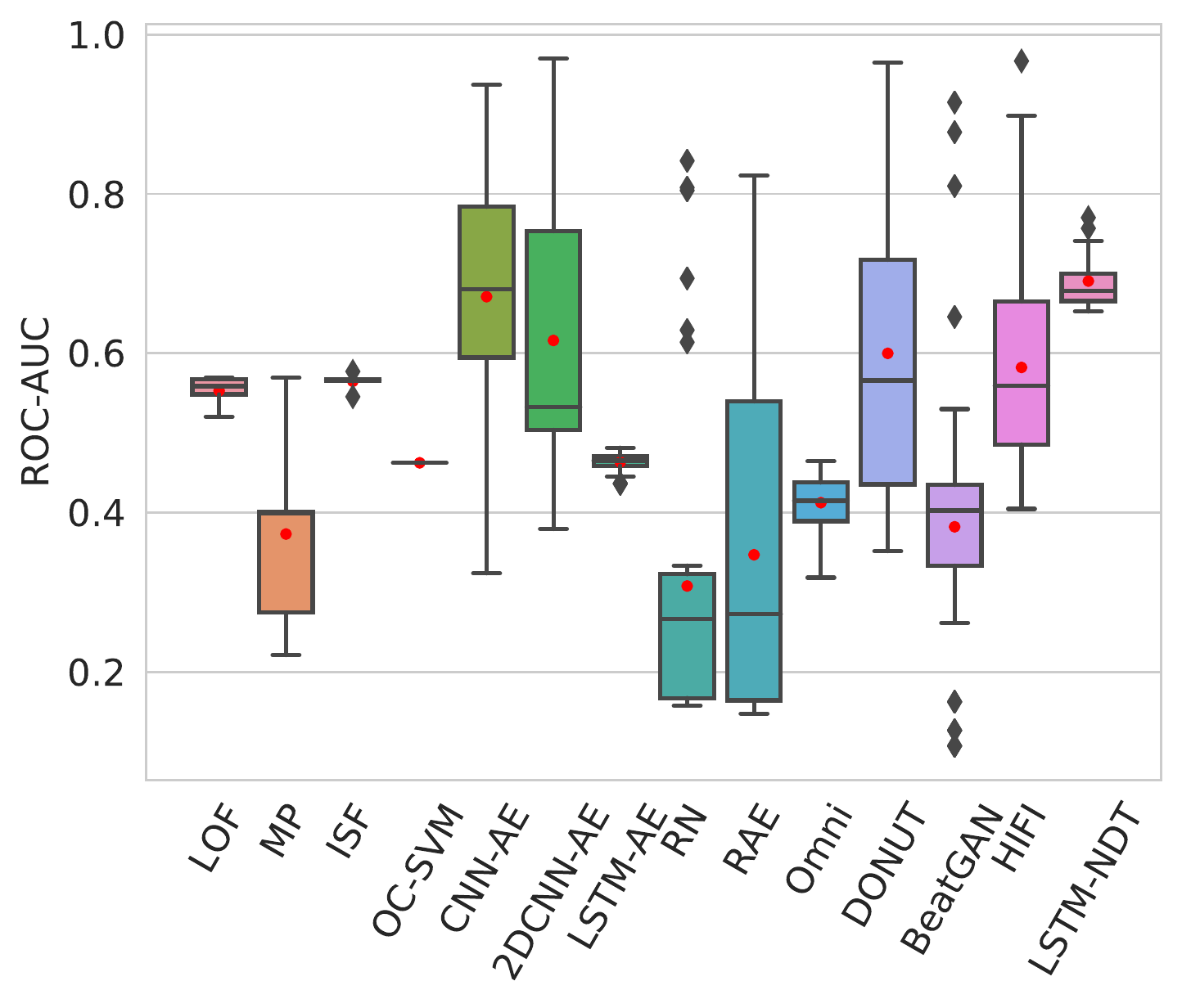}\label{fig:NAB-grok-Nonstationary-ROC}}
\vskip -9pt
\caption{ROC-AUC (Stationary vs. Non-stationary)}
\vspace{-0.5cm}
\label{fig:NAB_roc}
\end{figure*}

\begin{figure*}
\centering
\subfigure[NAB-art-stationary] {\includegraphics[width=0.21
\textwidth]{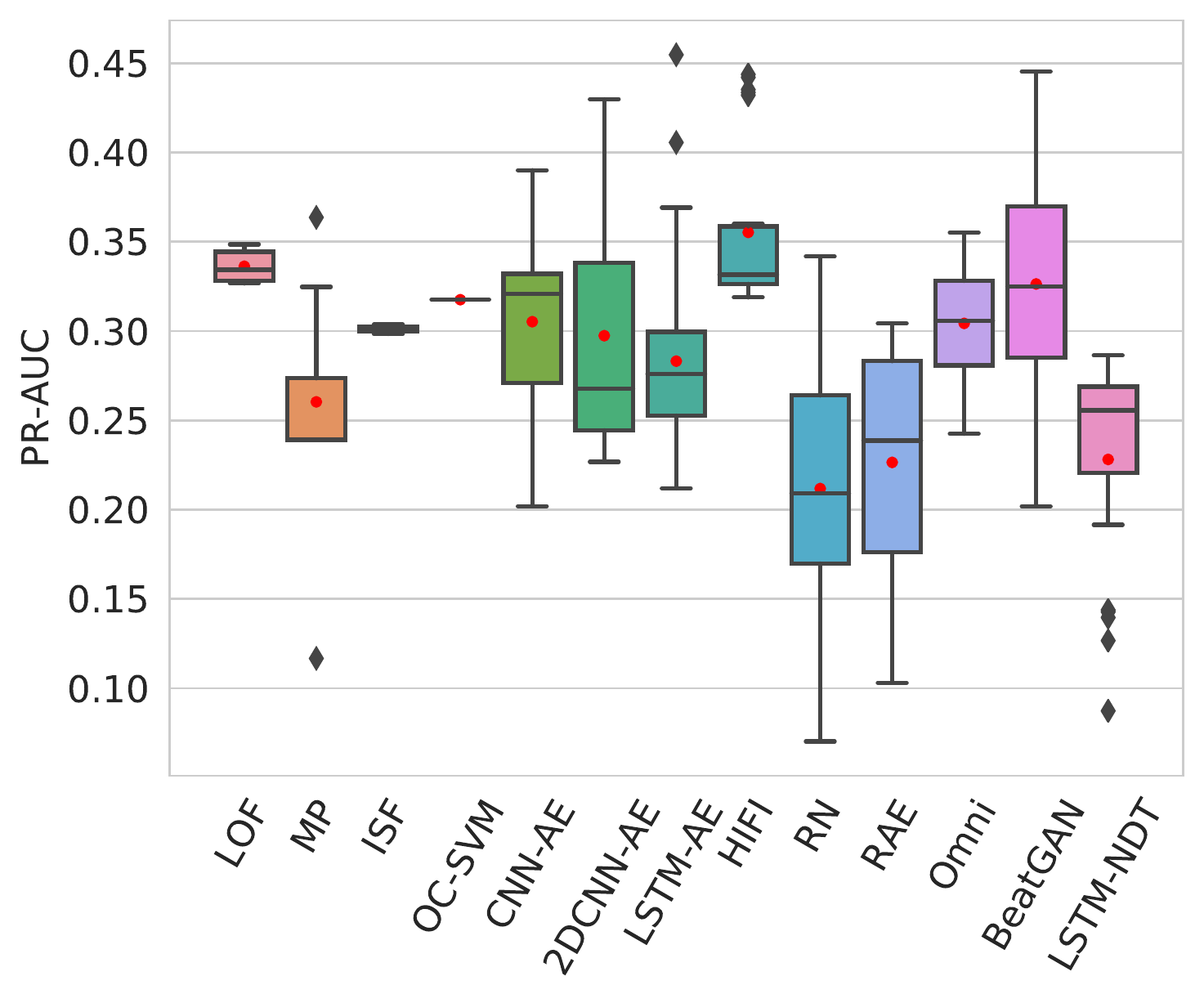}\label{fig:NAB-art-stationary-PR}}
\subfigure[NAB-cpu-stationary] {\includegraphics[width=0.21
\textwidth]{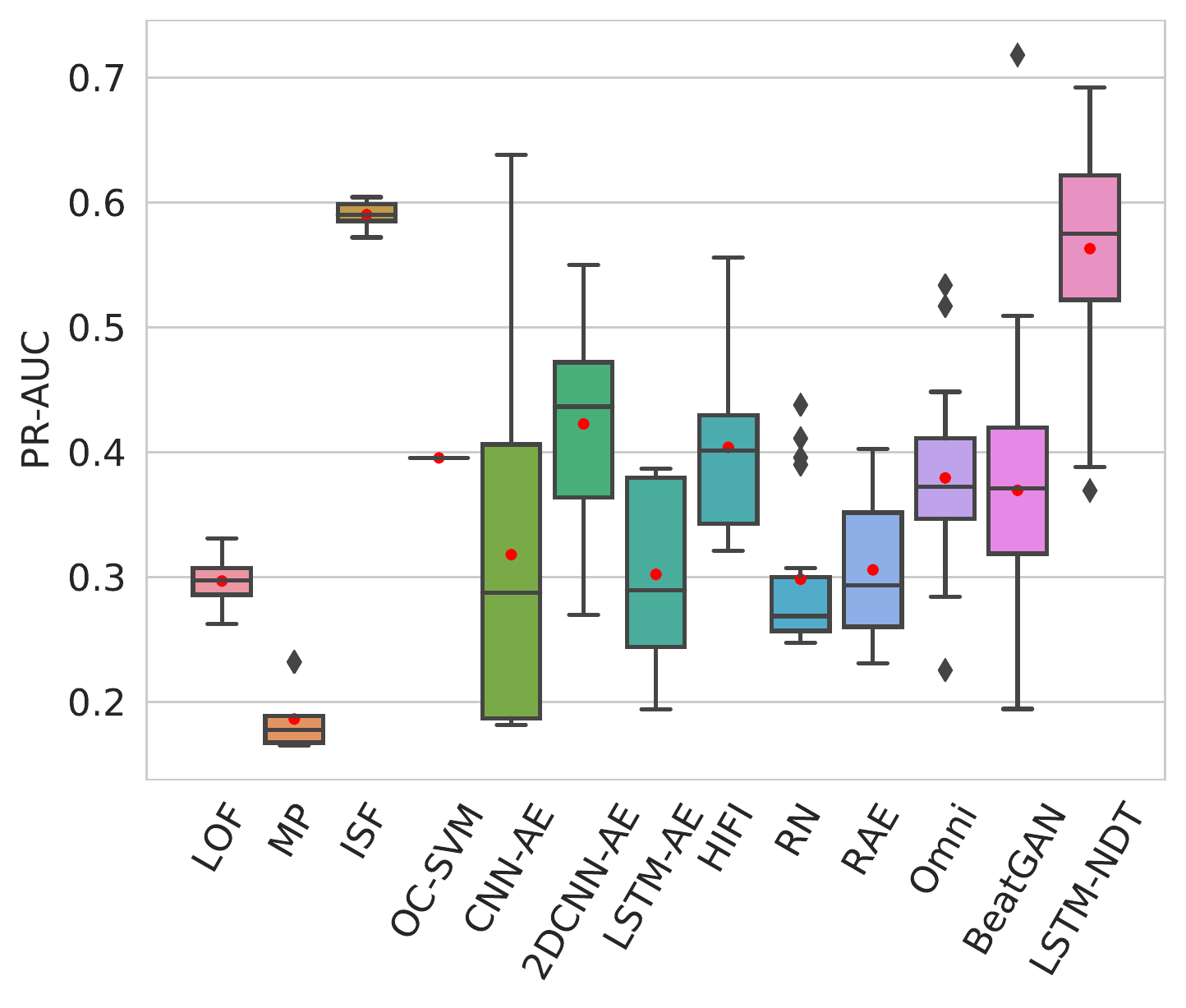}\label{fig:NAB-cpu-stationary-PR}}
\subfigure[NAB-ec2-stationary] {\includegraphics[width=0.21
\textwidth]{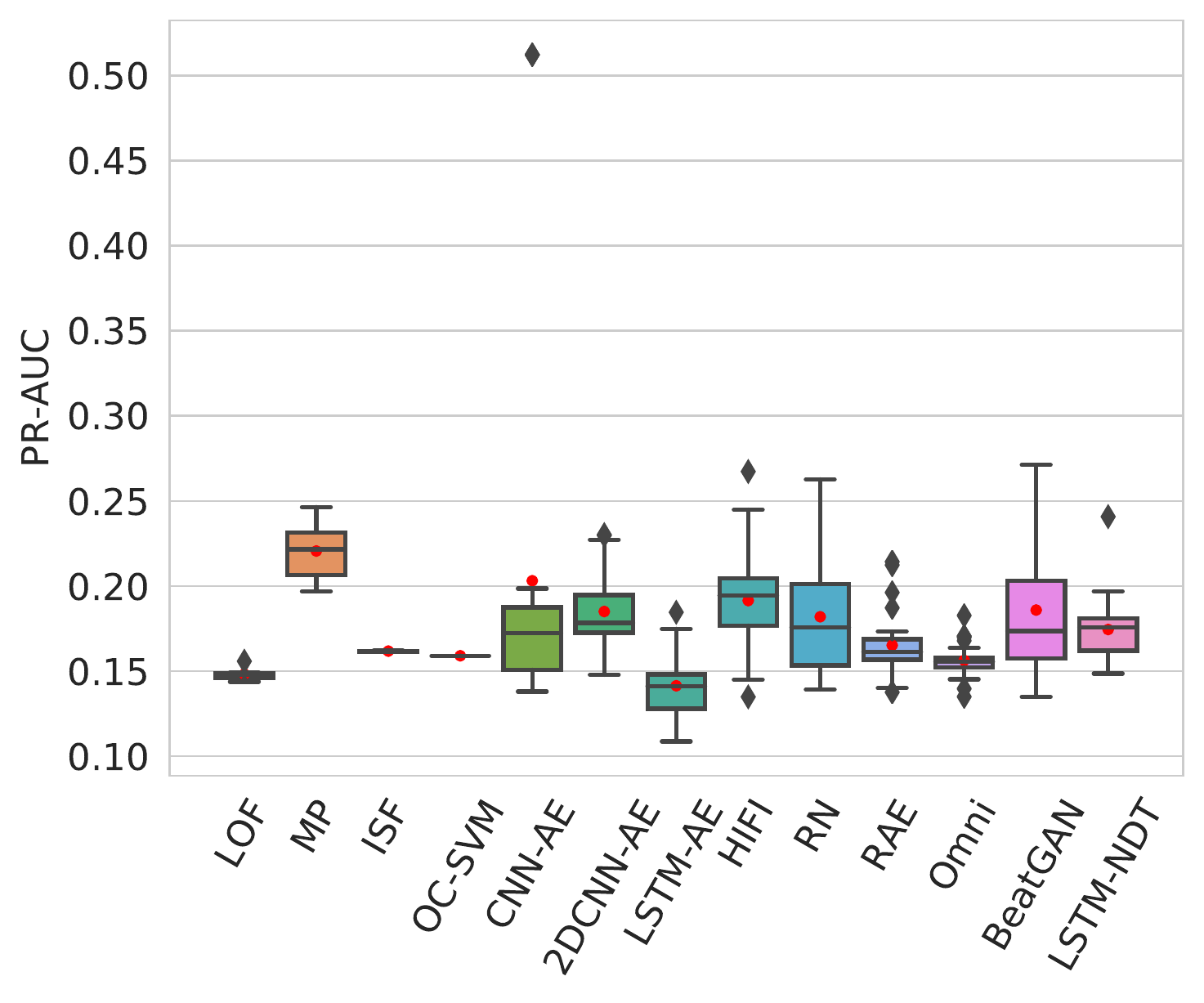}\label{fig:NAB-ec2-stationary-PR}}
\subfigure[NAB-elb-stationary] {\includegraphics[width=0.21
\textwidth]{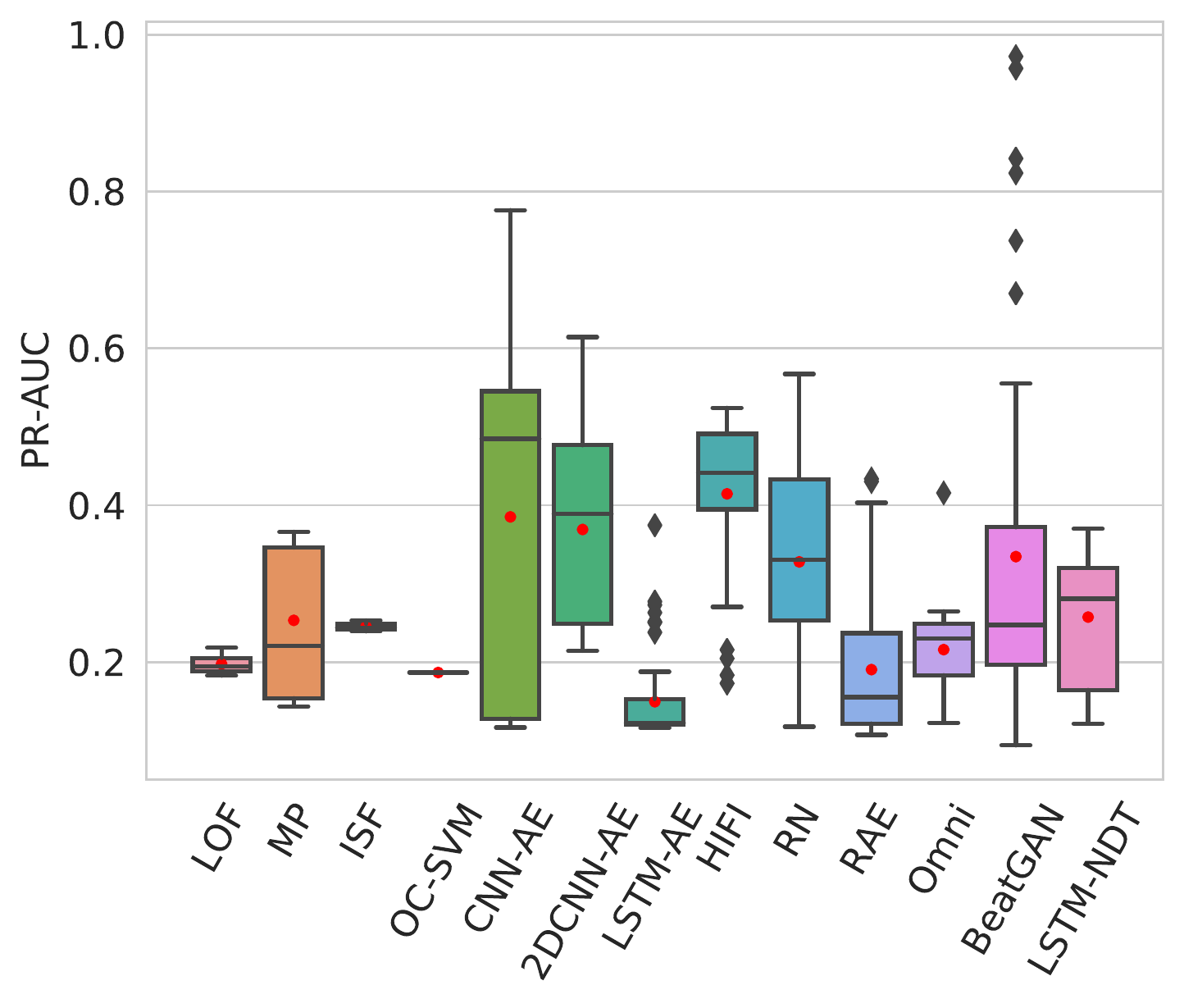}\label{fig:NAB-elb-stationary-PR}}
\subfigure[NAB-ambient-nonstationary] {\includegraphics[width=0.21
\textwidth]{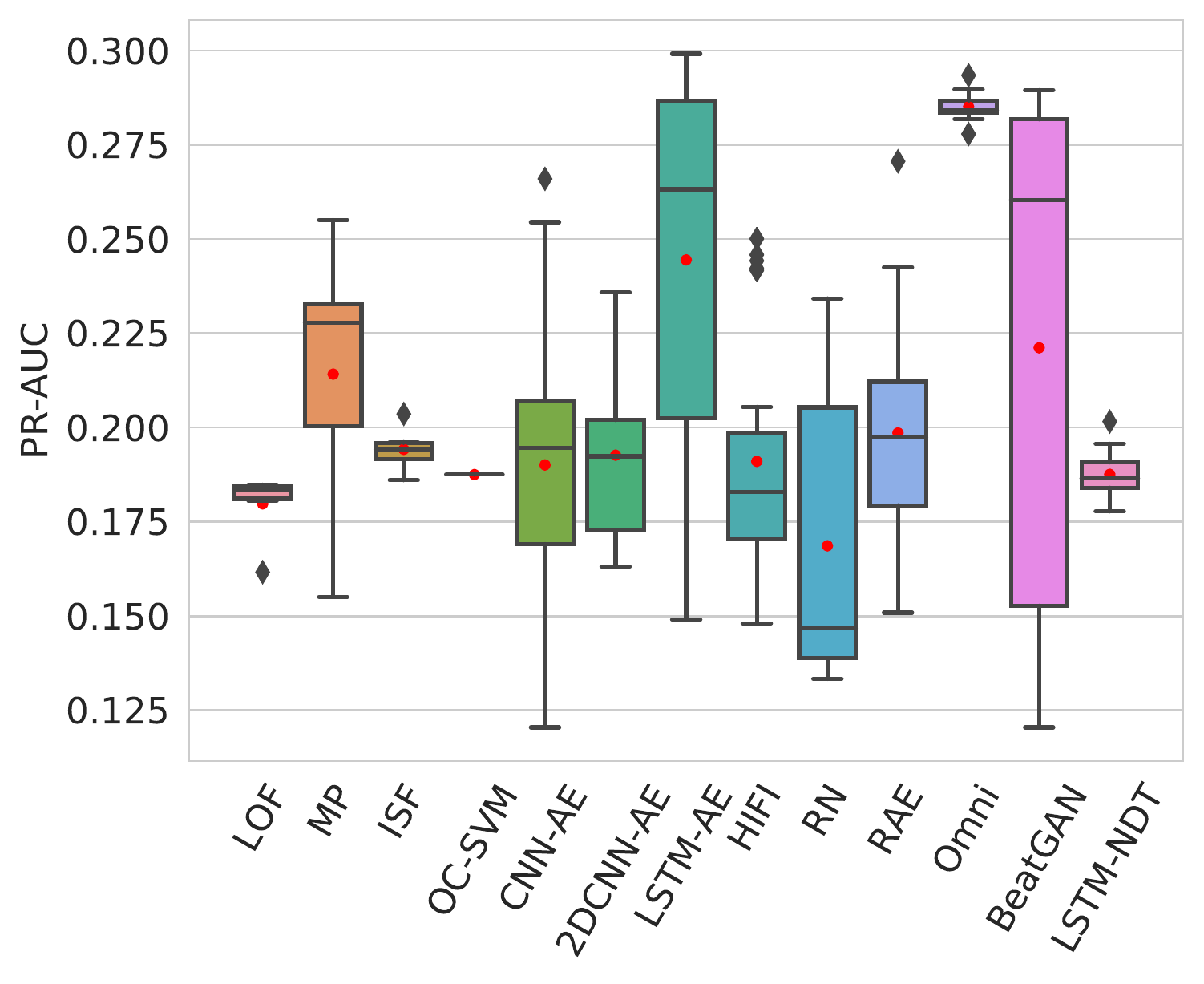}\label{fig:NAB-ambient-Nonstationary-PR}}
\subfigure[NAB-ec2-nonstationary] {\includegraphics[width=0.21
\textwidth]{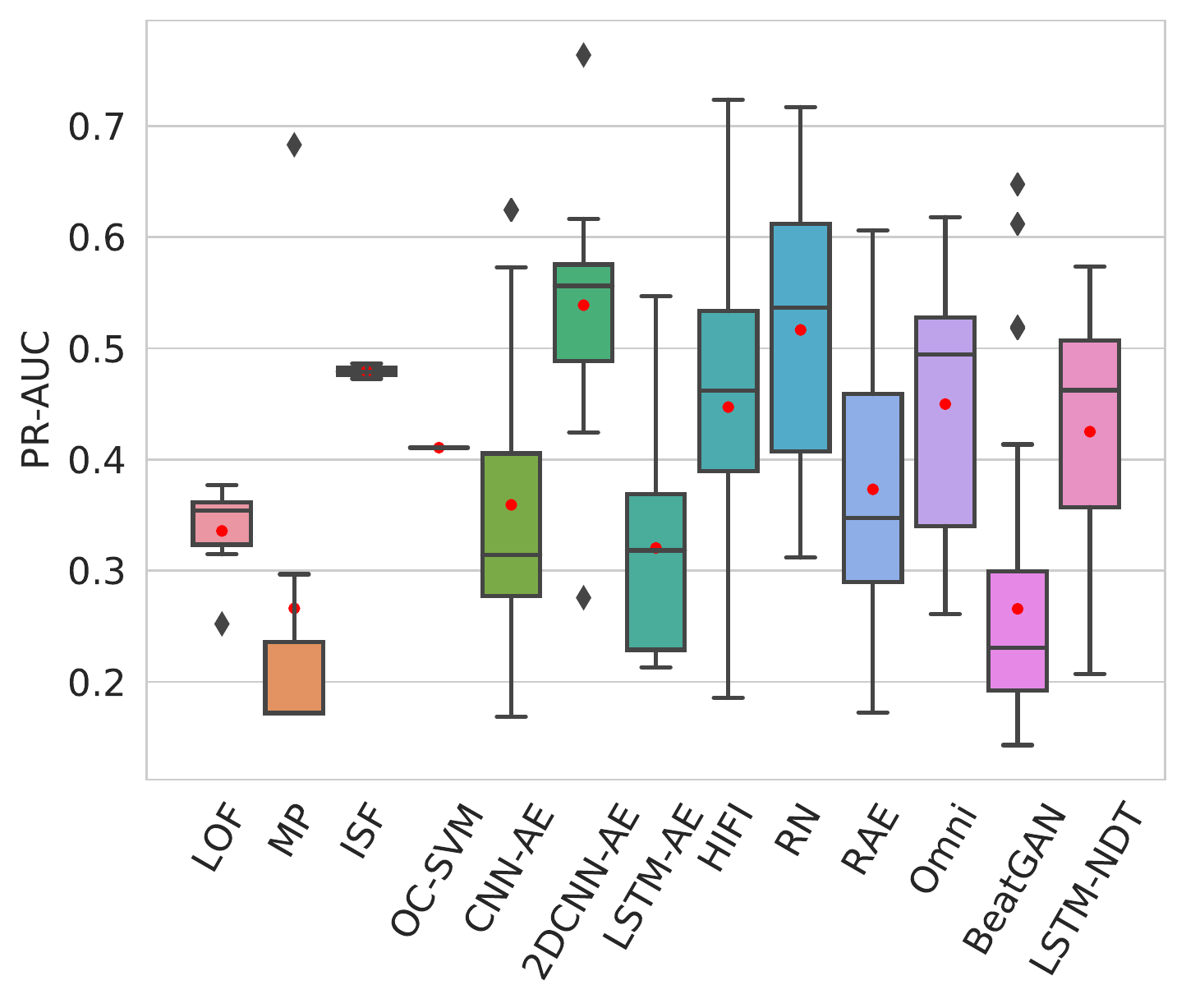}\label{fig:NAB-ec2-Nonstationary-PR}}
\subfigure[NAB-exchange-nonstationary] {\includegraphics[width=0.21
\textwidth]{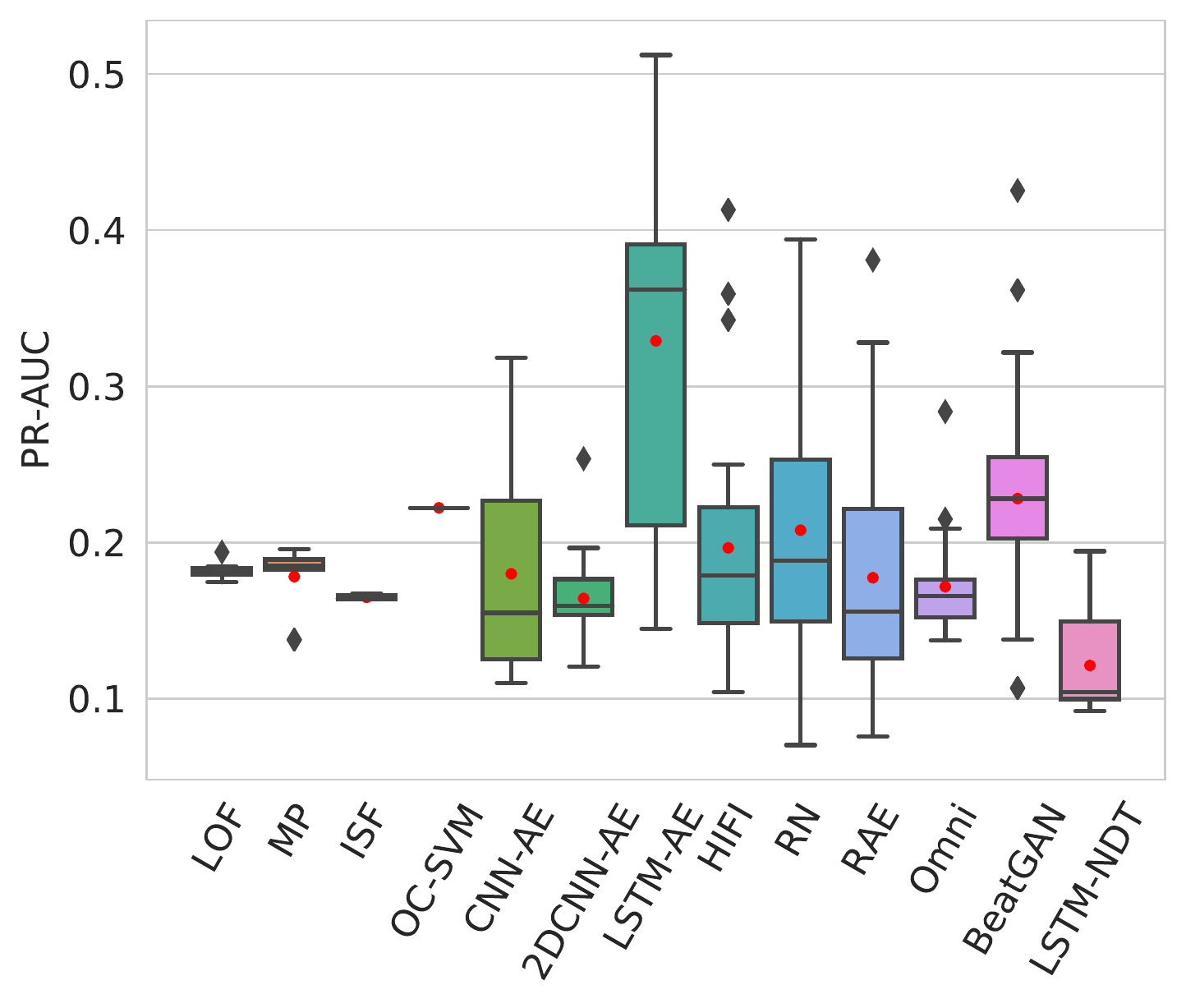}\label{fig:NAB-exchange-Nonstationary-PR}}
\subfigure[NAB-grok-nonstationary] {\includegraphics[width=0.21
\textwidth]{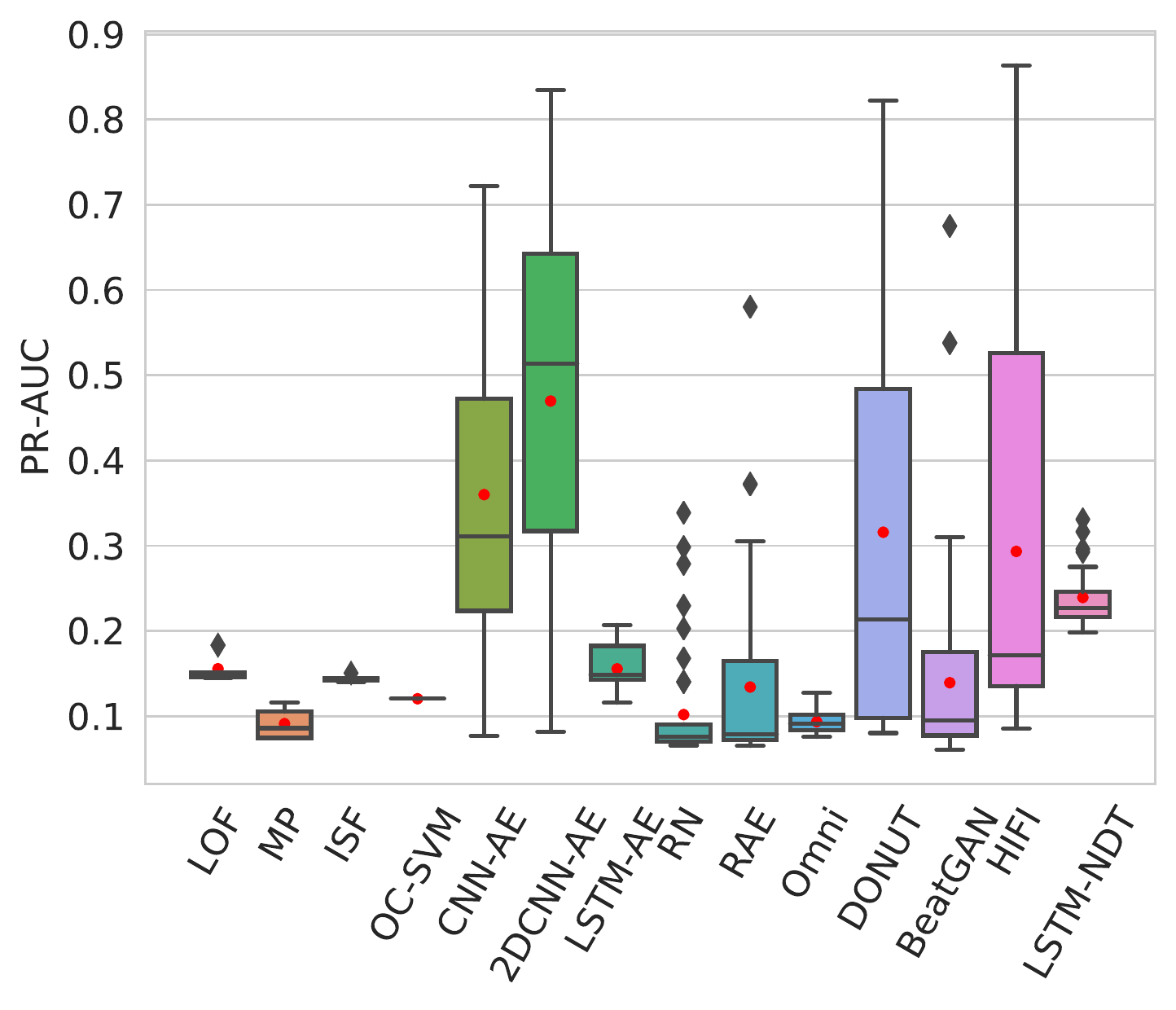}\label{fig:NAB-grok-Nonstationary-PR}}
\vskip -9pt
\caption{PR-AUC (Stationary vs. Non-stationary)}
\label{fig:NAB_pr}
\end{figure*}

\begin{figure*}
\centering
\subfigure[Precision (NAB)] {\includegraphics[width=0.21\textwidth]{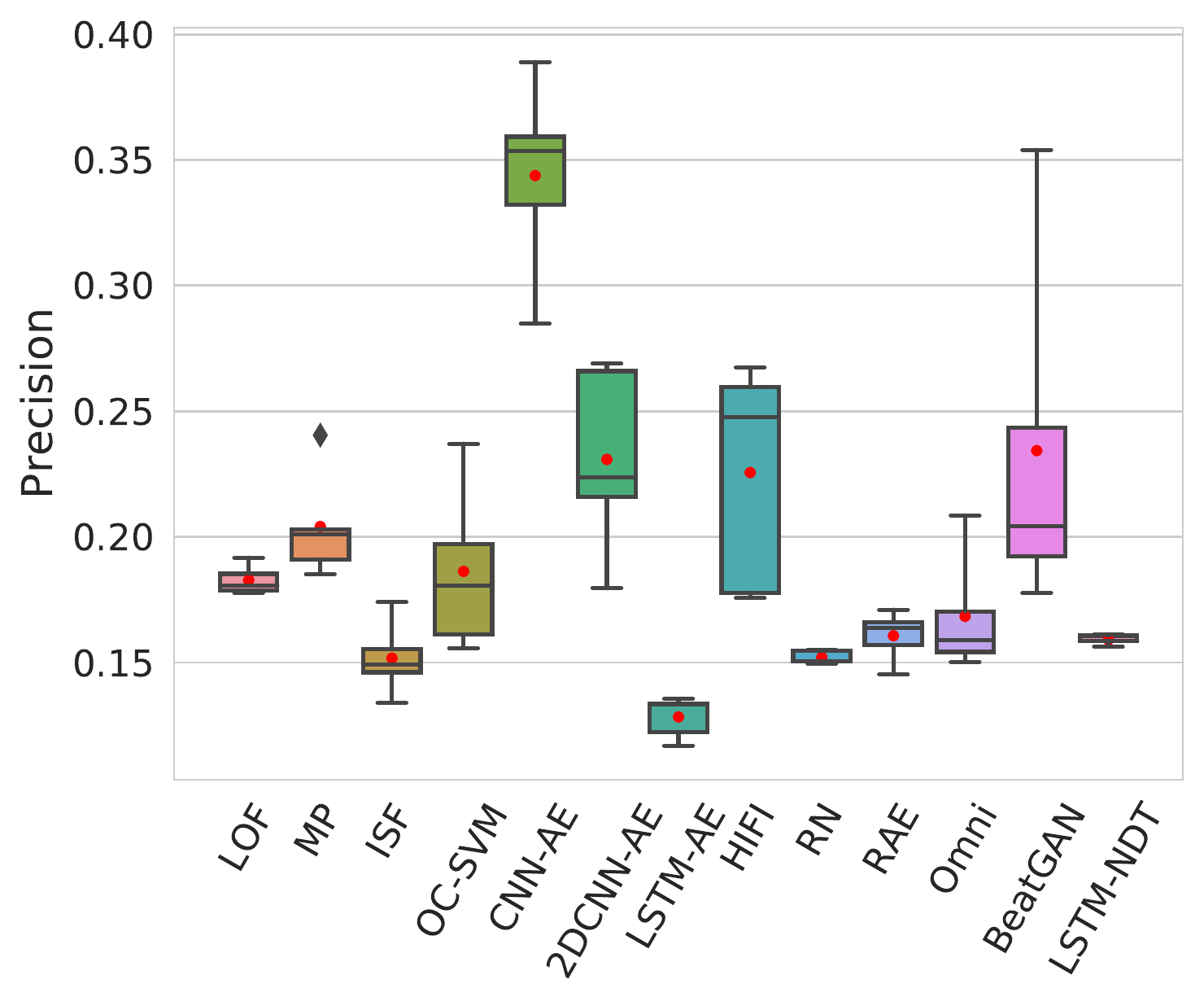}\label{fig:noise_NAB_Pre}}
\subfigure[Precision (MSL)] {\includegraphics[width=0.21\textwidth]{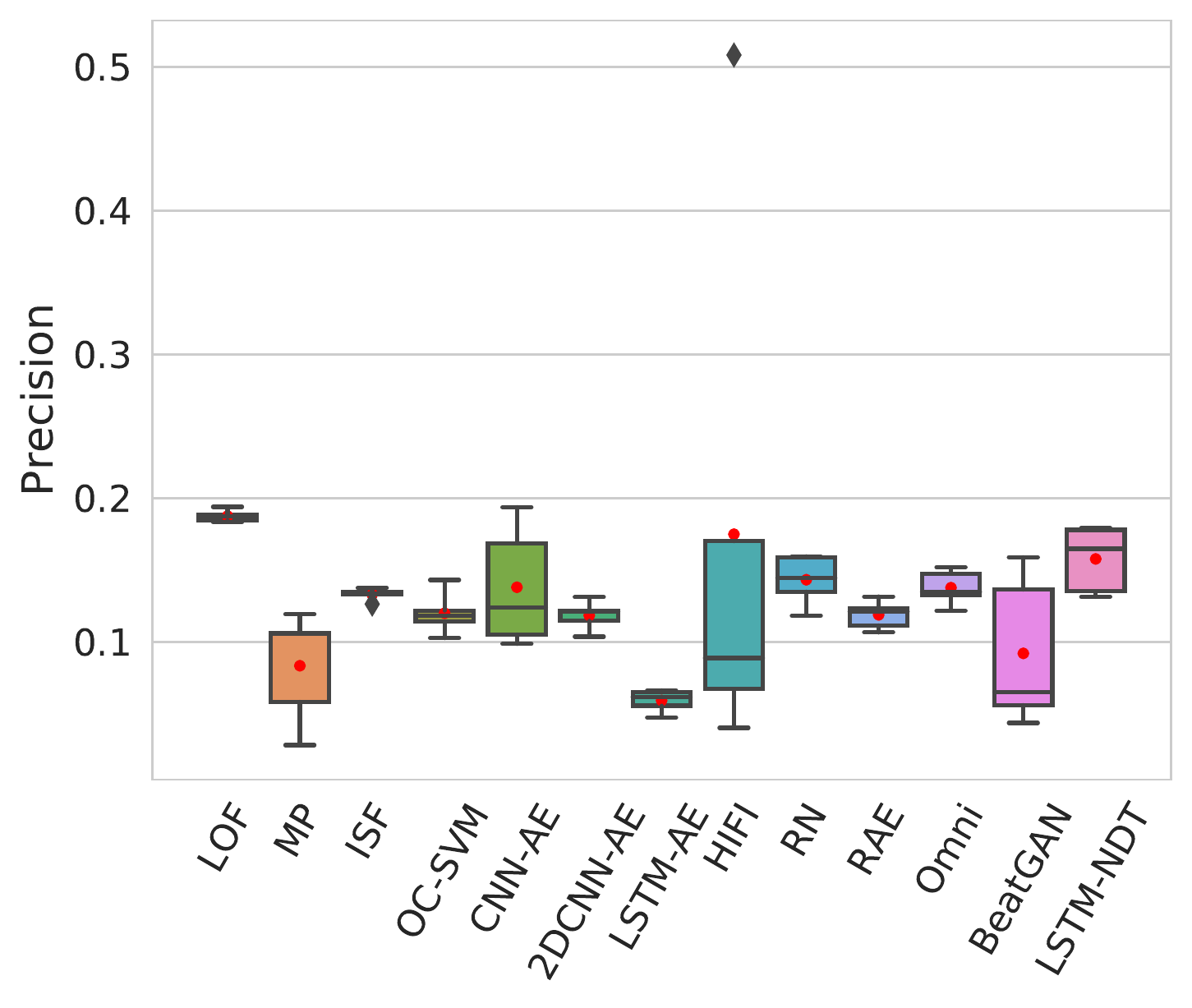}\label{fig:noise_MSL_Pre}}
\subfigure[Recall (NAB)] {\includegraphics[width=0.21\textwidth]{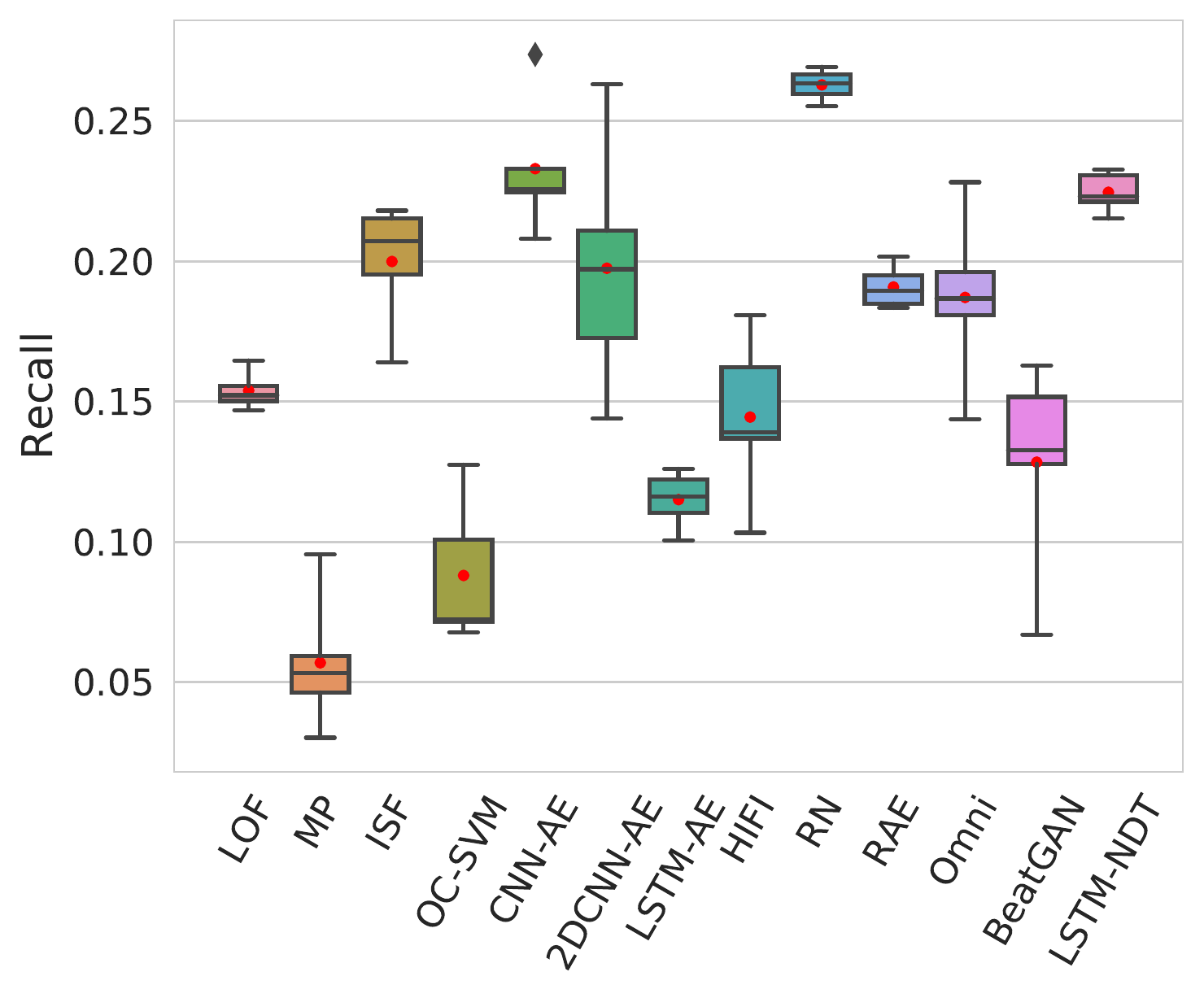}\label{fig:noise_NAB_Rec}}
\subfigure[Recall  (MSL)] {\includegraphics[width=0.21\textwidth]{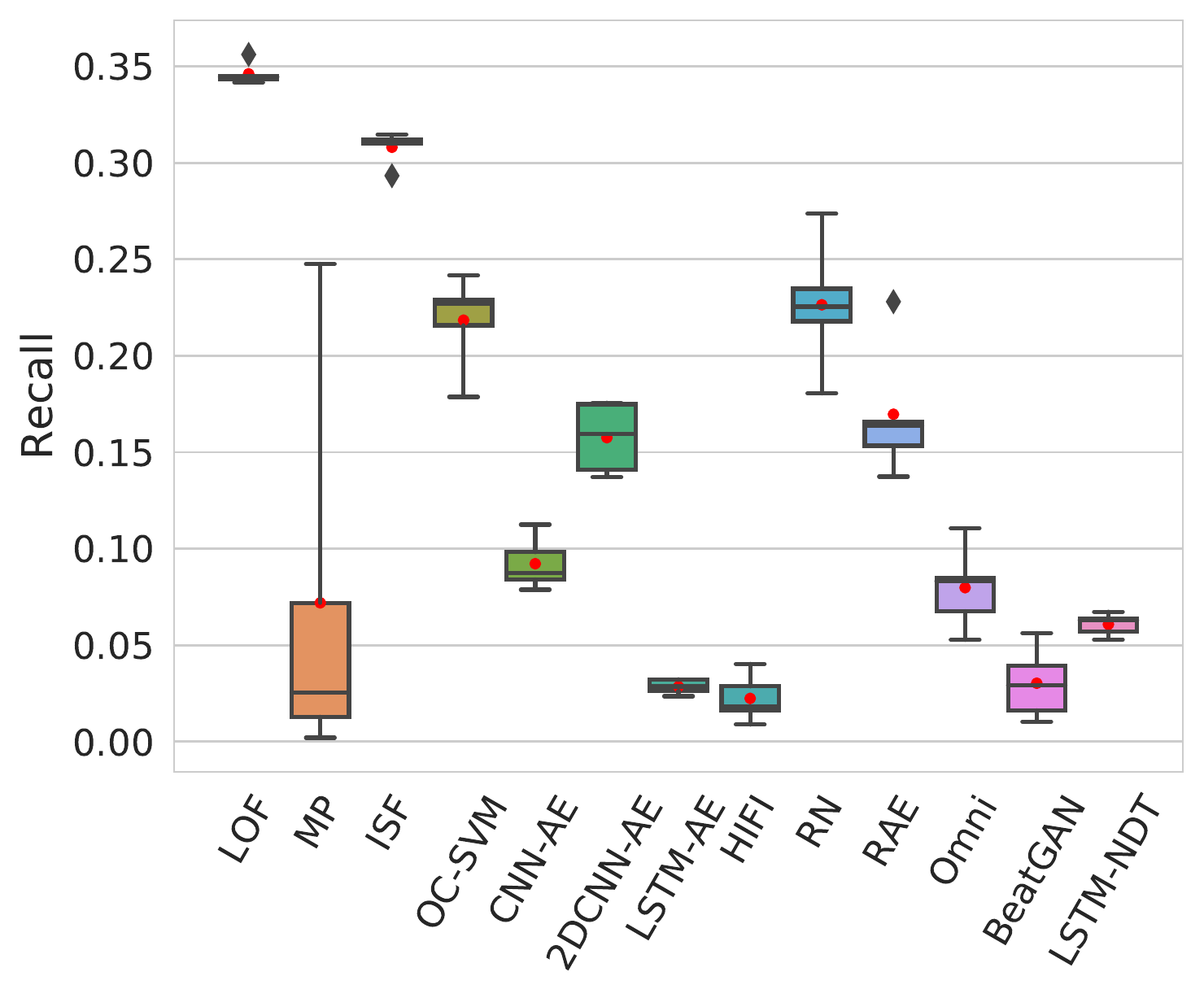}\label{fig:noise_MSL_Rec}}
\subfigure[F1 Score (NAB)] {\includegraphics[width=0.21\textwidth]{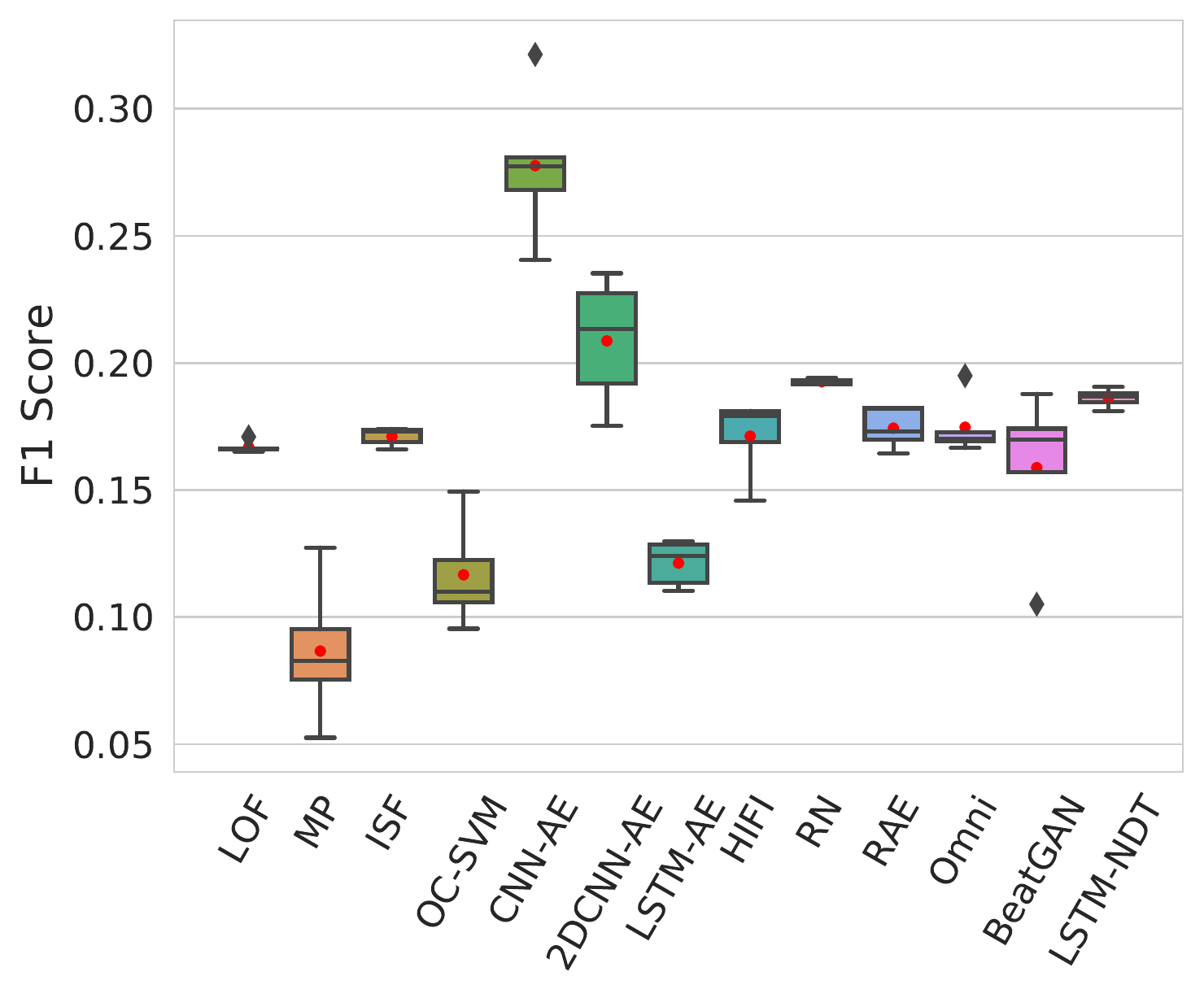}\label{fig:noise_NAB_f1}}
\subfigure[F1 Score (MSL)] {\includegraphics[width=0.21\textwidth]{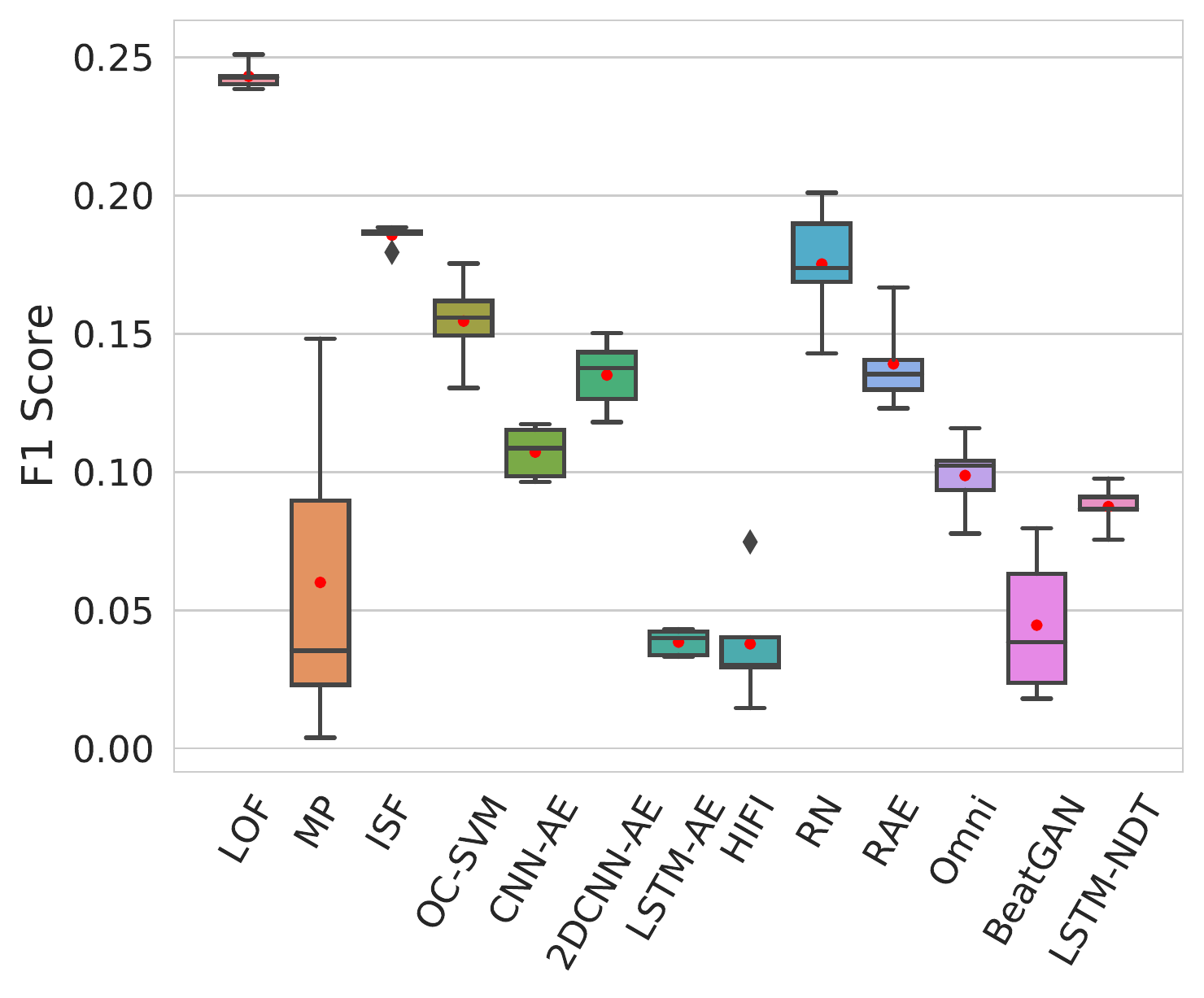}\label{fig:noise_MSL_f1}}
\subfigure[ROC-AUC (NAB)] {\includegraphics[width=0.21\textwidth]{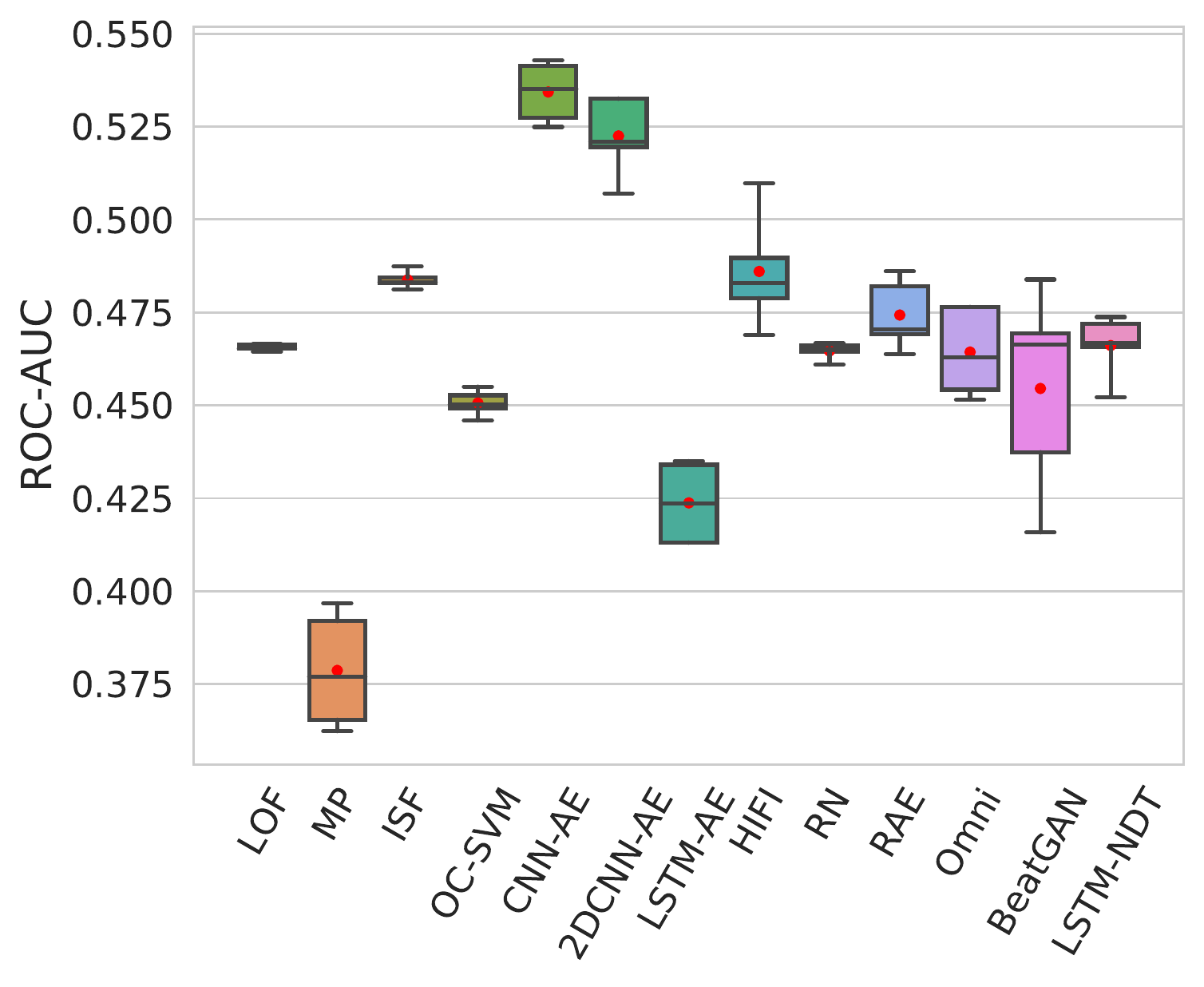}\label{fig:noise_NAB_ROC_AUC}}
\subfigure[ROC-AUC (MSL)] {\includegraphics[width=0.21\textwidth]{MSL_ROC-AUC.pdf}\label{fig:noise_MSL_ROC_AUC}}
\subfigure[PR-AUC (NAB)] {\includegraphics[width=0.21\textwidth]{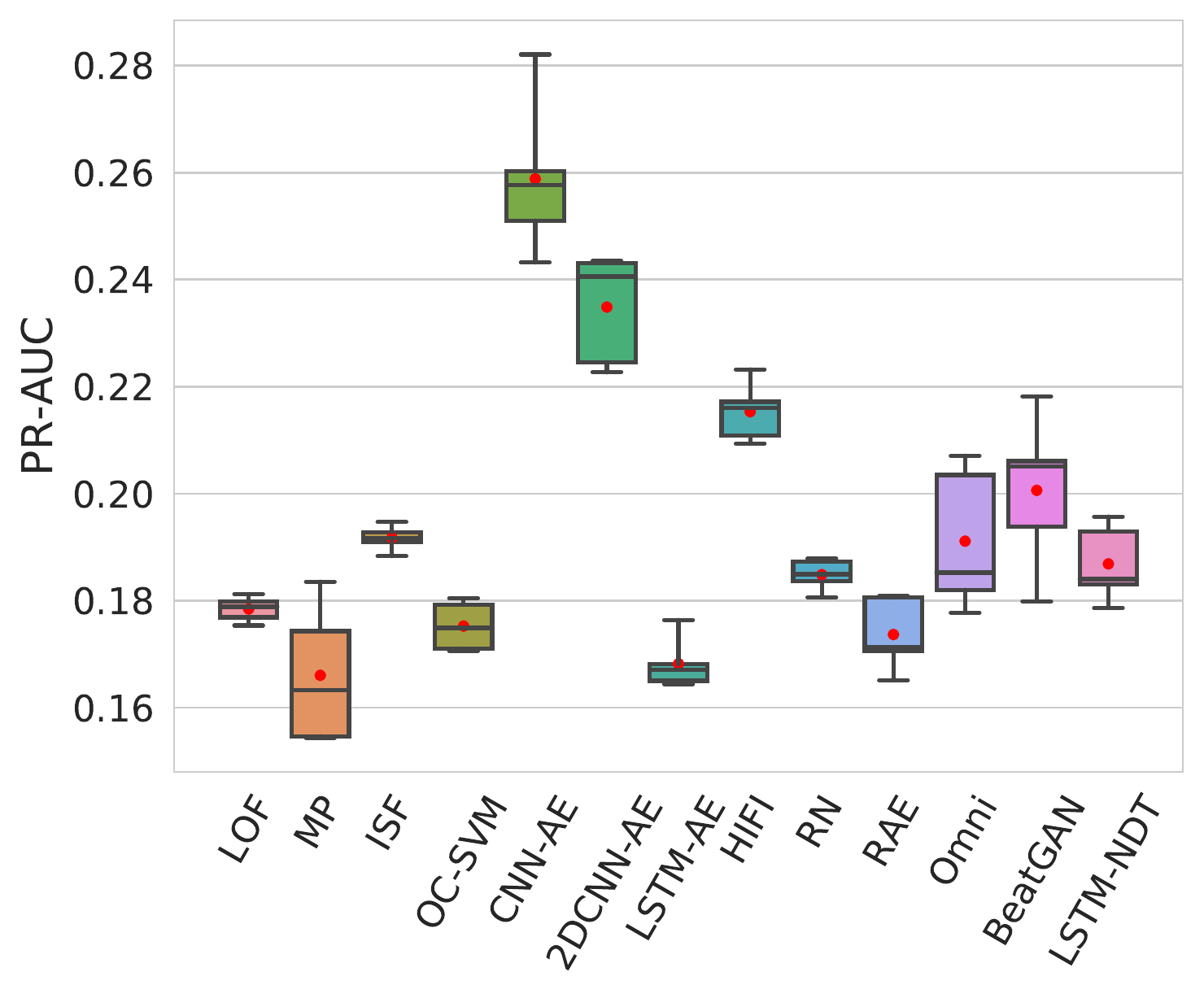}\label{fig:noise_NAB_PR_AUC}}
\subfigure[PR-AUC (MSL)] {\includegraphics[width=0.21\textwidth]{MSL_PR-AUC.pdf}\label{fig:noise_MSL_PR_AUC}}
\vskip -9pt
\caption{Robustness}
\label{fig:Robustness-f1}
\end{figure*}

\noindent\textbf{Recommendation Metric (\emph{RM}).}
For the accuracy and robustness results, we adopt Box-plots to show the distribution of detection accuracy results obtained for different hyperparameter combinations (cf.\ Figs.~\ref{fig:pre}--\ref{fig:NAB_pr}) or different noise proportions (cf.\ Fig.~\ref{fig:Robustness-f1}) for different datasets,
where red dots indicate mean values, boxes show quartiles of results, and  whiskers extend to show the rest of the distributions. The ranges of the whiskers are $[\max\{\mathit{Min}, Q1-3*\mathit{IQR}\},Q1]$ and $[Q3, \min\{\mathit{Max}, Q3+3*\mathit{IQR}\}]$, where $\mathit{Min}$ and $\mathit{Max}$ are the minimum and maximum values, respectively. Next,
$\mathit{IQR}$ is the difference between the values ranking in $25\%$ (denoted by $Q1$, the lower quartile) and $75\%$ (denoted by $Q3$, the upper quartile) among the full results, i.e., $\mathit{IQR} = Q3-Q1$, which denotes the box width in the Box-plots.

For method recommendation, two aspects are considered: the best performance a method can achieve and how hard it is to achieve the best performance.
We introduce a recommendation metric, $\mathit{RM}$, for selecting suitable methods considering these two aspects.
\begin{equation}
\label{equ:rm}
\mathit{RM} = \frac{\mathit{Max}}{\mathit{IQR}*(\mathit{Max}-\mathit{Median})+1},
\end{equation}
where 
$\mathit{Max}$ is the best result a method can obtain
and $\mathit{Median}$ is the median across all results.
Then, $\mathit{IQR}$ and $\mathit{Max}-\mathit{Median}$ indicate how hard it is to achieve the best result, and a larger $\mathit{IQR}*(\mathit{Max}-\mathit{Median})$ value indicates a higher difficulty to find the optimal hyperparameter configuration that achieves the best result.
Therefore, a higher $\mathit{RM}$ value means that the method is more likely to achieve better performance more easily.

\noindent\textbf{\emph{Training} and \emph{Testing Time}.}
For efficiency, \emph{training} and \emph{testing time} are used, where \emph{training time} is the average training time of each epoch, and \emph{testing time} is the average testing time across the whole testing set.

\subsection{Other Implementation Details}\label{sec:implementation}
All methods are implemented in \emph{Python 3.7.9}. LOF, ISF, and OC-SVM are implemented using \emph{Scikit-learn 1.19}, while the DL methods are implemented using PyTorch (\emph{v1.7.0 with Python 3.7.9}).
All experiments are run on a Linux (Ubuntu 16.04) machine with an NVIDIA GeForce RTX 3080 10GB and 256G memory.

\section{Experimental Results}\label{sec:exp}
We report on the outcomes of a study of effectiveness and efficiency of the representative methods according to the proposed taxonomies for data, methods, and evaluation strategies.
We provide three tables (cf. Tables~\ref{tab:effective}--\ref{tab:robustefficiency}) that aim to guide users in selecting suitable methods according to the data setting (with different dimensionality and stationarity) and performance needs (including accuracy, robustness, and training and testing time).

\subsection{Effectiveness}
\subsubsection{Accuracy}
We report the detection accuracy results obtained with different hyperparameter combinations.

\begin{figure}[H]
\centering
\subfigure[S5: \emph{one-dimensional}] {\includegraphics[width=0.21\textwidth]{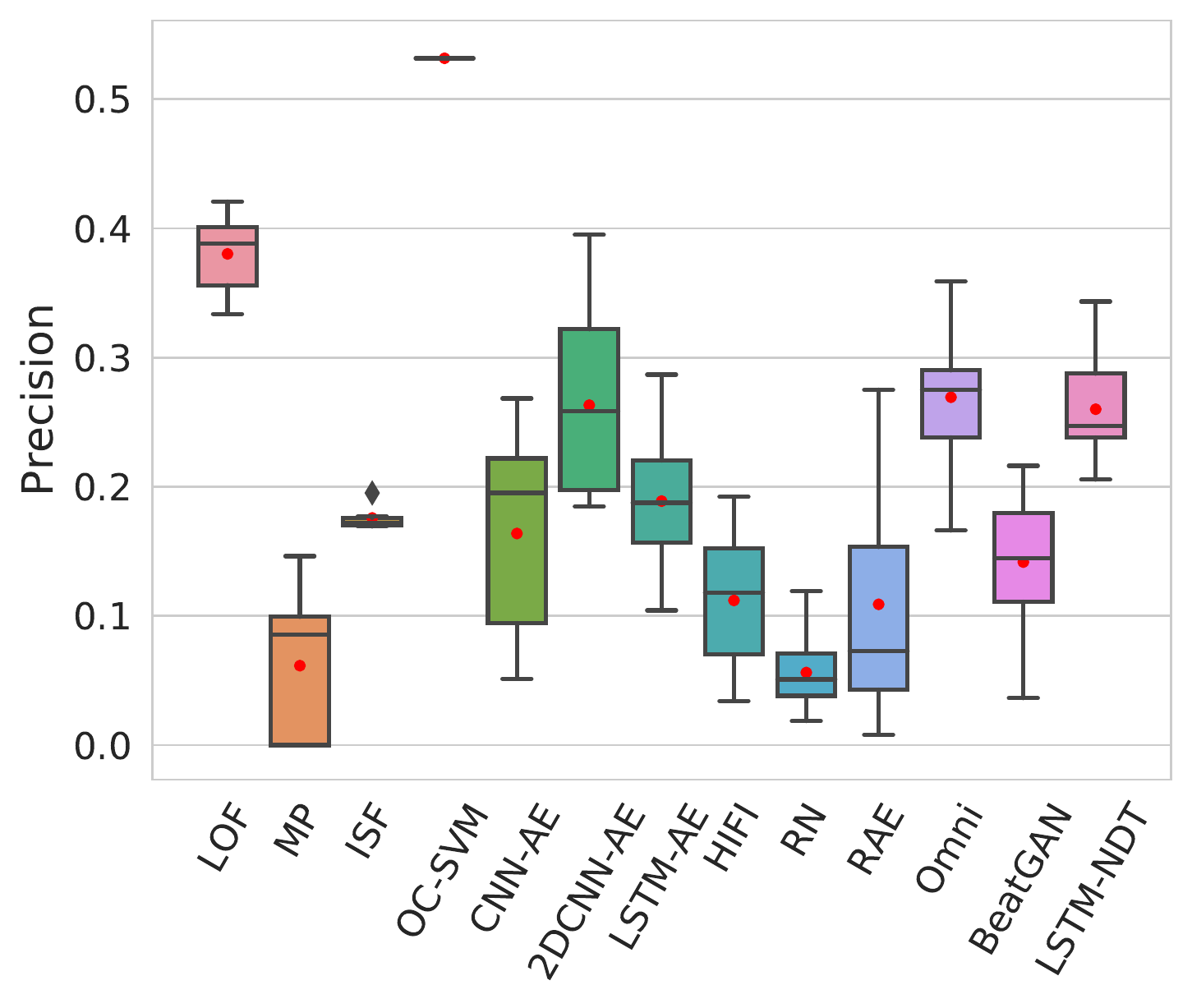}\label{fig:S5-pre}}
\subfigure[KPI: \emph{one-dimensional}] {\includegraphics[width=0.21\textwidth]{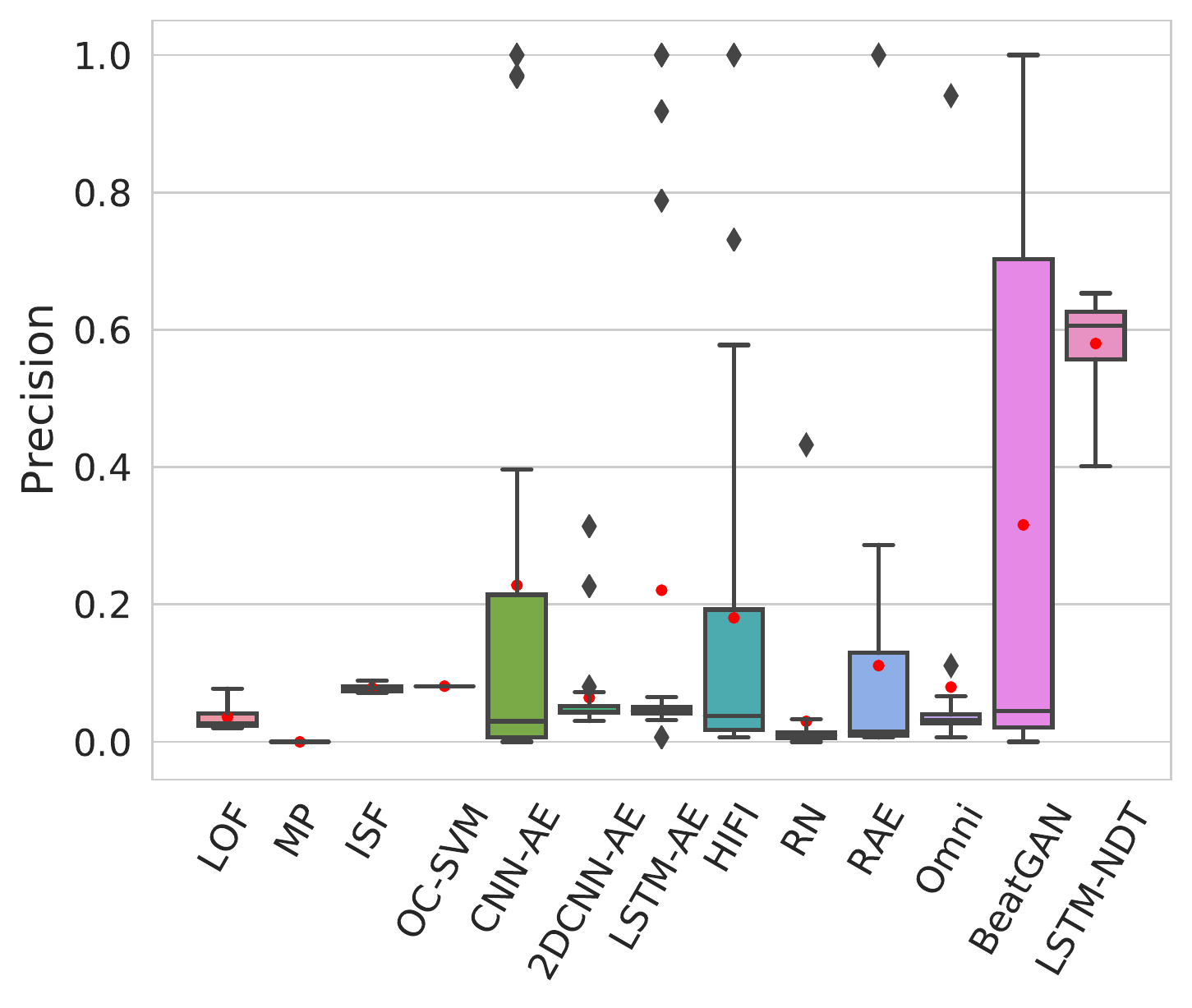}\label{fig:KPI-pre}}
\subfigure[ECG: \emph{2-dimensional}] {\includegraphics[width=0.21\textwidth]{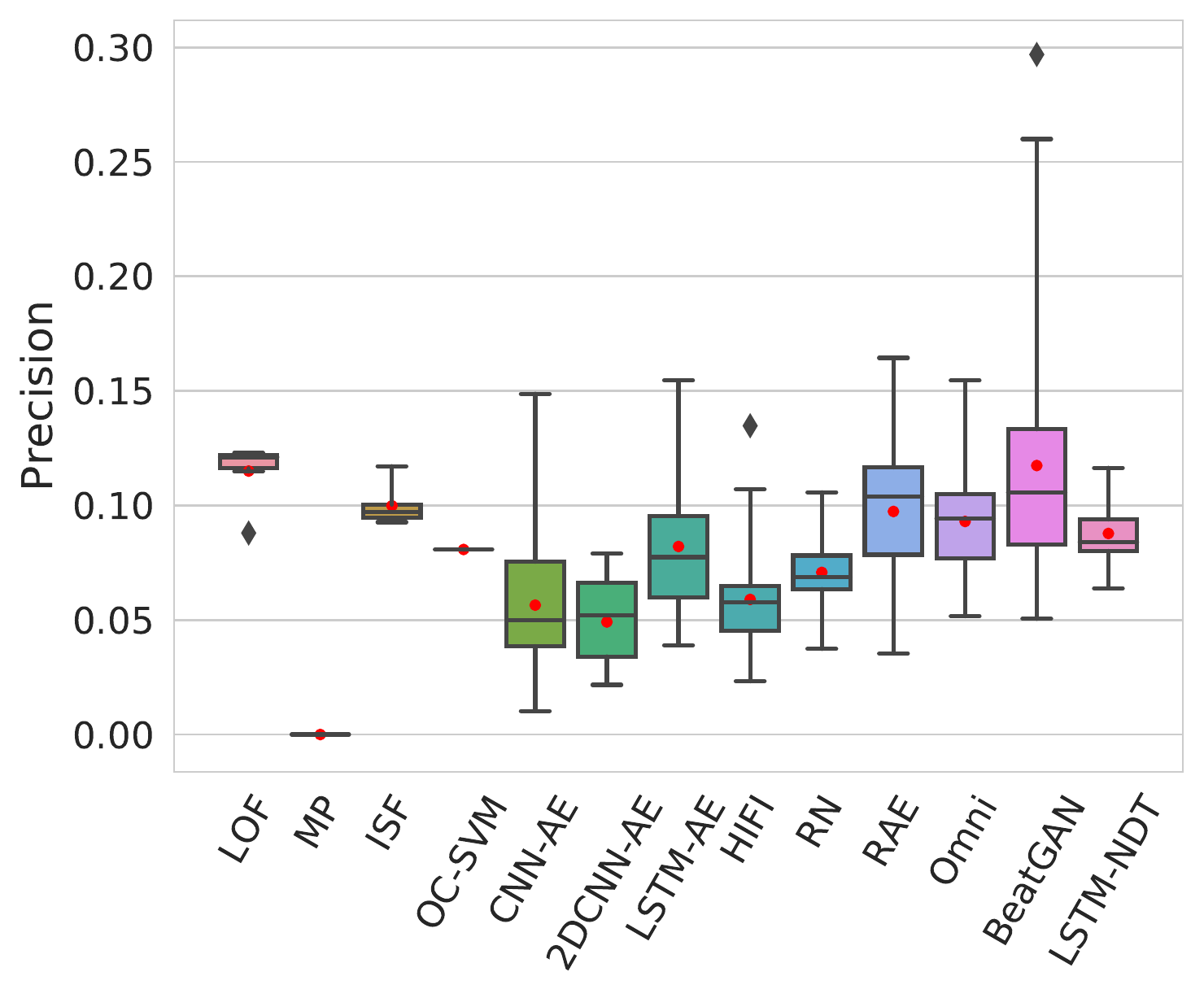}\label{fig:ECG-pre}}
\subfigure[NYC: \emph{3-dimensional}] {\includegraphics[width=0.21\textwidth]{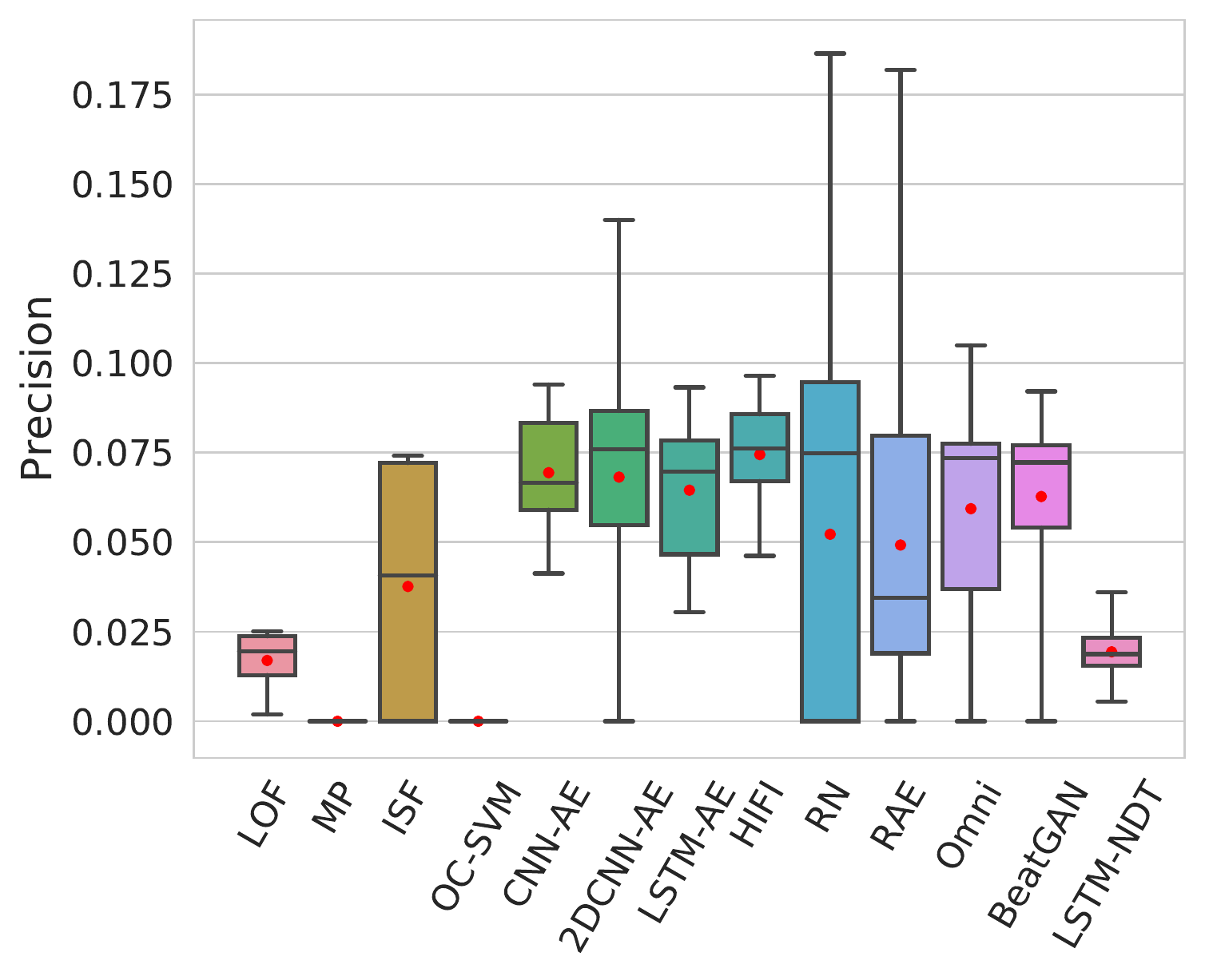}\label{fig:NYC-rec}}
\subfigure[SMAP: \emph{25-dimensional}] {\includegraphics[width=0.21\textwidth]{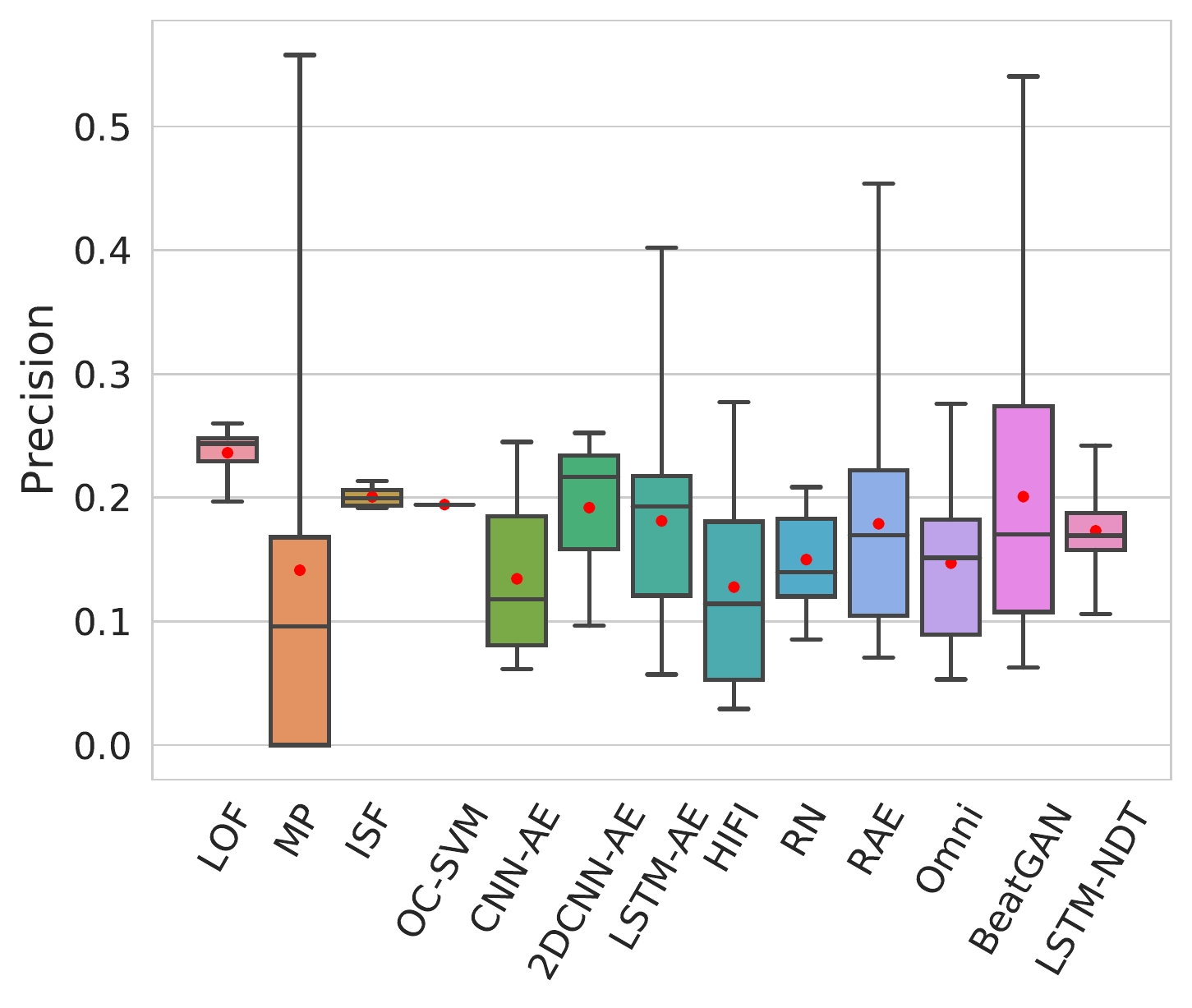}\label{fig:SMAP-pre}}
\subfigure[Credit: \emph{29-dimensional}] {\includegraphics[width=0.21\textwidth]{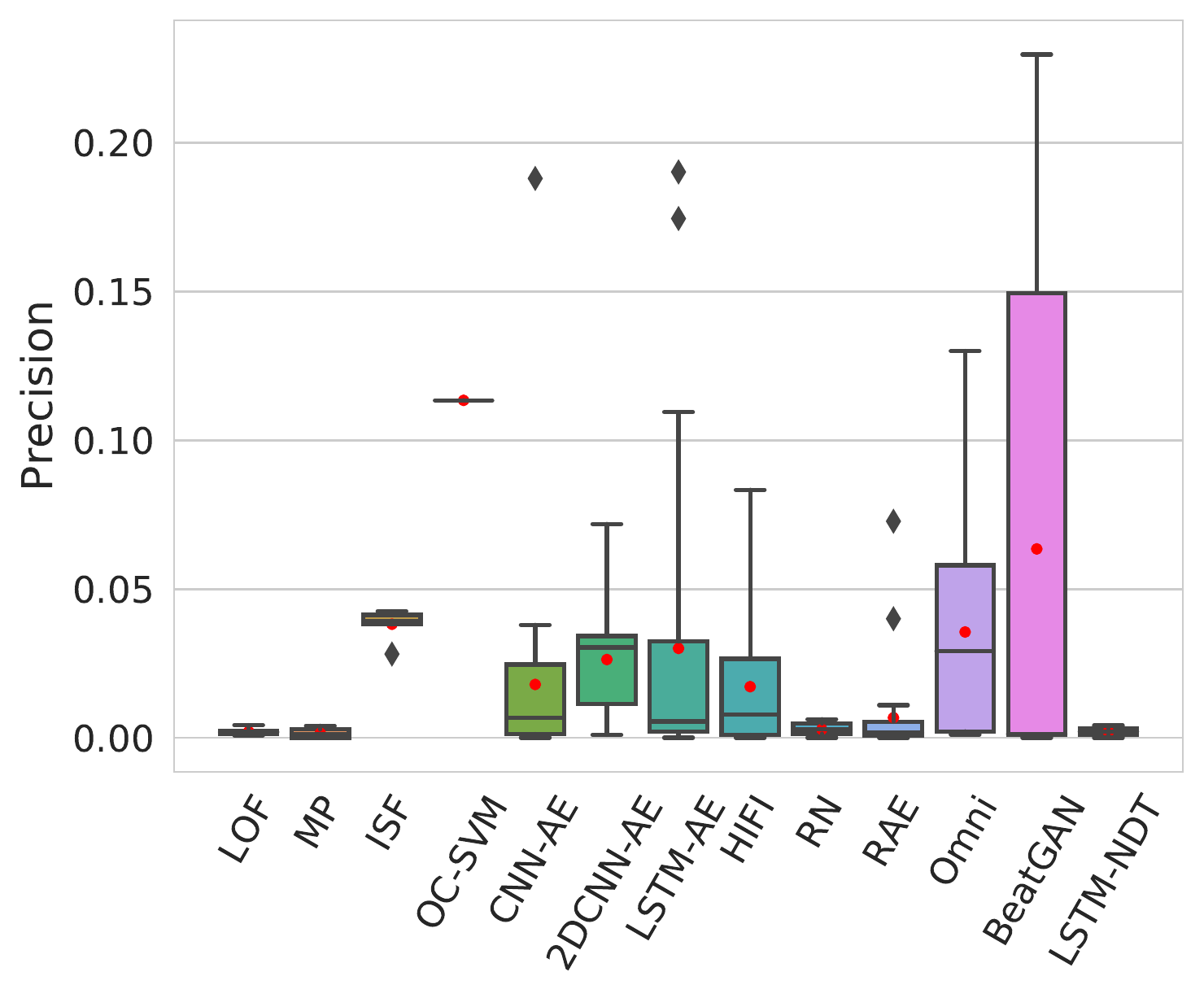}\label{fig:Credit-pre}}
\subfigure[SWaT: \emph{51-dimensional}] {\includegraphics[width=0.21\textwidth]{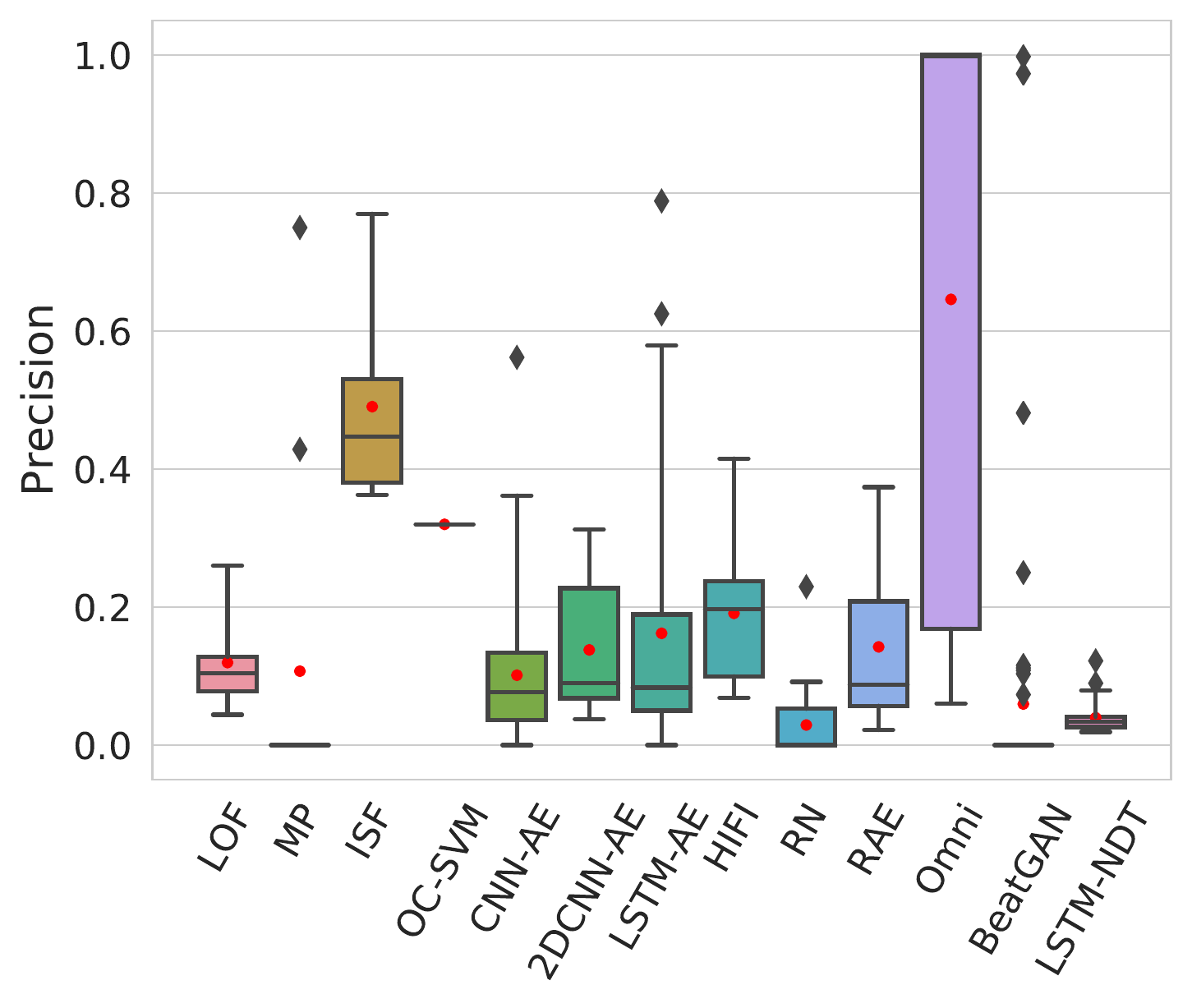}\label{fig:SWaT-pre}}
\subfigure[MSL: \emph{55-dimensional}] {\includegraphics[width=0.21\textwidth]{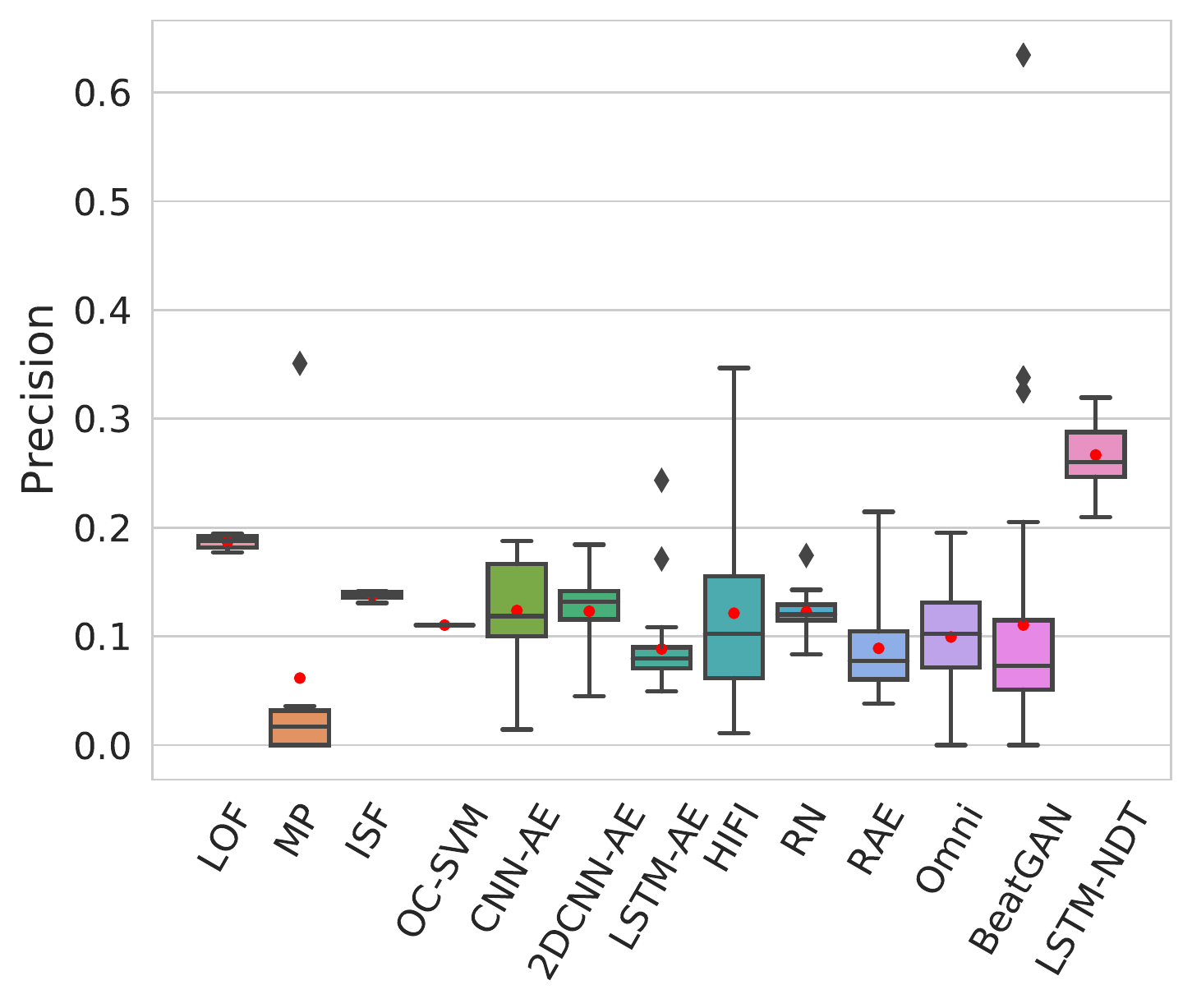}\label{fig:MSL-pre}}
\vskip -9pt
\caption{Precision (Univariate vs. Multivariate)}
\label{fig:pre}
\end{figure}

\noindent \textbf{Experiment 1: Univariate vs. Multivariate Data.}
To study the accuracy of the methods on datasets with different dimensions, we report the \emph{precision}, \emph{recall}, \emph{$\mathit{F1}$ score}, \emph{ROC-AUC}, and \emph{PR-AUC} on eight non-stationary datasets with different dimensions in Figs.~\ref{fig:pre}--\ref{fig:pr}. 

According to the best results the methods can achieve on these eight non-stationary datasets, we give an overall performance analysis in Table~\ref{tab:effective}, which compares the traditional (TR) and deep learning (DL) methods, the different TR methods, and the different DL methods.
We also give the average recommendation metric ($\mathit{RM}$) sorting of different methods in Table~\ref{tab:effective}. 

For brevity, we use shorthands to denote types of methods, e.g., we use ``Classification'' to denote classification-based method, and use the label, ``$>$'', to denote ``outperforms'', e.g., ``Classification $>$ Density'' means that the classification-based methods outperform the density-based methods.
Table~\ref{tab:effective} shows that for univariate datasets (i.e., S5 and KPI), the TR and DL methods are neck to neck in terms of obtaining the best \emph{precision};
when comparing the TR methods, the classification-based method (OC-SVM) outperforms the density-based method (LOF), followed by the similarity-based method (MP); and when comparing the DL methods, the prediction-based method outperforms the reconstruction-based methods in most cases,
as shown in Figs.~\ref{fig:S5-pre} and \ref{fig:KPI-pre}.
We also observe that in some cases, the TR methods are much less sensitive to hyperparameters compared with the DL methods, which is mainly because the DL methods have more hyperparameters, making them harder to train.
Another reason is that we report the results for the TR methods using only 1 or 5 hyperparameter combinations, while we report the results for the DL methods using more hyperparameter combinations (cf. Sec.~\ref{sec:expmethod}). 
Furthermore, we report the average $\mathit{RM}$ sorting for different methods in descending order.
The methods have different performance on the datasets with different dimensionality, and Table~\ref{tab:effective} provides guidance on method selection accordingly.

\noindent \textbf{Experiment 2: Stationary vs. Non-stationary.}
To study the accuracy of the methods on datasets with different stationarity, we use four stationary and four non-stationary subsets from NAB (shown in Table~\ref{tab:DimAndSta}) and report \emph{precision}, \emph{recall}, \emph{$\mathit{F1}$ score}, \emph{ROC-AUC}, and \emph{PR-AUC} in Figs.~\ref{fig:NAB_pre}--\ref{fig:NAB_pr}.
We also give a comprehensive analysis in Table~\ref{tab:accuracyNAB}, which consists of an overall performance analysis and an $\mathit{RM}$ sorting.
Methods have different performance under different stationarity. For example, the TR and the DL methods exhibit similar performance on stationary datasets in terms of \emph{precision}, while the DL methods perform better than the TR methods on non-stationary datasets.

\subsubsection{Robustness}
Robustness captures the ability of a method to maintain its performance in the face of noisy data.
Here, we fix the hyperparameters to the default values (e.g., the hidden size is set to $64$, and window size is set to $128$ for the DL methods) and
evaluate the robustness of all methods on NAB and MSL by changing the percentage (i.e., $1\%$, $2\%$, $3\%$, $4\%$, and $5\%$) of background noise. Recall that NAB is an univariate dataset consisting of both non-stationary and stationary data and that MSL is a multivariate dataset containing only non-stationary data.
We design a simple negative sampling method to generate out-of-range background noise.
Specifically, given a set of positive $k$-dimensional time series samples $T =  \langle S_1,S_2,...,S_n \rangle$, there are sufficient positive samples (normal observations) in the datasets, and only few negative samples, or anomalies, are expected.
The probability that any point drawn from the time series is normal is nearly one.
Therefore, we regard range $[\mathit{min}\{S^j\}, \mathit{max}\{S^j\}]$ as the normal range for the $j$th dimension ($1\le j \le k$), where $\mathit{min}\{S^j\}$ and $\mathit{max}\{S^j\}$ denote the minimum and maximum values of the $j$th dimension, and all values are normalized to the unit range. We generate negatives in ranges that are slightly outside the normal range.
Specifically, the values in the $j$th dimension of negative samples are chosen independently and uniformly from ranges $[\mathit{min}\{S^j\}-\delta, \mathit{min}\{S^j\})$ and $(\mathit{max}\{S^j\}, \mathit{max}\{S^j\}+\delta]$, where $\delta$ is a small positive number (set to 0.05). 

We give the robustness analysis based on all the accuracy metrics in Table~\ref{tab:robustefficiency}, consisting of an overall performance analysis in terms of box width and a box width sorting, where the box width is used to show the robustness,
i.e., the smaller the box width for a method, the more robust the method is.
The TR and DL methods exhibit neck to neck robustness performance. In TR methods, the density and partition based methods are robust,
and in most cases, the prediction-based method is more robust than the reconstruction-based ones.

\subsection{Efficiency}\label{sec:expEfficiency}
\begin{figure*}
\centering
\subfigure[Training Time (NAB)] {\includegraphics[width=0.21\textwidth]{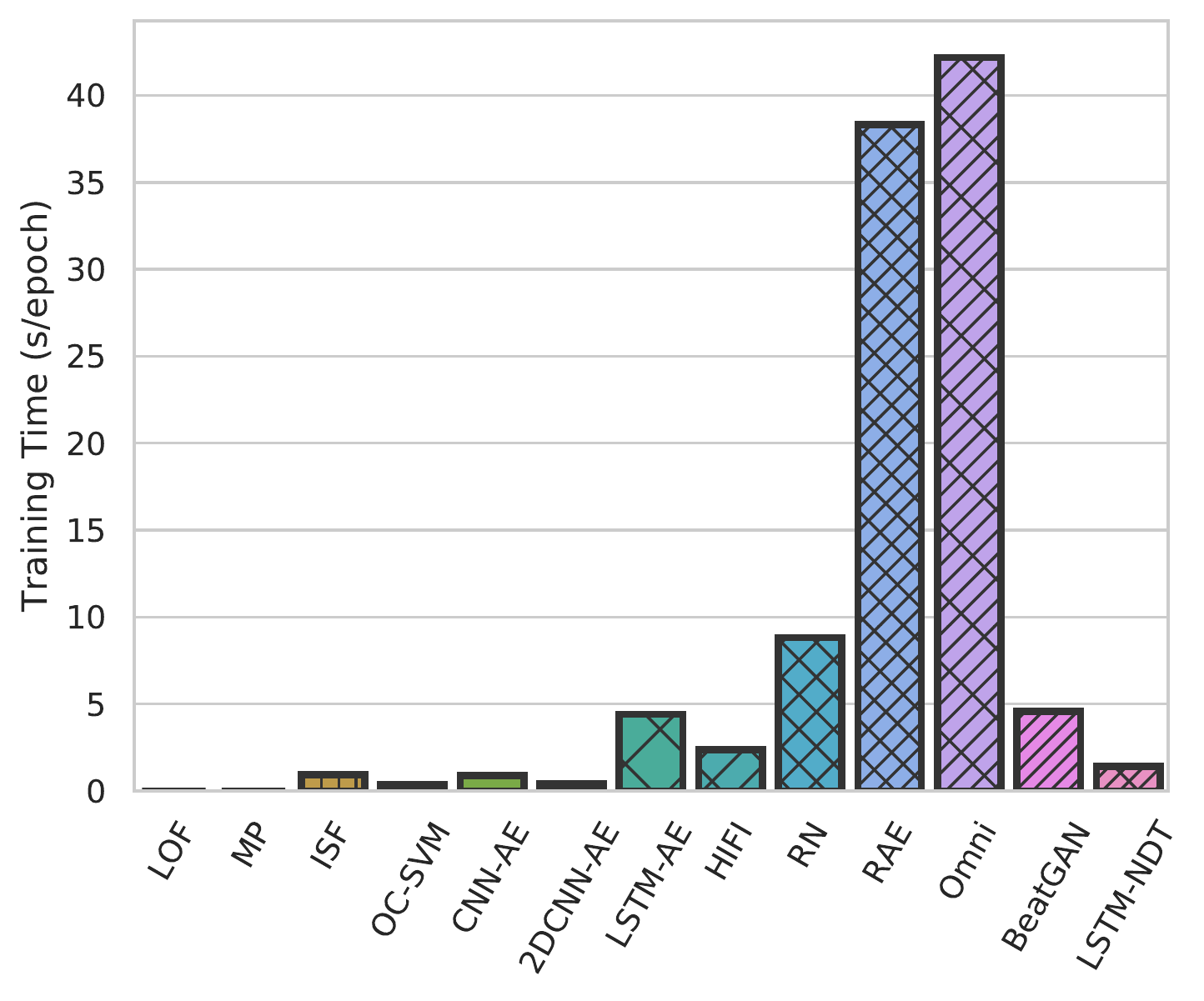}\label{fig:NAB-trainingTime}}
\subfigure[Training Time (MSL)] {\includegraphics[width=0.21\textwidth]{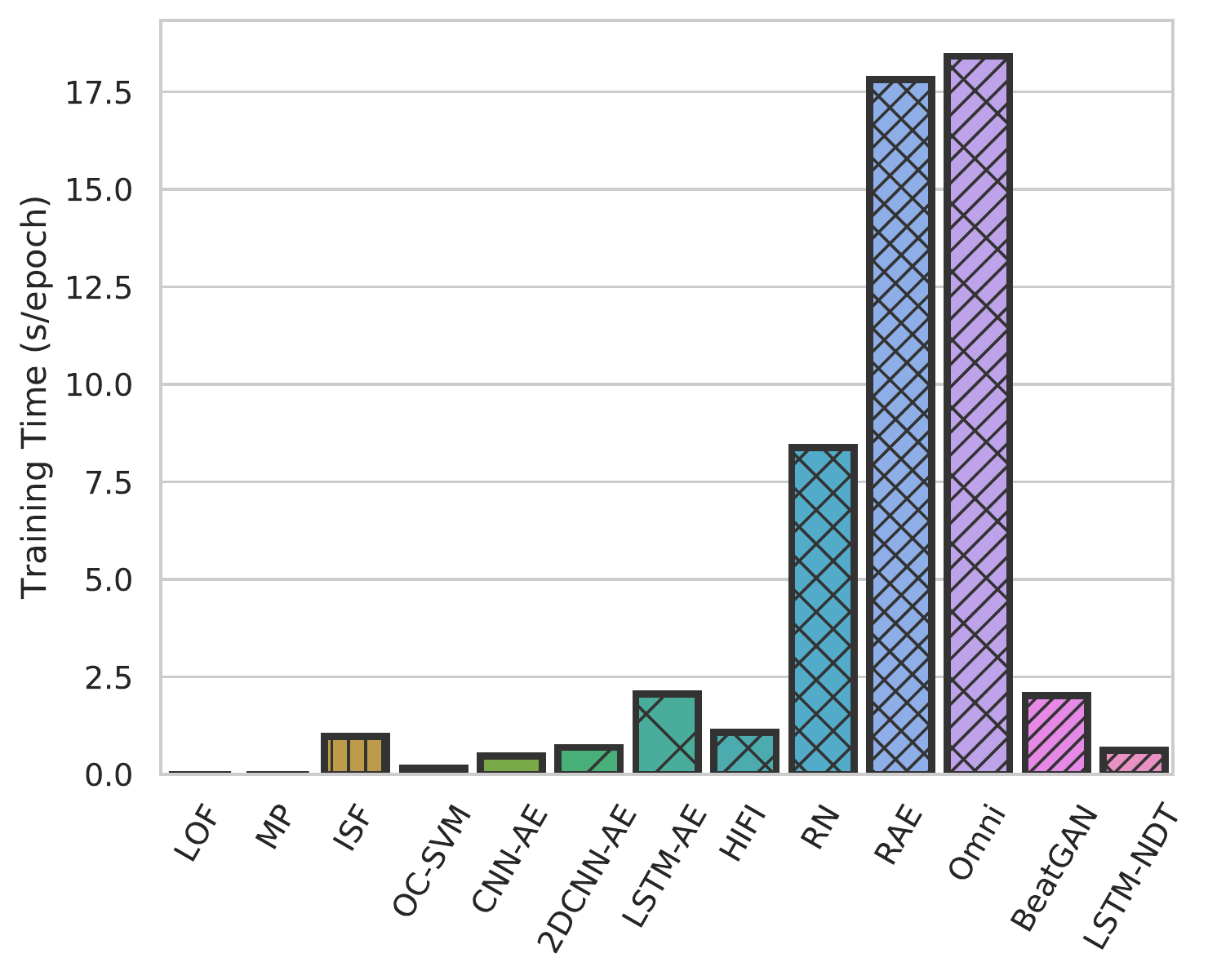}\label{fig:MSL-trainingTime}}
\subfigure[Testing Time (NAB)] {\includegraphics[width=0.21\textwidth]{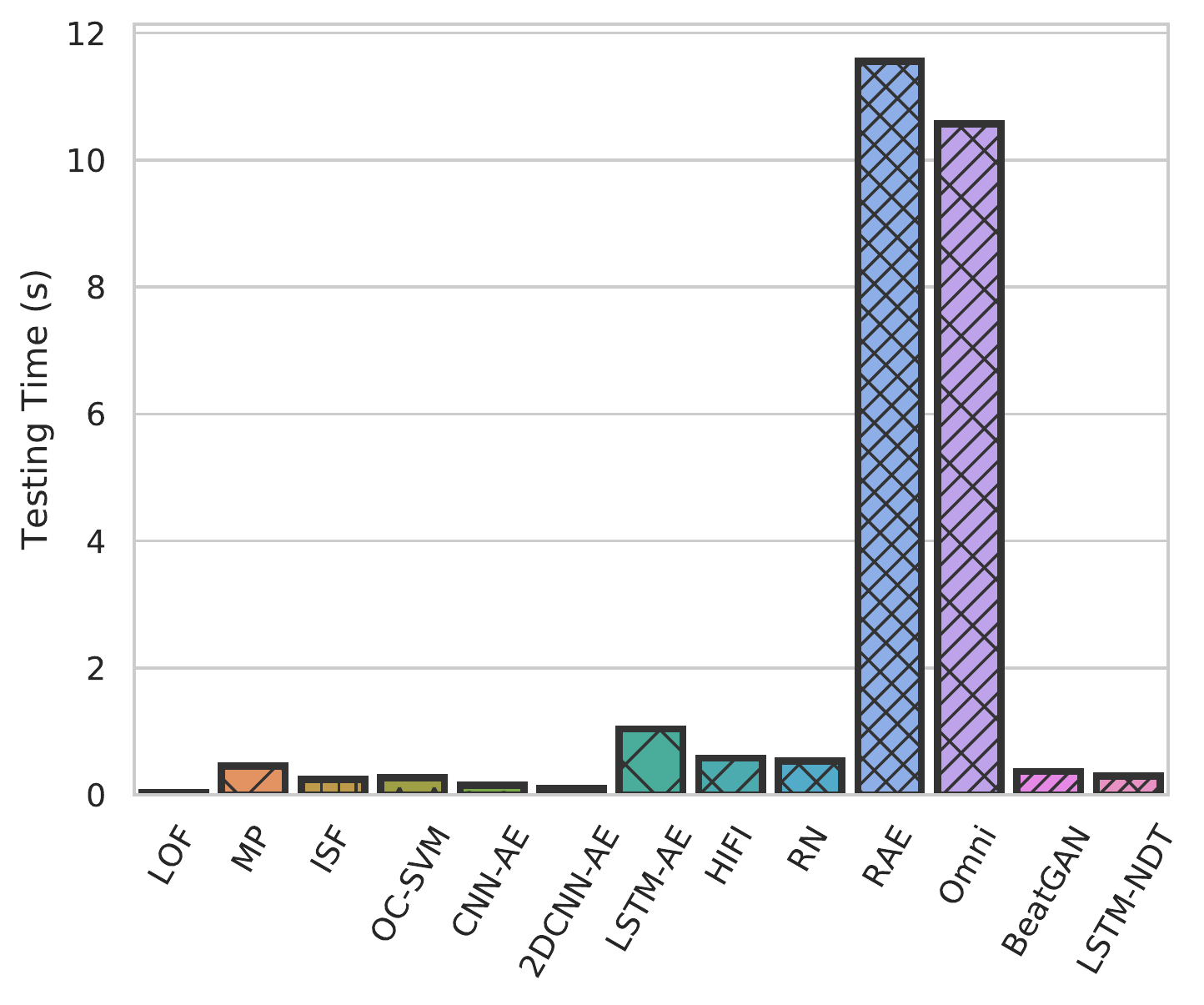}\label{fig:NAB-testingTime}}
\subfigure[Testing Time (MSL)] {\includegraphics[width=0.21\textwidth]{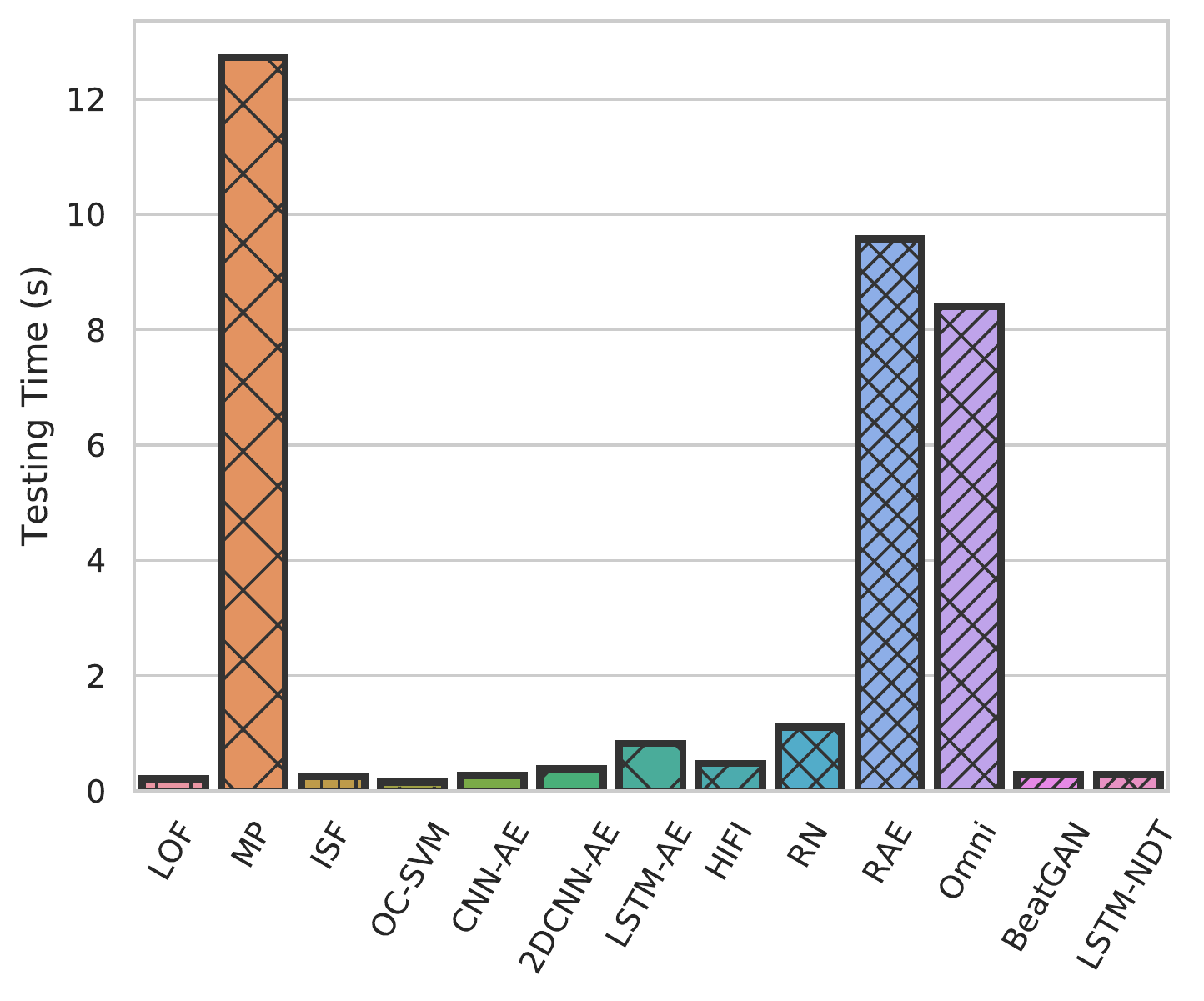}\label{fig:MSL-testingTime}}
\vskip -9pt
\caption{Efficiency}
\label{fig:trainingTime}
\end{figure*}

In this section, we still fix the hyperparameters at their default values and
report the anomaly detection efficiency, including \emph{training} and \emph{testing time}, on NAB and MSL in Fig.~\ref{fig:trainingTime}.
Omni incurs the highest training time on both datasets (cf. Figs.~\ref{fig:NAB-trainingTime} and \ref{fig:MSL-trainingTime}) since it adopts planar Normalizing Flow~\cite{rezende2015variational}, which is time-consuming because it uses a series of invertible mappings to learn non-Gaussian posterior distributions in a latent stochastic space.
This is also the reason why the testing time of Omni is relatively high---cf. Figs.~\ref{fig:NAB-testingTime} and \ref{fig:MSL-testingTime}.
RN and RAE expend long training times since they adopt ensemble techniques.
RAE consumes more time for training than RN since it adopts recurrent layers (i.e., RNN) that are trained recurrently, while RN uses linear layers that are trained directly without recursion. Due to the two factors, i.e., adopting ensemble techniques and recurrent layers, RAE has high testing time.
Without training, MP has the longest testing time on MSL, a 55-dimensional dataset. This occurs because
the computational complexity of MP depends on the dimensionality.
The overall performance analysis and method recommendation are shown in Table~\ref{tab:robustefficiency},
from which we can see that all methods except RN, RAE, and Omni are suitable for applications that need frequent periodic training, and all methods except MP, RAE, and Omni can support online anomaly detection applications.

\begin{table*}[!htbp]
\scriptsize
\centering
  \caption{Accuracy Analysis (Univariate vs. Multivariate)}
  \label{tab:effective}
  \renewcommand{\arraystretch}{1}
  \begin{tabular}{|p{0.8cm}|p{1.05cm}| p{0.9cm}|p{3.02cm}|p{3.02cm}|p{3.12cm}|p{2.92cm}|}\hline
    \multicolumn{3}{|c|}{\diagbox[width=4.2cm]{\textbf{ Metric \qquad}}{\textbf{Analysis}}{\textbf{Dataset}}} &
    \tabincell{c}{\tabincell{c}{\textbf{One-dimensional}\\\textbf{($\bm{D=1}$)}}}
   & \tabincell{c}{\tabincell{c}{\textbf{Low-dimensional}\\\textbf{($\bm{D\in(1,10] }$)}}}
   & \tabincell{c}{\tabincell{c}{\textbf{Intermediate-dimensional}\\\textbf{($\bm{D\in(10,50] }$)}}}
   & \tabincell{c}{\tabincell{c}{\textbf{High-dimensional}\\\textbf{($\bm{D>50}$)}}}
 \\
    \cline{1-7}
    \multirow{4}*{\tabincell{l}{\textbf{Precision}\\\textbf{(cf. Fig.~\ref{fig:pre})}}}   &\multirow{3}*{\tabincell{l}{\textbf{Overall}\\\textbf{performance}\\\textbf{(best results)}}} & TR vs. DL  & They are neck to neck. &  In most  cases,  DL $>$ TR.    & They are neck to neck.  & They are neck to neck. \\
\cline{3-7}
\multirow{4}*{}&\multirow{3}*{}&TR    & Classification   $>$ Density $>$  Similarity.    &  Density and Partition $>$ Similarity and Classification.      & All TR methods are neck to neck.    & Partition $>$ Classification.
\\
\cline{3-7}
\multirow{4}*{}&\multirow{3}*{}&DL    & In most cases, Prediction $>$ Reconstruction.    & All DL methods are neck to neck.     & In most cases, Reconstruction $>$ Prediction; GAN $>$ other Reconstruction.    & All DL methods are neck to neck.
\\
\cline{2-7}
    &\multicolumn{2}{l|}{$\bm{\mathit{RM}}$ \textbf{sorting (descendingly)}} & Omni, LSTM-AE, RAE, CNN-AE, HIFI, LSTM-NDT, BeatGAN, 2DCNN-AE,  OC-SVM, RN, LOF, ISF, MP &BeatGAN, RAE, RN, Omni,   LSTM-AE, CNN-AE, HIFI, 2DCNN-AE, ISF, LSTM-NDT, LOF, OC-SVM, MP &  BeatGAN, LSTM-AE, MP, RAE, Omni,  CNN-AE, HIFI, 2DCNN-AE, OC-SVM, LOF, ISF, LSTM-NDT, RN & BeatGAN, Omni, MP, LSTM-AE, ISF,  HIFI, CNN-AE, RAE, 2DCNN-AE, LOF, LSTM-NDT, OC-SVM, RN \\
    \cline{1-7}
    \multirow{4}*{\tabincell{l}{\textbf{Recall}\\\textbf{(cf. Fig.~\ref{fig:rec})}}}   &\multirow{3}*{\tabincell{l}{\textbf{Overall}\\\textbf{performance}\\\textbf{(best results)}}} & TR vs. DL  & They are neck to neck.  &  They are neck to neck.   & They are neck to neck.  & They are neck to neck.
\\
\cline{3-7}
\multirow{4}*{}&\multirow{3}*{}&TR    & Density and Partition $>$ Similarity and Classification.   &  Density is the best.      & Partition $>$ Density $>$ Similarity     & In most cases, Density is the best.
\\
\cline{3-7}
\multirow{4}*{}&\multirow{3}*{}&DL    & In most cases, AE and VAE $>$ GAN.     & In most cases, GAN $>$ other Reconstruction.      & In most  cases, VAE $>$ GAN and Transformer.     & All DL methods are neck to neck.
    \\
    \cline{2-7}
&\multicolumn{2}{l|}{$\bm{\mathit{RM}}$ \textbf{sorting (descendingly)}} & LOF, RAE, Omni, LSTM-AE, ISF, 2DCNN-AE, OC-SVM, HIFI, CNN-AE, LSTM-NDT,  RN, BeatGAN, MP &BeatGAN, RAE, LSTM-AE, Omni, RN, LSTM-NDT, CNN-AE, LOF,  HIFI, 2DCNN-AE, ISF, OC-SVM, MP &  ISF, 2DCNN-AE, OC-SVM,  Omni, LSTM-AE, BeatGAN, CNN-AE, HIFI, RN, RAE, LOF, LSTM-NDT, MP &  Omni, BeatGAN, LOF, 2DCNN-AE,  RN, LSTM-NDT, LSTM-AE, HIFI, OC-SVM, ISF, CNN-AE, RAE, MP \\
\cline{1-7}
\multirow{4}*{\tabincell{l}{\textbf{F1 Score}\\\textbf{(cf. Fig.~\ref{fig:f1})}}}   &\multirow{3}*{\tabincell{l}{\textbf{Overall}\\\textbf{performance}\\\textbf{(best results)}}} & TR vs. DL  & They are neck to neck.  &  They are neck to neck.    & They are neck to neck.   &  They are neck to neck.
\\
\cline{3-7}
\multirow{4}*{}&\multirow{3}*{}&TR    & Similarity is the worst.    &  Density $>$ Partition $>$ Similarity and Classification.     & Partition $>$ Density $>$ Similarity    & Similarity is the worst.
\\
\cline{3-7}
\multirow{4}*{}&\multirow{3}*{}&DL    & In most cases, Transformer $>$ GAN.    & In most cases, GAN $>$ Prediction.     & All DL methods are neck to neck.    & In most  cases, VAE $>$ AE.
    \\
    \cline{2-7}
&\multicolumn{2}{l|}{$\bm{\mathit{RM}}$ \textbf{sorting (descendingly)}} & 2DCNN-AE, OC-SVM, LOF, LSTM-NDT, CNN-AE, Omni, RAE, ISF, LSTM-AE, HIFI, BeatGAN,  RN, MP	&	BeatGAN, RAE, Omni, RN, LSTM-AE,  HIFI, LOF, LSTM-NDT, CNN-AE, 2DCNN-AE, ISF, OC-SVM, MP	&	BeatGAN, LSTM-AE, 2DCNN-AE,  CNN-AE, Omni, OC-SVM,  ISF, LSTM-NDT, RAE, RN, HIFI, LOF, MP	&	Omni, BeatGAN, ISF, OC-SVM, LOF, LSTM-NDT,  2DCNN-AE, RAE, HIFI, RN, CNN-AE, LSTM-AE, MP \\
\cline{1-7}
\multirow{4}*{\tabincell{l}{\textbf{ROC-AUC}\\\textbf{(cf. Fig.~\ref{fig:roc})}}}   &\multirow{3}*{\tabincell{l}{\textbf{Overall}\\\textbf{performance}\\\textbf{(best results)}}} & TR vs. DL  & They are neck to neck.  &  They are neck to neck.    & They are neck to neck.   &  They are neck to neck.
\\
\cline{3-7}
\multirow{4}*{}&\multirow{3}*{}&TR    & Density and Partition $>$ Similarity and Classification.    &  Similarity is the worst.     & In most cases, Partition $>$ Density and Similarity.   & In most cases, CL $>$ NL.
\\
\cline{3-7}
\multirow{4}*{}&\multirow{3}*{}&DL    & In most cases, Prediction $>$ Reconstruction; VAE $>$ GAN.    & All DL methods are neck to neck.   & All DL methods are neck to neck.    & In most cases, Transformer and VAE $>$ GAN.
    \\
    \cline{2-7}
&\multicolumn{2}{l|}{$\bm{\mathit{RM}}$ \textbf{sorting (descendingly)}} & ISF, LSTM-NDT, 2DCNN-AE, Omni, RAE, LOF, HIFI,  OC-SVM, CNN-AE, LSTM-AE, BeatGAN, RN, MP	&	OC-SVM, RAE, BeatGAN,  HIFI, ISF, Omni, LSTM-AE, CNN-AE, RN, 2DCNN-AE, LSTM-NDT, LOF, MP	&	ISF, OC-SVM, 2DCNN-AE,  Omni, LSTM-AE, HIFI, RAE, CNN-AE, LOF, RN, BeatGAN, MP, LSTM-NDT	&	Omni, ISF, HIFI, CNN-AE, LOF, OC-SVM, BeatGAN, 2DCNN-AE, RAE,  LSTM-AE, RN, LSTM-NDT, MP \\

\cline{1-7}
\multirow{4}*{\tabincell{l}{\textbf{PR-AUC}\\\textbf{(cf. Fig.~\ref{fig:pr})}}}   &\multirow{3}*{\tabincell{l}{\textbf{Overall}\\\textbf{performance}\\\textbf{(best results)}}} & TR vs. DL  & They are neck to neck.  &  They are neck to neck.    & They are neck to neck.   &  They are neck to neck.
\\
\cline{3-7}
\multirow{4}*{}&\multirow{3}*{}&TR    & All TR methods are neck to neck.    &  All TR methods are neck to neck.     & Partition $>$ Density.   & Classification $>$ Density.
\\
\cline{3-7}
\multirow{4}*{}&\multirow{3}*{}&DL    & In most cases, Prediction $>$ Reconstruction.    & In most cases, Reconstruction $>$ Prediction.   &  In most cases, GAN $>$ Transformer.    & All DL methods are neck to neck.
    \\
    \cline{2-7}
&\multicolumn{2}{l|}{$\bm{\mathit{RM}}$ \textbf{sorting (descendingly)}} & CNN-AE, LOF, ISF, OC-SVM, 2DCNN-AE, LSTM-NDT, Omni, MP, LSTM-AE, RAE, HIFI,  BeatGAN, RN	&	RAE, CNN-AE, BeatGAN,   HIFI, LSTM-AE, RN, Omni, LOF, ISF, 2DCNN-AE, OC-SVM, LSTM-NDT, MP	&	LSTM-AE, BeatGAN,  2DCNN-AE, Omni, CNN-AE, HIFI, OC-SVM, MP, RAE, ISF, LSTM-NDT, RN, LOF	&	BeatGAN, Omni, RN, CNN-AE, HIFI, OC-SVM, ISF, RAE, LSTM-AE,  LOF, 2DCNN-AE, MP, LSTM-NDT \\
\cline{1-7}
\end{tabular}
\end{table*}

\begin{table*}[!htbp]
\scriptsize
\centering
  \caption{Accuracy Analysis (Stationary vs. Non-stationary)}
  \label{tab:accuracyNAB}
  \renewcommand{\arraystretch}{1}
  \begin{tabular}{|p{0.9cm}|p{1.05cm}|p{0.9cm}|p{6.39cm}|p{6.39cm}|}\hline
    \multicolumn{3}{|c|}{\diagbox[width=4.2cm]{\textbf{ Metric \qquad}}{\textbf{Analysis}}{\textbf{Dataset}}} &
    \tabincell{c}{\textbf{Stationary}}
   & \tabincell{c}{\textbf{Non-stationary}}
 \\
    \cline{1-5}
    \multirow{4}*{\tabincell{l}{\textbf{Precision}\\\textbf{(cf. Fig.~\ref{fig:NAB_pre})}}}   &\multirow{3}*{\tabincell{l}{\textbf{Overall}\\\textbf{performance}\\\textbf{(best results)}}} & TR vs. DL  & They are neck to neck. &  In most  cases,  DL  $>$ TR.    \\
\cline{3-5}
\multirow{4}*{}&\multirow{3}*{}&TR    & Partition and Classification    $>$ Density.    &   Similarity is the worst.
\\
\cline{3-5}
\multirow{4}*{}&\multirow{3}*{}&DL    & In most cases, GAN $>$ VAE.    & In most cases, Reconstruction $>$ Prediction.
\\
\cline{2-5}    &\multicolumn{2}{l|}{$\bm{\mathit{RM}}$ \textbf{sorting (descendingly)}} & BeatGAN, CNN-AE, 2DCNN-AE, HIFI, LSTM-AE, Omni, RN, RAE,  LSTM-NDT, OC-SVM, ISF, MP, LOF	&LSTM-AE, BeatGAN, CNN-AE, RN, RAE, HIFI,  2DCNN-AE, Omni, OC-SVM, LSTM-NDT, ISF, LOF, MP \\
    \cline{1-5}
    \multirow{4}*{\tabincell{l}{\textbf{Recall}\\\textbf{(cf. Fig.~\ref{fig:NAB_rec})}}}   &\multirow{3}*{\tabincell{l}{\textbf{Overall}\\\textbf{performance}\\\textbf{(best results)}}} & TR vs. DL  & In most  cases,  DL  $>$ TR.  &  They are neck to neck.  \\
\cline{3-5}
\multirow{4}*{}&\multirow{3}*{}&TR    & In most  cases, Density $>$ Similarity and Classification.   &  Partition and Density $>$ Similarity.    \\
\cline{3-5}
\multirow{4}*{}&\multirow{3}*{}&DL    & All DL methods are neck to neck.   & All DL methods are neck to neck.
    \\
    \cline{2-5}
&\multicolumn{2}{l|}{$\bm{\mathit{RM}}$ \textbf{sorting (descendingly)}} & BeatGAN, CNN-AE, Omni, RN, 2DCNN-AE, LSTM-NDT,  LSTM-AE, HIFI, RAE, LOF, ISF, OC-SVM, MP	&RN, RAE, HIFI,  2DCNN-AE, LSTM-NDT, CNN-AE, BeatGAN, ISF, LOF, Omni, LSTM-AE, OC-SVM, MP \\
\cline{1-5}
\multirow{4}*{\tabincell{l}{\textbf{F1 Score}\\\textbf{(cf. Fig.~\ref{fig:NAB_f1})}}}   &\multirow{3}*{\tabincell{l}{\textbf{Overall}\\\textbf{performance}\\\textbf{(best results)}}} & TR vs. DL  & They are neck to neck.  &  They are neck to neck.  \\
\cline{3-5}
\multirow{4}*{}&\multirow{3}*{}&TR    & Density $>$ Similarity; Partition $>$ Classification.    &  Partition $>$ Similarity and Classification; Density $>$ Similarity.
\\
\cline{3-5}
\multirow{4}*{}&\multirow{3}*{}&DL    & All DL methods are neck to neck.    & All DL methods are neck to neck.
    \\
    \cline{2-5}
&\multicolumn{2}{l|}{$\bm{\mathit{RM}}$ \textbf{sorting (descendingly)}} & BeatGAN,  2DCNN-AE, RN, CNN-AE, LSTM-NDT, Omni, HIFI, RAE, LSTM-AE, ISF, LOF, OC-SVM, MP	&HIFI,  2DCNN-AE, RAE, BeatGAN, RN, CNN-AE, LSTM-NDT, Omni, ISF, LSTM-AE, LOF, OC-SVM, MP \\

\cline{1-5}
\multirow{4}*{\tabincell{l}{\textbf{ROC-AUC}\\\textbf{(cf. Fig.~\ref{fig:NAB_roc})}}}   &\multirow{3}*{\tabincell{l}{\textbf{Overall}\\\textbf{performance}\\\textbf{(best results)}}} & TR vs. DL  & They are neck to neck.  &  They are neck to neck.   \\
\cline{3-5}
\multirow{4}*{}&\multirow{3}*{}&TR    & All TR methods are neck to neck.     & All TR methods are neck to neck.
\\
\cline{3-5}
\multirow{4}*{}&\multirow{3}*{}&DL  & All DL methods are neck to neck.  & In most cases,  Reconstruction$>$ Prediction.         \\
    \cline{2-5}
&\multicolumn{2}{l|}{$\bm{\mathit{RM}}$ \textbf{sorting (descendingly)}} & BeatGAN, 2DCNN-AE, HIFI,  CNN-AE, Omni, RN, LSTM-NDT, RAE, ISF, LSTM-AE, LOF, OC-SVM, MP	&BeatGAN, HIFI,  RN, CNN-AE, 2DCNN-AE, RAE, LSTM-NDT, LSTM-AE, MP, Omni, ISF, LOF, OC-SVM \\
\cline{1-5}
\multirow{4}*{\tabincell{l}{\textbf{PR-AUC}\\\textbf{(cf. Fig.~\ref{fig:NAB_pr})}}}   &\multirow{3}*{\tabincell{l}{\textbf{Overall}\\\textbf{performance}\\\textbf{(best results)}}} & TR vs. DL  & They are neck to neck.  &  They are neck to neck.
\\
\cline{3-5}
\multirow{4}*{}&\multirow{3}*{}&TR    & All TR methods are neck to neck.    &  All TR methods are neck to neck.
\\
\cline{3-5}
\multirow{4}*{}&\multirow{3}*{}&DL    & All DL methods are neck to neck.    & In most cases, Reconstruction $>$ Prediction.      \\
    \cline{2-5}
&\multicolumn{2}{l|}{$\bm{\mathit{RM}}$ \textbf{sorting (descendingly)}} & 	BeatGAN, CNN-AE, 2DCNN-AE, HIFI,  RN, LSTM-NDT, Omni, LSTM-AE, RAE, ISF, MP, OC-SVM, LOF	& HIFI, 2DCNN-AE, BeatGAN, CNN-AE, RAE, RN, LSTM-AE, Omni, LSTM-NDT, MP, ISF, OC-SVM, LOF \\
\cline{1-5}
\end{tabular}
\end{table*}

\begin{table*}[!htbp]
\scriptsize
\centering
  \caption{Robustness and Efficiency Analysis}
  \label{tab:robustefficiency}
  \renewcommand{\arraystretch}{1}
  \begin{tabular}{|p{1.7cm}|p{1.7cm}|p{1cm}|p{11.8cm}|}\hline
    \multirow{4}*{\tabincell{l}{\textbf{Robustness}\\\textbf{(cf. Fig.~\ref{fig:Robustness-f1})}}}   &\multirow{3}*{\tabincell{l}{\textbf{Overall}\\\textbf{performance}\\\textbf{(box width)}}} & TR vs. DL  & They are neck to neck.     \\
\cline{3-4}
\multirow{4}*{}&\multirow{3}*{}&TR    & The density and partition based methods are robust, which means that their detection performance is not affected by noise.
\\
\cline{3-4}
\multirow{4}*{}&\multirow{3}*{}&DL    & In most cases, Prediction is more robust than Reconstruction.  \\
\cline{2-4}    &\multicolumn{2}{l|}{\tabincell{l}{\textbf{Box width sorting (ascendingly)}}} &LOF, ISF, RN, LSTM-NDT, LSTM-AE, RAE, Omni, OC-SVM, CNN-AE, 2DCNN-AE, MP, HIFI, BeatGAN \\
    \cline{1-4}
    \multirow{2}*{\tabincell{l}{\textbf{Training efficiency}\\\textbf{(cf. Fig.~\ref{fig:trainingTime})}}}   & \multicolumn{2}{l|}{\tabincell{l}{\textbf{Overall performance}}} &Ensemble models and Omni are the most time-consuming.
    \\
    \cline{2-4}
&\multicolumn{2}{l|}{\tabincell{l}{\textbf{Method recommendation}}} & All methods except RN, RAE, and Omni. \\
\cline{1-4}
\multirow{3}*{\tabincell{l}{\textbf{Testing efficiency}\\\textbf{(cf. Fig.~\ref{fig:trainingTime})}}}   & \multicolumn{2}{l|}{\tabincell{l}{\textbf{Overall performance}}} & 1) MP is the most time-consuming traditional method, and its testing time is the highest among all methods in the MSL dataset. 2) RAE and Omni are the most time-consuming among DL models.
    \\
    \cline{2-4}
&\multicolumn{2}{l|}{\tabincell{l}{\textbf{Method recommendation}}} & All methods except MP, RAE, and Omni. \\
\cline{1-4}
\end{tabular}
\end{table*}

\justifying

\subsection{The Whole Experimental Results}\label{sec:expresult}
We give the whole results in Tables~\ref{tab1}--\ref{tabn}, where ``-1'' denotes that no result is returned since the method runs out of memory.

\begin{table*}[]
\small
\caption{Results of LOF (Univariate vs. Multivariate)}
\begin{tabular}{lllllllllllllllllllll}
\hline
             & \multicolumn{5}{l}{NAB}                    & \multicolumn{5}{l}{S5}                     & \multicolumn{5}{l}{KPI}                    \\ \hline
n\_neighbors & Pre    & Rec    & F1     & ROC    & PR     & Pre    & Rec    & F1     & ROC    & PR     & Pre    & Rec    & F1     & ROC    & PR     \\ \hline
10           & 0.1715 & 0.1051 & 0.1303 & 0.5068 & 0.2139 & 0.3338 & 0.3084 & 0.3206 & 0.8345 & 0.6162 & 0.0776 & 0.0787 & 0.0781 & 0.5178 & 0.0442 \\ \hline
20           & 0.19   & 0.1075 & 0.1373 & 0.5192 & 0.2192 & 0.4208 & 0.3799 & 0.3993 & 0.86   & 0.6841 & 0.0412 & 0.1135 & 0.0604 & 0.5219 & 0.027  \\ \hline
30           & 0.1993 & 0.1188 & 0.1489 & 0.5181 & 0.2194 & 0.4006 & 0.3991 & 0.3999 & 0.871  & 0.7031 & 0.029  & 0.1559 & 0.0489 & 0.5252 & 0.0215 \\ \hline
40           & 0.1978 & 0.1301 & 0.1569 & 0.5211 & 0.2199 & 0.4013 & 0.4087 & 0.4049 & 0.8757 & 0.7145 & 0.0254 & 0.16   & 0.0439 & 0.52   & 0.0206 \\ \hline
50           & 0.1986 & 0.1415 & 0.1652 & 0.5218 & 0.2212 & 0.3758 & 0.4106 & 0.3924 & 0.8862 & 0.7212 & 0.0233 & 0.2404 & 0.0425 & 0.537  & 0.0207 \\ \hline
             & \multicolumn{5}{l}{ECG}                    & \multicolumn{5}{l}{NYC}                    & \multicolumn{5}{l}{SMAP}                   \\
\end{tabular}
\end{table*}

\clearpage 
\begin{table*}[]
\small
\caption{Results of MP (Univariate vs. Multivariate)}
\begin{tabular}{llllllllllllllll}
\hline
              & \multicolumn{5}{l}{NAB}                    & \multicolumn{5}{l}{S5}                     & \multicolumn{5}{l}{KPI}                    \\ \hline
pattern\_size & Pre    & Rec    & F1     & ROC    & PR     & Pre    & Rec    & F1     & ROC    & PR     & Pre    & Rec    & F1     & ROC    & PR     \\ \hline
16            & 0.191  & 0.0363 & 0.061  & 0.5006 & 0.2434 & 0.0937 & 0.1573 & 0.1174 & 0.55   & 0.0882 & 0      & 0      & 0      & 0.5007 & 0.5057 \\ \hline
32            & 0.2028 & 0.074  & 0.1084 & 0.5085 & 0.256  & 0.0856 & 0.1041 & 0.094  & 0.4497 & 0.0602 & 0      & 0      & 0      & 0.5027 & 0.5056 \\ \hline
64            & 0.2119 & 0.0725 & 0.108  & 0.485  & 0.2379 & 0.1463 & 0.0367 & 0.0587 & 0.3404 & 0.032  & 0      & 0      & 0      & 0.5125 & 0.5027 \\ \hline
128           & 0.3167 & 0.0475 & 0.0826 & 0.4976 & 0.2326 & 0.1049 & 0.0345 & 0.0519 & 0.2766 & 0.0284 & 0      & 0      & 0      & 0.5183 & 0.4946 \\ \hline
256           & 0.1928 & 0.0259 & 0.0456 & 0.5138 & 0.242  & 0      & 0      & 0      & 0.2978 & 0.0392 & 0      & 0      & 0      & 0.5154 & 0.4902 \\ \hline
              & \multicolumn{5}{l}{ECG}                    & \multicolumn{5}{l}{NYC}                    & \multicolumn{5}{l}{SMAP}                   \\
\bottomrule
\end{tabular}
\end{table*}
\clearpage 
\begin{table*}[]
\small
\caption{Results of ISF (Univariate vs. Multivariate)}
\begin{tabular}{llllllllllllllll}
\hline
              & \multicolumn{5}{l}{NAB}                    & \multicolumn{5}{l}{S5}                     & \multicolumn{5}{l}{KPI}                    \\ \hline
n\_estimators & Pre    & Rec    & F1     & ROC    & PR     & Pre    & Rec    & F1     & ROC    & PR     & Pre    & Rec    & F1     & ROC    & PR     \\ \hline
20            & 0.1451 & 0.1951 & 0.1664 & 0.5411 & 0.2269 & 0.1952 & 0.4355 & 0.2696 & 0.9239 & 0.697  & 0.0814 & 0.1137 & 0.0949 & 0.6101 & 0.0341 \\ \hline
40            & 0.1939 & 0.1472 & 0.1673 & 0.5457 & 0.2307 & 0.177  & 0.4349 & 0.2516 & 0.9149 & 0.7138 & 0.0782 & 0.1373 & 0.0996 & 0.6096 & 0.031  \\ \hline
60            & 0.1558 & 0.2003 & 0.1753 & 0.5423 & 0.2302 & 0.1728 & 0.4381 & 0.2478 & 0.9235 & 0.7079 & 0.0715 & 0.152  & 0.0972 & 0.6222 & 0.0311 \\ \hline
80            & 0.1531 & 0.1986 & 0.1729 & 0.5426 & 0.2291 & 0.1713 & 0.4361 & 0.246  & 0.9189 & 0.7162 & 0.0795 & 0.1402 & 0.1015 & 0.6197 & 0.0303 \\ \hline
100           & 0.152  & 0.2    & 0.1727 & 0.5435 & 0.2305 & 0.1693 & 0.4349 & 0.2437 & 0.9222 & 0.7187 & 0.0891 & 0.1347 & 0.1072 & 0.6324 & 0.0342 \\ \hline
              & \multicolumn{5}{l}{ECG}                    & \multicolumn{5}{l}{NYC}                    & \multicolumn{5}{l}{SMAP}                   \\
\bottomrule
\end{tabular}
\end{table*}
\clearpage 
\begin{table*}[]
\small
\caption{Results of OC-SVM (Univariate vs. Multivariate)}
\begin{tabular}{llllllllllllllll}
\hline
       & \multicolumn{5}{l}{NAB}                    & \multicolumn{5}{l}{S5}                     & \multicolumn{5}{l}{KPI}                    \\ \hline
kernel & Pre    & Rec    & F1     & ROC    & PR     & Pre    & Rec    & F1     & ROC    & PR     & Pre    & Rec    & F1     & ROC    & PR     \\ \hline
rbf    & 0.2296 & 0.0791 & 0.1177 & 0.5134 & 0.2149 & 0.5317 & 0.3959 & 0.4539 & 0.8681 & 0.7263 & 0.0811 & 0.0476 & 0.06   & 0.4984 & 0.0187 \\ \hline
       & \multicolumn{5}{l}{ECG}                    & \multicolumn{5}{l}{NYC}                    & \multicolumn{5}{l}{SMAP}                   \\ \hline
       & Pre    & Rec    & F1     & ROC    & PR     & Pre    & Rec    & F1     & ROC    & PR     & Pre    & Rec    & F1     & ROC    & PR     \\ \hline
rbf    & 0.0808 & 0.2337 & 0.1201 & 0.5722 & 0.1364 & 0      & 0      & 0      & 0.7354 & 0.0446 & 0.1944 & 0.1588 & 0.1748 & 0.5336 & 0.1961 \\ \hline
       & \multicolumn{5}{l}{Credit}                 & \multicolumn{5}{l}{SWaT}                   & \multicolumn{5}{l}{MSL}                    \\ \hline
       & Pre    & Rec    & F1     & ROC    & PR     & Pre    & Rec    & F1     & ROC    & PR     & Pre    & Rec    & F1     & ROC    & PR     \\ \hline
rbf    & 0.1134 & 0.713  & 0.1957 & 0.9441 & 0.188  & 0.3199 & 0.2698 & 0.2927 & 0.7094 & 0.2261 & 0.1103 & 0.2047 & 0.1434 & 0.5802 & 0.1942 \\ \bottomrule
\end{tabular}
\end{table*}
\clearpage 
\begin{table*}[]
\small
\caption{Results of DL Methods (Univariate vs. Multivariate; Sliding window size=16, Hidden size=16)}
\begin{tabular}{llllllllllllllll}
\hline
         & \multicolumn{5}{l}{NAB}                    & \multicolumn{5}{l}{S5}                     & \multicolumn{5}{l}{KPI}                    \\ \hline
         & Pre    & Rec    & F1     & ROC    & PR     & Pre    & Rec    & F1     & ROC    & PR     & Pre    & Rec    & F1     & ROC    & PR     \\ \hline
CNN-AE   & 0.1535 & 0.2009 & 0.174  & 0.5333 & 0.2278 & 0.2406 & 0.1992 & 0.218  & 0.704  & 0.4155 & 0.9678 & 0.0351 & 0.0677 & 0.3993 & 0.0453 \\ \hline
2DCNN-AE & 0.1455 & 0.1627 & 0.1536 & 0.5222 & 0.2224 & 0.3951 & 0.3582 & 0.3758 & 0.8518 & 0.5442 & 0.0303 & 0.0488 & 0.0374 & 0.6139 & 0.0189 \\ \hline
LSTM-AE  & 0.1626 & 0.1758 & 0.1689 & 0.5239 & 0.2035 & 0.1699 & 0.408  & 0.2399 & 0.6866 & 0.4133 & 0.0391 & 0.1367 & 0.0608 & 0.6148 & 0.0563 \\ \hline
HIFI     & 0.173  & 0.1789 & 0.1759 & 0.5466 & 0.2487 & 0.1922 & 0.2024 & 0.1972 & 0.7489 & 0.4027 & 0.0185 & 0.0263 & 0.0217 & 0.4521 & 0.0232 \\ \hline
RN       & 0.1534 & 0.2075 & 0.1764 & 0.5465 & 0.2373 & 0.0776 & 0.1451 & 0.1011 & 0.6074 & 0.093  & 0.4325 & 0.0106 & 0.0207 & 0.5468 & 0.0191 \\ \hline
RAE      & 0.1325 & 0.1277 & 0.1301 & 0.5173 & 0.2118 & 0.2439 & 0.2545 & 0.2491 & 0.7345 & 0.279  & 0.2252 & 0.0956 & 0.1342 & 0.6468 & 0.0608 \\
\end{tabular}
\end{table*}
\clearpage 
\begin{table*}[]
\small
\caption{Results of DL Methods (Univariate vs. Multivariate; Sliding window size=16, Hidden size=32)}
\begin{tabular}{llllllllllllllll}
\hline
         & \multicolumn{5}{l}{NAB}                    & \multicolumn{5}{l}{S5}                     & \multicolumn{5}{l}{KPI}                    \\ \hline
         & Pre    & Rec    & F1     & ROC    & PR     & Pre    & Rec    & F1     & ROC    & PR     & Pre    & Rec    & F1     & ROC    & PR     \\ \hline
CNN-AE   & 0.1575 & 0.1921 & 0.1731 & 0.5447 & 0.2401 & 0.2532 & 0.2018 & 0.2246 & 0.7063 & 0.4105 & 1      & 0.0252 & 0.0491 & 0.3963 & 0.0477 \\ \hline
2DCNN-AE & 0.1566 & 0.2095 & 0.1792 & 0.5492 & 0.248  & 0.3725 & 0.3646 & 0.3685 & 0.8778 & 0.5669 & 0.0324 & 0.045  & 0.0377 & 0.6123 & 0.0192 \\ \hline
LSTM-AE  & 0.149  & 0.2017 & 0.1714 & 0.5095 & 0.2094 & 0.2366 & 0.1948 & 0.2137 & 0.6438 & 0.3812 & 0.0317 & 0.1844 & 0.0541 & 0.631  & 0.0575 \\ \hline
HIFI     & 0.1765 & 0.1708 & 0.1736 & 0.5559 & 0.255  & 0.178  & 0.2056 & 0.1908 & 0.7241 & 0.3802 & 0.0171 & 0.0229 & 0.0196 & 0.4354 & 0.018  \\ \hline
RN       & 0.1524 & 0.2068 & 0.1755 & 0.5248 & 0.2289 & 0.0613 & 0.1272 & 0.0828 & 0.5816 & 0.0813 & 0.0158 & 0.0489 & 0.0239 & 0.4208 & 0.0311 \\ \hline
RAE      & 0.1481 & 0.187  & 0.1653 & 0.5071 & 0.2035 & 0.2713 & 0.2794 & 0.2753 & 0.7764 & 0.3735 & 1      & 0.0073 & 0.0145 & 0.4349 & 0.0379 \\
\bottomrule
\end{tabular}
\end{table*}
\clearpage 
\begin{table*}[]
\small
\caption{Results of DL Methods (Univariate vs. Multivariate; Sliding window size=16, Hidden size=64)}
\begin{tabular}{llllllllllllllll}
\hline
         & \multicolumn{5}{l}{NAB}                    & \multicolumn{5}{l}{S5}                     & \multicolumn{5}{l}{KPI}                    \\ \hline
         & Pre    & Rec    & F1     & ROC    & PR     & Pre    & Rec    & F1     & ROC    & PR     & Pre    & Rec    & F1     & ROC    & PR     \\ \hline
CNN-AE   & 0.1442 & 0.199  & 0.1672 & 0.5348 & 0.233  & 0.2684 & 0.2184 & 0.2408 & 0.7158 & 0.42   & 0.03   & 0.0575 & 0.0395 & 0.5875 & 0.0204 \\ \hline
2DCNN-AE & 0.1673 & 0.2738 & 0.2077 & 0.5626 & 0.2551 & 0.2661 & 0.3269 & 0.2934 & 0.8432 & 0.5162 & 0.3139 & 0.0465 & 0.081  & 0.6527 & 0.0627 \\ \hline
LSTM-AE  & 0.1516 & 0.2    & 0.1725 & 0.5191 & 0.2095 & 0.23   & 0.2005 & 0.2143 & 0.6576 & 0.4066 & 0.0373 & 0.1496 & 0.0597 & 0.6178 & 0.0554 \\ \hline
HIFI     & 0.168  & 0.1537 & 0.1606 & 0.5386 & 0.2375 & 0.1343 & 0.1833 & 0.155  & 0.6964 & 0.3106 & 0.0306 & 0.0184 & 0.023  & 0.4221 & 0.0133 \\ \hline
RN       & 0.1427 & 0.0942 & 0.1135 & 0.462  & 0.1888 & 0.1053 & 0.062  & 0.0781 & 0.5952 & 0.0938 & 0.0054 & 0.1224 & 0.0104 & 0.4191 & 0.0321 \\ \hline
RAE      & 0.1373 & 0.1728 & 0.153  & 0.5028 & 0.197  & 0.2751 & 0.2852 & 0.2801 & 0.7962 & 0.3894 & 0.287  & 0.0825 & 0.1282 & 0.6979 & 0.071  \\
\bottomrule
\end{tabular}
\end{table*}
\clearpage 
\begin{table*}[]
\small
\caption{Results of DL Methods (Univariate vs. Multivariate; Sliding window size=16, Hidden size=128)}
\begin{tabular}{llllllllllllllll}
\hline
         & \multicolumn{5}{l}{NAB}                    & \multicolumn{5}{l}{S5}                     & \multicolumn{5}{l}{KPI}                    \\ \hline
         & Pre    & Rec    & F1     & ROC    & PR     & Pre    & Rec    & F1     & ROC    & PR     & Pre    & Rec    & F1     & ROC    & PR     \\ \hline
CNN-AE   & 0.163  & 0.2147 & 0.1853 & 0.5287 & 0.2247 & 0.2219 & 0.2018 & 0.2114 & 0.6804 & 0.3769 & 0.0433 & 0.0488 & 0.0459 & 0.5998 & 0.027  \\ \hline
2DCNN-AE & 0.1798 & 0.1318 & 0.1521 & 0.535  & 0.2284 & 0.322  & 0.2893 & 0.3047 & 0.8123 & 0.4452 & 0.0442 & 0.0553 & 0.0491 & 0.6002 & 0.0322 \\ \hline
LSTM-AE  & 0.1543 & 0.1702 & 0.1619 & 0.522  & 0.2118 & 0.2617 & 0.2503 & 0.2559 & 0.7302 & 0.4467 & 0.7882 & 0.0311 & 0.0598 & 0.6652 & 0.0773 \\ \hline

HIFI     & 0.1616 & 0.1546 & 0.158  & 0.5421 & 0.2263 & 0.0962 & 0.145  & 0.1156 & 0.7035 & 0.2485 & 0.0077 & 0.0854 & 0.0142 & 0.4464 & 0.0097 \\ \hline
RN       & 0.1412 & 0.0948 & 0.1135 & 0.4825 & 0.1944 & 0.1191 & 0.0716 & 0.0895 & 0.435  & 0.0853 & 0.0055 & 0.1542 & 0.0106 & 0.4154 & 0.0345 \\ \hline
RAE      & 0.1507 & 0.1909 & 0.1684 & 0.5557 & 0.2357 & 0.2618 & 0.2698 & 0.2657 & 0.7934 & 0.3472 & 0.2083 & 0.0705 & 0.1054 & 0.7031 & 0.0572 \\
\bottomrule
\end{tabular}
\end{table*}

\justifying
\clearpage
\begin{table*}[]
\small
\caption{Results of DL Methods (Univariate vs. Multivariate; Sliding window size=16, Hidden size=256)}
\begin{tabular}{llllllllllllllll}
\hline
         & \multicolumn{5}{l}{NAB}                    & \multicolumn{5}{l}{S5}                     & \multicolumn{5}{l}{KPI}                    \\ \hline
         & Pre    & Rec    & F1     & ROC    & PR     & Pre    & Rec    & F1     & ROC    & PR     & Pre    & Rec    & F1     & ROC    & PR     \\ \hline
CNN-AE   & 0.1559 & 0.1919 & 0.172  & 0.5031 & 0.2114 & 0.2081 & 0.219  & 0.2134 & 0.6887 & 0.4231 & 0.1678 & 0.055  & 0.0828 & 0.596  & 0.0483 \\ \hline
2DCNN-AE & 0.1427 & 0.1747 & 0.1571 & 0.5472 & 0.2271 & 0.2229 & 0.3084 & 0.2588 & 0.868  & 0.5395 & 0.0521 & 0.0327 & 0.0402 & 0.6173 & 0.0255 \\ \hline
LSTM-AE  & 0.1482 & 0.1918 & 0.1672 & 0.5275 & 0.218  & 0.2238 & 0.2005 & 0.2115 & 0.6896 & 0.4147 & 0.0352 & 0.1644 & 0.0579 & 0.6235 & 0.0597 \\ \hline
HIFI     & 0.1514 & 0.129  & 0.1393 & 0.5042 & 0.2101 & 0.1181 & 0.1807 & 0.1428 & 0.7194 & 0.2812 & 0.0074 & 0.1001 & 0.0137 & 0.4528 & 0.0098 \\ \hline
RN       & 0.141  & 0.1461 & 0.1435 & 0.4996 & 0.2539 & 0.0646 & 0.0601 & 0.0623 & 0.4572 & 0.0806 & 0.033  & 0.0207 & 0.0254 & 0.5947 & 0.0179 \\ \hline
RAE      & 0.1444 & 0.1844 & 0.162  & 0.524  & 0.2146 & 0.2619 & 0.2622 & 0.262  & 0.783  & 0.3599 & 0.2559 & 0.0813 & 0.1234 & 0.69   & 0.0646 \\
\bottomrule
\end{tabular}
\end{table*}

\begin{table*}[]
\small
\caption{Results of DL Methods (Univariate vs. Multivariate; Sliding window size=32, Hidden size=16)}
\begin{tabular}{llllllllllllllll}
\hline
         & \multicolumn{5}{l}{NAB}                    & \multicolumn{5}{l}{S5}                     & \multicolumn{5}{l}{KPI}                    \\ \hline
         & Pre    & Rec    & F1     & ROC    & PR     & Pre    & Rec    & F1     & ROC    & PR     & Pre    & Rec    & F1     & ROC    & PR     \\ \hline
CNN-AE   & 0.1449 & 0.1931 & 0.1656 & 0.5208 & 0.2334 & 0.2196 & 0.2088 & 0.2141 & 0.727  & 0.4221 & 0.0276 & 0.0599 & 0.0378 & 0.5673 & 0.0193 \\ \hline
2DCNN-AE & 0.1623 & 0.2163 & 0.1854 & 0.5393 & 0.2408 & 0.3234 & 0.3742 & 0.347  & 0.8836 & 0.5442 & 0.0374 & 0.0352 & 0.0363 & 0.6039 & 0.0197 \\ \hline
LSTM-AE  & 0.1438 & 0.1238 & 0.133  & 0.4771 & 0.1875 & 0.1506 & 0.3314 & 0.2071 & 0.4961 & 0.3541 & 0.0411 & 0.0366 & 0.0387 & 0.6027 & 0.0206 \\ \hline
HIFI     & 0.2544 & 0.1624 & 0.1982 & 0.572  & 0.283  & 0.1774 & 0.2344 & 0.2019 & 0.7657 & 0.4165 & 0.0389 & 0.058  & 0.0465 & 0.4976 & 0.0264 \\ \hline
RN       & 0.1485 & 0.2342 & 0.1818 & 0.5095 & 0.2381 & 0.0509 & 0.0889 & 0.0648 & 0.5048 & 0.0479 & 0.0061 & 0.1731 & 0.0118 & 0.4195 & 0.0347 \\ \hline
RAE      & 0.1509 & 0.2281 & 0.1816 & 0.5449 & 0.2381 & 0.0938 & 0.3142 & 0.1444 & 0.6143 & 0.1634 & 0.1294 & 0.0885 & 0.1051 & 0.6491 & 0.0476 \\
\bottomrule
\end{tabular}
\end{table*}

\begin{table*}[]
\small
\caption{Univariate vs. Multivariate. Sliding window size=32, Hidden size=32.}
\begin{tabular}{llllllllllllllll}
\hline
         & \multicolumn{5}{l}{NAB}                    & \multicolumn{5}{l}{S5}                     & \multicolumn{5}{l}{KPI}                    \\ \hline
         & Pre    & Rec    & F1     & ROC    & PR     & Pre    & Rec    & F1     & ROC    & PR     & Pre    & Rec    & F1     & ROC    & PR     \\ \hline
CNN-AE   & 0.1475 & 0.1882 & 0.1654 & 0.5318 & 0.2374 & 0.2305 & 0.2229 & 0.2266 & 0.7446 & 0.4291 & 0.9696 & 0.0372 & 0.0716 & 0.4004 & 0.0469 \\ \hline
2DCNN-AE & 0.2301 & 0.161  & 0.1895 & 0.5565 & 0.2611 & 0.3019 & 0.3723 & 0.3334 & 0.8896 & 0.5545 & 0.0345 & 0.0357 & 0.0351 & 0.6034 & 0.0188 \\ \hline
LSTM-AE  & 0.1346 & 0.135  & 0.1348 & 0.5214 & 0.2091 & 0.2095 & 0.2273 & 0.2181 & 0.5821 & 0.3673 & 0.065  & 0.0374 & 0.0475 & 0.6041 & 0.0239 \\ \hline
HIFI     & 0.1708 & 0.1971 & 0.183  & 0.5664 & 0.2657 & 0.1875 & 0.2542 & 0.2158 & 0.751  & 0.3903 & 0.5777 & 0.0116 & 0.0227 & 0.4043 & 0.0208 \\ \hline
RN       & 0.128  & 0.0939 & 0.1083 & 0.4699 & 0.2025 & 0.0402 & 0.0705 & 0.0512 & 0.5031 & 0.0509 & 0.0061 & 0.1729 & 0.0118 & 0.4198 & 0.0334 \\ \hline
RAE      & 0.1495 & 0.1942 & 0.169  & 0.5148 & 0.2191 & 0.121  & 0.1074 & 0.1138 & 0.6008 & 0.1948 & 0.2027 & 0.1023 & 0.1359 & 0.6983 & 0.0656 \\
\bottomrule
\end{tabular}
\end{table*}

\begin{table*}[]
\small
\caption{Results of DL Methods (Univariate vs. Multivariate; Sliding window size=32, Hidden size=64)}
\begin{tabular}{llllllllllllllll}
\hline
         & \multicolumn{5}{l}{NAB}                    & \multicolumn{5}{l}{S5}                     & \multicolumn{5}{l}{KPI}                    \\ \hline
         & Pre    & Rec    & F1     & ROC    & PR     & Pre    & Rec    & F1     & ROC    & PR     & Pre    & Rec    & F1     & ROC    & PR     \\ \hline
CNN-AE   & 0.1317 & 0.1937 & 0.1568 & 0.5291 & 0.2349 & 0.2271 & 0.2171 & 0.222  & 0.7306 & 0.4253 & 0.0293 & 0.0569 & 0.0387 & 0.5774 & 0.0193 \\ \hline
2DCNN-AE & 0.2535 & 0.2198 & 0.2355 & 0.593  & 0.2784 & 0.2053 & 0.3308 & 0.2534 & 0.8554 & 0.5233 & 0.0723 & 0.0391 & 0.0508 & 0.5238 & 0.0358 \\ \hline
LSTM-AE  & 0.1338 & 0.1403 & 0.137  & 0.51   & 0.2007 & 0.1968 & 0.2069 & 0.2017 & 0.5736 & 0.3465 & 0.0482 & 0.034  & 0.0399 & 0.6023 & 0.0208 \\ \hline
HIFI     & 0.16   & 0.1711 & 0.1654 & 0.5673 & 0.258  & 0.1525 & 0.2197 & 0.18   & 0.7054 & 0.2884 & 0.0108 & 0.0241 & 0.0149 & 0.4652 & 0.0102 \\ \hline
RN       & 0.1318 & 0.0973 & 0.112  & 0.4647 & 0.1999 & 0.0314 & 0.0804 & 0.0451 & 0.3751 & 0.0461 & 0.0082 & 0.0401 & 0.0137 & 0.4145 & 0.031  \\ \hline
RAE      & 0.1469 & 0.1479 & 0.1474 & 0.5132 & 0.2283 & 0.2016 & 0.2121 & 0.2067 & 0.6869 & 0.2836 & 0.0062 & 0.1829 & 0.012  & 0.4247 & 0.0348 \\
\bottomrule
\end{tabular}
\end{table*}

\begin{table*}[]
\small
\caption{Univariate vs. Multivariate. Sliding window size=32, Hidden size=128.}
\begin{tabular}{llllllllllllllll}
\hline
         & \multicolumn{5}{l}{NAB}                    & \multicolumn{5}{l}{S5}                     & \multicolumn{5}{l}{KPI}                    \\ \hline
         & Pre    & Rec    & F1     & ROC    & PR     & Pre    & Rec    & F1     & ROC    & PR     & Pre    & Rec    & F1     & ROC    & PR     \\ \hline
CNN-AE   & 0.1467 & 0.2312 & 0.1795 & 0.5411 & 0.251  & 0.2356 & 0.235  & 0.2353 & 0.7477 & 0.4508 & 0.3963 & 0.0379 & 0.0691 & 0.4837 & 0.0505 \\ \hline
2DCNN-AE & 0.1667 & 0.1556 & 0.1609 & 0.5511 & 0.2449 & 0.2584 & 0.2963 & 0.276  & 0.8252 & 0.4313 & 0.0499 & 0.0244 & 0.0328 & 0.6098 & 0.0197 \\ \hline
LSTM-AE  & 0.1272 & 0.1241 & 0.1256 & 0.5104 & 0.2063 & 0.1876 & 0.2586 & 0.2174 & 0.6728 & 0.4233 & 0.0503 & 0.0362 & 0.0421 & 0.6024 & 0.021  \\ \hline
HIFI     & 0.1571 & 0.1722 & 0.1643 & 0.538  & 0.2427 & 0.1188 & 0.1711 & 0.1402 & 0.6583 & 0.2293 & 0.0119 & 0.0565 & 0.0196 & 0.495  & 0.0113 \\ \hline
RN       & 0.1363 & 0.1698 & 0.1513 & 0.4981 & 0.2567 & 0.0801 & 0.2253 & 0.1182 & 0.5528 & 0.0872 & 0.0124 & 0.0078 & 0.0095 & 0.576  & 0.0138 \\ \hline
RAE      & 0.1572 & 0.1527 & 0.1549 & 0.5375 & 0.2325 & 0.1534 & 0.1581 & 0.1557 & 0.6456 & 0.2187 & 0.0065 & 0.1954 & 0.0127 & 0.4279 & 0.0348 \\
\bottomrule
\end{tabular}
\end{table*}

\begin{table*}[]
\small
\caption{Results of DL Methods (Univariate vs. Multivariate; Sliding window size=32, Hidden size=256)}
\begin{tabular}{llllllllllllllll}
\hline
         & \multicolumn{5}{l}{NAB}                    & \multicolumn{5}{l}{S5}                     & \multicolumn{5}{l}{KPI}                    \\ \hline
         & Pre    & Rec    & F1     & ROC    & PR     & Pre    & Rec    & F1     & ROC    & PR     & Pre    & Rec    & F1     & ROC    & PR     \\ \hline
CNN-AE   & 0.2414 & 0.1836 & 0.2086 & 0.5459 & 0.2535 & 0.1953 & 0.2107 & 0.2027 & 0.7158 & 0.3949 & 0.0625 & 0.0647 & 0.0636 & 0.6187 & 0.0421 \\ \hline
2DCNN-AE & 0.1564 & 0.2006 & 0.1758 & 0.5633 & 0.2431 & 0.204  & 0.3161 & 0.2479 & 0.8626 & 0.5412 & 0.0579 & 0.0366 & 0.0449 & 0.6171 & 0.0304 \\ \hline
LSTM-AE  & 0.1314 & 0.1275 & 0.1294 & 0.4969 & 0.2027 & 0.166  & 0.2433 & 0.1974 & 0.6418 & 0.3772 & 0.0499 & 0.0368 & 0.0424 & 0.6023 & 0.021  \\ \hline
HIFI     & 0.1503 & 0.1562 & 0.1532 & 0.5361 & 0.2343 & 0.1281 & 0.1928 & 0.154  & 0.6475 & 0.2554 & 0.0183 & 0.1229 & 0.0319 & 0.535  & 0.0149 \\ \hline
RN       & 0.1436 & 0.2379 & 0.1791 & 0.5156 & 0.2111 & 0.0813 & 0.2167 & 0.1182 & 0.5598 & 0.04   & 0.0124 & 0.0078 & 0.0095 & 0.576  & 0.0138 \\ \hline
RAE      & 0.1482 & 0.1853 & 0.1647 & 0.5206 & 0.2359 & 0.1342 & 0.4466 & 0.2063 & 0.6284 & 0.2167 & 0.052  & 0.0622 & 0.0566 & 0.6369 & 0.0239 \\
\bottomrule
\end{tabular}
\end{table*}

\begin{table*}[]
\small
\caption{Results of DL Methods (Univariate vs. Multivariate; Sliding window size=64, Hidden size=16)}
\begin{tabular}{llllllllllllllll}
\hline
         & \multicolumn{5}{l}{NAB}                    & \multicolumn{5}{l}{S5}                     & \multicolumn{5}{l}{KPI}                    \\ \hline
         & Pre    & Rec    & F1     & ROC    & PR     & Pre    & Rec    & F1     & ROC    & PR     & Pre    & Rec    & F1     & ROC    & PR     \\ \hline
CNN-AE   & 0.1704 & 0.1947 & 0.1818 & 0.5177 & 0.2417 & 0.1994 & 0.1967 & 0.198  & 0.7102 & 0.3922 & 0.0064 & 0.1801 & 0.0123 & 0.4045 & 0.0436 \\ \hline
2DCNN-AE & 0.1815 & 0.2388 & 0.2062 & 0.5587 & 0.2615 & 0.3414 & 0.3404 & 0.3409 & 0.8476 & 0.5205 & 0.0428 & 0.0278 & 0.0337 & 0.603  & 0.0194 \\ \hline
LSTM-AE  & 0.159  & 0.0804 & 0.1068 & 0.5019 & 0.222  & 0.1708 & 0.3404 & 0.2275 & 0.4382 & 0.2657 & 0.9182 & 0.0098 & 0.0194 & 0.4574 & 0.0381 \\ \hline
HIFI     & 0.2194 & 0.1847 & 0.2005 & 0.5571 & 0.2762 & 0.1651 & 0.2043 & 0.1826 & 0.7097 & 0.2948 & 0.0786 & 0.0305 & 0.0439 & 0.4647 & 0.0306 \\ \hline
RN       & 0.1262 & 0.1    & 0.1116 & 0.4581 & 0.2135 & 0.0499 & 0.0346 & 0.0409 & 0.4914 & 0.0464 & 0.006  & 0.1535 & 0.0115 & 0.4174 & 0.0289 \\ \hline
RAE      & 0.2127 & 0.1349 & 0.1651 & 0.5109 & 0.2303 & 0.0589 & 0.2018 & 0.0912 & 0.5747 & 0.1388 & 0.0938 & 0.0971 & 0.0954 & 0.6726 & 0.0409 \\
\bottomrule
\end{tabular}
\end{table*}

\begin{table*}[]
\small
\caption{Results of DL Methods (Univariate vs. Multivariate; Sliding window size=64, Hidden size=32)}
\begin{tabular}{llllllllllllllll}
\hline
         & \multicolumn{5}{l}{NAB}                    & \multicolumn{5}{l}{S5}                     & \multicolumn{5}{l}{KPI}                    \\ \hline
         & Pre    & Rec    & F1     & ROC    & PR     & Pre    & Rec    & F1     & ROC    & PR     & Pre    & Rec    & F1     & ROC    & PR     \\ \hline
CNN-AE   & 0.2002 & 0.1885 & 0.1942 & 0.5102 & 0.2219 & 0.2114 & 0.2031 & 0.2072 & 0.7322 & 0.3918 & 0.0065 & 0.192  & 0.0126 & 0.4552 & 0.0442 \\ \hline
2DCNN-AE & 0.2045 & 0.2703 & 0.2329 & 0.5587 & 0.2657 & 0.3027 & 0.3346 & 0.3179 & 0.8561 & 0.494  & 0.0357 & 0.0323 & 0.0339 & 0.6031 & 0.019  \\ \hline
LSTM-AE  & 0.1668 & 0.0778 & 0.1061 & 0.4728 & 0.1943 & 0.1551 & 0.258  & 0.1938 & 0.3534 & 0.21   & 0.0429 & 0.0317 & 0.0364 & 0.6034 & 0.0198 \\ \hline
HIFI     & 0.2184 & 0.1929 & 0.2048 & 0.5609 & 0.2845 & 0.1199 & 0.1641 & 0.1386 & 0.6539 & 0.2448 & 0.0391 & 0.0643 & 0.0486 & 0.6279 & 0.0258 \\ \hline
RN       & 0.1346 & 0.0967 & 0.1126 & 0.4558 & 0.2066 & 0.0186 & 0.0482 & 0.0269 & 0.3925 & 0.0436 & 0.006  & 0.1297 & 0.0114 & 0.431  & 0.0289 \\ \hline
RAE      & 0.1271 & 0.1268 & 0.127  & 0.4662 & 0.1989 & 0.0726 & 0.2459 & 0.1121 & 0.5541 & 0.1401 & 0.0081 & 0.2446 & 0.0156 & 0.4398 & 0.0105 \\
\bottomrule
\end{tabular}
\end{table*}

\begin{table*}[]
\small
\caption{Results of DL Methods (Univariate vs. Multivariate; Sliding window size=64, Hidden size=64)}
\begin{tabular}{llllllllllllllll}
\hline
         & \multicolumn{5}{l}{NAB}                    & \multicolumn{5}{l}{S5}                     & \multicolumn{5}{l}{KPI}                    \\ \hline
         & Pre    & Rec    & F1     & ROC    & PR     & Pre    & Rec    & F1     & ROC    & PR     & Pre    & Rec    & F1     & ROC    & PR     \\ \hline
CNN-AE   & 0.1756 & 0.1854 & 0.1803 & 0.5048 & 0.231  & 0.1976 & 0.189  & 0.1932 & 0.7132 & 0.3711 & 0.0065 & 0.1871 & 0.0126 & 0.432  & 0.046  \\ \hline
2DCNN-AE & 0.2405 & 0.3063 & 0.2695 & 0.5834 & 0.2915 & 0.1974 & 0.2867 & 0.2339 & 0.8143 & 0.3897 & 0.0499 & 0.057  & 0.0532 & 0.5546 & 0.0382 \\ \hline
LSTM-AE  & 0.1536 & 0.0996 & 0.1208 & 0.5047 & 0.2055 & 0.2104 & 0.1347 & 0.1643 & 0.3611 & 0.1863 & 0.0431 & 0.0326 & 0.0372 & 0.6038 & 0.0199 \\ \hline
HIFI     & 0.2156 & 0.1888 & 0.2013 & 0.5523 & 0.2582 & 0.104  & 0.1558 & 0.1247 & 0.6664 & 0.1706 & 0.042  & 0.0778 & 0.0545 & 0.5595 & 0.0367 \\ \hline
RN       & 0.1391 & 0.1933 & 0.1618 & 0.4963 & 0.22   & 0.0503 & 0.1386 & 0.0738 & 0.5053 & 0.0711 & 0.0132 & 0.0083 & 0.0101 & 0.5618 & 0.0131 \\ \hline
RAE      & 0.1425 & 0.1545 & 0.1482 & 0.4948 & 0.2086 & 0.0984 & 0.1087 & 0.1033 & 0.5661 & 0.143  & 0.0171 & 0.0168 & 0.017  & 0.4825 & 0.011  \\
\bottomrule
\end{tabular}
\end{table*}

\begin{table*}[]
\small
\caption{Results of DL Methods (Univariate vs. Multivariate; Sliding window size=64, Hidden size=128)}
\begin{tabular}{llllllllllllllll}
\hline
         & \multicolumn{5}{l}{NAB}                    & \multicolumn{5}{l}{S5}                     & \multicolumn{5}{l}{KPI}                    \\ \hline
         & Pre    & Rec    & F1     & ROC    & PR     & Pre    & Rec    & F1     & ROC    & PR     & Pre    & Rec    & F1     & ROC    & PR     \\ \hline
CNN-AE   & 0.1896 & 0.1885 & 0.1891 & 0.5124 & 0.2302 & 0.1792 & 0.1826 & 0.1809 & 0.7228 & 0.3701 & 0.9707 & 0.0354 & 0.0682 & 0.3995 & 0.044  \\ \hline
2DCNN-AE & 0.177  & 0.1692 & 0.173  & 0.5367 & 0.2329 & 0.2111 & 0.2401 & 0.2247 & 0.8071 & 0.3265 & 0.0437 & 0.0266 & 0.0331 & 0.6043 & 0.0194 \\ \hline
LSTM-AE  & 0.1376 & 0.0925 & 0.1106 & 0.458  & 0.1847 & 0.2232 & 0.1571 & 0.1844 & 0.4255 & 0.2548 & 0.0443 & 0.0306 & 0.0362 & 0.6033 & 0.02   \\ \hline
HIFI     & 0.1806 & 0.1566 & 0.1677 & 0.5264 & 0.2361 & 0.0884 & 0.1315 & 0.1057 & 0.6236 & 0.1307 & 0.0375 & 0.0752 & 0.0501 & 0.5896 & 0.0322 \\ \hline
RN       & 0.1371 & 0.2263 & 0.1707 & 0.5069 & 0.2037 & 0.0453 & 0.1141 & 0.0648 & 0.557  & 0.033  & 0.0132 & 0.0083 & 0.0101 & 0.5618 & 0.0131 \\ \hline
RAE      & 0.1419 & 0.132  & 0.1368 & 0.5173 & 0.2311 & 0.0538 & 0.1902 & 0.0839 & 0.5282 & 0.1146 & 0.0893 & 0.0996 & 0.0942 & 0.6909 & 0.0446 \\
\bottomrule
\end{tabular}
\end{table*}

\begin{table*}[]
\small
\caption{Results of DL Methods (Univariate vs. Multivariate; Sliding window size=64, Hidden size=256)}
\begin{tabular}{llllllllllllllll}
\hline
         & \multicolumn{5}{l}{NAB}                    & \multicolumn{5}{l}{S5}                     & \multicolumn{5}{l}{KPI}                    \\ \hline
         & Pre    & Rec    & F1     & ROC    & PR     & Pre    & Rec    & F1     & ROC    & PR     & Pre    & Rec    & F1     & ROC    & PR     \\ \hline
CNN-AE   & 0.1808 & 0.2496 & 0.2097 & 0.5262 & 0.2581 & 0.1645 & 0.1705 & 0.1675 & 0.6856 & 0.2788 & 0.2607 & 0.1471 & 0.1881 & 0.5188 & 0.1067 \\ \hline
2DCNN-AE & 0.1647 & 0.2165 & 0.1871 & 0.5641 & 0.2528 & 0.1969 & 0.304  & 0.239  & 0.8401 & 0.44   & 0.0575 & 0.0363 & 0.0445 & 0.6134 & 0.0268 \\ \hline
LSTM-AE  & 0.1375 & 0.0905 & 0.1091 & 0.4898 & 0.1968 & 0.1925 & 0.1762 & 0.184  & 0.5367 & 0.3106 & 0.0469 & 0.0307 & 0.0371 & 0.6034 & 0.0201 \\ \hline
HIFI     & 0.1896 & 0.1469 & 0.1655 & 0.5511 & 0.2511 & 0.0902 & 0.1398 & 0.1097 & 0.6152 & 0.1483 & 0.0294 & 0.0246 & 0.0268 & 0.4005 & 0.0198 \\ \hline
RN       & -1     & -1     & -1     & -1     & -1     & -1     & -1     & -1     & -1     & -1     & -1     & -1     & -1     & -1     & -1     \\ \hline
RAE      & 0.2145 & 0.1212 & 0.1549 & 0.5321 & 0.2275 & 0.0818 & 0.0883 & 0.0849 & 0.5312 & 0.1235 & 0.0933 & 0.0891 & 0.0911 & 0.6893 & 0.0397 \\
\bottomrule
\end{tabular}
\end{table*}

\begin{table*}[]
\small
\caption{Results of DL Methods (Univariate vs. Multivariate; Sliding window size=128, Hidden size=16)}
\begin{tabular}{llllllllllllllll}
\hline
         & \multicolumn{5}{l}{NAB}                    & \multicolumn{5}{l}{S5}                     & \multicolumn{5}{l}{KPI}                    \\ \hline
         & Pre    & Rec    & F1     & ROC    & PR     & Pre    & Rec    & F1     & ROC    & PR     & Pre    & Rec    & F1     & ROC    & PR     \\ \hline
CNN-AE   & 0.4084 & 0.1233 & 0.1894 & 0.4964 & 0.2731 & 0.1141 & 0.1181 & 0.1161 & 0.6126 & 0.2062 & 0.0065 & 0.1862 & 0.0126 & 0.4293 & 0.0497 \\ \hline
2DCNN-AE & 0.2734 & 0.2097 & 0.2373 & 0.5715 & 0.2718 & 0.3872 & 0.3595 & 0.3728 & 0.8471 & 0.5596 & 0.0433 & 0.0288 & 0.0346 & 0.6035 & 0.0195 \\ \hline
LSTM-AE  & 0.1201 & 0.0778 & 0.0944 & 0.4892 & 0.1923 & 0.1724 & 0.251  & 0.2044 & 0.3765 & 0.2304 & 0.0066 & 0.1897 & 0.0128 & 0.4233 & 0.0387 \\ \hline
HIFI     & 0.2051 & 0.1713 & 0.1867 & 0.5573 & 0.2725 & 0.153  & 0.189  & 0.1691 & 0.6856 & 0.2706 & 0.7309 & 0.0177 & 0.0345 & 0.3761 & 0.0373 \\ \hline
RN       & 0.1552 & 0.09   & 0.1139 & 0.4441 & 0.1858 & 0.0195 & 0.0113 & 0.0143 & 0.4232 & 0.0327 & 0      & 0      & 0      & 0.4401 & 0.0114 \\ \hline
RAE      & 0.1785 & 0.1015 & 0.1294 & 0.4956 & 0.2076 & 0.0333 & 0.1408 & 0.0539 & 0.5432 & 0.0809 & 0.0092 & 0.1181 & 0.0171 & 0.4591 & 0.0121 \\
\bottomrule
\end{tabular}
\end{table*}

\begin{table*}[]
\small
\caption{Results of DL Methods (Univariate vs. Multivariate; Sliding window size=128, Hidden size=32)}
\begin{tabular}{llllllllllllllll}
\hline
         & \multicolumn{5}{l}{NAB}                    & \multicolumn{5}{l}{S5}                     & \multicolumn{5}{l}{KPI}                    \\ \hline
         & Pre    & Rec    & F1     & ROC    & PR     & Pre    & Rec    & F1     & ROC    & PR     & Pre    & Rec    & F1     & ROC    & PR     \\ \hline
CNN-AE   & 0.1647 & 0.1485 & 0.1562 & 0.4738 & 0.2106 & 0.1026 & 0.1066 & 0.1046 & 0.6147 & 0.2042 & 0.0304 & 0.055  & 0.0391 & 0.5152 & 0.0172 \\ \hline
2DCNN-AE & 0.2402 & 0.1969 & 0.2164 & 0.5673 & 0.2778 & 0.2977 & 0.3059 & 0.3017 & 0.8213 & 0.4612 & 0.0399 & 0.0308 & 0.0347 & 0.6038 & 0.0194 \\ \hline
LSTM-AE  & 0.1435 & 0.0989 & 0.1171 & 0.4715 & 0.2032 & 0.1568 & 0.2542 & 0.1939 & 0.3615 & 0.1914 & 0.0066 & 0.1951 & 0.0128 & 0.4481 & 0.0381 \\ \hline
HIFI     & 0.3228 & 0.1319 & 0.1873 & 0.5501 & 0.2648 & 0.0966 & 0.1507 & 0.1177 & 0.631  & 0.1839 & 0.1927 & 0.0445 & 0.0723 & 0.4187 & 0.0355 \\ \hline
RN       & 0.1454 & 0.1942 & 0.1663 & 0.5091 & 0.2373 & 0.0512 & 0.0958 & 0.0667 & 0.5133 & 0.0524 & 0.0098 & 0.0061 & 0.0075 & 0.5356 & 0.012  \\ \hline
RAE      & 0.1487 & 0.0798 & 0.1039 & 0.4664 & 0.1821 & 0.0429 & 0.1676 & 0.0684 & 0.6105 & 0.118  & 0.0137 & 0.0755 & 0.0231 & 0.5261 & 0.0122 \\
\bottomrule
\end{tabular}
\end{table*}

\begin{table*}[]
\small
\caption{Results of DL Methods (Univariate vs. Multivariate; Sliding window size=128, Hidden size=64)}
\begin{tabular}{llllllllllllllll}
\hline
         & \multicolumn{5}{l}{NAB}                    & \multicolumn{5}{l}{S5}                     & \multicolumn{5}{l}{KPI}                    \\ \hline
         & Pre    & Rec    & F1     & ROC    & PR     & Pre    & Rec    & F1     & ROC    & PR     & Pre    & Rec    & F1     & ROC    & PR     \\ \hline
CNN-AE   & 0.1528 & 0.1676 & 0.1599 & 0.4877 & 0.2272 & 0.0945 & 0.1137 & 0.1032 & 0.6147 & 0.1985 & 0.0284 & 0.0588 & 0.0383 & 0.5768 & 0.0195 \\ \hline
2DCNN-AE & 0.3577 & 0.2556 & 0.2981 & 0.6286 & 0.3101 & 0.1859 & 0.2746 & 0.2217 & 0.7851 & 0.3759 & 0.0803 & 0.0507 & 0.0622 & 0.5376 & 0.0396 \\ \hline
LSTM-AE  & 0.1744 & 0.0829 & 0.1124 & 0.4693 & 0.1958 & 0.1044 & 0.1731 & 0.1302 & 0.3235 & 0.1775 & 1      & 0.0173 & 0.034  & 0.3953 & 0.0474 \\ \hline
HIFI     & 0.1714 & 0.1405 & 0.1545 & 0.5471 & 0.2484 & 0.0613 & 0.1054 & 0.0775 & 0.5999 & 0.1299 & 0.0512 & 0.0362 & 0.0424 & 0.4459 & 0.0277 \\ \hline
RN       & 0.1216 & 0.2025 & 0.152  & 0.4899 & 0.1868 & 0.0362 & 0.0901 & 0.0517 & 0.5214 & 0.026  & 0.0098 & 0.0061 & 0.0075 & 0.5356 & 0.012  \\ \hline
RAE      & 0.1584 & 0.1102 & 0.13   & 0.4927 & 0.1949 & 0.0512 & 0.0683 & 0.0586 & 0.5416 & 0.1126 & 0.0075 & 0.1999 & 0.0144 & 0.4484 & 0.0095 \\
\bottomrule
\end{tabular}
\end{table*}

\begin{table*}[]
\small
\caption{Results of DL Methods (Univariate vs. Multivariate; Sliding window size=128, Hidden size=128)}
\begin{tabular}{llllllllllllllll}
\hline
         & \multicolumn{5}{l}{NAB}                    & \multicolumn{5}{l}{S5}                     & \multicolumn{5}{l}{KPI}                    \\ \hline
         & Pre    & Rec    & F1     & ROC    & PR     & Pre    & Rec    & F1     & ROC    & PR     & Pre    & Rec    & F1     & ROC    & PR     \\ \hline
CNN-AE   & 0.1519 & 0.14   & 0.1457 & 0.4835 & 0.2157 & 0.0822 & 0.1028 & 0.0914 & 0.5925 & 0.1791 & 0.028  & 0.0592 & 0.0381 & 0.5211 & 0.0177 \\ \hline
2DCNN-AE & 0.1589 & 0.2333 & 0.1891 & 0.5684 & 0.2513 & 0.194  & 0.2197 & 0.206  & 0.7449 & 0.277  & 0.0454 & 0.0262 & 0.0332 & 0.603  & 0.0196 \\ \hline
LSTM-AE  & 0.1331 & 0.0805 & 0.1003 & 0.4892 & 0.1975 & 0.1885 & 0.1213 & 0.1476 & 0.347  & 0.199  & 1      & 0.0126 & 0.0249 & 0.3953 & 0.0477 \\ \hline
HIFI     & 0.1855 & 0.133  & 0.1549 & 0.56   & 0.2575 & 0.0521 & 0.092  & 0.0665 & 0.5943 & 0.1183 & 0.0064 & 0.1814 & 0.0124 & 0.4307 & 0.0236 \\ \hline
RN       & -1     & -1     & -1     & -1     & -1     & -1     & -1     & -1     & -1     & -1     & -1     & -1     & -1     & -1     & -1     \\ \hline
RAE      & 0.1589 & 0.0915 & 0.1161 & 0.498  & 0.1885 & 0.0412 & 0.0613 & 0.0492 & 0.524  & 0.0817 & 0.0096 & 0.0917 & 0.0174 & 0.4759 & 0.0106 \\
\bottomrule
\end{tabular}
\end{table*}

\begin{table*}[]
\small
\caption{Results of DL Methods (Univariate vs. Multivariate; Sliding window size=128, Hidden size=256)}
\begin{tabular}{llllllllllllllll}
\hline
         & \multicolumn{5}{l}{NAB}                    & \multicolumn{5}{l}{S5}                     & \multicolumn{5}{l}{KPI}                    \\ \hline
         & Pre    & Rec    & F1     & ROC    & PR     & Pre    & Rec    & F1     & ROC    & PR     & Pre    & Rec    & F1     & ROC    & PR     \\ \hline
CNN-AE   & 0.1882 & 0.2223 & 0.2039 & 0.5568 & 0.2516 & 0.1221 & 0.1315 & 0.1267 & 0.6227 & 0.2236 & 0.0411 & 0.0316 & 0.0357 & 0.5033 & 0.0171 \\ \hline
2DCNN-AE & 0.2194 & 0.1633 & 0.1873 & 0.5729 & 0.2546 & 0.1847 & 0.2612 & 0.2164 & 0.7893 & 0.3843 & 0.0443 & 0.028  & 0.0343 & 0.6034 & 0.0196 \\ \hline
LSTM-AE  & 0.1399 & 0.0721 & 0.0951 & 0.4617 & 0.1909 & 0.149  & 0.3359 & 0.2064 & 0.4861 & 0.2915 & 1      & 0.0206 & 0.0403 & 0.3956 & 0.0451 \\ \hline
RN       & -1     & -1     & -1     & -1     & -1     & -1     & -1     & -1     & -1     & -1     & -1     & -1     & -1     & -1     & -1     \\ \hline
RAE      & 0.1504 & 0.1052 & 0.1238 & 0.5258 & 0.2095 & 0.0481 & 0.0704 & 0.0572 & 0.5383 & 0.1102 & 0.0085 & 0.1112 & 0.0158 & 0.4587 & 0.0101 \\ \hline
Omni     & 0.1371 & 0.121  & 0.1285 & 0.5048 & 0.2028 & 0.1558 & 0.4368 & 0.2296 & 0.4665 & 0.2516 & 0.0064 & 0.1814 & 0.0124 & 0.4059 & 0.0463 \\
\bottomrule
\end{tabular}
\end{table*}

\begin{table*}[]
\small
\caption{Results of DL Methods (Univariate vs. Multivariate; Sliding window size=256, Hidden size=16)}
\begin{tabular}{llllllllllllllll}
\hline
         & \multicolumn{5}{l}{NAB}                    & \multicolumn{5}{l}{S5}                     & \multicolumn{5}{l}{KPI}                    \\ \hline
         & Pre    & Rec    & F1     & ROC    & PR     & Pre    & Rec    & F1     & ROC    & PR     & Pre    & Rec    & F1     & ROC    & PR     \\ \hline
CNN-AE   & 0.1508 & 0.0077 & 0.0147 & 0.4564 & 0.4267 & 0.0707 & 0.0755 & 0.073  & 0.5675 & 0.1171 & 0      & 0      & 0      & 0.5    & 0.5057 \\ \hline
2DCNN-AE & 0.1828 & 0.261  & 0.215  & 0.5544 & 0.2566 & 0.3461 & 0.3636 & 0.3547 & 0.856  & 0.5427 & 0.0435 & 0.0286 & 0.0345 & 0.6037 & 0.0196 \\ \hline
LSTM-AE  & 0.1497 & 0.0849 & 0.1084 & 0.4948 & 0.2039 & 0.2204 & 0.2785 & 0.246  & 0.2866 & 0.1602 & 0.0451 & 0.0283 & 0.0348 & 0.6029 & 0.0198 \\ \hline
HIFI     & 0.2768 & 0.1416 & 0.1874 & 0.5709 & 0.2814 & 0.129  & 0.1498 & 0.1386 & 0.6438 & 0.2098 & 0.2229 & 0.0342 & 0.0593 & 0.3994 & 0.0316 \\ \hline
RN       & 0.1016 & 0.1835 & 0.1308 & 0.5298 & 0.2043 & 0.0511 & 0.0542 & 0.0526 & 0.495  & 0.0591 & 0.0163 & 0.0102 & 0.0125 & 0.5062 & 0.0112 \\ \hline
RAE      & 0.1084 & 0.1249 & 0.116  & 0.4913 & 0.1814 & 0.0649 & 0.2308 & 0.1013 & 0.3346 & 0.1482 & 0.0155 & 0.031  & 0.0207 & 0.5441 & 0.0125 \\
\bottomrule
\end{tabular}
\end{table*}

\begin{table*}[]
\small
\caption{Results of DL Methods (Univariate vs. Multivariate; Sliding window size=256, Hidden size=32)}
\begin{tabular}{llllllllllllllll}
\hline
         & \multicolumn{5}{l}{NAB}                    & \multicolumn{5}{l}{S5}                     & \multicolumn{5}{l}{KPI}                    \\ \hline
         & Pre    & Rec    & F1     & ROC    & PR     & Pre    & Rec    & F1     & ROC    & PR     & Pre    & Rec    & F1     & ROC    & PR     \\ \hline
CNN-AE   & 0.1613 & 0.0809 & 0.1077 & 0.48   & 0.227  & 0.0514 & 0.0787 & 0.0622 & 0.6068 & 0.1211 & 1      & 0.0252 & 0.0491 & 0.3963 & 0.0477 \\ \hline
2DCNN-AE & 0.2226 & 0.3056 & 0.2576 & 0.5688 & 0.2635 & 0.2727 & 0.3118 & 0.2909 & 0.8379 & 0.4709 & 0.0436 & 0.0282 & 0.0342 & 0.6035 & 0.0195 \\ \hline
LSTM-AE  & 0.1701 & 0.0982 & 0.1245 & 0.4933 & 0.2076 & 0.2869 & 0.1146 & 0.1638 & 0.2995 & 0.1749 & 0.045  & 0.0286 & 0.035  & 0.6033 & 0.0198 \\ \hline
HIFI     & 0.2602 & 0.1067 & 0.1514 & 0.5437 & 0.2553 & 0.0703 & 0.1114 & 0.0862 & 0.6168 & 0.1391 & 0.3063 & 0.0202 & 0.0379 & 0.3301 & 0.0313 \\ \hline
RN       & 0.0964 & 0.1647 & 0.1216 & 0.4942 & 0.174  & 0.0332 & 0.0739 & 0.0459 & 0.4662 & 0.0299 & 0.0163 & 0.0102 & 0.0125 & 0.5062 & 0.0112 \\ \hline
RAE      & 0.1538 & 0.0733 & 0.0993 & 0.4746 & 0.173  & 0.0304 & 0.049  & 0.0375 & 0.5297 & 0.0933 & 0.0086 & 0.0284 & 0.0132 & 0.5028 & 0.0112 \\
\bottomrule
\end{tabular}
\end{table*}

\clearpage
\begin{table*}[]
\small
\caption{Results of DL Methods (Univariate vs. Multivariate; Sliding window size=256, Hidden size=64)}
\begin{tabular}{llllllllllllllll}
\hline
         & \multicolumn{5}{l}{NAB}                    & \multicolumn{5}{l}{S5}                     & \multicolumn{5}{l}{KPI}                    \\ \hline
         & Pre    & Rec    & F1     & ROC    & PR     & Pre    & Rec    & F1     & ROC    & PR     & Pre    & Rec    & F1     & ROC    & PR     \\ \hline
CNN-AE   & 0.1716 & 0.1045 & 0.1299 & 0.4852 & 0.2279 & 0.0791 & 0.0922 & 0.0852 & 0.6332 & 0.1408 & 0.0649 & 0.0644 & 0.0647 & 0.4973 & 0.019  \\ \hline
2DCNN-AE & 0.2707 & 0.2852 & 0.2778 & 0.6222 & 0.2966 & 0.1947 & 0.2298 & 0.2108 & 0.7675 & 0.2767 & 0.2267 & 0.1674 & 0.1926 & 0.6219 & 0.1313 \\ \hline
LSTM-AE  & 0.1605 & 0.0885 & 0.1141 & 0.5013 & 0.2035 & 0.1651 & 0.2036 & 0.1823 & 0.2639 & 0.1399 & 0.0445 & 0.029  & 0.0351 & 0.6035 & 0.0197 \\ \hline
HIFI     & 0.2494 & 0.1142 & 0.1566 & 0.5007 & 0.2321 & 0.0528 & 0.0992 & 0.0689 & 0.6226 & 0.112  & 0.0102 & 0.0447 & 0.0166 & 0.4053 & 0.0298 \\ \hline
RN       & -1     & -1     & -1     & -1     & -1     & -1     & -1     & -1     & -1     & -1     & -1     & -1     & -1     & -1     & -1     \\ \hline
RAE      & 0.1115 & 0.0628 & 0.0804 & 0.5124 & 0.1805 & 0.0104 & 0.0757 & 0.0183 & 0.4932 & 0.0498 & 0.0107 & 0.0286 & 0.0156 & 0.4866 & 0.011  \\
\bottomrule
\end{tabular}
\end{table*}

\begin{table*}[]
\small
\caption{Results of DL Methods (Univariate vs. Multivariate; Sliding window size=256, Hidden size=128)}
\begin{tabular}{llllllllllllllll}
\hline
         & \multicolumn{5}{l}{NAB}                    & \multicolumn{5}{l}{S5}                     & \multicolumn{5}{l}{KPI}                    \\ \hline
         & Pre    & Rec    & F1     & ROC    & PR     & Pre    & Rec    & F1     & ROC    & PR     & Pre    & Rec    & F1     & ROC    & PR     \\ \hline
CNN-AE   & 0.2041 & 0.1144 & 0.1466 & 0.4724 & 0.2238 & 0.0667 & 0.1204 & 0.0858 & 0.6399 & 0.1404 & 0.0064 & 0.1826 & 0.0125 & 0.4051 & 0.0478 \\ \hline
2DCNN-AE & 0.1699 & 0.2676 & 0.2078 & 0.5617 & 0.2454 & 0.1945 & 0.2305 & 0.211  & 0.7786 & 0.2968 & 0.0438 & 0.0276 & 0.0339 & 0.6035 & 0.0196 \\ \hline
LSTM-AE  & 0.1603 & 0.1198 & 0.1371 & 0.5267 & 0.215  & 0.1547 & 0.2369 & 0.1872 & 0.3072 & 0.1804 & 0.0447 & 0.0286 & 0.0348 & 0.6035 & 0.0197 \\ \hline
HIFI     & 0.2211 & 0.1162 & 0.1524 & 0.5187 & 0.2289 & 0.0529 & 0.1031 & 0.0699 & 0.6485 & 0.1024 & 1      & 0.0162 & 0.0319 & 0.3993 & 0.0359 \\ \hline
RN       & -1     & -1     & -1     & -1     & -1     & -1     & -1     & -1     & -1     & -1     & -1     & -1     & -1     & -1     & -1     \\ \hline
RAE      & 0.1128 & 0.0357 & 0.0543 & 0.5448 & 0.1874 & 0.0111 & 0.0559 & 0.0185 & 0.4811 & 0.0423 & 0.0097 & 0.0629 & 0.0168 & 0.4772 & 0.0106 \\
\bottomrule
\end{tabular}
\end{table*}

\begin{table*}[]
\small
\caption{Results of DL Methods (Univariate vs. Multivariate; Sliding window size=256, Hidden size=256)}
\begin{tabular}{llllllllllllllll}
\hline
         & \multicolumn{5}{l}{NAB}                    & \multicolumn{5}{l}{S5}                     & \multicolumn{5}{l}{KPI}                    \\ \hline
         & Pre    & Rec    & F1     & ROC    & PR     & Pre    & Rec    & F1     & ROC    & PR     & Pre    & Rec    & F1     & ROC    & PR     \\ \hline
CNN-AE   & 0.284  & 0.22   & 0.2479 & 0.5528 & 0.2766 & 0.0597 & 0.089  & 0.0715 & 0.6135 & 0.1182 & 0.0066 & 0.1876 & 0.0128 & 0.4137 & 0.0431 \\ \hline
2DCNN-AE & 0.1711 & 0.2051 & 0.1866 & 0.5449 & 0.2243 & 0.2012 & 0.2414 & 0.2194 & 0.7882 & 0.3123 & 0.0443 & 0.0279 & 0.0342 & 0.6037 & 0.0196 \\ \hline
LSTM-AE  & 0.178  & 0.1083 & 0.1347 & 0.5167 & 0.2268 & 0.1399 & 0.2881 & 0.1883 & 0.4514 & 0.2454 & 0.0452 & 0.0284 & 0.0349 & 0.6034 & 0.0198 \\ \hline
HIFI     & 0.2494 & 0.1242 & 0.1658 & 0.543  & 0.2621 & 0.0477 & 0.1005 & 0.0647 & 0.6605 & 0.098  & 0.029  & 0.0587 & 0.0388 & 0.578  & 0.019  \\ \hline
RN       & -1     & -1     & -1     & -1     & -1     & -1     & -1     & -1     & -1     & -1     & -1     & -1     & -1     & -1     & -1     \\ \hline
RAE      & 0.13   & 0.0859 & 0.1034 & 0.5018 & 0.176  & 0.0081 & 0.0353 & 0.0131 & 0.5025 & 0.0423 & 0.0084 & 0.1593 & 0.0159 & 0.4602 & 0.0102 \\
\bottomrule
\end{tabular}
\end{table*}

\clearpage   \begin{table*}[]
\footnotesize
\caption{Results of LOF (Stationary vs. Non-stationary)}
\begin{tabular}{llllllllllllllll}
\hline
             & \multicolumn{5}{l}{NAB-ambient-Nonstationary}  & \multicolumn{5}{l}{NAB-art-stationary}      & \multicolumn{5}{l}{NAB-cpu-stationary}        \\ \hline
n\_neighbors & Pre      & Rec     & F1      & ROC    & PR     & Pre     & Rec    & F1     & ROC    & PR     & Pre      & Rec     & F1     & ROC    & PR     \\ \hline
10           & 0.1444   & 0.0746  & 0.0984  & 0.4853 & 0.1616 & 0.1648  & 0.099  & 0.1237 & 0.2739 & 0.1893 & 0.2416   & 0.1969  & 0.217  & 0.4777 & 0.2625 \\ \hline
20           & 0.2477   & 0.0746  & 0.1146  & 0.4974 & 0.1846 & 0.1765  & 0.0951 & 0.1236 & 0.2716 & 0.1901 & 0.2585   & 0.1969  & 0.2236 & 0.517  & 0.2836 \\ \hline
30           & 0.1963   & 0.058   & 0.0896  & 0.516  & 0.1806 & 0.1789  & 0.0951 & 0.1242 & 0.2639 & 0.1836 & 0.3039   & 0.2503  & 0.2745 & 0.5226 & 0.2932 \\ \hline
40           & 0.2203   & 0.0718  & 0.1083  & 0.5372 & 0.1842 & 0.1939  & 0.1158 & 0.145  & 0.2588 & 0.1782 & 0.3081   & 0.2664  & 0.2857 & 0.5294 & 0.3019 \\ \hline
50           & 0.181    & 0.058   & 0.0879  & 0.5317 & 0.1848 & 0.1912  & 0.1278 & 0.1532 & 0.2592 & 0.1782 & 0.3199   & 0.2323  & 0.2691 & 0.5317 & 0.3094 \\ \hline
             & \multicolumn{5}{l}{NAB-ec2-Nonstationary}      & \multicolumn{5}{l}{NAB-ec2-stationary}      & \multicolumn{5}{l}{NAB-elb-stationary}        \\
\hline
\end{tabular}
\end{table*}

\clearpage   \begin{table*}[]
\footnotesize
\caption{Results of MP (Stationary vs. Non-stationary)}
\begin{tabular}{llllllllllllllll}
\hline
              & \multicolumn{5}{l}{NAB-ambient-Nonstationary}  & \multicolumn{5}{l}{NAB-art-stationary}      & \multicolumn{5}{l}{NAB-cpu-stationary}        \\ \hline
pattern\_size & Pre      & Rec     & F1      & ROC    & PR     & Pre     & Rec    & F1     & ROC    & PR     & Pre      & Rec     & F1     & ROC    & PR     \\ \hline
16            & 0        & 0       & 0       & 0.5679 & 0.2003 & 0.1598  & 0.1067 & 0.128  & 0.247  & 0.1983 & 0        & 0       & 0      & 0.2035 & 0.2321 \\ \hline
32            & 0        & 0       & 0       & 0.639  & 0.2277 & 0.1273  & 0.1449 & 0.1355 & 0.2228 & 0.1771 & 0        & 0       & 0      & 0.1721 & 0.1891 \\ \hline
64            & 0        & 0       & 0       & 0.6281 & 0.2326 & 0       & 0      & 0      & 0.1951 & 0.1492 & 0        & 0       & 0      & 0.1676 & 0.1778 \\ \hline
128           & 0        & 0       & 0       & 0.6297 & 0.255  & 0       & 0      & 0      & 0.1675 & 0.1304 & 0        & 0       & 0      & 0.1146 & 0.1655 \\ \hline
256           & 0        & 0       & 0       & 0.44   & 0.1551 & 0.0512  & 0.0136 & 0.0215 & 0.1296 & 0.0636 & 0        & 0       & 0      & 0.0996 & 0.1675 \\ \hline
              & \multicolumn{5}{l}{NAB-ec2-Nonstationary}      & \multicolumn{5}{l}{NAB-ec2-stationary}      & \multicolumn{5}{l}{NAB-elb-stationary}        \\
\hline
\end{tabular}
\end{table*}

\clearpage   \begin{table*}[]
\footnotesize
\caption{Results of ISF (Stationary vs. Non-stationary)}
\begin{tabular}{llllllllllllllll}
\hline
              & \multicolumn{5}{l}{NAB-ambient-Nonstationary}  & \multicolumn{5}{l}{NAB-art-stationary}      & \multicolumn{5}{l}{NAB-cpu-stationary}        \\ \hline
n\_estimators & Pre      & Rec     & F1      & ROC    & PR     & Pre     & Rec    & F1     & ROC    & PR     & Pre      & Rec     & F1     & ROC    & PR     \\ \hline
20            & 0.1866   & 0.2072  & 0.1963  & 0.4941 & 0.1961 & 0.2329  & 0.0817 & 0.121  & 0.2599 & 0.1628 & 0.7419   & 0.1228  & 0.2108 & 0.7882 & 0.5858 \\ \hline
40            & 0.2611   & 0.163   & 0.2007  & 0.5149 & 0.191  & 0.167   & 0.0769 & 0.1053 & 0.2613 & 0.1636 & 0.7143   & 0.1068  & 0.1858 & 0.7939 & 0.6005 \\ \hline
60            & 0.2551   & 0.1713  & 0.205   & 0.5358 & 0.1953 & 0.2272  & 0.0802 & 0.1186 & 0.2621 & 0.1656 & 0.7902   & 0.1609  & 0.2673 & 0.7894 & 0.5943 \\ \hline
80            & 0.1841   & 0.2238  & 0.202   & 0.539  & 0.2036 & 0.166   & 0.0812 & 0.1091 & 0.2625 & 0.1638 & 0.6114   & 0.2657  & 0.3704 & 0.793  & 0.5851 \\ \hline
100           & 0.2226   & 0.174   & 0.1953  & 0.5076 & 0.1932 & 0.1512  & 0.0798 & 0.1044 & 0.2638 & 0.1653 & 0.5449   & 0.1943  & 0.2864 & 0.7935 & 0.5721 \\ \hline
              & \multicolumn{5}{l}{NAB-ec2-Nonstationary}      & \multicolumn{5}{l}{NAB-ec2-stationary}      & \multicolumn{5}{l}{NAB-elb-stationary}        \\
\hline
\end{tabular}
\end{table*}

\clearpage   \begin{table*}[]
\footnotesize
\caption{Results of OC-SVM (Stationary vs. Non-stationary)}
\begin{tabular}{llllllllllllllll}
\hline
       & \multicolumn{5}{l}{NAB-ambient-Nonstationary}  & \multicolumn{5}{l}{NAB-art-stationary}      & \multicolumn{5}{l}{NAB-cpu-stationary}        \\ \hline
kernel & Pre      & Rec     & F1      & ROC    & PR     & Pre     & Rec    & F1     & ROC    & PR     & Pre      & Rec     & F1     & ROC    & PR     \\ \hline
rbf    & 0.3306   & 0.1105  & 0.1656  & 0.4935 & 0.1875 & 0.3529  & 0.0259 & 0.0483 & 0.2652 & 0.1732 & 0.6067   & 0.2049  & 0.3064 & 0.4073 & 0.3956 \\ \hline
       & \multicolumn{5}{l}{NAB-ec2-Nonstationary}      & \multicolumn{5}{l}{NAB-ec2-stationary}      & \multicolumn{5}{l}{NAB-elb-stationary}        \\ \hline
       & Pre      & Rec     & F1      & ROC    & PR     & Pre     & Rec    & F1     & ROC    & PR     & Pre      & Rec     & F1     & ROC    & PR     \\ \hline
rbf    & 0.7656   & 0.0814  & 0.1471  & 0.6032 & 0.4106 & 0.1964  & 0.1349 & 0.16   & 0.4726 & 0.1446 & 0.4074   & 0.055   & 0.0969 & 0.4409 & 0.1869 \\ \hline
       & \multicolumn{5}{l}{NAB-exchange-Nonstationary} & \multicolumn{5}{l}{NAB-exchange-stationary} & \multicolumn{5}{l}{NAB-grok-Nonstationary}    \\ \hline
       & Pre      & Rec     & F1      & ROC    & PR     & Pre     & Rec    & F1     & ROC    & PR     & Pre      & Rec     & F1     & ROC    & PR     \\
\hline
\end{tabular}
\end{table*}

\begin{table*}[]
\footnotesize
\caption{Results of DL Methods (Stationary vs. Non-stationary; Sliding window size=16, Hidden size=16)}
\begin{tabular}{llllllllllllllll}
\hline
         & \multicolumn{5}{l}{NAB-ambient-Nonstationary}  & \multicolumn{5}{l}{NAB-art-stationary}      & \multicolumn{5}{l}{NAB-cpu-stationary}        \\ \hline
         & Pre      & Rec     & F1      & ROC    & PR     & Pre     & Rec    & F1     & ROC    & PR     & Pre      & Rec     & F1     & ROC    & PR     \\ \hline
CNN-AE   & 0.1893   & 0.0884  & 0.1205  & 0.5754 & 0.1939 & 0.0818  & 0.2465 & 0.1228 & 0.5407 & 0.3129 & 0.4768   & 0.3491  & 0.4031 & 0.6479 & 0.3933 \\ \hline
2DCNN-AE & 0        & 0       & 0       & 0.5342 & 0.202  & 0.093   & 0.3268 & 0.1448 & 0.4804 & 0.2677 & 0.309    & 0.0708  & 0.1152 & 0.4552 & 0.27   \\ \hline
LSTM-AE  & 0.2787   & 0.047   & 0.0804  & 0.5667 & 0.2035 & 0.0961  & 0.2556 & 0.1397 & 0.4919 & 0.2758 & 0.9375   & 0.01    & 0.0198 & 0.4024 & 0.2858 \\ \hline
HIFI     & 0.2157   & 0.1519  & 0.1783  & 0.5478 & 0.191  & 0.091   & 0.2691 & 0.136  & 0.5305 & 0.3284 & 0.5139   & 0.1729  & 0.2587 & 0.5195 & 0.3274 \\ \hline
RN       & 0.2083   & 0.0138  & 0.0259  & 0.5679 & 0.2122 & 0.087   & 0.1496 & 0.11   & 0.5152 & 0.2743 & 0.2839   & 0.3645  & 0.3192 & 0.4529 & 0.2523 \\ \hline
RAE      & 0.1437   & 0.0691  & 0.0933  & 0.5302 & 0.1764 & 0.0738  & 0.1281 & 0.0936 & 0.4802 & 0.2714 & 0.1939   & 0.2096  & 0.2015 & 0.4302 & 0.2379 \\
\bottomrule
\end{tabular}
\end{table*}

\begin{table*}[]
\footnotesize
\caption{Results of DL Methods (Stationary vs. Non-stationary; Sliding window size=16, Hidden size=32)}
\begin{tabular}{llllllllllllllll}
\hline
         & \multicolumn{5}{l}{NAB-ambient-Nonstationary}  & \multicolumn{5}{l}{NAB-art-stationary}      & \multicolumn{5}{l}{NAB-cpu-stationary}        \\ \hline
         & Pre      & Rec     & F1      & ROC    & PR     & Pre     & Rec    & F1     & ROC    & PR     & Pre      & Rec     & F1     & ROC    & PR     \\ \hline
CNN-AE   & 0.1847   & 0.1602  & 0.1716  & 0.4956 & 0.165  & 0.0884  & 0.2686 & 0.1331 & 0.5588 & 0.3369 & 0.1527   & 0.0354  & 0.0575 & 0.3275 & 0.2076 \\ \hline
2DCNN-AE & 0.25     & 0.0166  & 0.0311  & 0.5476 & 0.1917 & 0.0908  & 0.3301 & 0.1425 & 0.5406 & 0.315  & 0.4757   & 0.2156  & 0.2967 & 0.5744 & 0.3404 \\ \hline
LSTM-AE  & 0.253    & 0.058   & 0.0944  & 0.5787 & 0.2024 & 0.0912  & 0.2902 & 0.1388 & 0.4885 & 0.2802 & 0.2333   & 0.0674  & 0.1046 & 0.3588 & 0.2316 \\ \hline
HIFI     & 0.2252   & 0.1381  & 0.1712  & 0.5461 & 0.1878 & 0.0914  & 0.2585 & 0.135  & 0.5254 & 0.3241 & 0.5099   & 0.1716  & 0.2567 & 0.5087 & 0.3212 \\ \hline
RN       & 0.3253   & 0.0746  & 0.1213  & 0.5775 & 0.2146 & 0.093   & 0.3026 & 0.1423 & 0.5245 & 0.3417 & 0.2651   & 0.3451  & 0.2999 & 0.4416 & 0.2476 \\ \hline
RAE      & 0.3122   & 0.2044  & 0.2471  & 0.5808 & 0.2246 & 0.07    & 0.1939 & 0.1028 & 0.5148 & 0.2893 & 0.2587   & 0.2717  & 0.2651 & 0.4825 & 0.2661 \\
\hline
\end{tabular}
\end{table*}

\begin{table*}[]
\footnotesize
\caption{Results of DL Methods (Stationary vs. Non-stationary; Sliding window size=16, Hidden size=64)}
\begin{tabular}{llllllllllllllll}
\hline
         & \multicolumn{5}{l}{NAB-ambient-Nonstationary}  & \multicolumn{5}{l}{NAB-art-stationary}      & \multicolumn{5}{l}{NAB-cpu-stationary}        \\ \hline
         & Pre      & Rec     & F1      & ROC    & PR     & Pre     & Rec    & F1     & ROC    & PR     & Pre      & Rec     & F1     & ROC    & PR     \\ \hline
CNN-AE   & 0.1982   & 0.1188  & 0.1485  & 0.5335 & 0.1782 & 0.0793  & 0.2556 & 0.121  & 0.5495 & 0.3206 & 0.3771   & 0.2283  & 0.2844 & 0.5541 & 0.3119 \\ \hline
2DCNN-AE & 0.2165   & 0.116   & 0.1511  & 0.5632 & 0.1932 & 0.0598  & 0.21   & 0.0931 & 0.4123 & 0.2288 & 0.428    & 0.3892  & 0.4077 & 0.576  & 0.338  \\ \hline
LSTM-AE  & 0.2083   & 0.0552  & 0.0873  & 0.5734 & 0.1989 & 0.0784  & 0.2456 & 0.1189 & 0.515  & 0.2771 & 0.2325   & 0.0688  & 0.1061 & 0.3821 & 0.2439 \\ \hline
HIFI     & 0.2222   & 0.1271  & 0.1617  & 0.5548 & 0.1899 & 0.0881  & 0.259  & 0.1315 & 0.5438 & 0.3337 & 0.5195   & 0.1776  & 0.2647 & 0.5224 & 0.3277 \\ \hline
RN       & 0.2792   & 0.1188  & 0.1667  & 0.5792 & 0.2133 & 0.0633  & 0.1949 & 0.0956 & 0.4232 & 0.2721 & 0.2947   & 0.3772  & 0.3309 & 0.4651 & 0.2587 \\ \hline
RAE      & 0.2437   & 0.1326  & 0.1717  & 0.5551 & 0.1956 & 0.0403  & 0.1627 & 0.0646 & 0.4305 & 0.2092 & 0.2396   & 0.1649  & 0.1953 & 0.4517 & 0.2529 \\
\hline
\end{tabular}
\end{table*}

\clearpage   \begin{table*}[]
\footnotesize
\caption{Results of DL Methods (Stationary vs. Non-stationary; Sliding window size=16, Hidden size=128)}
%
\end{table*}

\clearpage   \begin{table*}[]
\footnotesize
\caption{Results of DL Methods (Stationary vs. Non-stationary; Sliding window size=16, Hidden size=256)}
%
\end{table*}

\clearpage   \begin{table*}[]
\footnotesize
\caption{Results of DL Methods (Stationary vs. Non-stationary; Sliding window size=32, Hidden size=16)}
%
\end{table*}

\clearpage   \begin{table*}[]
\footnotesize
\caption{Results of DL Methods (Stationary vs. Non-stationary; Sliding window size=32, Hidden size=64£©}
%
\end{table*}

\justifying

\clearpage   \begin{table*}[]
\vspace{-0.5cm}
\footnotesize
\caption{Results of DL Methods (Stationary vs. Non-stationary; Sliding window size=128, Hidden size=128)}
\vskip -9pt
%
\end{table*}

\justifying
\clearpage   \begin{table*}[]
\footnotesize
\vspace{-0.5cm}
\caption{Results of DL Methods (Stationary vs. Non-stationary; Sliding window size=128, Hidden size=256)}
%
\end{table*}
\clearpage

\clearpage   \begin{table*}[]
\footnotesize
\caption{Results of DL Methods (Stationary vs. Non-stationary; Sliding window size=256, Hidden size=32)}
%
\end{table*}
\clearpage

\clearpage   \begin{table*}[]
\footnotesize
\caption{Results of DL Methods (Stationary vs. Non-stationary; Sliding window size=256, Hidden size=64)}
%
\end{table*}
\clearpage

\begin{table*}[t]
\footnotesize
\caption{Results of DL Methods (Stationary vs. Non-stationary; Sliding window size=256, Hidden size=128)}
%
\end{table*}
\clearpage

\begin{table*}[t]
\footnotesize
\caption{Results of DL Methods (Stationary vs. Non-stationary; Sliding window size=256, Hidden size=256)}
%
\end{table*}
\clearpage

\clearpage   \begin{table*}[]
\small
\caption{Robustness (Noise Proportions equals to 1\%)}
%
\end{table*}

\clearpage   \begin{table*}[]
\small
\caption{Robustness (Noise Proportions equals to 2\%)}
%
\end{table*}

\clearpage   \begin{table*}[]
\small
\caption{Robustness (Noise Proportions equals to 3\%)}
%
\end{table*}

\clearpage   \begin{table*}[]
\small
\caption{Robustness (Noise Proportions equals to 4\%)}
%
\end{table*}

\justifying
\section{Conclusion and Future Directions}\label{sec:conclu}
A wealth of time series anomaly detection methods exist, and selecting the most suitable ones for particular application settings is challenging as the performance of the proposed methods vary widely across different settings.
In this study, we provide a comprehensive overview of time series
anomaly detection and evaluate judiciously chosen state-of-the-art traditional and
deep-learning-based methods,
based on which we offer recommendations for selecting suitable methods for different anomaly detection applications.
This study indicates a need for new research to advance the state of the art in time series anomaly detection, as well as  for research on integrated solutions to the discovery and explanation of anomalies.

Four specific directions are promising.

First, although DL has pushed the limits of what is possible in time series anomaly detection, it is too early to conclude that the traditional techniques that had been undergoing progressive development in the years prior to the rise of DL have become obsolete. This paper provides systematic taxonomies for data, methods, and evaluation strategies, and it evaluates the state-of-the-art traditional and DL methods.
In the future, it is of interest to explore how the two sides of anomaly detection can be combined to improve anomaly detection performance and to tackle problems not amenable to DL methods.

Second, the boundary between normal and anomalous behaviors is often not precisely defined and may evolve over time. The lack of well-defined representative normal boundaries poses challenges for both traditional and DL methods.

Third, other factors need to be explored that might affect the choice of methods, such as the way in which data is monitored and the environment in which anomaly detection is deployed.

Fourth, future efforts should also consider the creation of a larger high-quality anomaly detection database that can serve as a general benchmark for validating novel proposals.

\clearpage
\bibliographystyle{ACM-Reference-Format}
\bibliography{0main}

\end{document}